\documentclass{article}

\PassOptionsToPackage{numbers, compress}{natbib}
\usepackage[preprint]{neurips_2026}

\usepackage{amsmath,amsfonts,bm}

\def\eqref#1{equation~\ref{#1}}

\def\1{\bm{1}}

\DeclareMathAlphabet{\mathsfit}{\encodingdefault}{\sfdefault}{m}{sl}
\SetMathAlphabet{\mathsfit}{bold}{\encodingdefault}{\sfdefault}{bx}{n}

\usepackage[utf8]{inputenc} 
\usepackage[T1]{fontenc}    
\usepackage{hyperref}       
\usepackage{url}            
\usepackage{booktabs}       
\usepackage{amsfonts}       
\usepackage{algorithm}
\usepackage{algorithmic}
\usepackage{bm}
\usepackage{nicefrac}       
\usepackage{microtype}      
\usepackage{xcolor}         
\usepackage{multirow}
\usepackage{graphicx}
\usepackage{subcaption}
\usepackage{cleveref}

\usepackage{amsmath}
\usepackage{amssymb}
\usepackage{amsthm}
\usepackage{enumitem}
\usepackage{mathtools} 

\newtheorem{proposition}{Proposition}[section]

\newtheorem{assumption}{Assumption}

\crefname{assumption}{Assumption}{Assumptions}
\Crefname{assumption}{Assumption}{Assumptions}

\title{Correcting Gradient-Based Circuit Localization via Interaction-Aware Backpropagation}

\author{%
Joakim Edin$^{1,3}$ \quad Casper L. Christensen $^{1}$ \quad Róbert Csordás $^{2}$ \quad Tuukka Ruotsalo$^{3,4}$ \\
\textbf{Zhengxuan Wu}$^2$ \quad \textbf{Maria Maistro}$^3$ \quad  \textbf{Jing Huang}$^2$ \quad \textbf{Lars Maaløe}$^1$\\
$^1$Corti \quad $^2$Stanford University \quad $^3$Copenhagen University $^4$LUT University\\
\texttt{je@corti.ai}
}

\begin{document}

\maketitle

\begin{abstract}
Circuit localization methods aim to identify the subset of model components responsible for specific behaviors in large language models, enabling detailed mechanistic analysis. Most existing methods assume components act independently and estimate importance by perturbing each component in isolation. However, components in neural networks interact, and ignoring these interactions leads to systematic misestimation of component importance.
We find that one particularly problematic interaction is attention self-repair, in which softmax redistribution causes gradients for influential attention scores to vanish as other positions with similar values compensate.
We introduce Gradient Interaction Modifications (GIM), a technique that explicitly accounts for feature interactions during backpropagation. GIM achieves state-of-the-art performance on the circuit localization track of the Mechanistic Interpretability Benchmark and outperforms existing gradient-based methods on feature attribution across diverse tasks. By accounting for interaction effects and explaining why prior methods underestimate component importance, GIM enables more faithful mechanistic analysis of large language models. GIM is available as a Python package at \href{https://github.com/corticph/gim}{https://github.com/corticph/gim}.
\end{abstract}

\begin{figure*}[h]
    \centering
    \includegraphics[width=0.83\linewidth]{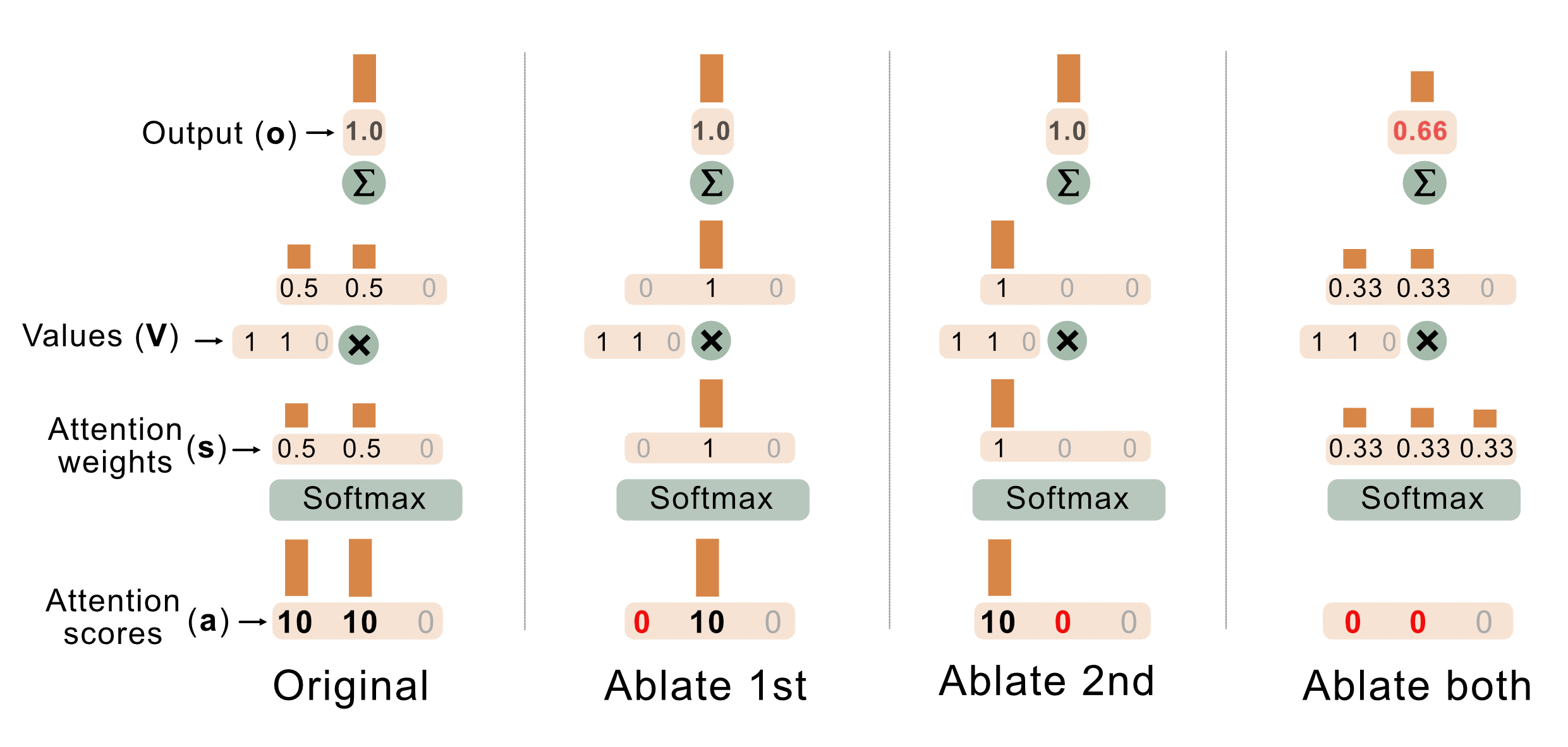}
    \caption{Attention self-repair. When multiple values associated with large attention weights contain similar information, ablating one attention score has little effect on the output because the softmax activation function compensates by increasing the weight of other positions. This results in perturbation-based and gradient-based explanation methods underestimating the importance of components that contribute to the attention scores through the keys and queries.}
    \label{fig:self-repair}
\end{figure*}

\section{Introduction}
Large language models (LLMs) have demonstrated remarkable capabilities, yet our understanding of their inner mechanisms remains limited. This opacity hinders our ability to explain model outputs, catch hallucinations, and ensure reliable behavior. The field of mechanistic interpretability aims to address this by deciphering how models process information internally. One key technique is \textit{circuit localization}, which isolates the subset of model components (e.g., attention heads, neurons, connections) responsible for a specific behavior, enabling researchers to study them in detail~\citep{wangInterpretabilityWildCircuit2022}.

Most circuit localization methods assume components are independent, measuring each component's importance by perturbing it individually and observing the effect on the output~\citep{conmyAutomatedCircuitDiscovery2023, mengLocatingEditingFactual}. However, components in neural networks interact; perturbing multiple components simultaneously can yield different effects than perturbing them individually. When interactions are ignored, methods may systematically misestimate component importance.

We propose \textbf{Gradient Interaction Modifications (GIM)}, a gradient-based attribution method that estimates the impact of perturbing multiple interacting components simultaneously. We develop GIM by analyzing how different operations in transformers introduce feature interactions, then design targeted backpropagation modifications to account for them. Through this process, we discover that the attention mechanism introduces an OR-gate-like interaction: when multiple positions have large attention weights and similar value vectors, perturbing any single attention score has minimal effect because the softmax redistributes weight to redundant positions (\Cref{fig:self-repair}). We denote this phenomenon \textbf{attention self-repair} and prove that it causes gradients to vanish even for influential attention scores. Our experiments show that this occurs across all inputs and LLMs.

GIM addresses attention self-repair through \textbf{temperature-adjusted softmax gradients (TSG)}, which approximate the effect of perturbing multiple attention scores simultaneously. GIM also incorporates \textbf{layernorm freeze}~\citep{rushingExplorationsSelfRepairLanguage2024} and \textbf{gradient normalization}~\citep{achtibatAttnLRPAttentionAwareLayerWise2024} to handle interactions in normalization layers and multiplicative operations.

Our main contributions are:
\begin{enumerate}
    \item We introduce \textbf{GIM}, a gradient-based attribution method that accounts for component interactions. GIM achieves state-of-the-art performance on the circuit localization track of the Mechanistic Interpretability Benchmark~\citep{muellerMIBMechanisticInterpretability2025}.
    \item We identify \textbf{attention self-repair}, an OR-gate-like interaction where softmax redistribution causes attribution methods to underestimate component importance, and prove it leads to vanishing gradients.
    \item We show that GIM can also be used for feature attribution, outperforming existing methods across several LLMs and tasks.
\end{enumerate}

\section{Related Work}

\subsection{Circuit Localization}\label{sec:rw_cl}

A neural network can be viewed as a computational graph, where nodes are components (e.g., attention heads, MLPs) and edges represent information flowing between them. Circuit localization methods estimate the importance of either nodes or edges for a specific behavior. By isolating the most important nodes or edges, we obtain a subgraph called a \textit{circuit}, which can then be studied in detail. We focus on edge-level methods, which provide finer granularity than node-level by revealing how information flows between components. Edge Activation Patching (EActP)~\citep{conmyAutomatedCircuitDiscovery2023} estimates edge importance by measuring the change in model output when replacing an edge's activation with a counterfactual activation. However, this requires one forward pass per edge, which is computationally expensive for large models that comprise billions of edges. Edge Attribution Patching (EAP)~\citep{syedAttributionPatchingOutperforms2023} is a faster approximation that leverages gradients to estimate all edge importances in a single backward pass. EAP first runs a forward pass on a counterfactual example to cache activations $\tilde{a}_i$, then runs a forward and backward pass on the original example. The importance of edge $(i, j)$ from upstream node $i$ to downstream node $j$ is:
\begin{equation}\label{eq:eap}
    I_{ij} = (a_i - \tilde{a}_i) \cdot \frac{\partial z}{\partial h_j}
\end{equation}
where $a_i$ is the activation of node $i$ on the original example, $h_j$ is the input to node $j$, and $z$ is the model output. EAP-IG~\citep{hannaHaveFaithFaithfulness2024} extends EAP with integrated gradients~\citep{shrikumarLearningImportantFeatures2017}. Information Flow Routes (IFR)~\citep{ferrandoInformationFlowRoutes2024} tracks information flow by analyzing attention patterns across the layers. Uniform Gradient Sampling (UGS)~\citep{liOptimalAblationInterpretability2024} uses gradient descent to learn a pruning mask over edges, optimizing for sparsity while maintaining model behavior. Hybrid-ens is an ensemble method that combines EAP-IG with a edge-masking method similar to UGS~\cite{mondorfBlackboxNLP2025MIBShared2025}. All these approaches estimate the effect of perturbing each edge individually, while we estimate the effect of perturbing multiple edges simultaneously.

\subsection{Feature Interactions} \label{sec:rw_fi}

Feature interactions occur when the joint effect of perturbing multiple components differs from the sum of their individual effects~\citep{tsangHowDoesThis2020}. A well-studied interaction in language models is \textit{self-repair}: when a component is ablated, other components compensate, making the ablated component appear unimportant~\citep{wangInterpretabilityWildCircuit2022, mcgrathHydraEffectEmergent2023}. This is a form of logical OR-gate behaviour. Prior work has identified backup attention heads~\citep{mcdougallCopySuppressionComprehensively2023}, MLP erasure neurons~\citep{mcgrathHydraEffectEmergent2023}, and layer normalization rescaling~\citep{rushingExplorationsSelfRepairLanguage2024} as sources of self-repair. Layer normalization rescaling occurs because ablating a component reduces the magnitude of activations, causing layer normalization to amplify the remaining components' contributions. We identify a new source: softmax redistribution within the attention mechanism.

\subsection{Gradient Modifications for Feature Interactions}\label{sec:rw_gm}

First-order gradients measure the effect of infinitesimal perturbations to each edge independently, failing to capture feature interactions. Previous work has attempted to leverage higher-order gradients to estimate interactions, but these methods require a prohibitive amount of memory and computation~\citep{janizekExplainingExplanationsAxiomatic}. Instead, several studies modify standard backpropagation to better account for feature interactions. TransformerLRP~\citep{aliXAITransformersBetter2022} treats layer normalization and attention weights as constants during backpropagation, thereby avoiding the interactions introduced by these operations. However, not backpropagating through attention weights discards potentially crucial information. \citet{achtibatAttnLRPAttentionAwareLayerWise2024} propose normalizing the gradients of multiplicative interactions by dividing by the number of inputs. We refer to this method as \textit{Gradient Norm} and call treating layer normalization as a constant \textit{Layernorm Freeze}. We show that combining these two methods yields considerably larger gains than using either individually. Furthermore, we propose a new method, \textit{Temperature-adjusted Softmax Gradients (TSG)}, to address attention self-repair.

\section{Attention Self-Repair}\label{sec:self-repair-intro}
Self-repair in the attention layer is when an attention score strongly influences the output, yet perturbing it has minimal effect. This effect deems both perturbation- and gradient-based attribution methods inaccurate. In this section, we show when and why attention self-repair occurs and its consequences on gradient-based explanation methods.

In transformer models, attention mechanisms control information flow between hidden states at different positions. Position $i$ gathers information from all positions using the following computations:
\begin{align}\label{eq:attention}
\begin{split}
  &\bm{a} = \bm{Q}_i \cdot \bm{K}^T \quad \text{(Attention scores for position $i$)} \\
  &\bm{s} = \text{Softmax}(\bm{a}) \quad \text{(Attention weights for position $i$)} \\
  &\bm{o} = \bm{s} \cdot \bm{V} \quad \text{(Attention output for position $i$)}
\end{split}
\end{align}
where $\bm{Q}_i$ is the query for position $i$, $\bm{K}$ are the keys, and $\bm{V}$ are the values. The softmax function normalizes attention scores into a probability distribution:
\begin{align} \label{eq:softmax}
\text{Softmax}(\bm{a}) = \frac{e^{\bm{a}/\tau}}{\sum_{k} e^{a_{k}/\tau}}
\end{align}
where $\tau$ is the temperature parameter. For clarity, we distinguish between softmax input ``attention scores'' $\bm{a}$ and softmax output ``attention weights'' $\bm{s}$. 

Attention mechanisms can be viewed as information routing systems, where each value vector $\bm{V}_j$
contains information from position $j$, and the attention weights $\bm{s}_j$ determine how much information to copy from that position. 

Self-repair occurs when multiple values with large attention weights contain similar information. When an attention score $\bm{a}_j$ is ablated, the softmax redistributes weights primarily to positions with the highest remaining scores. If these positions contain values $\bm{V}_j$
that contribute similar information (their cosine similarity is close to one), the output remains virtually unchanged.

\Cref{fig:self-repair} illustrates this with attention weights $[0.5,0.5,0]$ and value vectors $[1,1,0]$. Note that the values are identical
at the positions with non-zero attention weights. Individually ablating either attention score (i.e., softmax input) leaves the output unchanged, as the softmax shifts the weight to the other position with the same value. Only ablating both scores simultaneously affects the output, as the softmax shifts the attention weights to the position with a different value.

\subsection{Attention self-repair yields vanishing attention-score gradients}
\label{sec:self-repair-proof}
We now prove that under attention self-repair, the attention-score
gradient $\partial z/\partial \bm{a}_j$ becomes near zero even when the
softmax-weight gradient $\partial z/\partial \bm{s}_j$ is large. This
may seem counter-intuitive: if an infinitely small perturbation on the
softmax weight impacts the model's output, we would expect the same
from perturbing the attention score. However, the gradient signal is lost when backpropagating through the softmax function.
 
Let $z\in\mathbb{R}$ denote a model output logit (e.g., the logit of
the predicted token), and let $\bm{a},\bm{s}\in\mathbb{R}^n$ with
$\bm{s}=\mathrm{Softmax}(\bm{a})$ as in \Cref{eq:attention}; we set
$\tau=1$ without loss of generality, as the gradient below is rescaled
by $1/\tau$ for $\tau>0$. Define
\begin{equation}\label{eq:grad}
g_k \;:=\; \frac{\partial z}{\partial \bm{s}_k}.
\end{equation}
The softmax derivatives are
$\partial \bm{s}_j/\partial \bm{a}_j = \bm{s}_j(1-\bm{s}_j)$ and
$\partial \bm{s}_k/\partial \bm{a}_j = -\bm{s}_k\bm{s}_j$ for $k\neq j$.
Splitting the chain-rule sum into the diagonal $k=j$ term and the
remaining off-diagonal terms gives
\begin{equation*}
\frac{\partial z}{\partial \bm{a}_j}
\;=\; g_j\,\bm{s}_j(1-\bm{s}_j)\;-\;\sum_{k\neq j}g_k\,\bm{s}_k\bm{s}_j
\;=\; \bm{s}_j\Bigl((1-\bm{s}_j)\,g_j \;-\; \sum_{k\neq j}\bm{s}_k\,g_k\Bigr).
\end{equation*}
Since softmax weights sum to one, $1-\bm{s}_j = \sum_{k\neq j}\bm{s}_k$;
substituting this in the first term yields the symmetric form
\begin{equation}\label{eq:softmax_grad}
\frac{\partial z}{\partial \bm{a}_j}
\;=\; \bm{s}_j\!\sum_{k\neq j}\bm{s}_k\,(g_j-g_k),
\end{equation}
The gradient is small whenever $\bm{s}_j$ is small or the sum is
small. Each term in the sum is small whenever $\bm{s}_k$ is small or
$g_j-g_k$ is small. Since softmax weights are strictly positive,
$\bm{s}_k$ is never exactly zero, but it can be arbitrarily close.
 
For a threshold $\epsilon>0$, we call positions \emph{significant} if
$\bm{s}_k>\epsilon$ and \emph{negligible} otherwise; let
$\mathcal{I}_\epsilon=\{k:\bm{s}_k>\epsilon\}$ be the set of significant positions and
$\mathcal{I}_\epsilon^c$ its complement.
\begin{assumption}[Equivalent contributions on $\mathcal{I}_\epsilon$]\label{ass:repair}
$g_j=g_k$ for every $j,k\in\mathcal{I}_\epsilon$.
\end{assumption}
\Cref{ass:repair} is implied by (but weaker than)
$\bm{V}_j=\bm{V}_k$ for $j,k\in\mathcal{I}_\epsilon$, capturing the
cosine-similarity-close-to-one regime that motivates self-repair.
\begin{proposition}[Self-repair makes the attention-score gradient small]
\label{prop:self-repair}
Under \Cref{ass:repair}, for every $j\in\mathcal{I}_\epsilon$,
\[
\Bigl|\frac{\partial z}{\partial \bm{a}_j}\Bigr|
\;\leq\;
\epsilon\,\bm{s}_j\!\sum_{k\in\mathcal{I}_\epsilon^c}|g_j-g_k|.
\]
\end{proposition}
The smaller the $\epsilon$ for which \Cref{ass:repair} holds, the
tighter the bound; in the idealized limit $\epsilon\to 0$, the gradient
vanishes.
\begin{proof}
Fix $j\in\mathcal{I}_\epsilon$ and split the sum in
\eqref{eq:softmax_grad} over significant and negligible $k$:
\[
\sum_{k\neq j}\bm{s}_k(g_j-g_k)
\;=\;
\underbrace{\sum_{k\in\mathcal{I}_\epsilon\setminus\{j\}}\!\!\bm{s}_k(g_j-g_k)}_{=\,0\text{ by \Cref{ass:repair}}}
\;+\;\sum_{k\in\mathcal{I}_\epsilon^c}\bm{s}_k(g_j-g_k).
\]
Since $\bm{s}_k\leq\epsilon$ on $\mathcal{I}_\epsilon^c$, the second
sum is bounded in absolute value by
$\epsilon\sum_{k\in\mathcal{I}_\epsilon^c}|g_j-g_k|$. Multiplying by
$\bm{s}_j$ gives the claim.
\end{proof}
In practice, $\epsilon$ will be small but not zero. Furthermore, $g_j$ and $g_k$ will be close but rarely identical, so the attention-score gradient becomes small but does not vanish exactly.

Mechanically, $\bm{a}_j$ has opposing effects on the softmax: it positively influences $\bm{s}_j$ but negatively influences every other
$\bm{s}_k$, and vice versa. When $g_j$ and $g_k$ are similar (i.e., $\partial z/\partial \bm{s}_j$ and $\partial z/\partial \bm{s}_k$), the
gain from boosting $\bm{s}_j$ cancels the loss from suppressing $\bm{s}_k$, and the gradient disappears. However, $s_j$ and $s_k$ still have a positive influence on the output: simultaneously suppressing both would impact it. This motivates the gradient modification in \Cref{sec:tsg}.

\section{Gradient Interaction Modifications}

To address feature interactions in transformer models, we introduce  \textbf{Gradient Interaction Modifications (GIM)}, an approach that combines three complementary modifications of standard back-propagation. This section describes the three modifications: \textbf{Temperature-adjusted Softmax Gradients (TSG)}, \textbf{Layernorm Freeze}, and \textbf{Gradient Norm}.

\subsection{Temperature-adjusted softmax gradients}\label{sec:tsg}
TSG modifies backpropagation through attention to address the
self-repair problem from \Cref{sec:self-repair-proof}. The motivation
is captured by Pearl's ``firing squad'' analogy~\citep{pearlCausality2009}:
two soldiers fire at a prisoner, and we want each soldier's causal
importance. Preventing each from firing in turn misleadingly suggests
both are innocent, since either shot alone is fatal; their true
responsibility is revealed only by jointly preventing both.

Attention self-repair has the same OR-gate structure: ablating any
single attention score with significant softmax weight leaves the
output unchanged because the others compensate. Standard gradients
measure the effect of perturbing each score in isolation and so become
small at exactly the influential scores (\Cref{prop:self-repair}).
TSG instead approximates the gradient that would arise under
simultaneous perturbation of multiple attention scores. We developed
TSG empirically, selecting techniques that best approximated joint
ablation while producing faithful explanations; we validate the
approximation in \Cref{sec:results}.

During backpropagation, TSG recomputes the softmax with an elevated
temperature ($T > 1$) before computing the softmax gradients. This
flattens the distribution: large weights shrink and small weights grow,
enlarging the set $\mathcal{I}_\epsilon$ of significant positions from
\Cref{prop:self-repair}. The proposition's bound is small only when
\Cref{ass:repair} holds on $\mathcal{I}_\epsilon$, requiring that the
contributions $g_k$ agree across $\mathcal{I}_\epsilon$. Enlarging
$\mathcal{I}_\epsilon$ admits positions whose contributions are
unlikely to match, so the pairwise differences $g_j-g_k$ in
\eqref{eq:softmax_grad} no longer all vanish, and the gradient
survives.

\subsection{Layernorm freeze}
Layer normalization introduces interactions between components. In transformers, the representation at each position is a sum of outputs from the components in the previous layers (attention heads, MLPs, and the residual stream). Layer normalization is applied to this summed representation, computing $\sigma$ across all dimensions. When one component's contribution changes, $\sigma$ changes, which in turn rescales the contributions from every other component. This creates a zero-sum dynamic: increasing one component's magnitude causes layernorm to downweight all components proportionally. When a component is ablated, the reduced activation magnitude causes layernorm to amplify the remaining components' contributions. This is a form of self-repair identified by \citet{rushingExplorationsSelfRepairLanguage2024}. Standard gradients capture this compensation, causing them to underestimate the importance of ablated components. To address this interaction, Layernorm Freeze treats the normalization factor $\sigma$ as a constant during backpropagation, measuring each component's direct contribution before normalization rescaling occurs.

\subsection{Gradient Norm} 
Multiplicative operations introduce interactions when multiple inputs depend on the same upstream variable. Consider computing the attention matrix $A = QK^T$, where $Q = XW_Q$ and $K = XW_K$ both depend on input $X$. Our goal is to estimate the change in $A$ when we replace $X$ with a counterfactual $\tilde{X}$. Following EAP, we approximate this change using a first-order Taylor expansion:

\begin{equation}
A(X) - A(\tilde{X}) \approx (X - \tilde{X}) \cdot \frac{\partial A}{\partial X}
\end{equation}
By the product rule, the gradient is:
\begin{equation}
\frac{\partial A}{\partial X} = W_Q K^T + Q W_K^T
\end{equation}
However, this gradient double-counts the contribution of $X$. The first term $W_Q K^T$ captures the effect of $X$ through $Q$ while holding $K$ constant, and the second term $Q W_K^T$ captures the effect through $K$ while holding $Q$ constant. In reality, perturbing $X$ affects both $Q$ and $K$ simultaneously.

To illustrate, consider a concrete example with $\tilde{X} = 0$. Let $X=3$, $W_Q=4$, $W_K=5$. Then $Q = 12$, $K = 15$, and $A = QK = 180$. The true change when setting $X$ to zero is $A(X) - A(\tilde{X}) = 180 - 0 = 180$. However, the gradient-based estimate gives:
\begin{equation}
\begin{aligned}
(X - \tilde{X}) \cdot \frac{\partial A}{\partial X} = 3 \cdot (W_Q K + Q W_K) 
= 3 \cdot (4 \cdot 15 + 12 \cdot 5)  
= 3 \cdot 120 = 360
\end{aligned}
\end{equation}
This is exactly twice the true change. To correct this overcounting, Gradient Norm divides the gradient by the number of multiplicative inputs. We apply this correction (dividing by 2) at three locations in the transformer: attention-value multiplication, query-key multiplication, and MLP gate-projection multiplication. This approach aligns with the ``uniform rule'' proposed by \citet{achtibatAttnLRPAttentionAwareLayerWise2024}, who provide formal justification via Taylor decomposition and Shapley values.


\section{Experimental Setup}

Our experiments are designed to evaluate three aspects of our contribution: (1) comparing GIM against existing circuit localization methods on faithfulness metrics, (2) comparing GIM against existing gradient-based feature attribution methods on faithfulness metrics, and (3) empirically validating that attention self-repair occurs in LLMs and that TSG effectively approximates joint ablation effects. This section describes the experimental setup for each evaluation.

\textbf{Circuit localization:}
To use GIM for circuit localization, we employ Edge Attribution Patching (EAP) but replace standard backpropagation with GIM (see \Cref{app:gim_feature_attribution} for more details). 

We evaluate GIM on the circuit localization track in the Mechanistic Interpretability Benchmark (MIB)~\citep{muellerMIBMechanisticInterpretability2025}. This benchmark evaluates the faithfulness of circuits identified by circuit localization methods across multiple tasks and LLMs. The tasks are Indirect Object Identification (IOI), Arithmetic, Multiple-Choice Question Answering (MCQA), and AI2 Reasoning Challenge (ARC). The LLMs are GPT-2 120M, Qwen-2.5 0.5B, Gemma-2 2B, and LLama-3.1 8B. We use the public test set.

We compare GIM with the circuit localization methods evaluated by \citet{muellerMIBMechanisticInterpretability2025}. These methods include Edge Activation Patching (EActP), Edge Attribution Patching (EAP), EAP with integrated gradients (EAP-IG), Hybrid-ens,  Information Flow Routes (IFR), and Uniform Gradient Sampling (UGS) (see \Cref{sec:rw_cl}). 

MIB evaluates methods using the integrated-circuit performance ratio (CPR) and the integrated-circuit model distance (CMD) metrics. Similar to \citet{muellerMIBMechanisticInterpretability2025}, we used the absolute values of the attribution scores when computing the CMD scores.  We refer to \Cref{app:cpr_cmd,app:cl_datasets} for more details about the datasets and metrics.

\textbf{Feature attributions}
For our experiments, we use seven instruction-tuned LLMs spanning three model families of various sizes: Gemma-2 (2B, 9B), LLaMA-3.2 (1B, 3B), LLaMA-3.1 (8B), and Qwen-2.5 (1.5B, 3B)~\citep{teamGemma2Improving2024,grattafioriLlama3Herd2024, qwenQwen25TechnicalReport2025}. 

We use six datasets spanning four tasks: BoolQ (question-answering), FEVER and SciFact (fact verification), Twitter sentiment classification and Movie review (sentiment classification), and HateXplain (hate speech detection). These datasets were selected based on their diversity and prevalence in prior attribution studies~\citep{ lyuFaithfulModelExplanation2024}. We provide more details on the datasets in \Cref{app:datasets}.

We evaluate the faithfulness of the attributions using Comprehensiveness and Sufficiency~\citep{deyoungERASERBenchmarkEvaluate2020}. We describe these metrics in \cref{app:comp_suff}.

We compare GIM with five gradient-based methods:  GradientXInput (GxI)~\citep{sundararajanAxiomaticAttributionDeep2017a}, Integrated gradients (IG)~\citep{sundararajanAxiomaticAttributionDeep2017a}, DeepLIFT~\citep{shrikumarLearningImportantFeatures2017}, TransformerLRP~\citep{aliXAITransformersBetter2022}, and AttnLRP~\citep{achtibatAttnLRPAttentionAwareLayerWise2024}. We did not evaluate perturbation-based methods because they were too computationally expensive.

We compute feature attributions with GIM using the same approach as for circuit localization, but computing importance scores for token embeddings rather than edges between model components (see \Cref{app:gim_feature_attribution} for details). We used the embedding of the whitespace token as the counterfactual. We used a temperature of 2 for TSG for both circuit localization and feature attribution, selected based on results obtained with Gemma-2 2B on the FEVER and HateXplain datasets.

\textbf{Attention self-repair in LLMs:}
We investigate whether attention self-repair occurs in real LLMs and
whether TSG accurately approximates joint attention-score perturbation.
To find concrete instances in which \Cref{ass:repair} holds
approximately, we treat $\epsilon$ as a hyperparameter and set it to
$0.01$. We focus on the top $1\%$ most important attention weights as
ranked by $g_j\,\bm{s}_j$, and keep only those vectors for which
$\mathcal{I}_\epsilon = \{k : \bm{s}_k > \epsilon\}$ contains at least
two positions. We then test \Cref{ass:repair} by computing the
coefficient of variation of $g_j$ across $\mathcal{I}_\epsilon$,
classifying the case as self-repair when it falls below~$0.1$. These
thresholds serve only to produce a countable set of instances for
plotting.

\section{Results}\label{sec:results}

\begin{table*}[t]
\centering
\caption{\textbf{CPR} scores across circuit localization methods and ablation types. All evaluations were performed using counterfactual ablations. Higher scores are better. Arithmetic scores are averaged across addition and subtraction. We \textbf{bold} and \underline{underline} the best and second-best methods per column.}
\label{tab:cpr}
\resizebox{\linewidth}{!}{%
\begin{tabular}{lcccccccccccc}
\toprule
 & \textbf{Average} & \multicolumn{4}{c}{IOI} & \multicolumn{1}{c}{Arithmetic} & \multicolumn{3}{c}{MCQA} & \multicolumn{2}{c}{ARC (E)} & \multicolumn{1}{c}{ARC (C)}\\
\cmidrule(lr){3-6} \cmidrule(lr){7-7} \cmidrule(lr){8-10} \cmidrule(lr){11-12}\cmidrule(lr){13-13}
  \textbf{Method} & & GPT-2 & Qwen-2.5 & Gemma-2 & Llama-3.1 & Llama-3.1 & Qwen-2.5 & Gemma-2 & Llama-3.1 & Gemma-2 & Llama-3.1 & Llama-3.1 \\
\midrule
Random & 0.27 & 0.25 & 0.28 & 0.30 & 0.25 & 0.25 & 0.27 & 0.32 & 0.26 & 0.32 & 0.26 & 0.25 \\
\midrule
UGS & - & 0.97 & 0.98 & - & - & - & \underline{1.17} & - & - & - & - & - \\
EActP & - & \textbf{2.30} & 1.21 & - & - & - & 0.85 & - & - & - & - & - \\
Hybrid-ens & - & 2.23 & \underline{1.88} & - & - & - & 1.22 & \\ 
IFR & 0.40 & 0.58 & 0.31 & 0.25 & 0.09 & 0.89 & 0.40 & 0.38 & 0.52 & 0.34 & 0.36 & 0.24 \\
EAP & 0.93 & 1.20 & 0.26 & 1.29 & 0.85 & 0.55 & 0.85 & 1.49 & 1.00 & 1.08 & 0.80 & 0.82 \\
EAP-IG-activations & 1.30 & 1.82 & 1.63 & 2.07 & 1.60 & 0.98 & 0.77 & 1.57 & 0.79 & \underline{1.70} & 0.71 & 0.63 \\
EAP-IG-inputs & \underline{1.56} & 1.85 & 1.63 & \underline{3.20} & \underline{2.08} & \underline{0.99} & 1.16 & \underline{1.64} & \underline{1.05} & 1.53 & \underline{1.04} & \underline{0.98} \\
\midrule
\textbf{GIM} & \textbf{2.13} & \underline{2.24} & \textbf{2.02}  & \textbf{3.54} & \textbf{2.54} & \textbf{1.09} & \textbf{2.47} & \textbf{2.52} & \textbf{1.65} & \textbf{2.36} & \textbf{1.39} & \textbf{1.65} \\
\bottomrule
\end{tabular}}
\end{table*}

\textbf{GIM is more faithful than other circuit localization methods:} \Cref{tab:cpr} compares circuit localization methods using the CPR metric. GIM significantly outperforms the other methods on most tasks and models. Notably, GIM achieves the highest average CPR score across all model sizes, demonstrating that the method scales effectively. The improvements are particularly pronounced on tasks requiring multi-step reasoning, such as Arithmetic and ARC. \Cref{tab:cmd} compares circuit localization methods using the CMD metric. Here, GIM performs comparably to EAP-IG. Since CMD uses the absolute values of attribution scores, whereas CPR preserves the sign, GIM's dominance in CPR could be explained by GIM's superior ability to distinguish between negative and positive attributions.

\textbf{GIM is often more faithful than other gradient-based feature attribution methods:} \Cref{tab:fa_results} compares GIM against five other gradient-based attribution methods across multiple datasets and model architectures. GIM significantly outperforms traditional methods, such as GradientXInput, Integrated Gradients, and DeepLIFT, in nearly all cases. While AttnLRP outperforms GIM on several dataset-model pairs, GIM achieves the highest comprehensiveness and sufficiency scores for the majority of combinations, establishing it as the most consistently faithful method overall.

\begin{figure*}[b]
     \centering
     \begin{subfigure}[b]{0.3\textwidth}
         \centering
         \caption{}
         \includegraphics[width=0.95\textwidth]{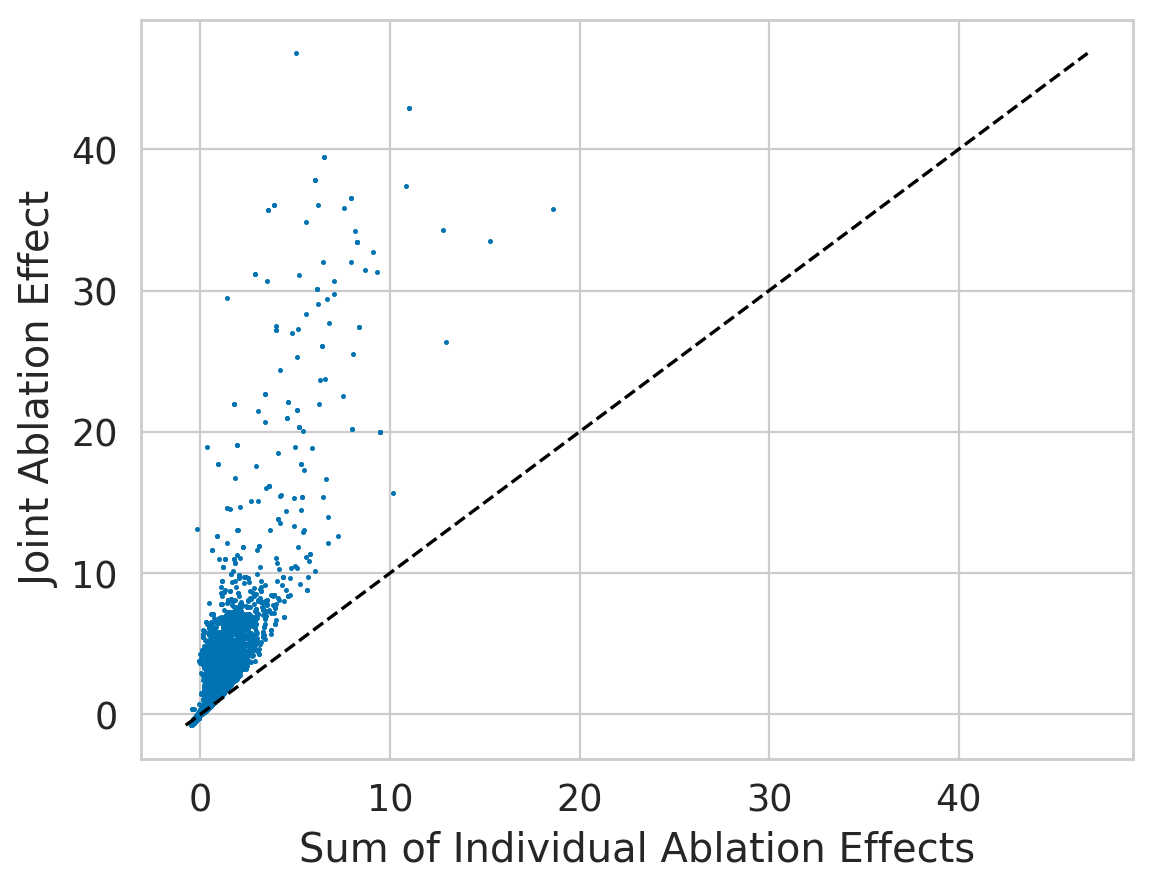}
         
         \label{fig:ablation_effects}
     \end{subfigure}
     \hfill
     \begin{subfigure}[b]{0.3\textwidth}
         \centering
         \caption{}
         \includegraphics[width=0.95\textwidth]{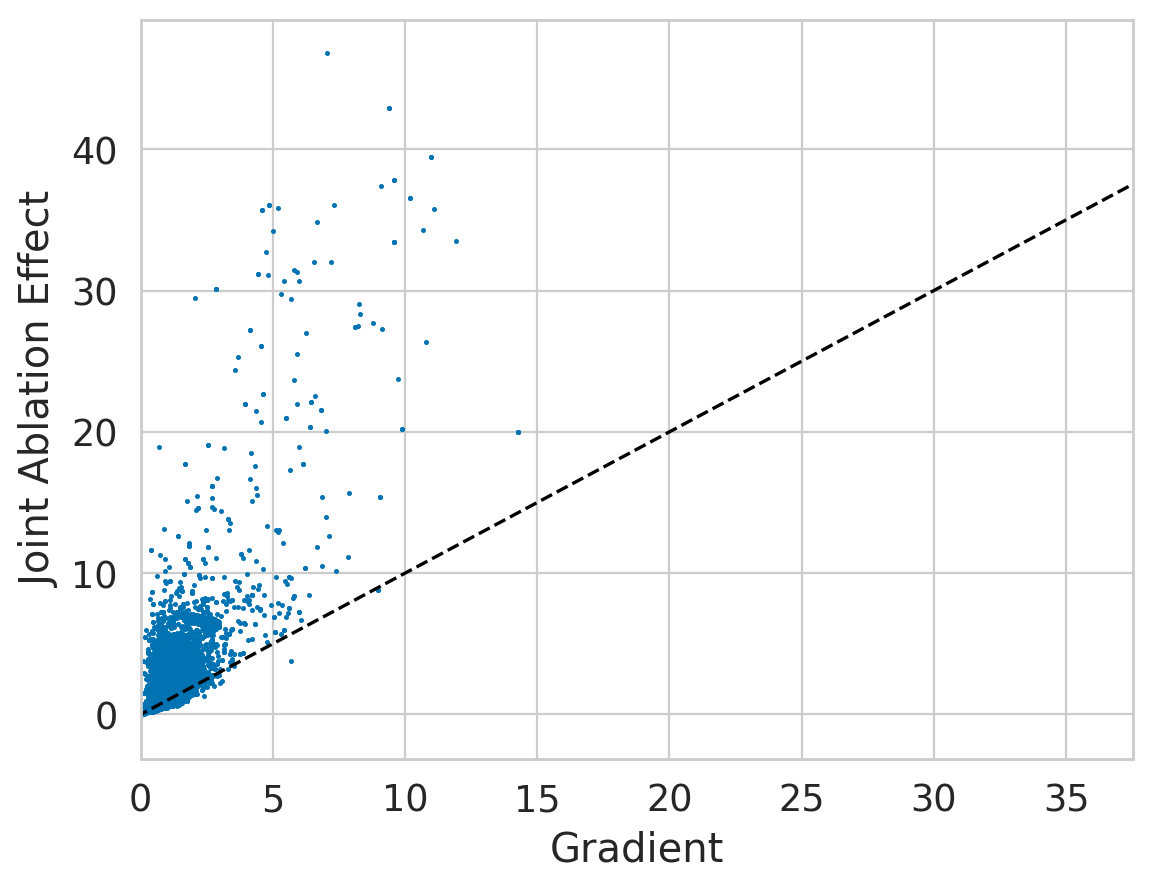}
         
         \label{fig:grad_vs_joint}
     \end{subfigure}
     \hfill
     \begin{subfigure}[b]{0.3\textwidth}
         \centering
         \caption{}
         \includegraphics[width=0.95\textwidth]{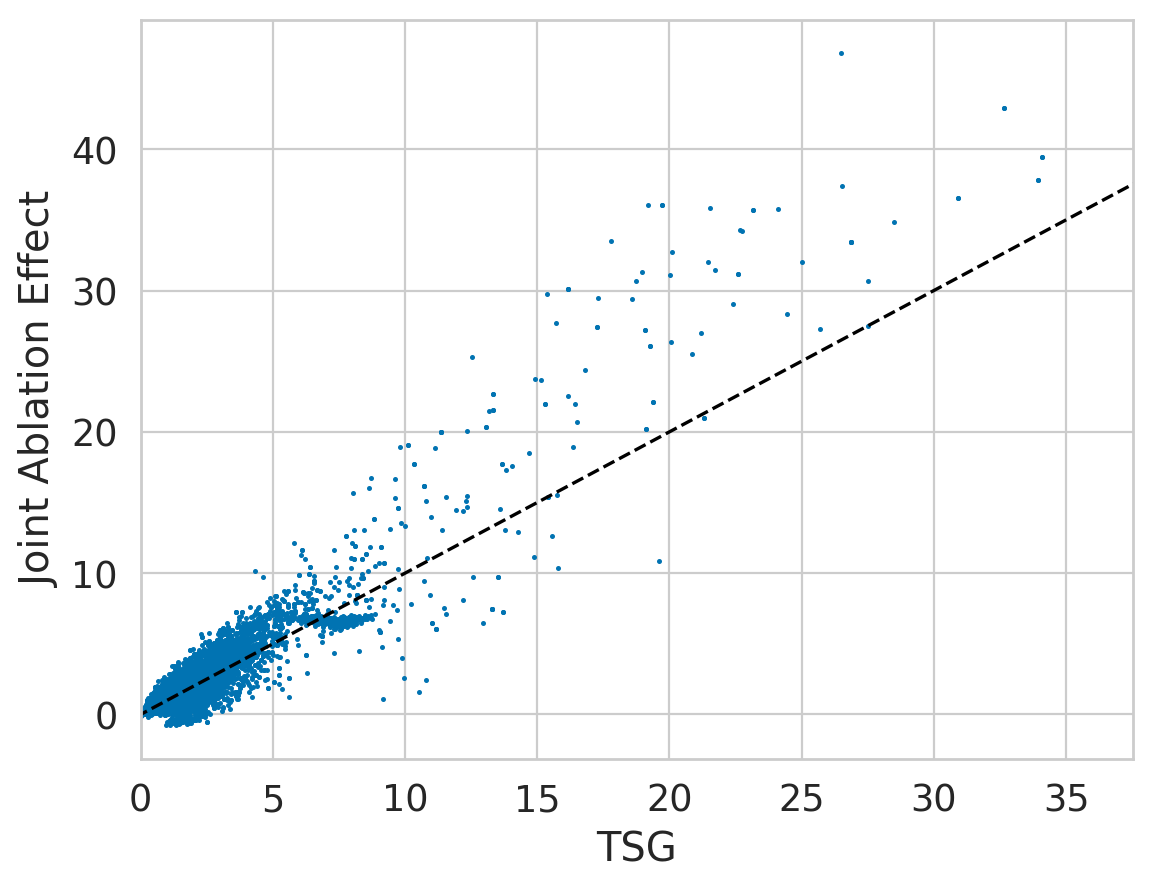}
         
         \label{fig:tsg_vs_joint}
     \end{subfigure}

        \caption{The attention self-repair effect and how temperature-adjusted softmax gradients approximate the joint ablation effect for LLAMA-3.2 1B on FEVER. The figures compare ablating the two largest attention scores jointly with a) ablating them separately, b) gradients, and c) TSG.}
        \label{fig:self-repair-proof}
\end{figure*}

\textbf{Self-repair is prevalent in attention mechanisms:} Figure~\ref{fig:ablation_effects} compares the effect of ablating the two largest attention scores simultaneously (y-axis) versus the sum of their individual ablation effects (x-axis) for the samples we identified as self-repair. The consistent gap above the diagonal provides clear empirical evidence of self-repair: individual attention scores show disproportionately small effects when ablated separately. These results empirically validate our theoretical analysis from Section~\ref{sec:self-repair-proof}. We provide results for other models and datasets in \Cref{app:self-repair}. We identified self-repair in all inputs, on average, 65--1200 times per input, depending on the model and dataset.

\textbf{TSG accurately approximates the joint effect:} Figure~\ref{fig:grad_vs_joint} and \ref{fig:tsg_vs_joint} compare standard gradient attribution and TSG with the joint effect. As intended, TSG more accurately approximates the effect of perturbing multiple attention scores simultaneously. We provide similar results for other models and datasets in \Cref{app:self-repair}.

\begin{table*}[t]
\caption{Comparison of the faithfulness of gradient-based feature attribution methods. The best scores for each model-dataset pair are shown in bold. AttnLRP has not been implemented for Gemma.}
\label{tab:fa_results}
\centering
\resizebox{\linewidth}{!}{%
\begin{tabular}{llccccccc|ccccccc}
\toprule
 &  & \multicolumn{7}{c}{\textbf{Comprehensiveness $\uparrow$}} & \multicolumn{7}{c}{\textbf{Sufficiency $\downarrow$}} \\
 &  & \multicolumn{2}{c}{Gemma}& \multicolumn{3}{c}{LLAMA} & \multicolumn{2}{c}{Qwen} & \multicolumn{2}{c}{Gemma}& \multicolumn{3}{c}{LLAMA} & \multicolumn{2}{c}{Qwen} \\
 \cmidrule(lr){3-4}\cmidrule(lr){5-7}\cmidrule(lr){8-9}\cmidrule(lr){10-11} \cmidrule(lr){12-14} \cmidrule(lr){15-16}
 &  & 2B & 9B & 1B & 3B & 8B & 1.5B & 3B & 2B & 9B & 1B & 3B & 8B & 1.5B & 3B \\
\midrule
\multirow[c]{6}{*}{BoolQ} & GxI & 0.09 (0.08) & 0.00 (0.02) & 0.18 (0.07) & 0.45 (0.07) & 0.39 (0.07) & 0.27 (0.13) & 0.57 (0.18) & 0.60 (0.09) & 0.43 (0.12) & 0.71 (0.07) & 0.39 (0.07) & 0.67 (0.06) & 0.54 (0.17) & 0.43 (0.11) \\
 & IG & 0.51 (0.08) & 0.20 (0.12) & 0.20 (0.09) & 0.52 (0.07) & 0.40 (0.06) & -0.00 (0.00) & 0.48 (0.22) & 0.11 (0.04) & 0.12 (0.11) & 0.43 (0.11) & 0.34 (0.03) & 0.58 (0.07) & 0.63 (0.03) & 0.54 (0.22) \\
 & DeepLIFT & 0.41 (0.15) & 0.15 (0.07) & 0.26 (0.09) & 0.36 (0.07) & 0.31 (0.08) & 0.20 (0.17) & 0.34 (0.12) & 0.37 (0.14) & 0.37 (0.13) & 0.58 (0.11) & 0.55 (0.09) & 0.74 (0.06) & 0.67 (0.14) & 0.77 (0.18) \\
 & AttnLRP & - & - & 0.66 (0.03) & 0.70 (0.02) & \bfseries 0.61 (0.04) & 0.67 (0.11) & \bfseries 0.65 (0.03) & — & — & 0.25 (0.06) & 0.17 (0.05) & 0.50 (0.05) & 0.15 (0.05) & 0.34 (0.11) \\
 & GIM & \bfseries 0.59 (0.02) & \bfseries 0.43 (0.04) & \bfseries 0.69 (0.03) & \bfseries 0.72 (0.03) & 0.60 (0.05) & \bfseries 0.68 (0.05) & 0.61 (0.03) & \bfseries 0.03 (0.04) & \bfseries 0.04 (0.03) & \bfseries 0.22 (0.05) & \bfseries 0.10 (0.04) & \bfseries 0.49 (0.04) & \bfseries 0.09 (0.04) & \bfseries 0.23 (0.10) \\
\midrule
\multirow[c]{6}{*}{FEVER} & GxI & 0.03 (0.19) & 0.06 (0.08) & 0.39 (0.07) & 0.53 (0.12) & 0.46 (0.06) & 0.43 (0.21) & 0.47 (0.13) & 0.47 (0.11) & 0.40 (0.10) & 0.56 (0.07) & 0.63 (0.09) & 0.71 (0.04) & 0.59 (0.06) & 0.73 (0.10) \\
 & IG & \bfseries 0.47 (0.10) & 0.32 (0.08) & 0.28 (0.08) & 0.51 (0.11) & 0.50 (0.10) & 0.03 (0.09) & 0.54 (0.14) & 0.02 (0.21) & 0.20 (0.08) & 0.51 (0.07) & 0.59 (0.09) & 0.62 (0.05) & 0.67 (0.03) & 0.61 (0.12) \\
 & DeepLIFT & 0.12 (0.19) & 0.20 (0.10) & 0.33 (0.07) & 0.47 (0.07) & 0.53 (0.06) & 0.31 (0.10) & 0.43 (0.14) & 0.43 (0.13) & 0.36 (0.12) & 0.58 (0.07) & 0.65 (0.07) & 0.76 (0.05) & 0.64 (0.10) & 0.76 (0.10) \\
 & AttnLRP & - & - & 0.60 (0.03) & \bfseries 0.75 (0.03) & 0.65 (0.04) & \bfseries 0.61 (0.04) & 0.68 (0.03) & - & - & 0.32 (0.06) & 0.44 (0.09) & 0.52 (0.03) & \bfseries 0.23 (0.08) & 0.49 (0.08) \\
 & GIM & 0.42 (0.11) & \bfseries 0.39 (0.05) & \bfseries 0.62 (0.03) & \bfseries 0.75 (0.02) & \bfseries 0.69 (0.03) & 0.51 (0.03) & \bfseries 0.68 (0.04) & \bfseries -0.01 (0.19) & \bfseries 0.09 (0.07) & \bfseries 0.26 (0.07) & \bfseries 0.39 (0.11) & \bfseries 0.51 (0.03) & 0.31 (0.09) & \bfseries 0.41 (0.08) \\
\midrule
\multirow[c]{6}{*}{HateXplain} & GxI & 0.07 (0.05) & 0.08 (0.05) & 0.53 (0.08) & 0.38 (0.14) & 0.44 (0.08) & 0.49 (0.29) & 0.53 (0.15) & 0.59 (0.04) & 0.42 (0.06) & 0.60 (0.06) & 0.67 (0.07) & 0.73 (0.03) & 0.63 (0.09) & 0.63 (0.10) \\
 & IG & 0.62 (0.03) & 0.37 (0.09) & 0.49 (0.08) & 0.64 (0.09) & 0.44 (0.05) & 0.00 (0.01) & 0.46 (0.11) & 0.35 (0.03) & 0.32 (0.07) & 0.58 (0.06) & 0.56 (0.06) & 0.61 (0.05) & 0.64 (0.03) & 0.72 (0.14) \\
 & DeepLIFT & 0.36 (0.05) & 0.29 (0.05) & 0.55 (0.07) & 0.36 (0.15) & 0.41 (0.09) & 0.29 (0.08) & 0.42 (0.11) & 0.52 (0.04) & 0.45 (0.05) & 0.59 (0.06) & 0.70 (0.08) & 0.73 (0.03) & 0.74 (0.05) & 0.73 (0.09) \\
 & AttnLRP & - & - & \bfseries 0.68 (0.04) & \bfseries 0.70 (0.04) & 0.60 (0.02) & \bfseries 0.85 (0.06) & 0.59 (0.09) & - & - & 0.30 (0.05) & 0.47 (0.04) & \bfseries 0.42 (0.04) & \bfseries 0.33 (0.05) & 0.71 (0.17) \\
 & GIM & \bfseries 0.64 (0.01) & \bfseries 0.48 (0.02) & \bfseries 0.68 (0.04) & 0.64 (0.08) & \bfseries 0.65 (0.02) & 0.80 (0.06) & \bfseries 0.63 (0.08) & \bfseries 0.26 (0.04) & \bfseries 0.14 (0.05) & \bfseries 0.26 (0.06) & \bfseries 0.46 (0.04) & 0.47 (0.06) & 0.35 (0.06) & \bfseries 0.51 (0.15) \\
\midrule
\multirow[c]{6}{*}{Movie} & GxI & 0.15 (0.13) & 0.04 (0.09) & 0.34 (0.10) & 0.51 (0.10) & 0.34 (0.11) & 0.42 (0.06) & 0.52 (0.10) & 0.61 (0.10) & 0.52 (0.10) & 0.61 (0.09) & 0.51 (0.07) & 0.71 (0.03) & 0.68 (0.12) & 0.76 (0.17) \\
 & IG & \bfseries 0.60 (0.06) & 0.38 (0.08) & 0.28 (0.12) & 0.69 (0.08) & 0.41 (0.08) & 0.00 (0.01) & 0.58 (0.21) & 0.17 (0.06) & 0.13 (0.08) & 0.49 (0.09) & 0.36 (0.11) & 0.52 (0.07) & 0.71 (0.03) & 0.56 (0.16) \\
 & DeepLIFT & 0.44 (0.11) & 0.22 (0.10) & 0.31 (0.10) & 0.62 (0.09) & 0.48 (0.07) & 0.49 (0.12) & 0.35 (0.14) & 0.35 (0.12) & 0.48 (0.07) & 0.60 (0.08) & 0.49 (0.07) & 0.71 (0.06) & 0.56 (0.09) & 0.77 (0.11) \\
 & AttnLRP & - & - & 0.69 (0.04) & 0.76 (0.02) & \bfseries 0.72 (0.02) & 0.78 (0.05) & \bfseries 0.77 (0.11) & - & - & 0.25 (0.06) & \bfseries 0.30 (0.05) & \bfseries 0.48 (0.03) & 0.19 (0.06) & 0.46 (0.16) \\
 & GIM & \bfseries 0.60 (0.03) & \bfseries 0.44 (0.03) & \bfseries 0.71 (0.04) & \bfseries 0.78 (0.02) & 0.71 (0.02) & \bfseries 0.83 (0.03) & 0.76 (0.09) & \bfseries 0.11 (0.06) & \bfseries 0.07 (0.04) & \bfseries 0.19 (0.06) & \bfseries 0.30 (0.05) & \bfseries 0.48 (0.03) & \bfseries 0.17 (0.05) & \bfseries 0.31 (0.19) \\
\midrule
\multirow[c]{6}{*}{SciFact} & GxI & 0.09 (0.19) & 0.05 (0.12) & 0.27 (0.07) & 0.42 (0.15) & 0.40 (0.08) & 0.34 (0.20) & 0.57 (0.20) & 0.55 (0.11) & 0.39 (0.13) & 0.57 (0.09) & 0.59 (0.09) & 0.72 (0.06) & 0.56 (0.11) & 0.63 (0.16) \\
 & IG & \bfseries 0.58 (0.14) & 0.32 (0.14) & 0.27 (0.09) & 0.46 (0.14) & 0.50 (0.08) & 0.01 (0.02) & 0.65 (0.19) & 0.18 (0.19) & 0.20 (0.18) & 0.46 (0.08) & 0.56 (0.11) & 0.57 (0.07) & 0.69 (0.03) & 0.54 (0.14) \\
 & DeepLIFT & 0.28 (0.25) & 0.27 (0.17) & 0.28 (0.07) & 0.43 (0.13) & 0.50 (0.09) & 0.25 (0.15) & 0.62 (0.22) & 0.47 (0.12) & 0.37 (0.16) & 0.58 (0.08) & 0.61 (0.08) & 0.75 (0.07) & 0.69 (0.16) & 0.65 (0.14) \\
 & AttnLRP & - & - & 0.64 (0.03) & 0.71 (0.03) & 0.65 (0.04) & \bfseries 0.67 (0.13) & \bfseries 0.74 (0.10) & - & - & 0.30 (0.07) & 0.35 (0.10) & \bfseries 0.49 (0.07) & \bfseries 0.22 (0.08) & 0.40 (0.12) \\
 & GIM & 0.57 (0.08) & \bfseries 0.40 (0.08) & \bfseries 0.67 (0.03) & \bfseries 0.74 (0.04) & \bfseries 0.67 (0.04) & 0.62 (0.08) & 0.74 (0.06) & \bfseries 0.04 (0.18) & \bfseries 0.08 (0.13) & \bfseries 0.22 (0.08) & \bfseries 0.31 (0.09) & 0.49 (0.08) & 0.26 (0.07) & \bfseries 0.38 (0.11) \\
\midrule
\multirow[c]{6}{*}{Twitter} & GxI & 0.40 (0.09) & 0.27 (0.06) & 0.41 (0.07) & 0.48 (0.10) & 0.45 (0.05) & 0.48 (0.07) & 0.73 (0.10) & 0.51 (0.07) & 0.37 (0.06) & 0.64 (0.05) & 0.75 (0.05) & 0.77 (0.03) & 0.73 (0.06) & 0.64 (0.04) \\
 & IG & \bfseries 0.53 (0.06) & 0.34 (0.06) & 0.41 (0.08) & 0.70 (0.07) & 0.55 (0.05) & 0.40 (0.05) & 0.67 (0.13) & 0.36 (0.06) & 0.33 (0.05) & 0.62 (0.05) & \bfseries 0.59 (0.05) & 0.66 (0.04) & 0.71 (0.09) & 0.66 (0.07) \\
 & DeepLIFT & 0.44 (0.06) & 0.23 (0.06) & 0.42 (0.06) & 0.65 (0.06) & 0.54 (0.05) & 0.51 (0.07) & \bfseries 0.75 (0.12) & 0.54 (0.05) & 0.44 (0.05) & 0.62 (0.05) & 0.67 (0.03) & 0.75 (0.03) & 0.75 (0.04) & 0.67 (0.05) \\
 & AttnLRP & - & - & 0.64 (0.04) & \bfseries 0.73 (0.03) & 0.68 (0.02) & 0.75 (0.07) & 0.66 (0.05) & - & - & 0.42 (0.04) & 0.61 (0.04) & 0.57 (0.03) & \bfseries 0.48 (0.04) & 0.66 (0.08) \\
 & GIM &  0.52 (0.06) & \bfseries 0.38 (0.05) & \bfseries 0.68 (0.03) & 0.72 (0.03) & \bfseries 0.69 (0.02) & \bfseries 0.76 (0.06) & 0.72 (0.06) & \bfseries 0.34 (0.04) & \bfseries 0.22 (0.03) & \bfseries 0.41 (0.04) & 0.61 (0.05) & \bfseries 0.56 (0.03) & 0.55 (0.04) & \bfseries 0.58 (0.04) \\
\bottomrule
\end{tabular}}
\end{table*}

\textbf{Ablation study:}\label{sec:ablation}
We evaluate the contribution of each modification in GIM through an ablation study on the feature attribution task. Although this paper emphasizes circuit localization, we use feature attribution because it is less computationally expensive while relying on the same gradient computations. Since GIM's modifications target specific operations during backpropagation, improvements in feature attribution should transfer to circuit localization.

\Cref{tab:ablate_mod} presents performance when each modification (TSG, Layernorm Freeze, and Grad Norm) is individually removed from the full GIM method. Removing any single modification degrades performance across most model-dataset pairs. When they do not, removing a modification rarely improves the performance. These findings confirm that each component contributes meaningfully.


\begin{table*}[t]
    \centering
    \caption{Performance when ablating each modification in GIM separately. Scores are shown as the mean (standard deviation). Average scores worse than GIM are shown in bold, while scores that are significantly worse are underlined (independent t-tests with 0.05$\leq \alpha$)}
    \label{tab:ablate_mod}
\resizebox{\linewidth}{!}{%
\begin{tabular}{llrrrrrrrrrr}
\toprule
 &  & \multicolumn{5}{c}{\textbf{Comprehensiveness $\uparrow$}} & \multicolumn{5}{c}{\textbf{Sufficiency $\downarrow$}} \\

 &  & \multicolumn{1}{c}{Gemma}& \multicolumn{2}{c}{LLAMA} & \multicolumn{2}{c}{Qwen}  & \multicolumn{1}{c}{Gemma} & \multicolumn{2}{c}{LLAMA} & \multicolumn{2}{c}{Qwen}\\
 \cmidrule(lr){3-3}\cmidrule(lr){4-5}\cmidrule(lr){6-7}\cmidrule(lr){8-8}\cmidrule(lr){9-10}\cmidrule(lr){11-12}
 &  & 2B & 1B & 3B & 1.5B & 3B & 2B & 1B &  3B & 1.5B & 3B\\
\midrule
\multirow[c]{4}{*}{BoolQ} & GIM & 0.59 (0.02) & 0.69 (0.03) & 0.72 (0.03) & 0.68 (0.05) & 0.61 (0.03) & 0.03 (0.04) & 0.22 (0.05) & 0.10 (0.04) & 0.09 (0.04) & 0.23 (0.10) \\
 & - LN freeze & \bfseries \underline{0.37 (0.14)} & \bfseries \underline{0.36 (0.09)} & \bfseries \underline{0.49 (0.14)} & \bfseries \underline{0.45 (0.07)} & 0.67 (0.07) & \bfseries \underline{0.46 (0.11)} & \bfseries \underline{0.57 (0.07)} & \bfseries \underline{0.58 (0.04)} & \bfseries \underline{0.68 (0.06)} & \bfseries \underline{0.39 (0.10)} \\
 & - Grad norm & \bfseries \underline{0.09 (0.09)} & \bfseries \underline{0.45 (0.07)} & \bfseries \underline{0.47 (0.07)} & \bfseries \underline{0.37 (0.29)} & \bfseries \underline{0.18 (0.10)} & \bfseries \underline{0.53 (0.11)} & \bfseries 0.24 (0.05) & \bfseries \underline{0.30 (0.12)} & \bfseries \underline{0.62 (0.14)} & \bfseries \underline{0.72 (0.19)} \\
 & - TSG & \bfseries \underline{0.53 (0.05)} & \bfseries 0.68 (0.03) & 0.72 (0.02) & 0.69 (0.07) & 0.66 (0.04) & \bfseries \underline{0.07 (0.05)} & \bfseries \underline{0.28 (0.04)} & \bfseries 0.12 (0.04) & \bfseries \underline{0.12 (0.05)} & \bfseries 0.29 (0.10) \\
\midrule
\multirow[c]{4}{*}{FEVER} & GIM & 0.42 (0.11) & 0.62 (0.03) & 0.75 (0.02) & 0.51 (0.03) & 0.68 (0.04) & -0.01 (0.19) & 0.26 (0.07) & 0.39 (0.11) & 0.31 (0.09) & 0.41 (0.08) \\
 & - LN freeze & \bfseries \underline{0.30 (0.11)} & \bfseries \underline{0.34 (0.08)} & \bfseries \underline{0.61 (0.05)} & 0.54 (0.06) & 0.70 (0.05) & \bfseries \underline{0.32 (0.12)} & \bfseries \underline{0.55 (0.05)} & \bfseries \underline{0.72 (0.05)} & \bfseries \underline{0.71 (0.07)} & \bfseries \underline{0.49 (0.08)} \\
 & - Grad norm & \bfseries \underline{0.10 (0.14)} & \bfseries \underline{0.49 (0.08)} & \bfseries \underline{0.38 (0.10)} & \bfseries \underline{0.33 (0.17)} & \bfseries \underline{0.43 (0.15)} & \bfseries \underline{0.38 (0.13)} & \bfseries \underline{0.38 (0.09)} & \bfseries \underline{0.67 (0.09)} & \bfseries \underline{0.65 (0.07)} & \bfseries \underline{0.72 (0.12)} \\
 & - TSG & \bfseries \underline{0.39 (0.13)} & \bfseries \underline{0.60 (0.03)} & 0.76 (0.03) & 0.62 (0.04) & \bfseries \underline{0.66 (0.04)} & -0.03 (0.21) & \bfseries \underline{0.29 (0.07)} & 0.39 (0.10) & 0.25 (0.08) & \bfseries \underline{0.46 (0.08)} \\
\midrule
\multirow[c]{4}{*}{HateXplain} & GIM & 0.64 (0.01) & 0.68 (0.04) & 0.64 (0.08) & 0.80 (0.06) & 0.63 (0.08) & 0.26 (0.04) & 0.26 (0.06) & 0.46 (0.04) & 0.35 (0.06) & 0.51 (0.15) \\
 & - LN freeze & \bfseries \underline{0.46 (0.02)} & \bfseries \underline{0.61 (0.04)} & 0.64 (0.03) & \bfseries \underline{0.54 (0.07)} & 0.81 (0.05) & \bfseries \underline{0.51 (0.04)} & \bfseries \underline{0.45 (0.06)} & \bfseries \underline{0.58 (0.06)} & \bfseries \underline{0.73 (0.06)} & 0.45 (0.05) \\
 & - Grad norm & \bfseries \underline{0.22 (0.05)} & \bfseries \underline{0.57 (0.13)} & \bfseries \underline{0.28 (0.12)} & \bfseries \underline{0.17 (0.10)} & \bfseries \underline{0.15 (0.08)} & \bfseries \underline{0.56 (0.06)} & \bfseries \underline{0.47 (0.09)} & \bfseries \underline{0.70 (0.05)} & \bfseries \underline{0.76 (0.06)} & \bfseries \underline{0.80 (0.07)} \\
 & - TSG & \bfseries \underline{0.63 (0.02)} & \bfseries \underline{0.67 (0.04)} & 0.64 (0.06) & 0.85 (0.09) & \bfseries \underline{0.50 (0.05)} & \bfseries \underline{0.31 (0.05)} & \bfseries \underline{0.37 (0.06)} & \bfseries \underline{0.52 (0.05)} & 0.28 (0.05) & \bfseries \underline{0.91 (0.11)} \\
\midrule
\multirow[c]{4}{*}{Movie} & GIM & 0.60 (0.03) & 0.71 (0.04) & 0.78 (0.02) & 0.83 (0.03) & 0.76 (0.09) & 0.11 (0.06) & 0.19 (0.06) & 0.30 (0.05) & 0.17 (0.05) & 0.31 (0.19) \\
 & - LN freeze & \bfseries 0.59 (0.07) & \bfseries \underline{0.63 (0.05)} & \bfseries \underline{0.49 (0.05)} & \bfseries \underline{0.61 (0.07)} & 0.88 (0.06) & \bfseries \underline{0.49 (0.05)} & \bfseries \underline{0.45 (0.05)} & \bfseries \underline{0.67 (0.06)} & \bfseries \underline{0.75 (0.11)} & \bfseries \underline{0.43 (0.10)} \\
 & - Grad norm & \bfseries \underline{0.27 (0.14)} & \bfseries \underline{0.62 (0.04)} & \bfseries \underline{0.62 (0.08)} & \bfseries \underline{0.37 (0.12)} & \bfseries \underline{0.35 (0.28)} & \bfseries \underline{0.56 (0.09)} & \bfseries \underline{0.25 (0.05)} & \bfseries \underline{0.35 (0.07)} & \bfseries \underline{1.10 (0.19)} & \bfseries \underline{0.76 (0.09)} \\
 & - TSG & \bfseries \underline{0.55 (0.04)} & \bfseries \underline{0.68 (0.03)} & 0.78 (0.02) & \bfseries \underline{0.82 (0.04)} & \bfseries 0.73 (0.12) & \bfseries \underline{0.16 (0.06)} & \bfseries \underline{0.26 (0.05)} & \bfseries 0.31 (0.05) & 0.13 (0.06) & \bfseries \underline{0.45 (0.19)} \\
\midrule
\multirow[c]{4}{*}{SciFact} & GIM & 0.57 (0.08) & 0.67 (0.03) & 0.74 (0.04) & 0.62 (0.08) & 0.74 (0.06) & 0.04 (0.18) & 0.22 (0.08) & 0.31 (0.09) & 0.26 (0.07) & 0.38 (0.11) \\
 & - LN freeze & \bfseries \underline{0.43 (0.09)} & \bfseries \underline{0.33 (0.09)} & \bfseries \underline{0.51 (0.08)} & \bfseries \underline{0.54 (0.07)} & 0.74 (0.10) & \bfseries \underline{0.43 (0.14)} & \bfseries \underline{0.53 (0.07)} & \bfseries \underline{0.68 (0.07)} & \bfseries \underline{0.75 (0.10)} & \bfseries \underline{0.42 (0.12)} \\
 & - Grad norm & \bfseries \underline{0.23 (0.13)} & \bfseries \underline{0.53 (0.07)} & \bfseries \underline{0.34 (0.11)} & \bfseries \underline{0.28 (0.11)} & \bfseries \underline{0.36 (0.27)} & \bfseries \underline{0.48 (0.12)} & \bfseries \underline{0.33 (0.07)} & \bfseries \underline{0.59 (0.08)} & \bfseries \underline{0.74 (0.08)} & \bfseries \underline{0.76 (0.15)} \\
 & - TSG & 0.60 (0.09) & \bfseries \underline{0.64 (0.03)} & \bfseries \underline{0.72 (0.04)} & 0.66 (0.09) & \bfseries \underline{0.72 (0.07)} & \bfseries 0.05 (0.18) & \bfseries \underline{0.25 (0.07)} & 0.30 (0.09) & 0.20 (0.08) & \bfseries \underline{0.40 (0.12)} \\
\midrule
\multirow[c]{4}{*}{Twitter} & GIM & 0.52 (0.06) & 0.68 (0.03) & 0.72 (0.03) & 0.76 (0.06) & 0.72 (0.06) & 0.34 (0.04) & 0.41 (0.04) & 0.61 (0.05) & 0.55 (0.04) & 0.58 (0.04) \\
 & - LN freeze & \bfseries \underline{0.47 (0.05)} & \bfseries \underline{0.48 (0.05)} & \bfseries \underline{0.55 (0.04)} & \bfseries \underline{0.63 (0.06)} & 0.80 (0.05) & \bfseries \underline{0.48 (0.04)} & \bfseries \underline{0.51 (0.04)} & \bfseries \underline{0.68 (0.02)} & \bfseries \underline{0.75 (0.04)} & \bfseries \underline{0.61 (0.04)} \\
 & - Grad norm & \bfseries \underline{0.37 (0.07)} & \bfseries \underline{0.47 (0.06)} & \bfseries \underline{0.52 (0.09)} & \bfseries \underline{0.42 (0.06)} & \bfseries \underline{0.51 (0.14)} & \bfseries \underline{0.56 (0.05)} & \bfseries \underline{0.63 (0.05)} & \bfseries \underline{0.75 (0.04)} & \bfseries \underline{0.87 (0.05)} & \bfseries \underline{0.73 (0.05)} \\
 & - TSG & 0.52 (0.06) & \bfseries \underline{0.64 (0.03)} & 0.73 (0.03) & 0.78 (0.06) & \bfseries \underline{0.65 (0.05)} & 0.32 (0.04) & \bfseries \underline{0.43 (0.04)} & 0.60 (0.03) & 0.47 (0.04) & \bfseries \underline{0.66 (0.08)} \\
\bottomrule
\end{tabular}}
\end{table*}

\section{Discussion}
\textbf{Should gradients estimate the joint effect?} We designed TSG to approximate the effect of perturbing multiple attention scores simultaneously to address the self-repair problem. However, is this always desirable?
Whether to approximate the effect of perturbing one or multiple attention scores depends on the causal question being asked. If a perturbation would cause changes in multiple interacting attention scores, then we should estimate the joint effect; otherwise, we should estimate the individual effect. However, when perturbing early-layer variables, determining which attention scores in deep layers are affected becomes difficult because the number of potential causal paths grows exponentially with depth. We naively assumed that perturbations would always impact all interacting attention scores. This assumption may fail for certain inputs, models, and types of perturbations. TSG is a naive solution to the self-repair problem; future work should investigate methods to estimate whether a perturbation will cause changes in multiple interacting attention scores, enabling a more principled selection between joint and individual effect estimation.

\textbf{Gradient Norm is an oversimplification.} The Gradient Norm correction assumes that both multiplicative inputs (e.g., $Q$ and $K$) fully depend on the variable being perturbed. This is accurate when estimating the effect of perturbing $X$ directly. However, $X$ is a sum of outputs from multiple components (attention heads, MLPs, etc.), and we want to estimate the effect of perturbing each component's output individually. A single component's output may project differently onto $W_Q$ and $W_K$, contributing more to one than the other. In such cases, dividing by 2 is too coarse a correction. Despite these simplifications, our ablation study in Section~\ref{sec:ablation} shows that Gradient Norm consistently improves attribution faithfulness, suggesting that correcting for multiplicative interactions, even approximately, is beneficial.

\textbf{Limitations} Our experiments demonstrate that GIM is a highly effective circuit localization method, achieving strong faithfulness scores across a range of models. GIM combines several modifications to standard gradient-based attribution, including temperature scaling, activation patching, and layer normalization freezing. While we provide motivation for each modification, and our ablation study in \Cref{sec:ablation} confirms that each component contributes to GIM's performance, developing a deeper theoretical understanding of why each modification improves faithfulness remains an interesting direction for future work.

\section{Conclusion}
We introduced Gradient Interaction Modifications (GIM), a technique that accounts for feature interactions during backpropagation through three complementary modifications: temperature-adjusted softmax gradients, layernorm freeze, and gradient normalization. In developing GIM, we identified attention self-repair, a phenomenon where softmax redistribution causes gradients to vanish for influential attention scores, leading interpretability methods to underestimate component importance. Our evaluations across multiple models and datasets show that GIM achieves state-of-the-art results on circuit localization and outperforms existing gradient-based methods on feature attribution. By revealing why existing methods fail and accounting for feature interactions, GIM enables more faithful mechanistic analysis of LLMs.




{\small
\bibliography{joakim_references}
}

\appendix

\section{Appendix / supplemental material}

\subsection{Evaluation Metrics}\label{app:metrics}
This section describes the evaluation metrics used in the main paper.
\subsubsection{Circuit localization metrics}\label{app:cpr_cmd}
We evaluate circuit localization methods using two complementary metrics from the Mechanistic Interpretability Benchmark~\citep{muellerMIBMechanisticInterpretability2025}: the \textbf{integrated circuit performance ratio (CPR)} and the \textbf{integrated circuit-model distance (CMD)}.

Given a circuit $\mathcal{C}$ and the full neural network $\mathcal{N}$, we first define faithfulness $f$ as:
\begin{equation}
    f(\mathcal{C}, \mathcal{N}; m) = \frac{m(\mathcal{C}) - m(\varnothing)}{m(\mathcal{N}) - m(\varnothing)}
\end{equation}
where $m$ is an evaluation metric that measures model performance on the task. Following~\citet{muellerMIBMechanisticInterpretability2025}, we use the logit difference between the correct and incorrect answers as $m$. The term $\varnothing$ denotes the empty circuit (all components ablated). This formulation assigns meaning to $f=0$ ($\mathcal{C}$ recovers none of the performance of $\mathcal{N}$) and $f=1$ ($\mathcal{C}$ achieves identical behavior to $\mathcal{N}$).

Rather than evaluating circuits at a single size threshold, both metrics marginalize over circuit sizes to capture Pareto optimality with respect to faithfulness and minimality.

\textbf{CPR} ($\uparrow$ the higher the better) is the area under the faithfulness curve with respect to circuit size:
\begin{equation}
    \text{CPR} = \int_0^1 f(\mathcal{C}_k) \, dk
\end{equation}
where $k$ is the proportion of edges from $\mathcal{N}$ in circuit $\mathcal{C}_k$. CPR prioritizes methods that locate components with a \emph{positive} effect on model performance. In practice, we approximate these integrals using a Riemann sum by measuring faithfulness at representative circuit sizes $k \in \{0.001, 0.002, 0.005, 0.01, 0.02, 0.05, 0.1, 0.2, 0.5, 1\}$ and computing the area using the trapezoidal rule.

\textbf{CMD} ($\downarrow$ the lower the better, with 0 being optimal) is the area between the faithfulness curve and 1:
\begin{equation}
    \text{CMD} = \int_0^1 |1 - f(\mathcal{C}_k)| \, dk
\end{equation}
CMD prioritizes methods that locate components with \emph{any strong effect} on model performance, including negative effects. When constructing circuits for CMD evaluation, the absolute values of importance scores are used, meaning that the ability to correctly identify the \emph{sign} of component importance is ignored. We note a limitation of this metric: methods that successfully identify components with negative importance will include them in the circuit (due to their high absolute scores), resulting in low $f$ and, consequently, high CMD. In other words, methods are penalized for correctly detecting negative components. We disagree with this aspect of the metric, but include CMD results in the appendix for transparency.

\subsubsection{Feature attribution metrics}\label{app:comp_suff}
Comprehensiveness and Sufficiency are popular evaluation metrics for feature attribution methods~\citep{deyoungERASERBenchmarkEvaluate2020}.
\textbf{Comprehensiveness} ($\uparrow$ the higher the better) measures the average output change after cumulatively ablating top-attributed features:
\begin{equation}
    \text{Comp}(f, \bm{x}, \bm{e}) = \frac{1}{Nf(\bm{x})} \sum_{i=1}^{N} f(\bm{x})-f(p(\bm{x}, \text{rank}(\bm{e})_{1:i}))
\end{equation}
where $f$ is the model, $\bm{x}$ is either the input embeddings or hidden representation, $\bm{e}$ contains attribution scores, $N$ is the number of input features, $p(\bm{x},\bm{i} )$ is the perturbation function, and $\text{rank}(\cdot)$ orders features by attribution. The perturbation function replaces the input embeddings $\bm{x}$ with a counterfactual embedding at the positions specified by $\bm{i}$. In our experiments, the counterfactual embedding was the embedding representing the whitespace character.

\textbf{Sufficiency} ($\downarrow$ the lower the better) is the opposite of Comprehensiveness. It measures the average output change after cumulatively ablating the lowest-attributed features:
\begin{equation}
    \text{Suff}(f, \bm{x}, \bm{e}) = \frac{1}{Nf(\bm{x})} \sum_{i=1}^{N} f(\bm{x})-f(p(\bm{x}, \text{rank}(-\bm{e})_{1:i}))
\end{equation}

\subsection{Datasets} \label{app:datasets}

\subsubsection{Circuit localization datasets}\label{app:cl_datasets}

We evaluate circuit localization methods on the Mechanistic Interpretability Benchmark (MIB)~\citep{muellerMIBMechanisticInterpretability2025}, which comprises four tasks spanning various reasoning types, difficulty levels, and answer formats. Each task includes fixed counterfactual inputs used for interventions.

\textbf{Indirect Object Identification (IOI)} is one of the most studied tasks in mechanistic interpretability~\citep{wangInterpretabilityWildCircuit2022}. IOI sentences follow the pattern ``When Mary and John went to the store, John gave an apple to \_\_'', containing a subject (``John'') and an indirect object (``Mary''). The model must complete the sentence with the indirect object. All names are constrained to tokenize to single tokens across all evaluated models.

\textbf{Arithmetic} follows~\citet{stolfoMechanisticInterpretationArithmetic2023} in defining the task as performing addition and subtraction with two operands of up to two digits each. Given a pair of numbers and an operator, the model must predict the result of the operation (e.g., ``What is the sum of 13 and 25?''). Six natural language templates are used for each operand pair to ensure robust behavior is isolated.

\textbf{Multiple-Choice Question Answering (MCQA)} uses a synthetic dataset designed to isolate a model's MCQA ability from task-specific knowledge~\citep{wiegreffeAnswerAssembleAce2025}. All information needed to answer the questions is contained in the prompt. Questions have four choices and concern the color of an object (e.g., ``A box is brown. What color is a box?'').

\textbf{AI2 Reasoning Challenge (ARC)}~\citep{clarkThinkYouHave2018} comprises grade-school-level multiple-choice science questions. This dataset represents a more realistic task used to benchmark state-of-the-art language models. The dataset is partitioned into Easy and Challenge subsets, which are analyzed separately due to large accuracy differences between them.

For all circuit localization tasks, counterfactual inputs are provided for each instance, where mappings from original inputs to counterfactuals are fixed to ensure consistency in evaluation. We present dataset statistics in \Cref{tab:circuit_dataset}.

\begin{table*}[t]
\centering
\caption{Circuit localization dataset statistics from the Mechanistic Interpretability Benchmar.}
\label{tab:circuit_dataset}
\begin{tabular}{lcccc}
\toprule
Dataset & Train & Validation & Test (Public) & Test (Private) \\
\midrule
IOI & 10000 & 10000 & 1000 & 1000 \\
MCQA & 110 & 50 & 50 & 50 \\
Arithmetic (+) & 34400 & 4920 & 1000 & 1000 \\
Arithmetic (-) & 17400 & 2484 & 1000 & 1000 \\
ARC (Easy) & 2251 & 570 & 1188 & 1188 \\
ARC (Challenge) & 1119 & 299 & 586 & 586 \\
\bottomrule
\end{tabular}
\end{table*}

\subsubsection{Feature attribution datasets}

\textbf{BoolQ} comprises yes/no questions about Wikipedia articles, which is used in the SuperGLUE and ERASER benchmarks~\citep{clarkBoolQExploringSurprising2019, wangSuperGLUEStickierBenchmark, deyoungERASERBenchmarkEvaluate2020}.

\textbf{FEVER} and \textbf{SciFact} are fact-verification datasets~\citep{thorneFEVERLargescaleDataset2018, waddenFactFictionVerifying2020}. Fact verification is classifying whether a claim is true, given a document containing the necessary information. The documents and claims in FEVER are generated from Wikipedia, while SciFact comprises scientific abstracts and claims about COVID-19. 

\textbf{Movie Reviews} and \textbf{Twitter Sentiment Extraction} are sentiment classification datasets, where the goal is to classify the sentiment of movie reviews and Twitter posts~\citep{zaidanUsingAnnotatorRationales2007, maggieTweetSentimentExtraction2020}. 

\textbf{HateXplain} comprises Twitter posts, where the task is to classify whether a post contains hate speech~\citep{mathewHateXplainBenchmarkDataset2021}.

We downsample our datasets because computing and evaluating feature attribution and circuit localization methods is computationally expensive. We create smaller subsets by randomly sampling 300 examples where all the models predicted the correct answer. We present an overview of the datasets in \Cref{tab:dataset}.

\begin{table*}[t]
\centering
\caption{Feature attribution dataset statistics.}
\label{tab:dataset}
\begin{tabular}{lccc}
\toprule
Name & Number of examples & AVG number of tokens & Task \\
\midrule
HateXplain & 300 & 314 & Hate speech detection \\
Movie Review & 199 & 930 & Sentiment classification \\
Twitter Sentiment & 300 & 78 & Sentiment classification \\
BoolQ & 199 & 1184 & Question-answering \\
FEVER & 300 & 336 & Fact-verification \\
SciFact & 209 & 448 & Fact-verification \\
\bottomrule
\end{tabular}
\end{table*}

\subsection{GIM Integration with Edge Attribution Patching}
\label{app:gim_eap_integration}

This section provides additional details on how Gradient Interaction Modifications (GIM) is integrated with Edge Attribution Patching (EAP) for circuit localization.

Recall from \Cref{sec:rw_cl} that EAP estimates the importance of an edge $(i, j)$ from upstream node $i$ to downstream node $j$ as:
\begin{equation}
    I_{ij} = (a_i - \tilde{a}_i) \cdot \frac{\partial z}{\partial h_j}
\end{equation}
where $a_i$ is the activation of node $i$ on the original example, $\tilde{a}_i$ is the activation on a counterfactual example, $h_j$ is the input to node $j$, and $z$ is the model output.

To apply GIM for circuit localization, we modify only the backward pass used to compute $\frac{\partial z}{\partial h_j}$. The counterfactual forward pass and the importance calculation formula remain unchanged from standard EAP. Specifically, during backpropagation we apply the following three modifications:

\begin{enumerate}
    \item \textbf{Temperature-adjusted Softmax Gradients (TSG):} Before computing the gradients through the attention mechanism, we recompute the softmax with an elevated temperature ($T=2$ in our experiments). This produces a more uniform distribution of attention weights, which mitigates the gradient cancellation caused by attention self-repair (\Cref{sec:self-repair-intro}).
    
    \item \textbf{Layernorm Freeze:} During the forward pass, we cache the normalization factor $\sigma$ for each layer normalization operation. During the backward pass, we divide the upstream gradient by $\sigma$ rather than backpropagating through the normalization computation. This prevents the self-repair effect introduced by layer normalization rescaling.
    
    \item \textbf{Gradient Norm:} At multiplicative interactions, we divide the gradient by the number of inputs involved in the multiplication. In transformer models, we apply this normalization (dividing by 2) at three locations: (1) the attention-value multiplication, (2) the query-key multiplication, and (3) the MLP gate-projection multiplication.
\end{enumerate}

Algorithm~\ref{alg:gim_eap} summarizes the complete procedure for computing edge importance scores using GIM-modified EAP.

\begin{algorithm}[h]
\caption{Edge Attribution Patching with GIM}
\label{alg:gim_eap}
\begin{algorithmic}[1]
\REQUIRE Original input $x$, counterfactual input $\tilde{x}$, model $f$, temperature $T$
\ENSURE Edge importance scores $\{I_{ij}\}$
\STATE Run forward pass on $\tilde{x}$, cache activations $\{\tilde{a}_i\}$
\STATE Run forward pass on $x$, cache activations $\{a_i\}$ and normalization factors $\{\sigma_l\}$
\STATE Compute output $z = f(x)$
\STATE \textbf{GIM-modified backward pass:}
\FOR{each layer $l$ from output to input}
    \IF{layer $l$ contains attention}
        \STATE Recompute softmax with temperature $T$: $s' = \text{Softmax}(a/T)$
        \STATE Compute gradients using $s'$
        \STATE Apply Gradient Norm (divide by 2) at attention-value and query-key products
    \ENDIF
    \IF{layer $l$ contains layer normalization}
        \STATE Apply Layernorm Freeze: divide gradient by cached $\sigma_l$
    \ENDIF
    \IF{layer $l$ contains MLP with gating}
        \STATE Apply Gradient Norm (divide by 2) at gate-projection product
    \ENDIF
\ENDFOR
\STATE Obtain gradients $\{\frac{\partial z}{\partial h_j}\}$ for all downstream nodes
\STATE Compute importance: $I_{ij} = (a_i - \tilde{a}_i) \cdot \frac{\partial z}{\partial h_j}$ for all edges $(i,j)$
\STATE \textbf{return} $\{I_{ij}\}$
\end{algorithmic}
\end{algorithm}

Note that the computational cost of GIM-modified EAP is comparable to standard EAP, requiring one counterfactual forward pass, one forward pass on the original input, and one backward pass. The only additional overhead comes from caching normalization factors and recomputing the softmax with adjusted temperature during backpropagation.

\subsection{GIM Integration for Feature Attribution}
\label{app:gim_feature_attribution}

This section describes how we adapt Gradient Interaction Modifications (GIM) for feature attribution at the token level, as used in our feature attribution experiments.

Feature attribution aims to assign importance scores to input tokens based on their contribution to the model's output. We adapt the EAP framework to operate on token embeddings rather than edges between model components. Specifically, we treat each token embedding as an upstream node and compute its importance for the model's prediction.

Given an input sequence with token embeddings $\{e_1, e_2, \ldots, e_n\}$, we compute the importance of token $i$ as:
\begin{equation}
    I_i = (e_i - \tilde{e}_i) \cdot \frac{\partial z}{\partial e_i}
\end{equation}
where $e_i$ is the embedding of token $i$, $\tilde{e}_i$ is a counterfactual embedding, $z$ is the model output (e.g., the logit for the predicted class), and $\cdot$ denotes the dot product. In our experiments, we use the embedding of the whitespace token as the counterfactual $\tilde{e}_i$ for all positions.

The GIM modifications (TSG, Layernorm Freeze, and Gradient Norm) are applied during the backward pass when computing $\frac{\partial z}{\partial e_i}$, following the same procedure as for circuit localization (\Cref{app:gim_eap_integration}).

The key difference from circuit localization is the granularity of analysis: circuit localization computes importance for edges between internal model components (e.g., attention heads, MLPs), while feature attribution computes importance for input tokens. However, the gradient modifications remain identical since both approaches require backpropagating through the same transformer operations.

This approach can be viewed as a GIM-modified variant of GradientXInput \citep{sundararajanAxiomaticAttributionDeep2017a}, where we replace the implicit zero baseline with an explicit counterfactual (the whitespace token embedding) and modify the backward pass to account for feature interactions.

\subsection{Experimental Procedure for Identifying Attention Self-Repair}
\label{app:self_repair_procedure}
This section provides a detailed description of the experimental
procedure used to identify instances of attention self-repair in LLMs
and to evaluate the effectiveness of Temperature-adjusted Softmax
Gradients (TSG); essentially, the procedure used to produce
\Cref{fig:grad_vs_joint}.
 
Given a model and input, we identify attention weight vectors
exhibiting self-repair using $\epsilon = 0.01$ as the threshold
separating significant from negligible attention weights in
\Cref{prop:self-repair}.
 
\paragraph{Step 1: Select important attention weight vectors.} Not all
attention weights are relevant for the model's prediction. We rank
attention weights by their importance $g_j \cdot \bm{s}_j$ (where
$g_j = \partial z / \partial \bm{s}_j$ is defined in \Cref{eq:grad})
and retain only attention weight vectors containing at least one weight
in the top $1\%$.
 
\paragraph{Step 2: Filter for $|\mathcal{I}_\epsilon| \geq 2$.}
Self-repair requires redundancy: when one attention score is ablated,
another position must compensate. From the selected vectors, we keep
only those for which
$\mathcal{I}_\epsilon = \{k : \bm{s}_k > \epsilon\}$ contains at least
two positions.
 
\paragraph{Step 3: Test \Cref{ass:repair}.} For each remaining vector,
we compute the coefficient of variation of $g_j$ across
$\mathcal{I}_\epsilon$:
\begin{equation}
    \mathrm{CV} \;=\; \frac{\sigma(\{g_j\}_{j \in \mathcal{I}_\epsilon})}{\mu(\{g_j\}_{j \in \mathcal{I}_\epsilon})}.
\end{equation}
If $\mathrm{CV} < 0.1$, we classify this attention weight vector as
exhibiting self-repair. A low coefficient of variation indicates that
the contributions $g_j$ across $\mathcal{I}_\epsilon$ are similar,
approximating \Cref{ass:repair} and producing the cancellation analyzed
in \Cref{sec:self-repair-proof}.
 
Algorithm~\ref{alg:self_repair} summarizes the complete procedure.
 
\begin{algorithm}[t]
\caption{Identifying Attention Self-Repair Instances}
\label{alg:self_repair}
\begin{algorithmic}[1]
\REQUIRE Model $f$, input $x$, threshold $\epsilon = 0.01$, CV threshold $\tau_{\mathrm{CV}} = 0.1$
\ENSURE Set of self-repair instances $\mathcal{S}$
\STATE $\mathcal{S} \leftarrow \emptyset$
\STATE Run a forward and backward pass to obtain $z$, attention weights $\{\bm{s}\}$, and gradients $\{g_k = \partial z / \partial \bm{s}_k\}$
\FOR{each attention head in each layer}
    \STATE Compute importance: $\mathrm{imp}_j = g_j \cdot \bm{s}_j$ for all positions $j$
    \IF{$\max_j(\mathrm{imp}_j)$ is in the top $1\%$ globally}
        \STATE $\mathcal{I}_\epsilon \leftarrow \{j : \bm{s}_j > \epsilon\}$
        \IF{$|\mathcal{I}_\epsilon| \geq 2$}
            \STATE Compute $\mathrm{CV} = \sigma(\{g_j\}_{j \in \mathcal{I}_\epsilon}) / \mu(\{g_j\}_{j \in \mathcal{I}_\epsilon})$
            \IF{$\mathrm{CV} < \tau_{\mathrm{CV}}$}
                \STATE $\mathcal{S} \leftarrow \mathcal{S} \cup \{(\mathrm{head}, \mathcal{I}_\epsilon, \{\bm{s}_j\}_{j \in \mathcal{I}_\epsilon})\}$
            \ENDIF
        \ENDIF
    \ENDIF
\ENDFOR
\STATE \textbf{return} $\mathcal{S}$
\end{algorithmic}
\end{algorithm}

\subsection{Additional circuit localization results}\label{app:mib}
In the main paper, we only present the circuit localization results using the CPR metric. In \Cref{tab:cmd} we present the results for CMD. As mentioned in \Cref{app:cpr_cmd}, we deem this metric flawed because it penalizes methods that correctly identify components with negative contributions. We include the results for this metric for transparency.

\begin{table*}
\centering
\caption{\textbf{CMD} scores across circuit localization methods and ablation types (lower is better). All evaluations were performed using counterfactual ablations. Higher scores are better. Arithmetic scores are averaged across addition and subtraction. We \textbf{bold} and \underline{underline} the best and second-best methods per column.}
\label{tab:cmd}
\resizebox{\linewidth}{!}{%
\begin{tabular}{lcccccccccccc}
\toprule
 & \textbf{Average} & \multicolumn{4}{c}{IOI} & \multicolumn{1}{c}{Arithmetic} & \multicolumn{3}{c}{MCQA} & \multicolumn{2}{c}{ARC (E)} & \multicolumn{1}{c}{ARC (C)}\\
\cmidrule(lr){3-6} \cmidrule(lr){7-7} \cmidrule(lr){8-10} \cmidrule(lr){11-12}\cmidrule(lr){13-13}
  \textbf{Method} &  &  GPT-2 & Qwen-2.5 & Gemma-2 & Llama-3.1 & Llama-3.1 & Qwen-2.5 & Gemma-2 & Llama-3.1 & Gemma-2 & Llama-3.1 & Llama-3.1 \\
\midrule
Random & 0.75 & 0.72 & 0.69 & 0.74 & 0.75 & 0.73 & 0.68 & 0.74 & 0.68 & 0.74 & 0.74 \\
\midrule

UGS & - & 0.03 & 0.03 & - & - & - & 0.20 & - & - & - & - & - \\
EActP & - & \textbf{0.02} & 0.49 & - & - & - & 0.36 & - & - & - & - & - \\
Hybrid-ens & - & \textbf{0.02} & 0.03 & - & - & - & \textbf{0.04} \\
IFR & 0.60 & 0.42 & 0.69 & 0.75 & 0.83 & 0.22 & 0.60 & 0.62 & 0.48 & 0.66 & 0.64 & 0.76 \\
EAP-IG-activations & 0.10 & \underline{0.03} & \textbf{0.01} & \textbf{0.03} & \textbf{0.01} & \textbf{0.00} & \underline{0.05} & 0.07 & \underline{0.13} & \textbf{0.04} & 0.30 & 0.37 \\
EAP & \underline{0.08} & 0.03 & 0.15 & 0.06 & \textbf{0.01} & 0.01 & 0.07 & 0.08 & \textbf{0.09} & \textbf{0.04} & \textbf{0.11} & \underline{0.18} \\
EAP-IG-inputs & \textbf{0.07} &  \underline{0.03} & \underline{0.02} & \underline{0.04} & \textbf{0.01} & \textbf{0.00} & 0.08 & \underline{0.06} & 0.14 & \textbf{0.04} & \textbf{0.11} & 0.22 \\

\midrule
\textbf{GIM} & \textbf{0.07} & \textbf{0.02} & \underline{0.02} & 0.05  & \textbf{0.01} & \textbf{0.00} & 0.10 & \textbf{0.03} & 0.21  & \textbf{0.04} & \underline{0.16} & \textbf{0.17} \\
\bottomrule
\end{tabular}}
\end{table*}

\subsection{Additional ablation study results}
In the main paper, we showed the effect of ablating each modification in GIM individually. In \Cref{tab:mod_results}, we study the impact of adding each modification to EAP (adding all three is equivalent to GIM). 

The modifications in GIM exhibit strong synergistic effects: their combined impact far exceeds the sum of their individual contributions. \Cref{tab:mod_results} demonstrates this by showing performance when adding each modification individually to EAP (note that adding all three is equivalent to GIM). Adding a single modification to EAP yields a smaller, less consistent impact than removing one from the complete GIM method. This asymmetry indicates that the modifications work synergistically: their benefits emerge primarily when combined.

\begin{table*}[h]
    \centering
    \caption{Performance when we add individual modifications to EAP. The gain from adding individual modifications is much smaller and more inconsistent than when combining them. Scores are shown as the mean (standard deviation). Average scores better than EAP are shown in bold, while scores that are significantly better are underlined (independent t-tests with 0.05$\leq \alpha$)}
    \label{tab:mod_results}
\resizebox{\linewidth}{!}{%
\begin{tabular}{llrrrrrrrrrr}
\toprule
 &  & \multicolumn{5}{c}{\textbf{Comprehensiveness $\uparrow$}} & \multicolumn{5}{c}{\textbf{Sufficiency $\downarrow$}} \\

 &  & \multicolumn{1}{c}{Gemma}& \multicolumn{2}{c}{LLAMA} & \multicolumn{2}{c}{Qwen}  & \multicolumn{1}{c}{Gemma} & \multicolumn{2}{c}{LLAMA} & \multicolumn{2}{c}{Qwen}\\
 \cmidrule(lr){3-3}\cmidrule(lr){4-5}\cmidrule(lr){6-7}\cmidrule(lr){8-8}\cmidrule(lr){9-10}\cmidrule(lr){11-12}
 &  & 2B & 1B & 3B & 1.5B & 3B & 2B & 1B &  3B & 1.5B & 3B\\

\midrule
\multirow[c]{4}{*}{BoolQ} & EAP & 0.09 (0.08) & 0.18 (0.07) & 0.45 (0.07) & 0.27 (0.13) & 0.57 (0.18) & 0.60 (0.09) & 0.71 (0.07) & 0.39 (0.07) & 0.54 (0.17) & 0.43 (0.11) \\
 & + LN freeze & 0.08 (0.08) & \bfseries \underline{0.50 (0.07)} & 0.45 (0.14) & \bfseries 0.32 (0.26) & 0.43 (0.19) & 0.65 (0.10) & \bfseries \underline{0.36 (0.10)} & 0.44 (0.16) & 0.55 (0.14) & 0.55 (0.15) \\
 & + Grad norm & \bfseries \underline{0.44 (0.08)} & \bfseries \underline{0.38 (0.08)} & \bfseries 0.48 (0.09) & \bfseries \underline{0.40 (0.16)} & \bfseries 0.63 (0.04) & \bfseries \underline{0.37 (0.09)} & \bfseries \underline{0.56 (0.08)} & 0.43 (0.08) & \bfseries 0.52 (0.09) & 0.43 (0.07) \\
 & + TSG & 0.09 (0.09) & \bfseries \underline{0.35 (0.11)} & 0.25 (0.11) & \bfseries 0.32 (0.19) & 0.47 (0.15) & \bfseries 0.54 (0.11) & \bfseries \underline{0.56 (0.11)} & 0.65 (0.09) & 0.72 (0.09) & 0.46 (0.10) \\
\midrule
\multirow[c]{4}{*}{FEVER} & EAP & 0.03 (0.19) & 0.39 (0.07) & 0.53 (0.12) & 0.43 (0.21) & 0.47 (0.13) & 0.47 (0.11) & 0.56 (0.07) & 0.63 (0.09) & 0.59 (0.06) & 0.73 (0.10) \\
 & + LN freeze & \bfseries \underline{0.18 (0.17)} & \bfseries \underline{0.49 (0.08)} & 0.52 (0.09) & 0.37 (0.15) & \bfseries \underline{0.58 (0.14)} & \bfseries \underline{0.38 (0.12)} & \bfseries \underline{0.48 (0.08)} & 0.69 (0.10) & \bfseries \underline{0.49 (0.09)} & \bfseries \underline{0.61 (0.06)} \\
 & + Grad norm & \bfseries \underline{0.25 (0.12)} & \bfseries \underline{0.40 (0.06)} & \bfseries \underline{0.62 (0.08)} & \bfseries \underline{0.53 (0.11)} & \bfseries \underline{0.72 (0.09)} & \bfseries \underline{0.36 (0.10)} & \bfseries \underline{0.54 (0.06)} & 0.73 (0.05) & \bfseries \underline{0.51 (0.07)} & \bfseries \underline{0.53 (0.07)} \\
 & + TSG & 0.03 (0.17) & \bfseries \underline{0.42 (0.09)} & 0.41 (0.12) & 0.31 (0.07) & 0.47 (0.13) & \bfseries \underline{0.45 (0.11)} & \bfseries \underline{0.54 (0.07)} & 0.70 (0.07) & 0.70 (0.07) & \bfseries \underline{0.71 (0.09)} \\
\midrule
\multirow[c]{4}{*}{HateXplain} & EAP & 0.07 (0.05) & 0.53 (0.08) & 0.38 (0.14) & 0.49 (0.29) & 0.53 (0.15) & 0.59 (0.04) & 0.60 (0.06) & 0.67 (0.07) & 0.63 (0.09) & 0.63 (0.10) \\
 & + LN freeze & \bfseries \underline{0.19 (0.07)} & 0.42 (0.17) & 0.31 (0.18) & 0.19 (0.08) & 0.18 (0.10) & \bfseries \underline{0.58 (0.04)} & 0.66 (0.06) & 0.77 (0.08) & 0.68 (0.05) & 0.78 (0.06) \\
 & + Grad norm & \bfseries \underline{0.39 (0.03)} & 0.40 (0.09) & \bfseries \underline{0.61 (0.04)} & 0.30 (0.05) & \bfseries \underline{0.83 (0.06)} & \bfseries \underline{0.58 (0.03)} & \bfseries 0.59 (0.09) & \bfseries \underline{0.65 (0.05)} & 0.67 (0.07) & \bfseries \underline{0.61 (0.04)} \\
 & + TSG & 0.03 (0.02) & 0.39 (0.11) & \bfseries \underline{0.47 (0.09)} & 0.41 (0.10) & 0.42 (0.13) & 0.67 (0.04) & 0.65 (0.06) & \bfseries \underline{0.62 (0.05)} & 0.75 (0.09) & 0.68 (0.08) \\
\midrule
\multirow[c]{4}{*}{Movie} & EAP & 0.15 (0.13) & 0.34 (0.10) & 0.51 (0.10) & 0.42 (0.06) & 0.52 (0.10) & 0.61 (0.10) & 0.61 (0.09) & 0.51 (0.07) & 0.68 (0.12) & 0.76 (0.17) \\
 & + LN freeze & \bfseries \underline{0.28 (0.15)} & \bfseries \underline{0.57 (0.06)} & \bfseries \underline{0.68 (0.07)} & 0.29 (0.10) & 0.26 (0.17) & \bfseries \underline{0.56 (0.11)} & \bfseries \underline{0.28 (0.07)} & \bfseries \underline{0.40 (0.10)} & 0.75 (0.14) & \bfseries 0.75 (0.08) \\
 & + Grad norm & \bfseries \underline{0.52 (0.07)} & 0.34 (0.08) & 0.49 (0.04) & 0.42 (0.13) & \bfseries \underline{0.73 (0.06)} & \bfseries \underline{0.50 (0.06)} & 0.65 (0.07) & 0.71 (0.05) & \bfseries \underline{0.55 (0.08)} & \bfseries \underline{0.51 (0.07)} \\
 & + TSG & 0.14 (0.11) & \bfseries \underline{0.57 (0.08)} & 0.27 (0.05) & \bfseries \underline{0.56 (0.11)} & 0.50 (0.10) & 0.65 (0.10) & \bfseries \underline{0.45 (0.09)} & 0.74 (0.05) & 0.68 (0.18) & 0.80 (0.18) \\
\midrule
\multirow[c]{4}{*}{SciFact} & EAP & 0.09 (0.19) & 0.27 (0.07) & 0.42 (0.15) & 0.34 (0.20) & 0.57 (0.20) & 0.55 (0.11) & 0.57 (0.09) & 0.59 (0.09) & 0.56 (0.11) & 0.63 (0.16) \\
 & + LN freeze & \bfseries \underline{0.16 (0.17)} & \bfseries \underline{0.52 (0.06)} & \bfseries \underline{0.49 (0.13)} & 0.23 (0.14) & \bfseries 0.59 (0.28) & \bfseries \underline{0.45 (0.15)} & \bfseries \underline{0.42 (0.09)} & \bfseries \underline{0.55 (0.10)} & 0.57 (0.11) & 0.65 (0.13) \\
 & + Grad norm & \bfseries \underline{0.41 (0.12)} & \bfseries \underline{0.36 (0.07)} & \bfseries \underline{0.48 (0.11)} & \bfseries \underline{0.43 (0.15)} & \bfseries \underline{0.76 (0.09)} & \bfseries \underline{0.43 (0.16)} & \bfseries \underline{0.53 (0.07)} & 0.69 (0.10) & \bfseries 0.55 (0.14) & \bfseries \underline{0.52 (0.09)} \\
 & + TSG & 0.09 (0.19) & \bfseries \underline{0.31 (0.11)} & 0.25 (0.11) & \bfseries 0.37 (0.14) & \bfseries \underline{0.63 (0.21)} & 0.56 (0.12) & \bfseries 0.56 (0.09) & 0.69 (0.09) & 0.80 (0.20) & 0.66 (0.13) \\
\midrule
\multirow[c]{4}{*}{Twitter} & EAP & 0.40 (0.09) & 0.41 (0.07) & 0.48 (0.10) & 0.48 (0.07) & 0.73 (0.10) & 0.51 (0.07) & 0.64 (0.05) & 0.75 (0.05) & 0.73 (0.06) & 0.64 (0.04) \\
 & + LN freeze & \bfseries \underline{0.45 (0.08)} & \bfseries \underline{0.50 (0.06)} & 0.46 (0.08) & 0.39 (0.05) & \bfseries \underline{0.75 (0.10)} & \bfseries \underline{0.48 (0.05)} & \bfseries \underline{0.56 (0.04)} & 0.75 (0.06) & \bfseries \underline{0.72 (0.06)} & 0.69 (0.05) \\
 & + Grad norm & \bfseries \underline{0.48 (0.07)} & 0.38 (0.09) & \bfseries \underline{0.55 (0.05)} & 0.39 (0.08) & 0.69 (0.06) & \bfseries \underline{0.47 (0.04)} & \bfseries \underline{0.62 (0.05)} & 0.75 (0.02) & \bfseries \underline{0.69 (0.03)} & 0.64 (0.04) \\
 & + TSG & 0.35 (0.06) & \bfseries \underline{0.50 (0.06)} & 0.47 (0.09) & \bfseries \underline{0.59 (0.06)} & 0.71 (0.08) & 0.59 (0.06) & \bfseries \underline{0.61 (0.05)} & 0.76 (0.05) & 0.78 (0.04) & 0.66 (0.05) \\
\bottomrule
\end{tabular}}
\end{table*}

\subsection{Licenses}\label{sec:licenses}
Here is a list of the assets used in our experiments and their Licences:

\begin{enumerate}
    \item BoolQ: CC-BY 3.0
    \item FEVER: CC-BY 3.0
    \item SciFact: CC BY 4.0.
    \item HateXplain: MIT License
    \item Movie review: CC-BY 3.0
    \item Twitter Sentiment Extraction: CC-BY 4.0
    \item LLAMA models: \hyperlink{https://huggingface.co/meta-llama/Llama-3.2-1B/blob/main/LICENSE.txt}{LLAMA3.2 COMMUNITY LICENSE AGREEMENT}
    \item Gemma models: \hyperlink{https://ai.google.dev/gemma/terms}{GEMMA LICENCE AGREEMENT}
    \item Qwen models: \hyperlink{https://huggingface.co/Qwen/Qwen2.5-72B-Instruct/blob/main/LICENSE}{Qwen LICENSE AGREEMENT}
\end{enumerate}

\subsection{Computational requirements}\label{sec:computation}
We ran all our experiments on an A100 80GB GPU. The feature attribution and self-repair experiments were fast, using less than an hour each for the smaller models (<3B) and around 2--3 hours for the large models. The experiments comparing circuit localization were slow, especially for integrated gradients. Integrated gradients used around half a day per dataset on the A100 GPU. 

\subsection{Impact Statement}\label{sec:impact}
This paper presents work whose goal is to advance the field of mechanistic interpretability. Improved methods for understanding the internal mechanisms of LLMs may contribute to AI safety and transparency. We do not foresee specific negative societal consequences arising from this work.

\subsection{Classification performance}
In \Cref{tab:classification}, we evaluate the LLMs' performance on the datasets. All models perform well across datasets except for LLAMA-3.2 1B, which always predicts positive sentiment on the Movie and Twitter datasets (high recall and low specificity). Since we in the main paper focus on evaluating explanation faithfulness, a strong performance is not crucial for accurate results. We only want accurate explanations of the models' inner mechanisms, whether that inner mechanism solves the tasks well or not.

\begin{table*}
\caption{Classification performance of the models across the datasets. Except for LLAMA3.1 1B on the sentiment classification datasets, where it always predicts positive sentiment, the models perform well across all tasks.}
\label{tab:classification}
\centering
\begin{tabular}{llrrrrr}
\toprule
 &  & Accuracy & Precision & Recall & F1 & Specificity \\
\midrule
\multirow[c]{7}{*}{HateXplain} & Gemma2.2 2B & 57.49 & 50.39 & 98.82 & 66.74 & 26.09 \\
 & Gemma2.2 9B & 65.12 & 55.31 & 100.00 & 71.22 & 38.62 \\
 & Llama3.2 1B & 50.51 & 46.14 & 87.54 & 60.43 & 22.38 \\
 & Llama3.2 3B & 63.74 & 54.65 & 93.94 & 69.10 & 40.79 \\
 & Llama3.1 8B & 54.94 & 48.93 & 100.00 & 65.71 & 20.72 \\
 & Qwen2 1.5B & 54.36 & 48.55 & 95.79 & 64.44 & 22.89 \\
 & Qwen2 3B & 69.62 & 59.50 & 92.76 & 72.50 & 52.05 \\
 \midrule
\multirow[c]{7}{*}{Movie} & Gemma2.2 2B & 82.41 & 100.00 & 64.65 & 78.53 & 100.00 \\
 & Gemma2.2 9B & 95.48 & 98.91 & 91.92 & 95.29 & 99.00 \\
 & Llama3.2 1B & 49.75 & 49.75 & 100.00 & 66.44 & 0.00 \\
 & Llama3.2 3B & 92.46 & 87.50 & 98.99 & 92.89 & 86.00 \\
 & Llama3.1 8B & 88.44 & 81.67 & 98.99 & 89.50 & 78.00 \\
 & Qwen2 1.5B & 85.93 & 80.34 & 94.95 & 87.04 & 77.00 \\
 & Qwen2 3B & 85.93 & 97.33 & 73.74 & 83.91 & 98.00 \\
 \midrule
\multirow[c]{7}{*}{Twitter} & Gemma2.2 2B & 84.28 & 89.69 & 79.11 & 84.07 & 89.98 \\
 & Gemma2.2 9B & 88.11 & 89.37 & 87.77 & 88.56 & 88.48 \\
 & Llama3.2 1B & 52.69 & 52.58 & 99.83 & 68.88 & 0.71 \\
 & Llama3.2 3B & 84.06 & 91.34 & 76.91 & 83.50 & 91.95 \\
 & Llama3.1 8B & 87.06 & 85.43 & 90.81 & 88.04 & 82.92 \\
 & Qwen2 1.5B & 77.79 & 73.92 & 89.06 & 80.79 & 65.35 \\
 & Qwen2 3B & 81.82 & 93.15 & 70.53 & 80.28 & 94.28 \\
 \midrule
\multirow[c]{7}{*}{BoolQ} & Gemma2.2 2B & 78.39 & 81.82 & 85.04 & 83.40 & 66.67 \\
 & Gemma2.2 9B & 82.41 & 87.10 & 85.04 & 86.06 & 77.78 \\
 & Llama3.2 1B & 65.33 & 69.86 & 80.31 & 74.73 & 38.89 \\
 & Llama3.2 3B & 67.84 & 87.95 & 57.48 & 69.52 & 86.11 \\
 & Llama3.1 8B & 76.88 & 89.32 & 72.44 & 80.00 & 84.72 \\
 & Qwen2 1.5B & 66.33 & 68.29 & 88.19 & 76.98 & 27.78 \\
 & Qwen2 3B & 64.32 & 87.84 & 51.18 & 64.68 & 87.50 \\
 \midrule
\multirow[c]{7}{*}{FEVER} & Gemma2.2 2B & 81.41 & 85.19 & 75.70 & 80.17 & 87.04 \\
 & Gemma2.2 9B & 93.41 & 91.59 & 95.48 & 93.49 & 91.36 \\
 & Llama3.2 1B & 62.22 & 61.22 & 65.12 & 63.11 & 59.36 \\
 & Llama3.2 3B & 92.31 & 94.08 & 90.17 & 92.09 & 94.41 \\
 & Llama3.1 8B & 94.37 & 94.98 & 93.60 & 94.29 & 95.13 \\
 & Qwen2 1.5B & 92.10 & 94.09 & 89.71 & 91.85 & 94.44 \\
 & Qwen2 3B & 89.30 & 96.92 & 81.01 & 88.25 & 97.47 \\
 \midrule
\multirow[c]{7}{*}{SciFact} & Gemma2.2 2B & 78.95 & 80.13 & 90.58 & 85.03 & 56.34 \\
 & Gemma2.2 9B & 86.60 & 92.31 & 86.96 & 89.55 & 85.92 \\
 & Llama3.2 1B & 49.76 & 77.97 & 33.33 & 46.70 & 81.69 \\
 & Llama3.2 3B & 82.78 & 91.13 & 81.88 & 86.26 & 84.51 \\
 & Llama3.1 8B & 83.73 & 94.83 & 79.71 & 86.61 & 91.55 \\
 & Qwen2 1.5B & 79.43 & 86.82 & 81.16 & 83.90 & 76.06 \\
 & Qwen2 3B & 77.03 & 97.87 & 66.67 & 79.31 & 97.18 \\
\bottomrule
\end{tabular}
\end{table*}

\subsection{Model prompts}
We designed unique prompt templates for each model-dataset pair. We tested different templates and chose the ones that resulted in the highest classification accuracy. Each prompt was designed such that the model would answer yes or no. The prompt templates are provided in our released code repository: \url{https://github.com/JoakimEdin/gim}

\subsection{TSG at different temperatures}
In \Cref{tab:temperatures}, we present the GIM results for different temperatures used in TSG. Note that we ran this experiment after having selected the temperature of 2 in the main paper. We selected this temperature based on the results for Gemma 2B on Fever and HateXplain. Each model-dataset pair has a sweetspot. Increasing the temperature to high levels degrades performance.
\begin{table*}[]
\caption{Faithfulness of GIM when using different temperatures for TSG across all models and datasets.}
\label{tab:temperatures}
\centering
\resizebox{0.8\linewidth}{!}{
\begin{tabular}{llcccccccccc}
    
\toprule
 &  & \multicolumn{5}{c}{\textbf{Comprehensiveness $\uparrow$}} & \multicolumn{5}{c}{\textbf{Sufficiency $\downarrow$}} \\

 &  & \multicolumn{1}{c}{Gemma}& \multicolumn{2}{c}{LLAMA} & \multicolumn{2}{c}{Qwen}  & \multicolumn{1}{c}{Gemma} & \multicolumn{2}{c}{LLAMA} & \multicolumn{2}{c}{Qwen}\\
 \cmidrule(lr){3-3}\cmidrule(lr){4-5}\cmidrule(lr){6-7}\cmidrule(lr){8-8}\cmidrule(lr){9-10}\cmidrule(lr){11-12}
 & Temp & 2B & 1B & 3B & 1.5B & 3B & 2B & 1B &  3B & 1.5B & 3B\\
\midrule
\multirow[c]{8}{*}{BoolQ} & 1 & 0.53 & 0.68 & 0.72 & 0.69 & \bfseries 0.66 & 0.07 & 0.28 & 0.12 & 0.12 & 0.29 \\
 & 1.5 & 0.57 & 0.68 & 0.72 & 0.71 & 0.64 & 0.05 & 0.23 & 0.10 & 0.17 & \bfseries 0.22 \\
 & 2 & 0.59 & 0.69 & \bfseries 0.72 & 0.68 & 0.61 & \bfseries 0.03 & 0.22 & \bfseries 0.10 & \bfseries 0.09 & 0.23 \\
 & 2.5 & 0.57 & 0.68 & 0.72 & 0.71 & 0.64 & 0.05 & 0.23 & 0.10 & 0.17 & \bfseries 0.22 \\
 & 3 & 0.60 & \bfseries 0.69 & 0.70 & 0.67 & 0.61 & 0.04 & 0.21 & 0.12 & 0.10 & 0.25 \\
 & 5 & 0.60 & 0.69 & 0.66 & \bfseries 0.73 & 0.59 & 0.05 & \bfseries 0.20 & 0.13 & 0.23 & 0.24 \\
 & 10 & 0.60 & 0.69 & 0.65 & 0.62 & 0.62 & 0.06 & 0.20 & 0.12 & 0.22 & 0.23 \\
 & 100 & \bfseries 0.61 & 0.69 & 0.65 & 0.66 & 0.63 & 0.06 & 0.21 & 0.13 & 0.22 & 0.26 \\
\midrule
\multirow[c]{8}{*}{FEVER} & 1 & 0.39 & 0.60 & \bfseries 0.76 & 0.62 & 0.66 & -0.03 & 0.29 & 0.39 & \bfseries 0.25 & 0.46 \\
 & 1.5 & 0.39 & 0.61 & 0.75 & \bfseries 0.64 & 0.67 & 0.03 & \bfseries 0.26 & \bfseries 0.38 & 0.35 & 0.43 \\
 & 2 & 0.42 & \bfseries 0.62 & 0.75 & 0.51 & \bfseries 0.68 & -0.01 & 0.26 & 0.39 & 0.31 & \bfseries 0.41 \\
 & 2.5 & 0.39 & 0.61 & 0.75 & \bfseries 0.64 & 0.67 & 0.03 & \bfseries 0.26 & \bfseries 0.38 & 0.35 & 0.43 \\
 & 3 & \bfseries 0.43 & 0.62 & 0.74 & 0.48 & 0.63 & \bfseries -0.13 & 0.27 & 0.40 & 0.33 & 0.42 \\
 & 5 & 0.42 & 0.61 & 0.74 & 0.55 & 0.63 & -0.13 & 0.27 & 0.41 & 0.48 & 0.44 \\
 & 10 & 0.42 & 0.60 & 0.75 & 0.57 & 0.65 & -0.13 & 0.27 & 0.41 & 0.48 & 0.46 \\
 & 100 & 0.42 & 0.60 & 0.75 & 0.57 & 0.67 & -0.12 & 0.27 & 0.41 & 0.47 & 0.48 \\
\midrule
\multirow[c]{8}{*}{HateXplain} & 1 & 0.63 & 0.67 & 0.64 & 0.85 & 0.50 & 0.31 & 0.37 & 0.52 & \bfseries 0.28 & 0.91 \\
 & 1.5 & \bfseries 0.64 & 0.68 & \bfseries 0.65 & 0.83 & 0.62 & \bfseries 0.23 & 0.28 & 0.48 & 0.43 & 0.59 \\
 & 2 & 0.64 & 0.68 & 0.64 & 0.80 & 0.63 & 0.26 & 0.26 & \bfseries 0.46 & 0.35 & 0.51 \\
 & 2.5 & \bfseries 0.64 & 0.68 & \bfseries 0.65 & 0.83 & 0.62 & \bfseries 0.23 & 0.28 & 0.48 & 0.43 & 0.59 \\
 & 3 & 0.61 & \bfseries 0.68 & 0.59 & 0.77 & 0.56 & 0.29 & 0.22 & 0.49 & 0.36 & 0.46 \\
 & 5 & 0.60 & 0.68 & 0.54 & 0.82 & 0.58 & 0.31 & \bfseries 0.20 & 0.51 & 0.48 & 0.45 \\
 & 10 & 0.61 & 0.68 & 0.56 & 0.77 & 0.68 & 0.33 & 0.21 & 0.52 & 0.52 & 0.45 \\
 & 100 & 0.61 & 0.68 & 0.58 & \bfseries 0.93 & \bfseries 0.76 & 0.33 & 0.24 & 0.52 & 0.51 & \bfseries 0.45 \\
\midrule
\multirow[c]{8}{*}{Movie} & 1 & 0.55 & 0.68 & 0.78 & 0.82 & 0.73 & 0.16 & 0.26 & 0.31 & \bfseries 0.13 & 0.45 \\
 & 1.5 & 0.58 & 0.70 & 0.78 & \bfseries 0.84 & 0.75 & 0.12 & \bfseries 0.18 & \bfseries 0.28 & 0.23 & 0.38 \\
 & 2 & 0.60 & 0.71 & \bfseries 0.78 & 0.83 & \bfseries 0.76 & \bfseries 0.11 & 0.19 & 0.30 & 0.17 & 0.31 \\
 & 2.5 & 0.58 & 0.70 & 0.78 & \bfseries 0.84 & 0.75 & 0.12 & \bfseries 0.18 & \bfseries 0.28 & 0.23 & 0.38 \\
 & 3 & 0.60 & 0.71 & 0.78 & 0.80 & 0.71 & 0.11 & 0.20 & 0.30 & 0.16 & 0.32 \\
 & 5 & 0.60 & 0.71 & 0.76 & 0.82 & 0.72 & 0.11 & 0.21 & 0.32 & 0.28 & \bfseries 0.31 \\
 & 10 & 0.60 & \bfseries 0.71 & 0.75 & 0.74 & 0.73 & 0.11 & 0.22 & 0.33 & 0.28 & 0.32 \\
 & 100 & \bfseries 0.61 & 0.71 & 0.75 & 0.74 & 0.70 & 0.14 & 0.23 & 0.34 & 0.28 & 0.42 \\
\midrule
\multirow[c]{8}{*}{SciFact} & 1 & \bfseries 0.60 & 0.64 & 0.72 & 0.66 & 0.72 & 0.05 & 0.25 & 0.30 & \bfseries 0.20 & 0.40 \\
 & 1.5 & 0.58 & 0.66 & 0.73 & \bfseries 0.70 & 0.73 & 0.06 & \bfseries 0.22 & \bfseries 0.30 & 0.34 & \bfseries 0.38 \\
 & 2 & 0.57 & \bfseries 0.67 & \bfseries 0.74 & 0.62 & \bfseries 0.74 & 0.04 & 0.22 & 0.31 & 0.26 & 0.38 \\
 & 2.5 & 0.58 & 0.66 & 0.73 & \bfseries 0.70 & 0.73 & 0.06 & \bfseries 0.22 & \bfseries 0.30 & 0.34 & \bfseries 0.38 \\
 & 3 & 0.56 & 0.66 & 0.73 & 0.55 & 0.68 & \bfseries 0.01 & 0.24 & 0.33 & 0.27 & 0.41 \\
 & 5 & 0.54 & 0.64 & 0.72 & 0.61 & 0.67 & 0.01 & 0.24 & 0.35 & 0.46 & 0.41 \\
 & 10 & 0.54 & 0.63 & 0.72 & 0.61 & 0.69 & 0.02 & 0.24 & 0.36 & 0.47 & 0.41 \\
 & 100 & 0.55 & 0.63 & 0.73 & 0.63 & 0.70 & 0.04 & 0.24 & 0.36 & 0.46 & 0.43 \\
\midrule
\multirow[c]{8}{*}{Twitter} & 1 & 0.52 & 0.64 & 0.73 & 0.78 & 0.65 & \bfseries 0.32 & 0.43 & \bfseries 0.60 & \bfseries 0.47 & 0.66 \\
 & 1.5 & 0.52 & 0.66 & \bfseries 0.73 & \bfseries 0.81 & 0.71 & 0.33 & 0.42 & 0.61 & 0.58 & 0.59 \\
 & 2 & 0.52 & 0.68 & 0.72 & 0.76 & \bfseries 0.72 & 0.34 & \bfseries 0.41 & 0.61 & 0.55 & 0.58 \\
 & 2.5 & 0.52 & 0.66 & \bfseries 0.73 & \bfseries 0.81 & 0.71 & 0.33 & 0.42 & 0.61 & 0.58 & 0.59 \\
 & 3 & 0.52 & \bfseries 0.68 & 0.69 & 0.76 & 0.70 & 0.35 & 0.42 & 0.63 & 0.56 & 0.58 \\
 & 5 & 0.52 & 0.67 & 0.69 & 0.79 & 0.69 & 0.35 & 0.43 & 0.64 & 0.60 & 0.58 \\
 & 10 & 0.53 & 0.66 & 0.69 & 0.67 & 0.69 & 0.35 & 0.44 & 0.65 & 0.63 & \bfseries 0.57 \\
 & 100 & \bfseries 0.54 & 0.65 & 0.69 & 0.67 & 0.72 & 0.35 & 0.45 & 0.65 & 0.63 & 0.58 \\
\bottomrule
\end{tabular}}
\end{table*}

\subsection{Comprehensiveness and Sufficiency per layer} \label{app:layer_aopc}
We analyze the performance of different methods per layer. We compare GIM with EAP~\citep{syedAttributionPatchingOutperforms2023}, ATP*\footnote{We only implemented ATP*'s softmax gradient fix component.}~\citep{kramarAtPEfficientScalable2024}, and EAP-IG~\citep{hannaHaveFaithFaithfulness2024} by computing comprehensiveness and sufficiency at each layer separately. We use the layer-average representations as the counterfactual.

\begin{figure*}[h]
     \centering
     \begin{subfigure}[b]{0.25\textwidth}
         \centering
         \includegraphics[width=\textwidth]{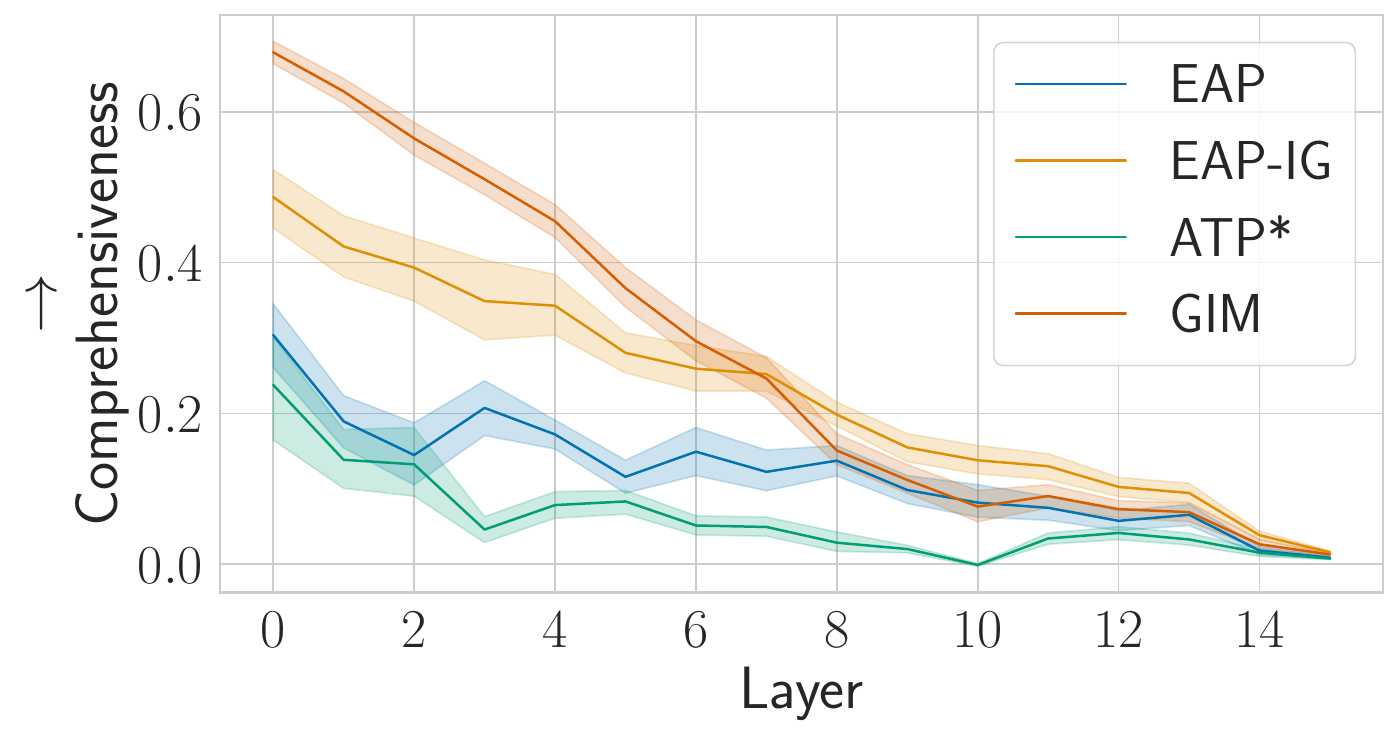}
         \includegraphics[width=\textwidth]{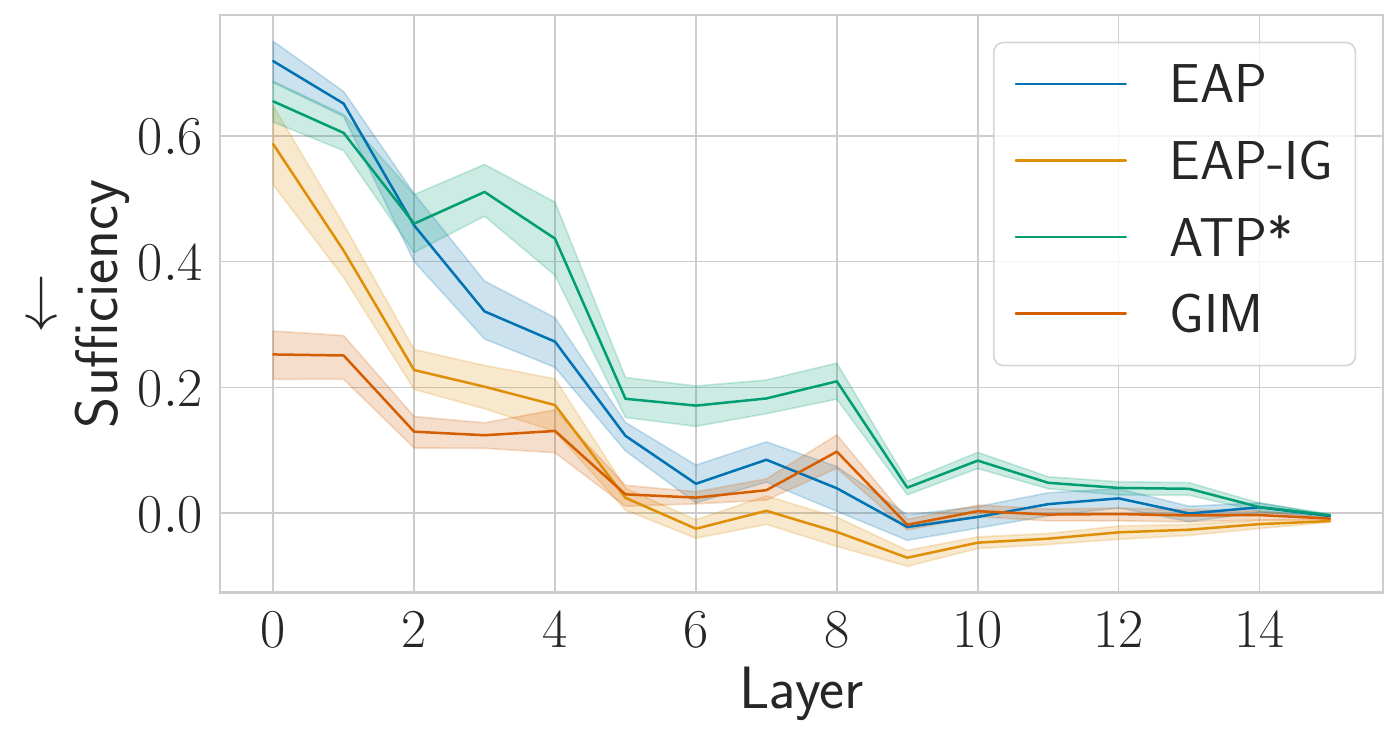}
         \caption{BoolQ}
         \label{fig:boolq}
     \end{subfigure}
     \hfill
     \begin{subfigure}[b]{0.25\textwidth}
         \centering
         \includegraphics[width=\textwidth]{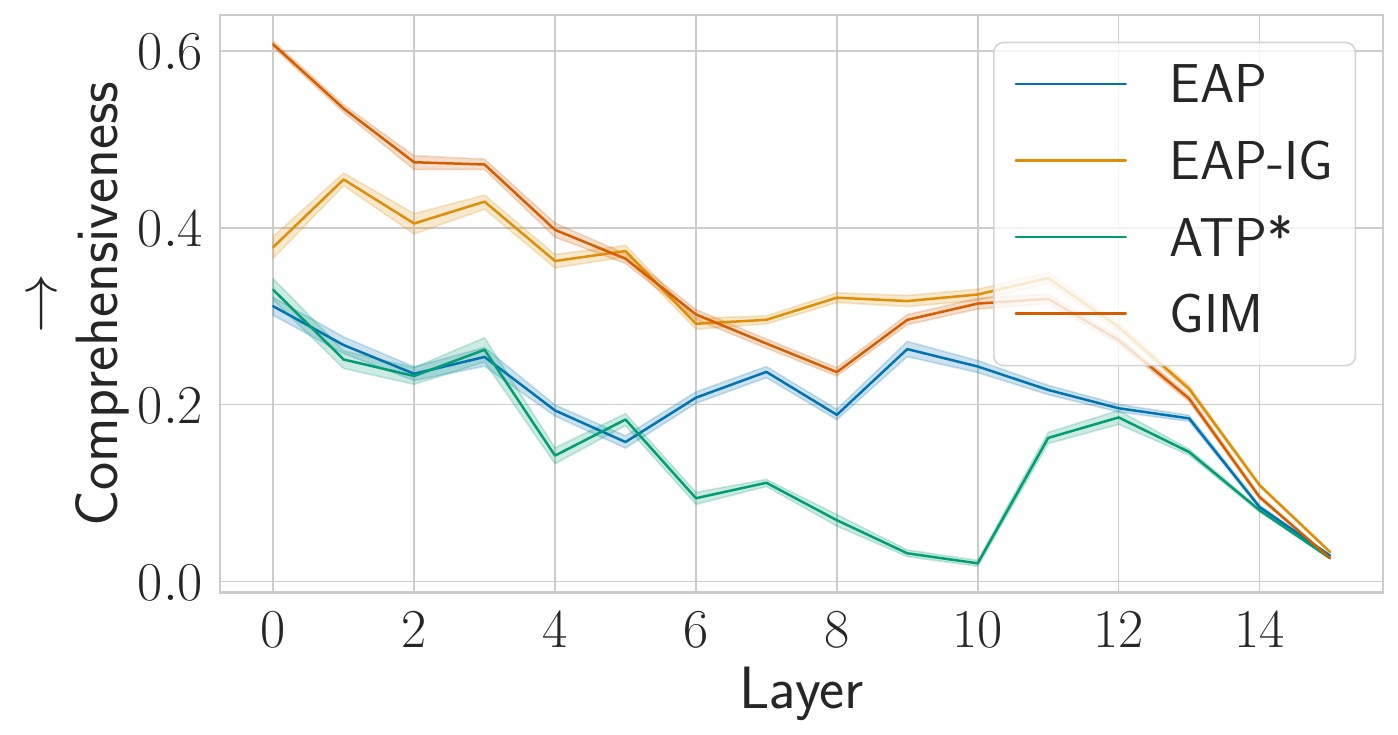}
         \includegraphics[width=\textwidth]{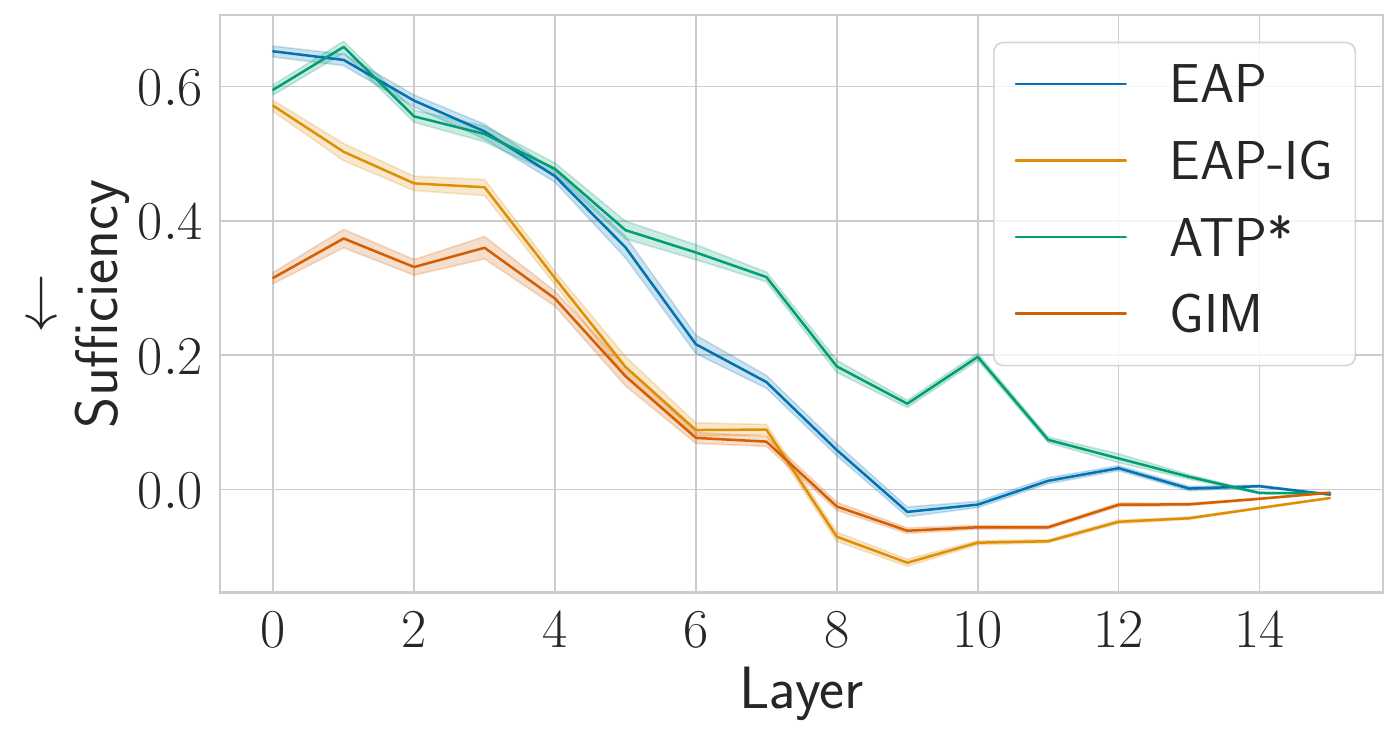}
         \caption{Fever}
     \end{subfigure}
     \hfill
     \begin{subfigure}[b]{0.25\textwidth}
         \centering
         \includegraphics[width=\textwidth]{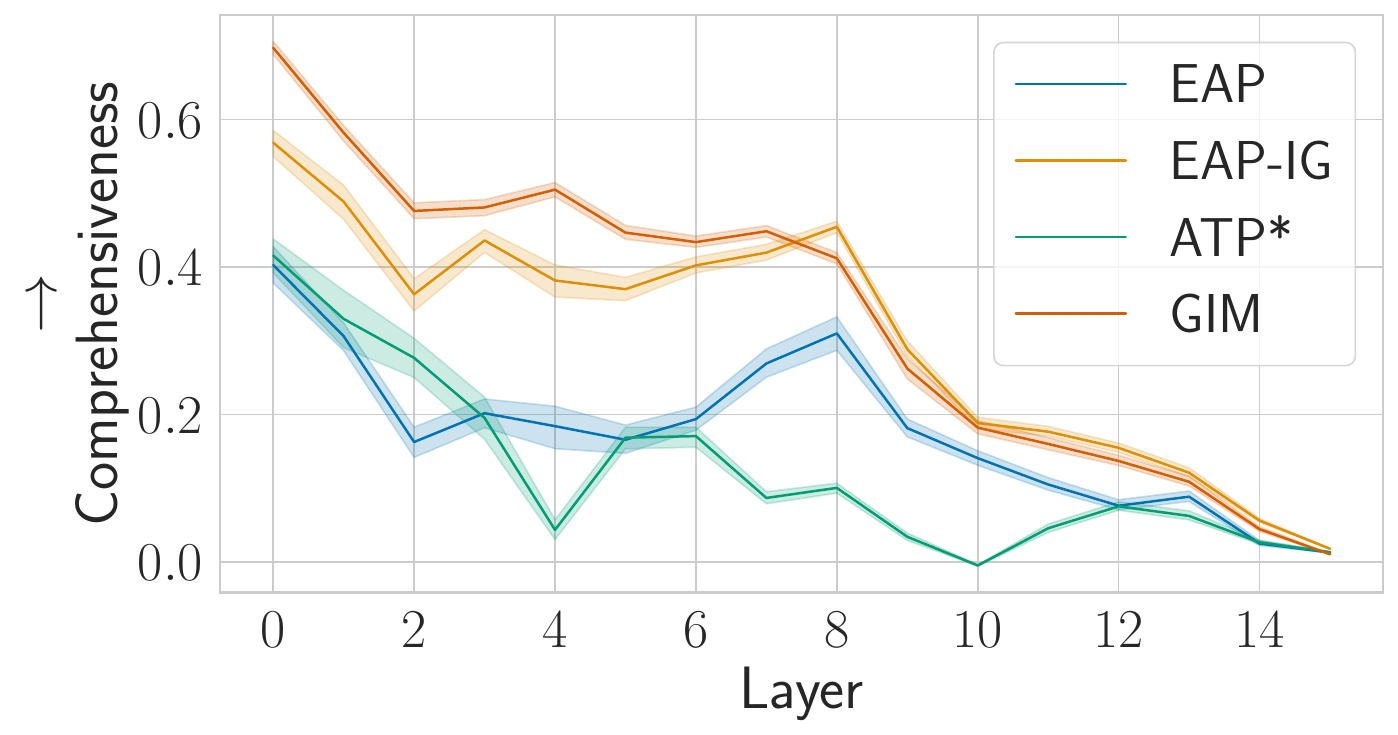}
         \includegraphics[width=\textwidth]{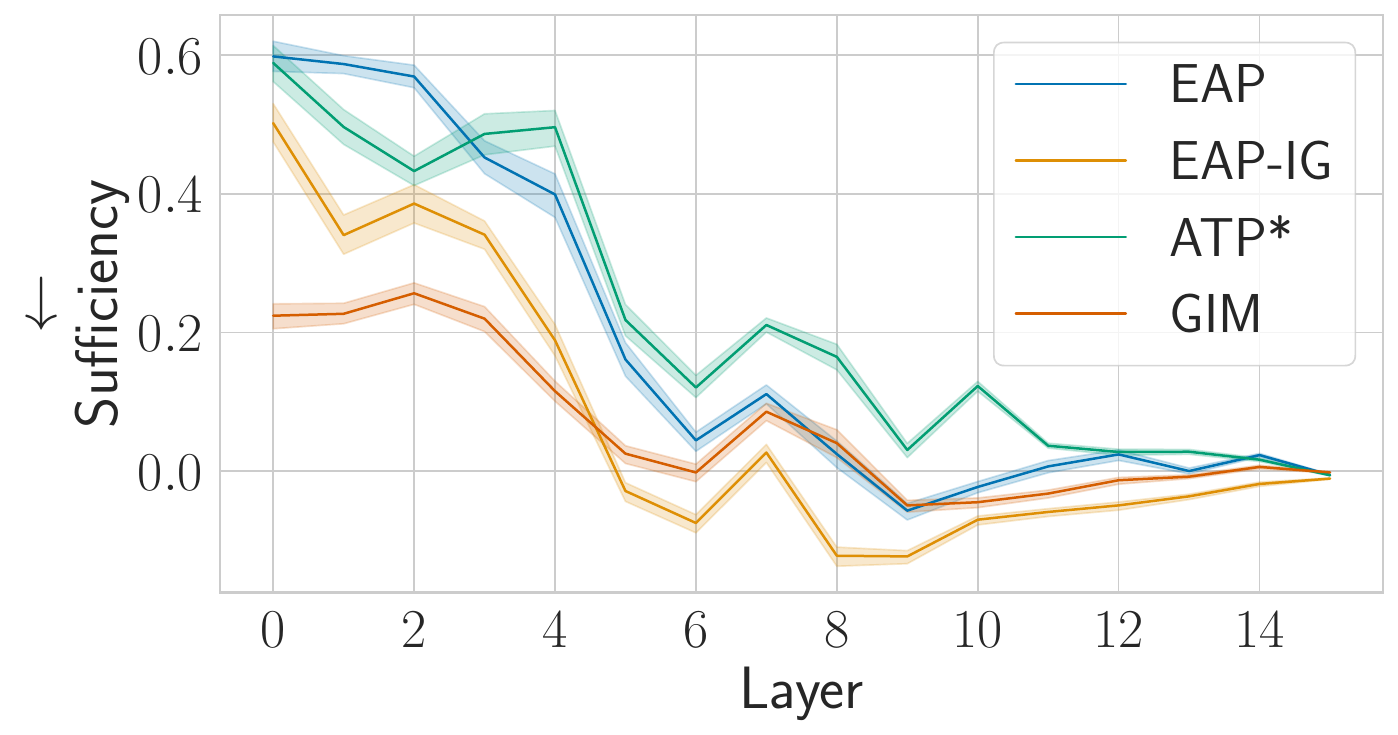}
         \caption{Movie review}

     \end{subfigure}
     \begin{subfigure}[b]{0.25\textwidth}
         \centering
         \includegraphics[width=\textwidth]{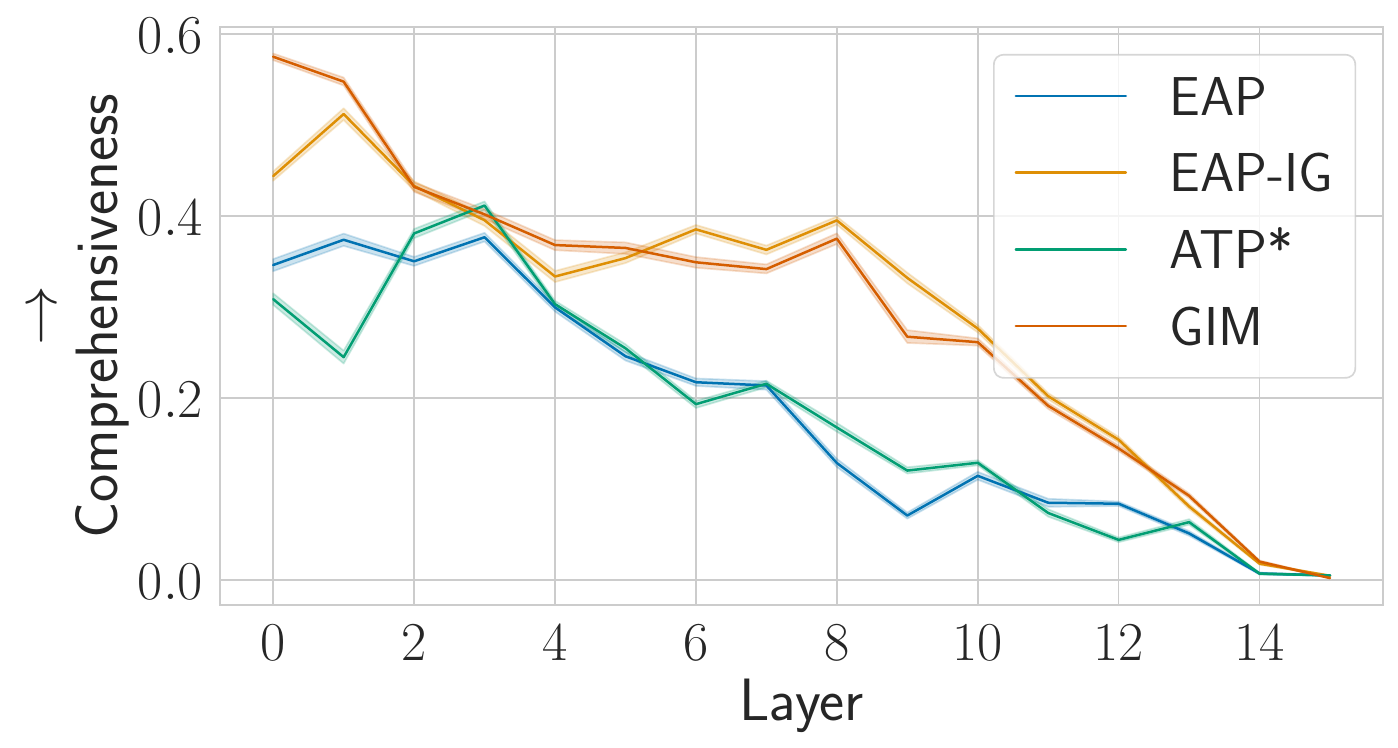}
         \includegraphics[width=\textwidth]{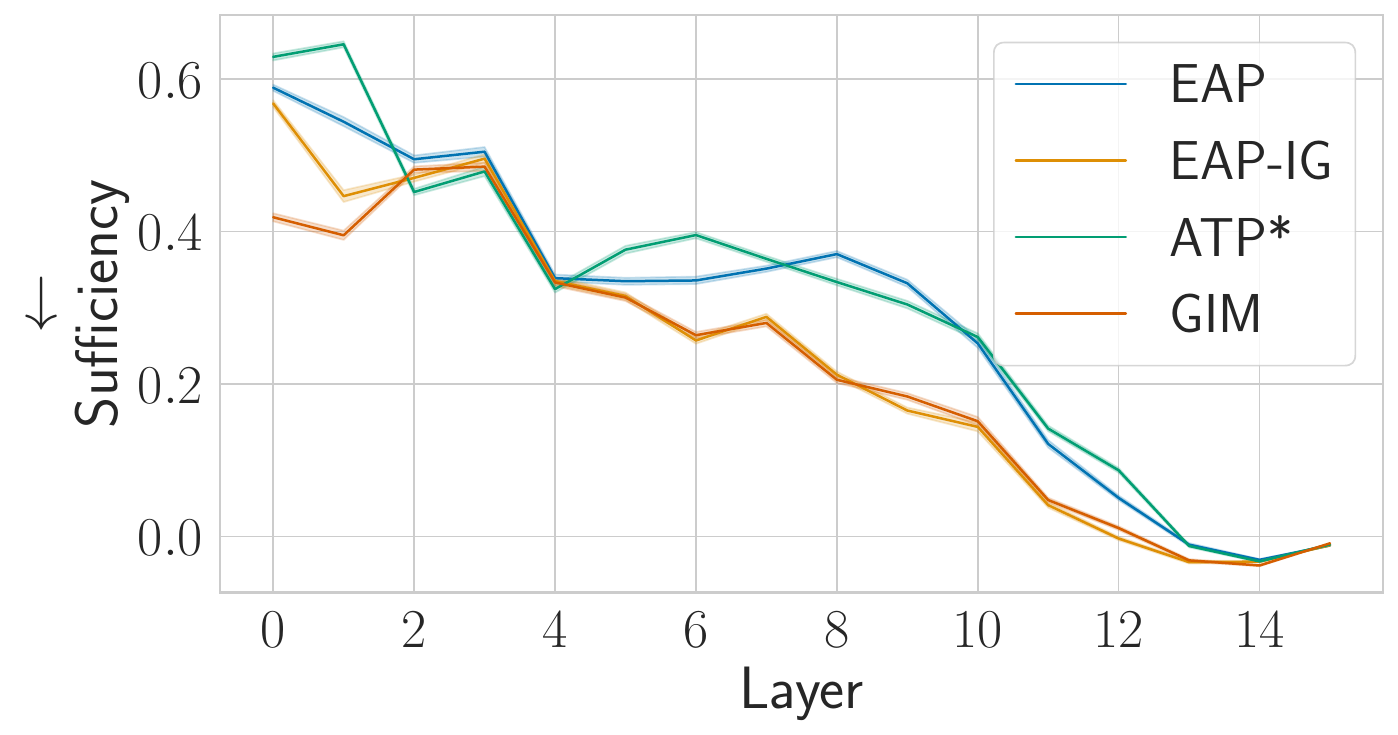}
         \caption{Twitter}
     \end{subfigure}
     \hfill
     \begin{subfigure}[b]{0.25\textwidth}
         \centering
         \includegraphics[width=\textwidth]{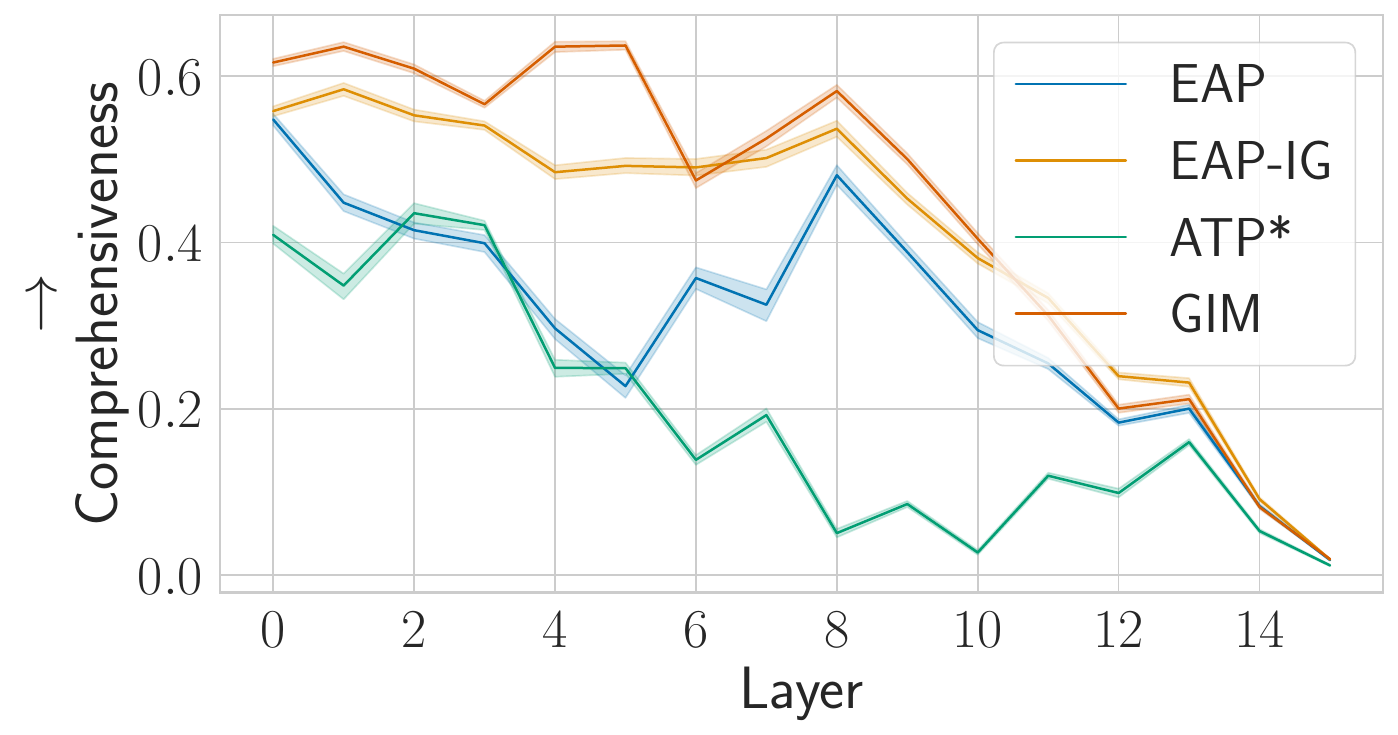}
         \includegraphics[width=\textwidth]{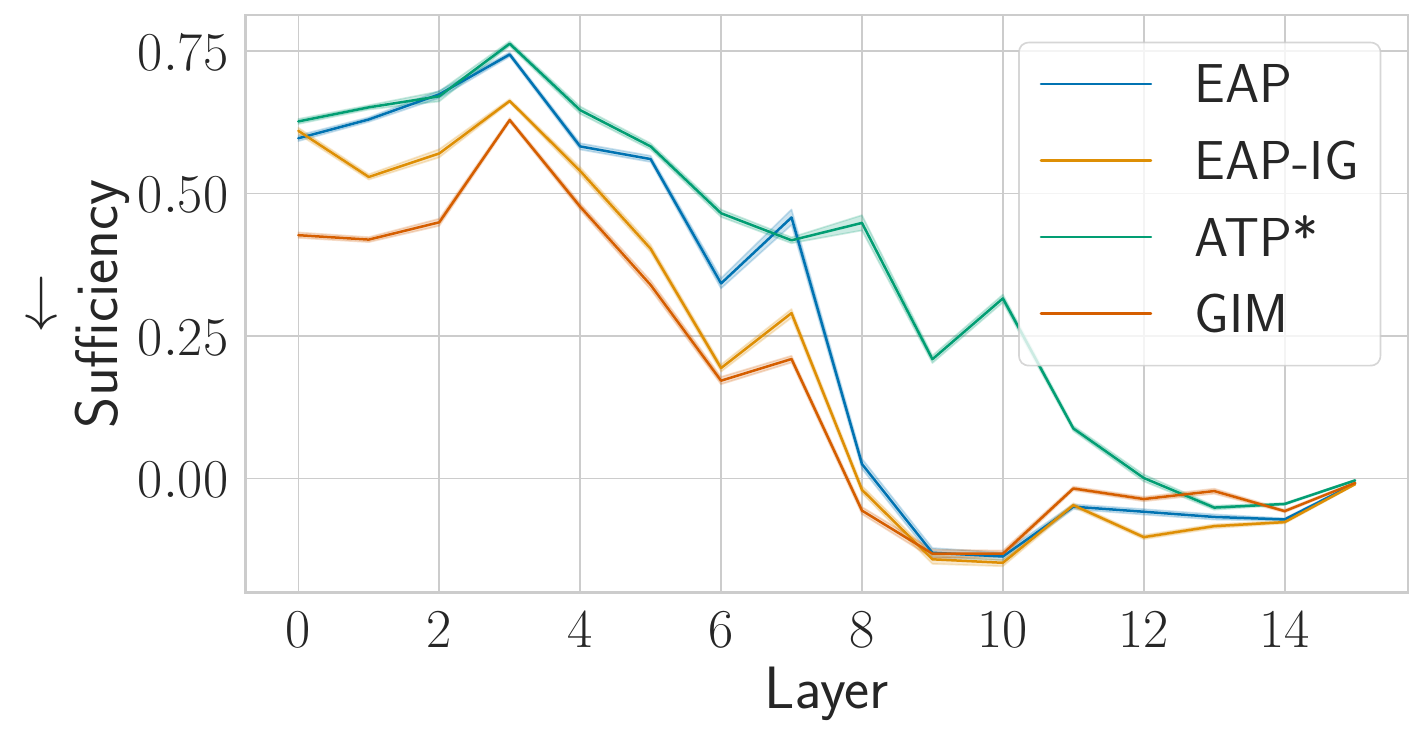}
         \caption{Hatexplain}
     \end{subfigure}
     \hfill
     \begin{subfigure}[b]{0.25\textwidth}
         \centering
         \includegraphics[width=\textwidth]{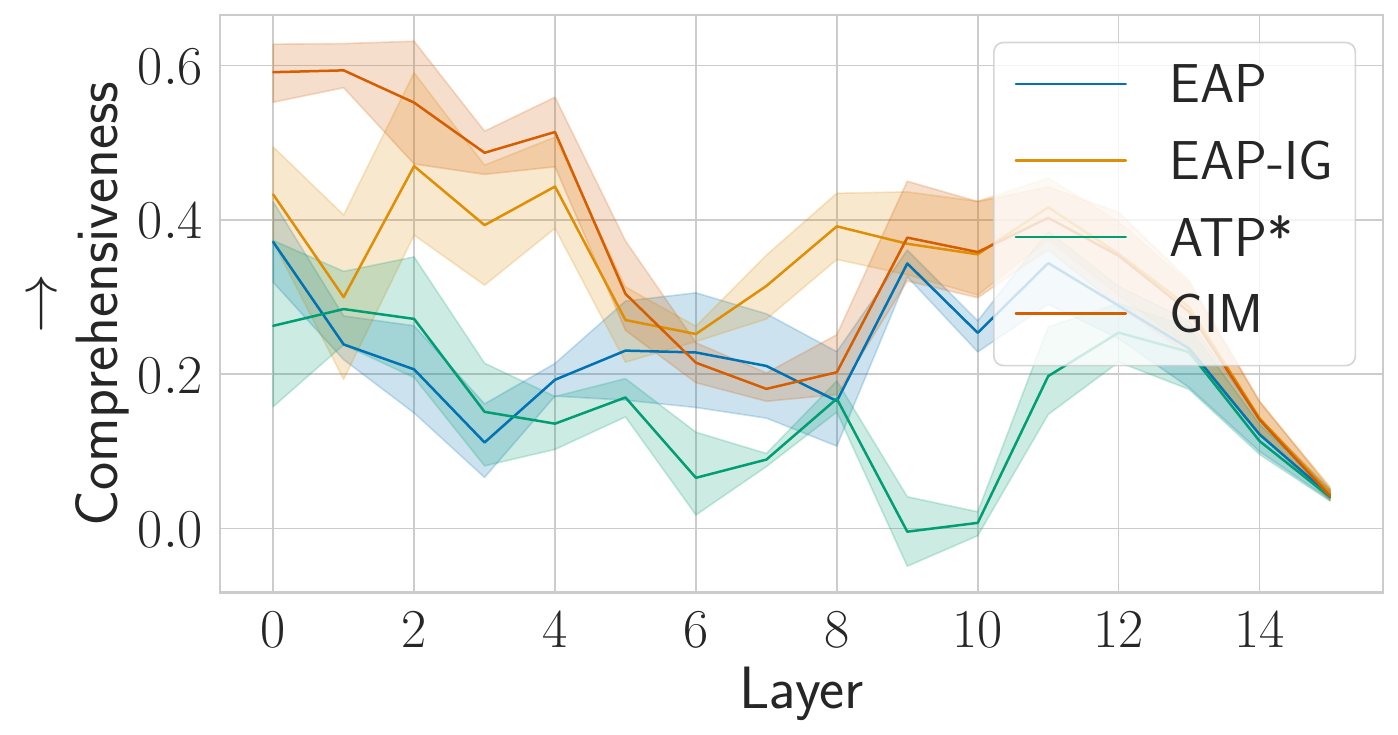}
         \includegraphics[width=\textwidth]{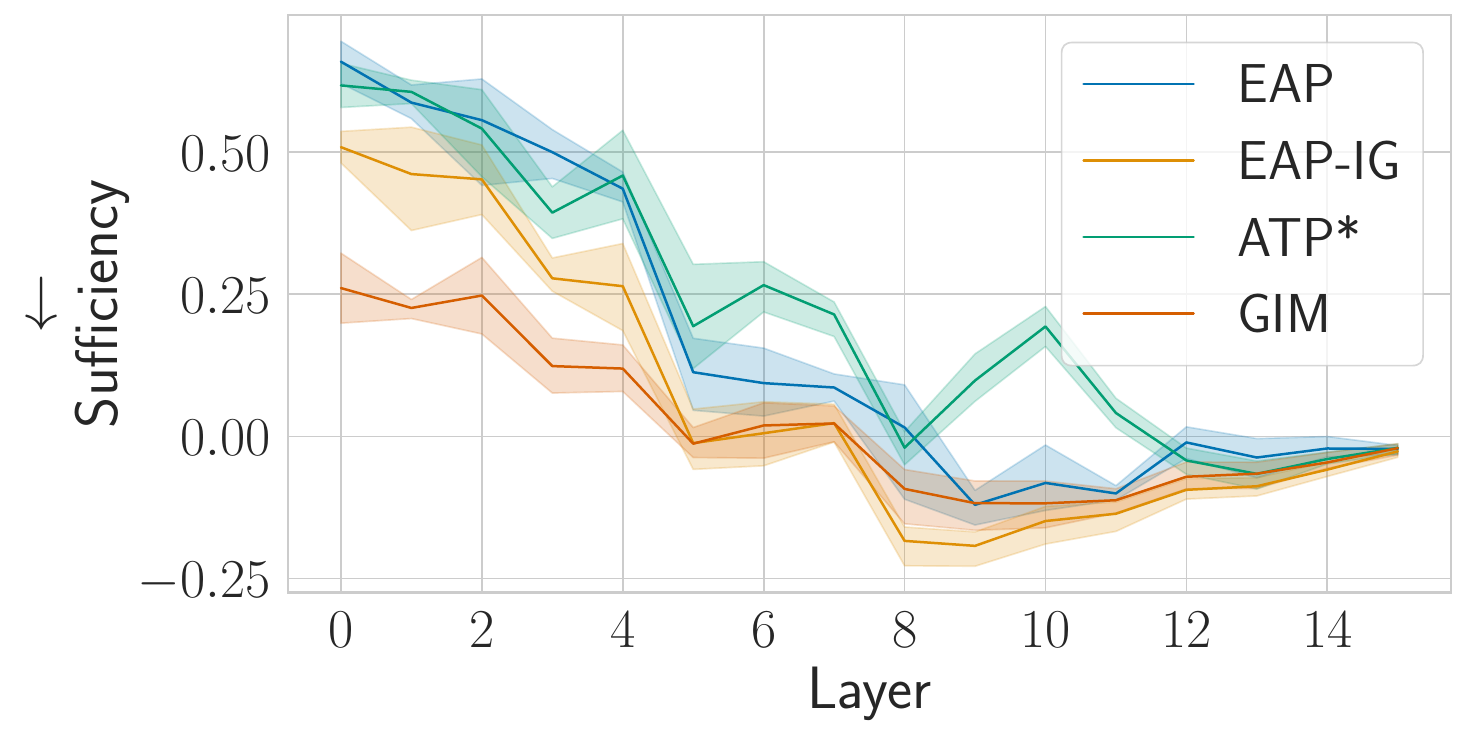}
         
         \caption{Scifact}
     \end{subfigure}
        \caption{Feature attribution methods per layer for LLAMA-3.2 1B (95\% CI). The top row depicts comprehensiveness per layer ($\uparrow$). The bottom row depicts sufficiency ($\downarrow$).}
        \label{fig:layer_aopc_llama1_2}
\end{figure*}

\begin{figure*}[h]
     \centering
     \begin{subfigure}[b]{0.25\textwidth}
         \centering
         \includegraphics[width=\textwidth]{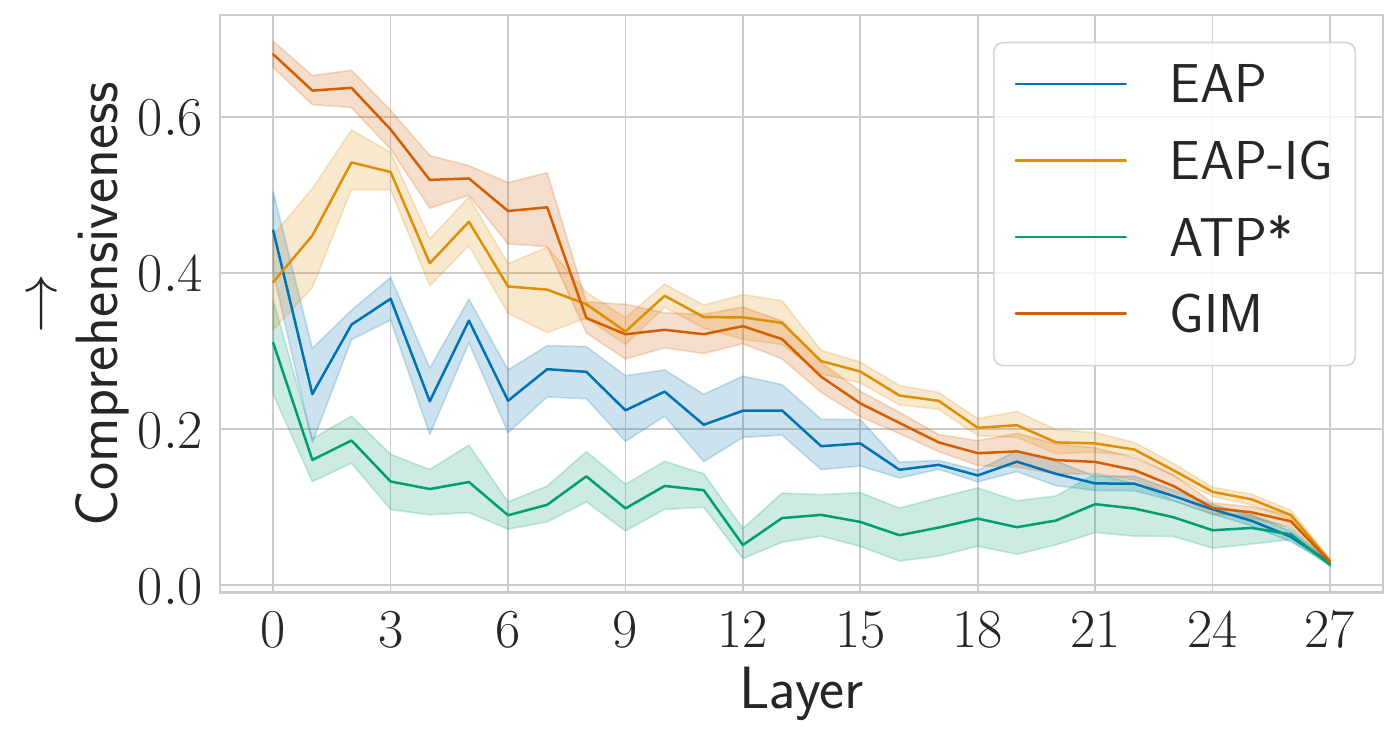}
         \includegraphics[width=\textwidth]{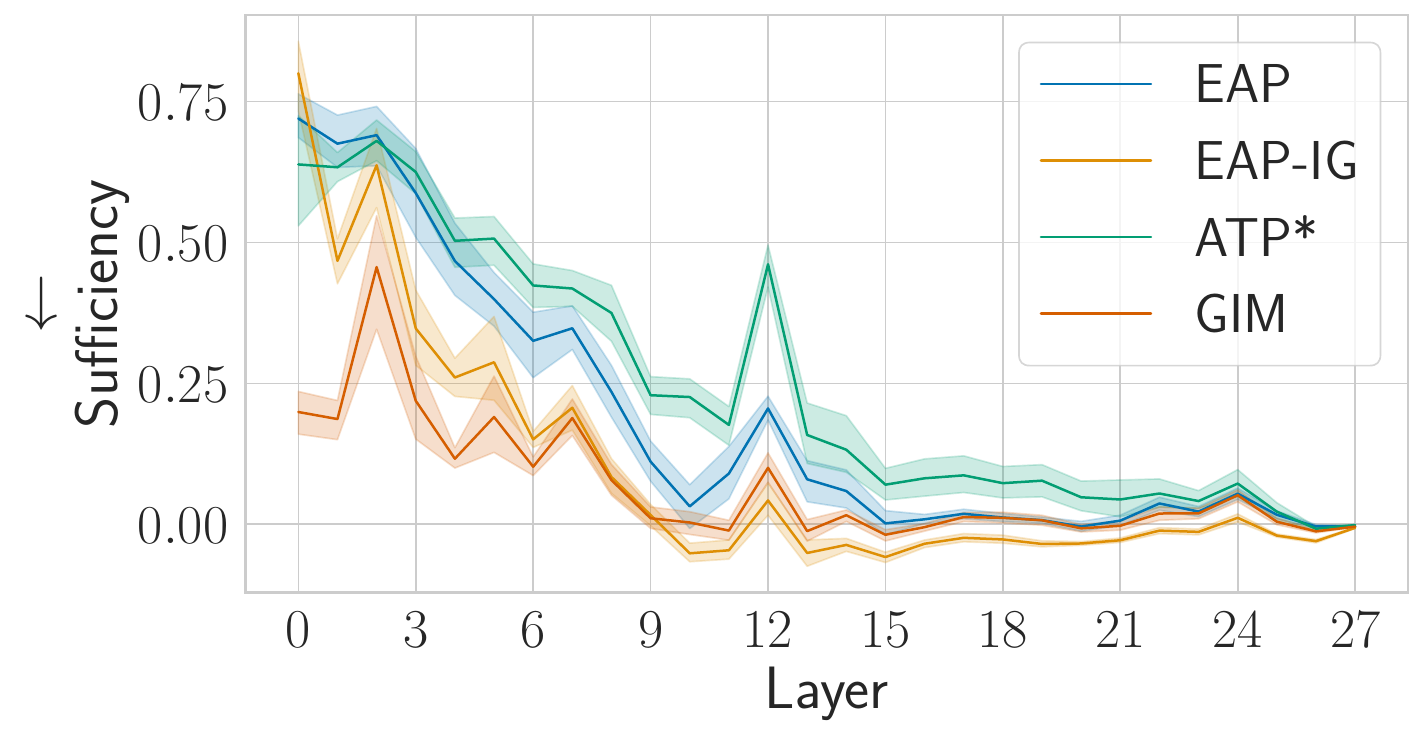}
         \caption{BoolQ}
     \end{subfigure}
     \hfill
     \begin{subfigure}[b]{0.25\textwidth}
         \centering
         \includegraphics[width=\textwidth]{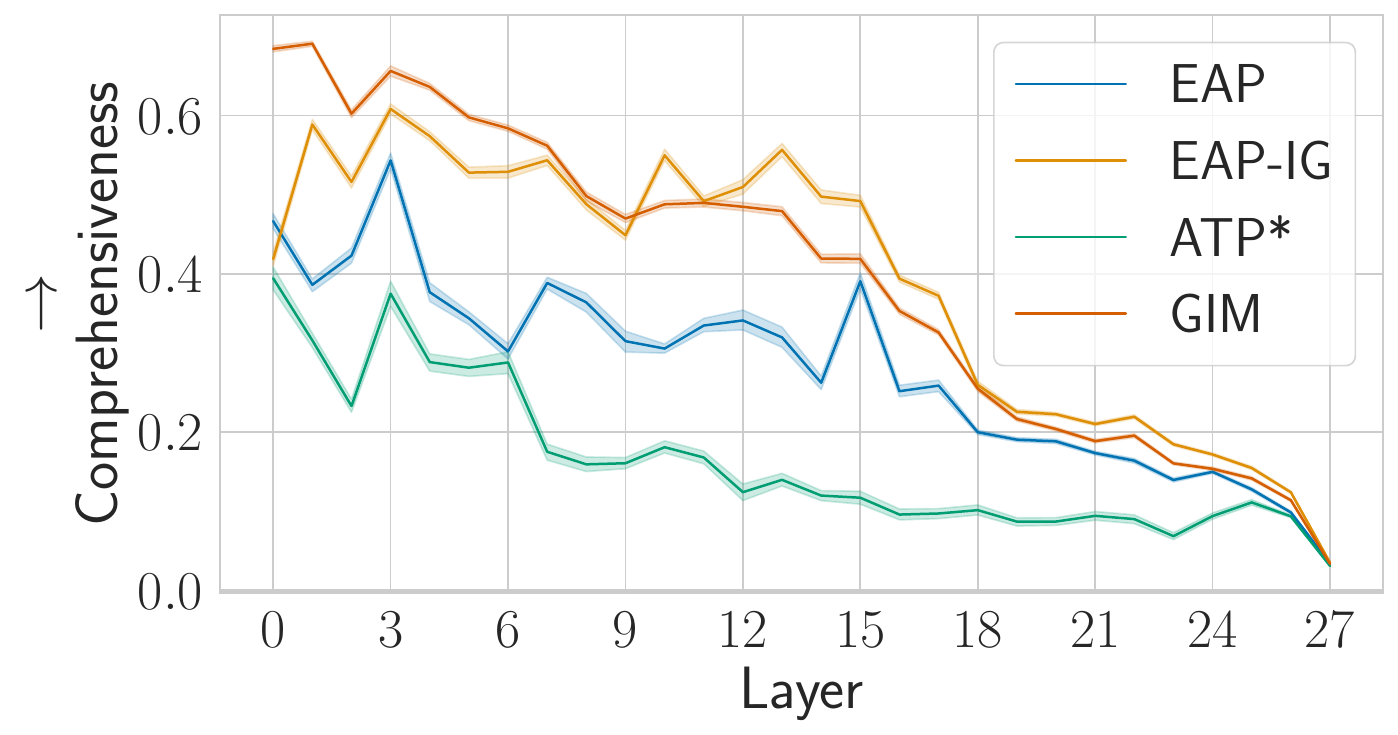}
         \includegraphics[width=\textwidth]{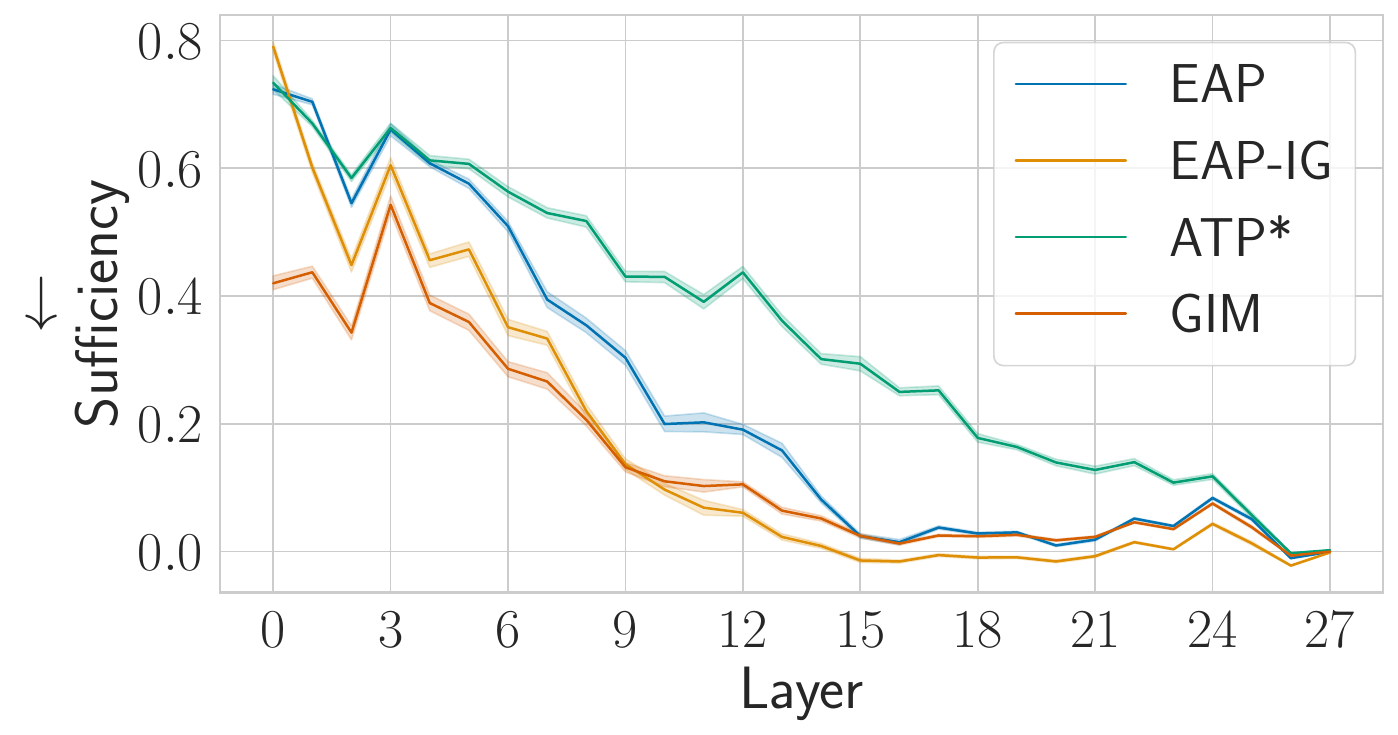}
         \caption{Fever}
     \end{subfigure}
     \hfill
     \begin{subfigure}[b]{0.25\textwidth}
         \centering
         \includegraphics[width=\textwidth]{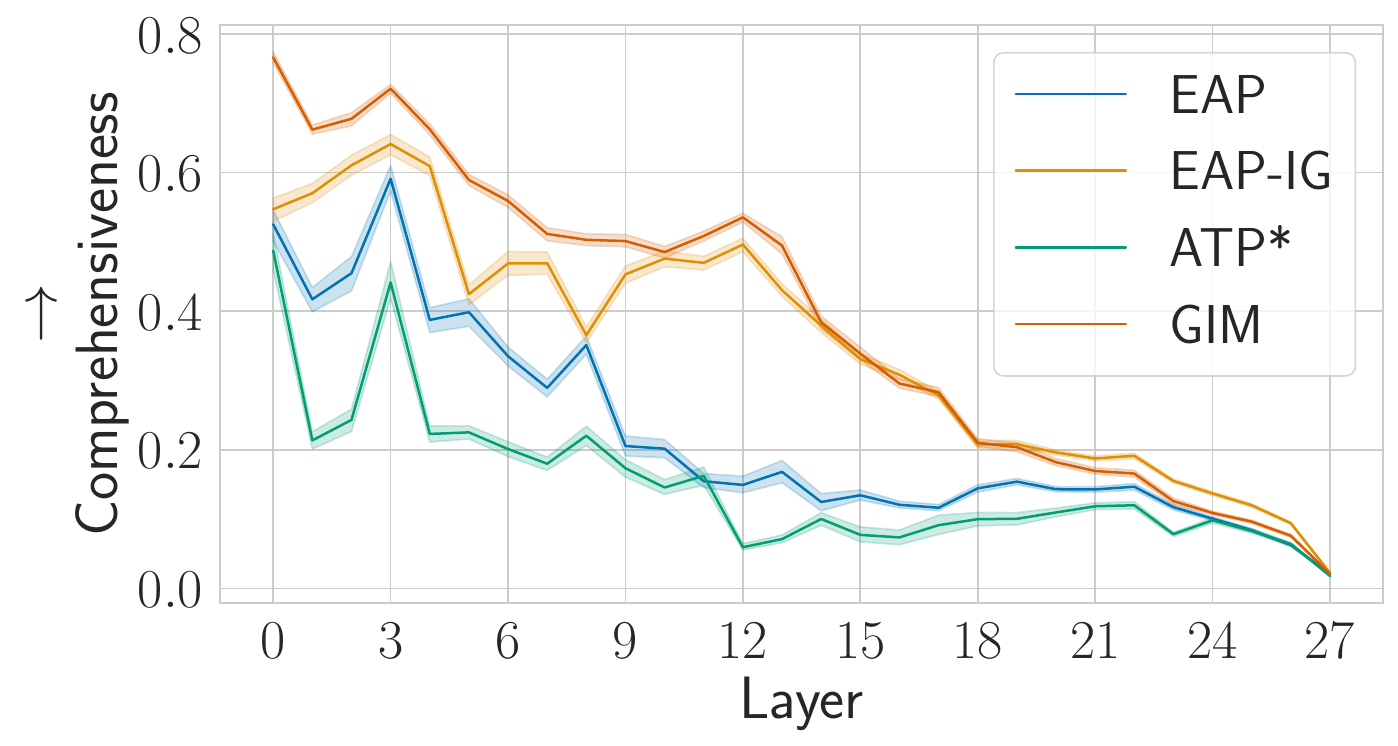}
         \includegraphics[width=\textwidth]{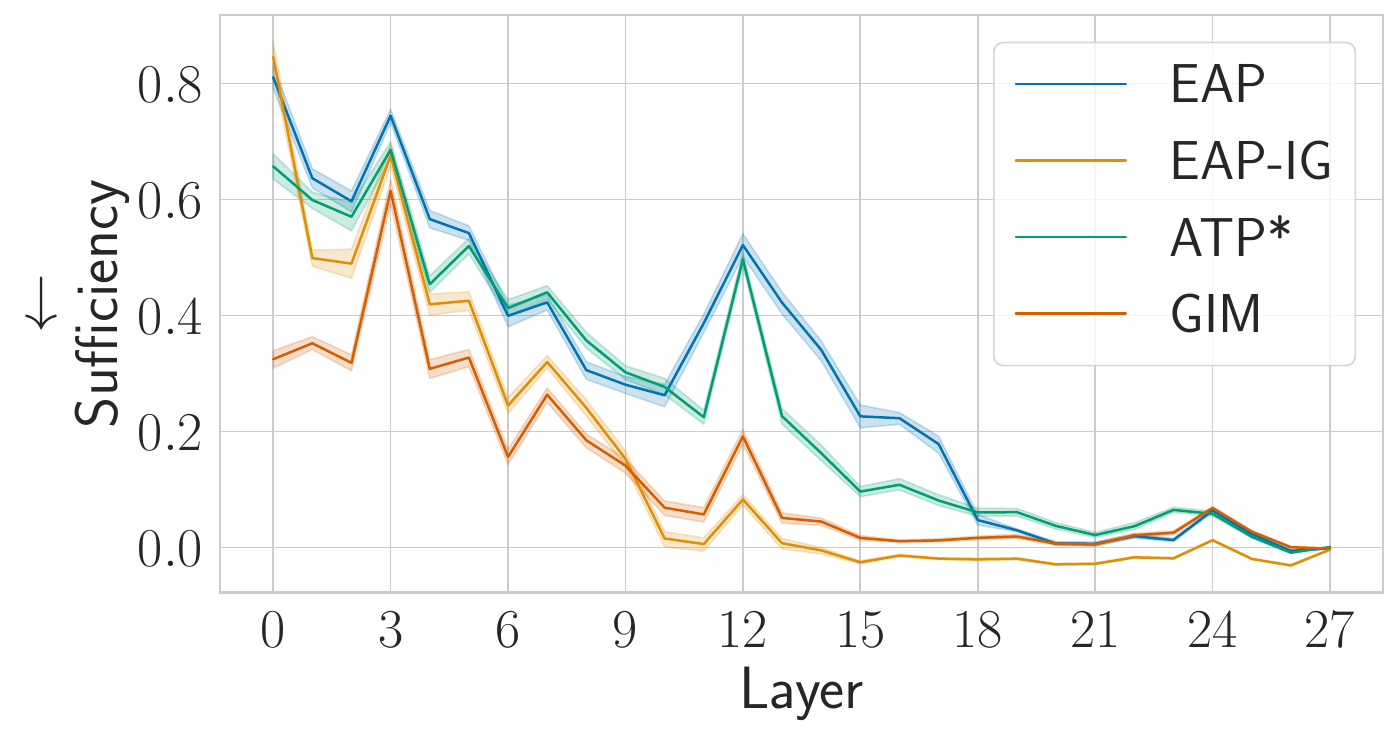}
         
         \caption{Movie review}
     \end{subfigure}
     \hfill
     \begin{subfigure}[b]{0.25\textwidth}
         \centering
         \includegraphics[width=\textwidth]{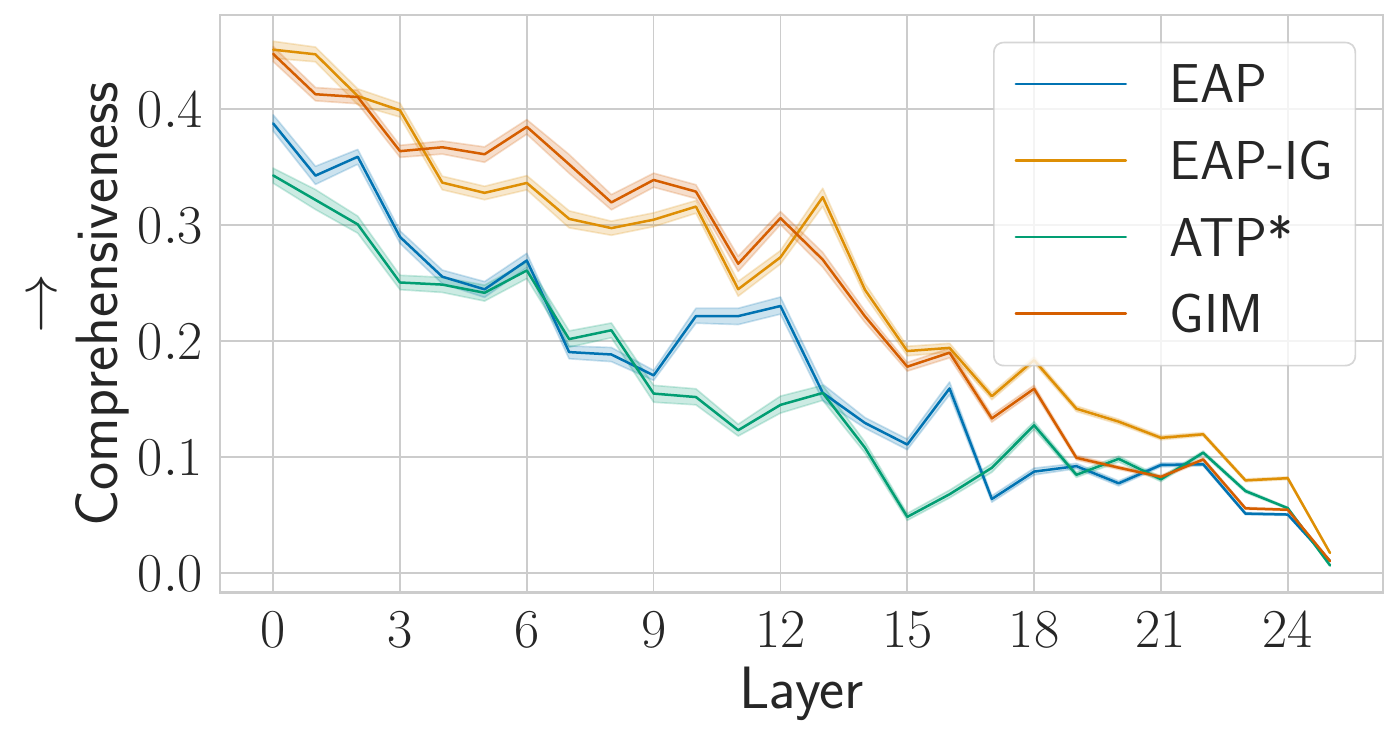}
         \includegraphics[width=\textwidth]{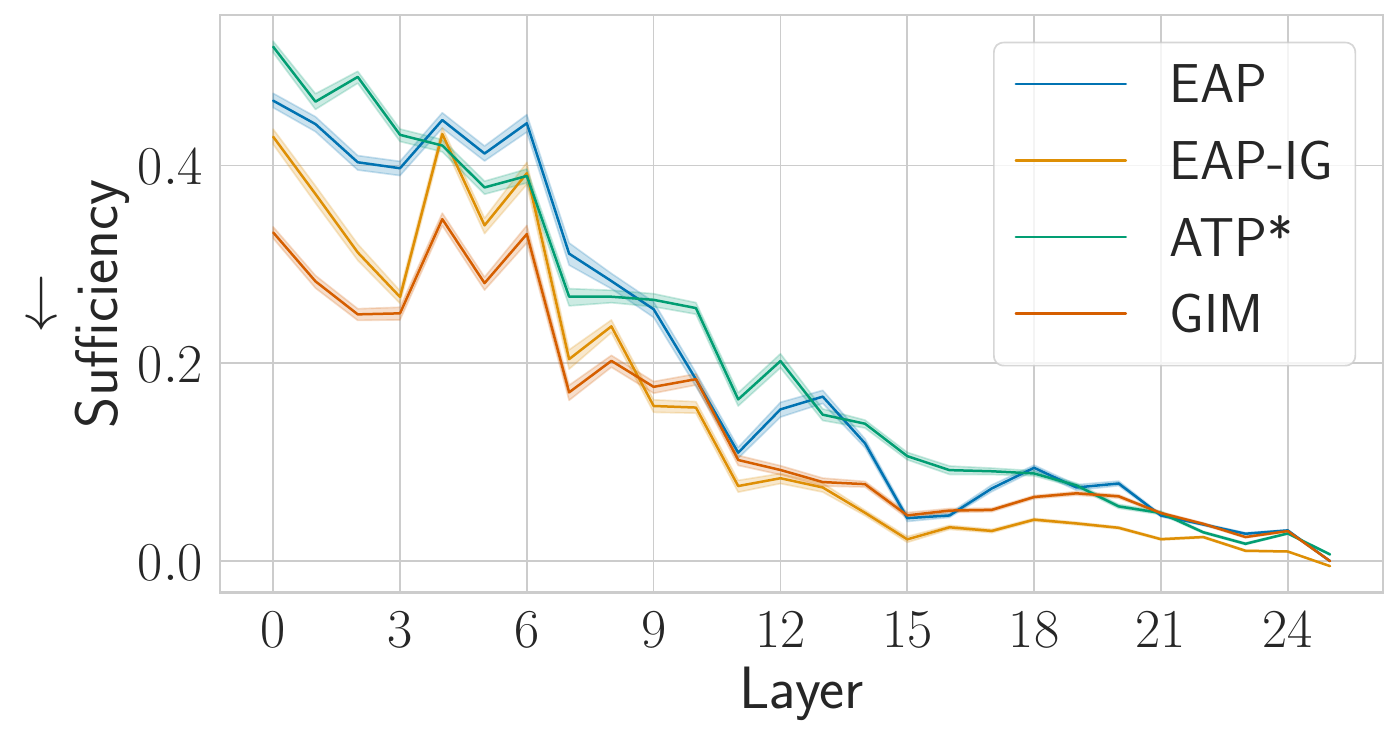}
         \caption{Twitter}
     \end{subfigure}
     \hfill
     \begin{subfigure}[b]{0.25\textwidth}
         \centering
         \includegraphics[width=\textwidth]{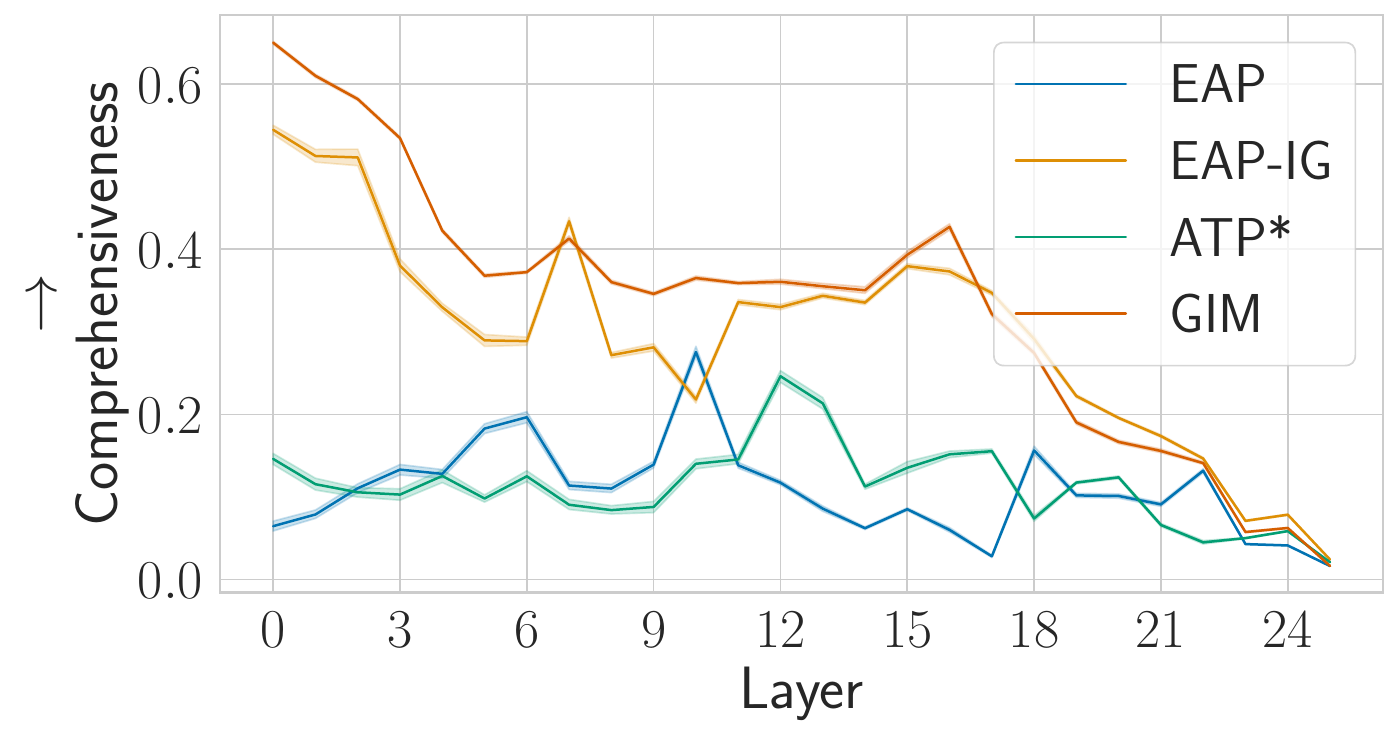}
         \includegraphics[width=\textwidth]{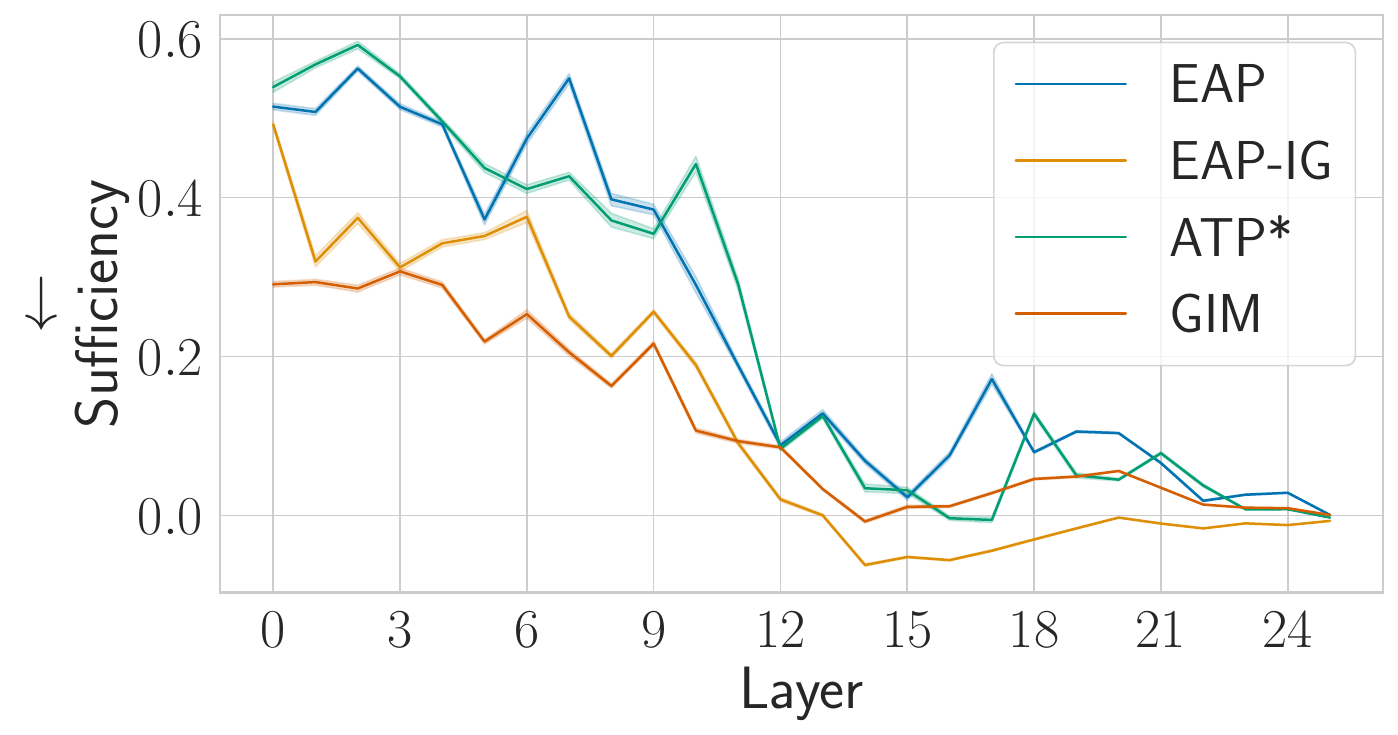}
         \caption{Hatexplain}
     \end{subfigure}
     \hfill
     \begin{subfigure}[b]{0.25\textwidth}
         \centering
         \includegraphics[width=\textwidth]{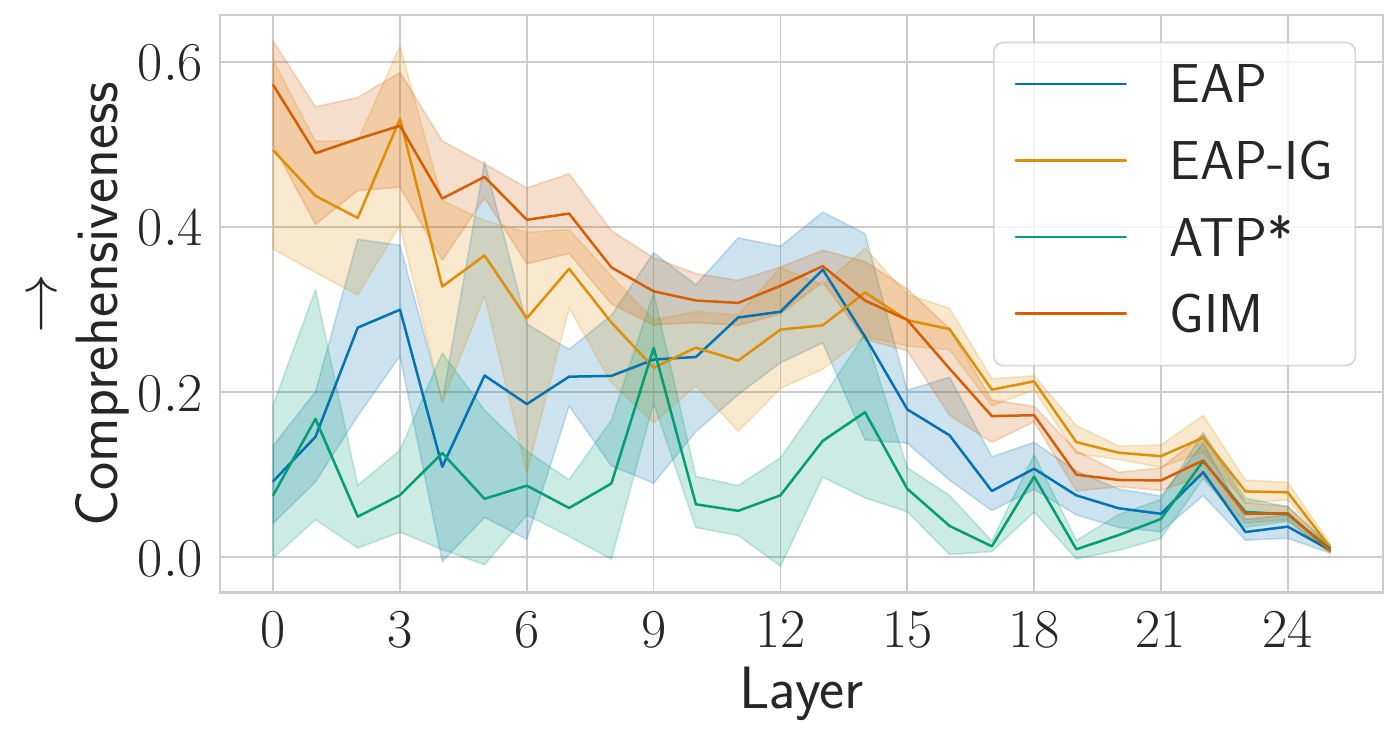}
         \includegraphics[width=\textwidth]{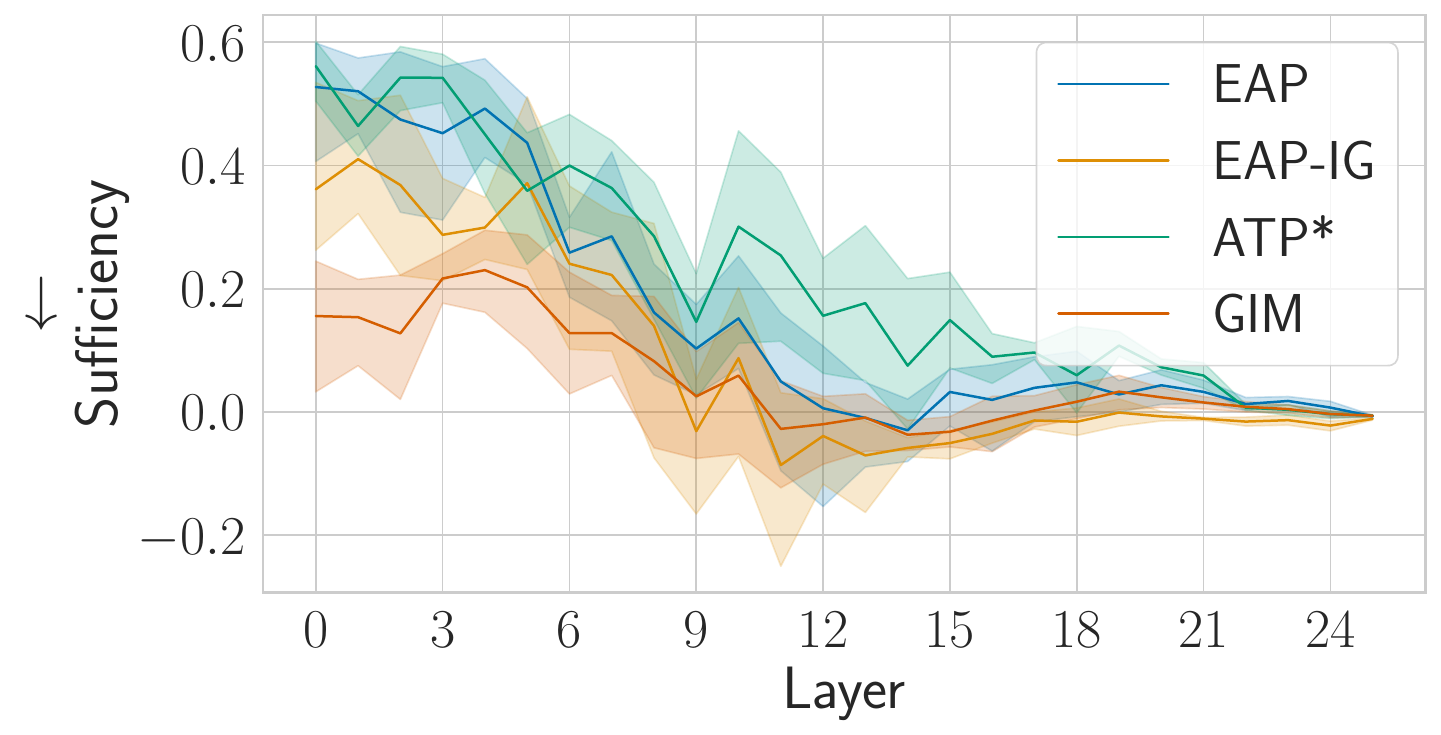}
         
         \caption{Scifact}
     \end{subfigure}
        \caption{Faithfulness per layer for \textbf{Gemma-2 2B}. The top row depicts comprehensiveness per layer, where higher is better. The bottom row depicts sufficiency, where lower is better.}
        \label{fig:layer_aopc_gemma2}
\end{figure*}

\begin{figure*}[h]
     \centering
     \begin{subfigure}[b]{0.25\textwidth}
         \centering
         \includegraphics[width=\textwidth]{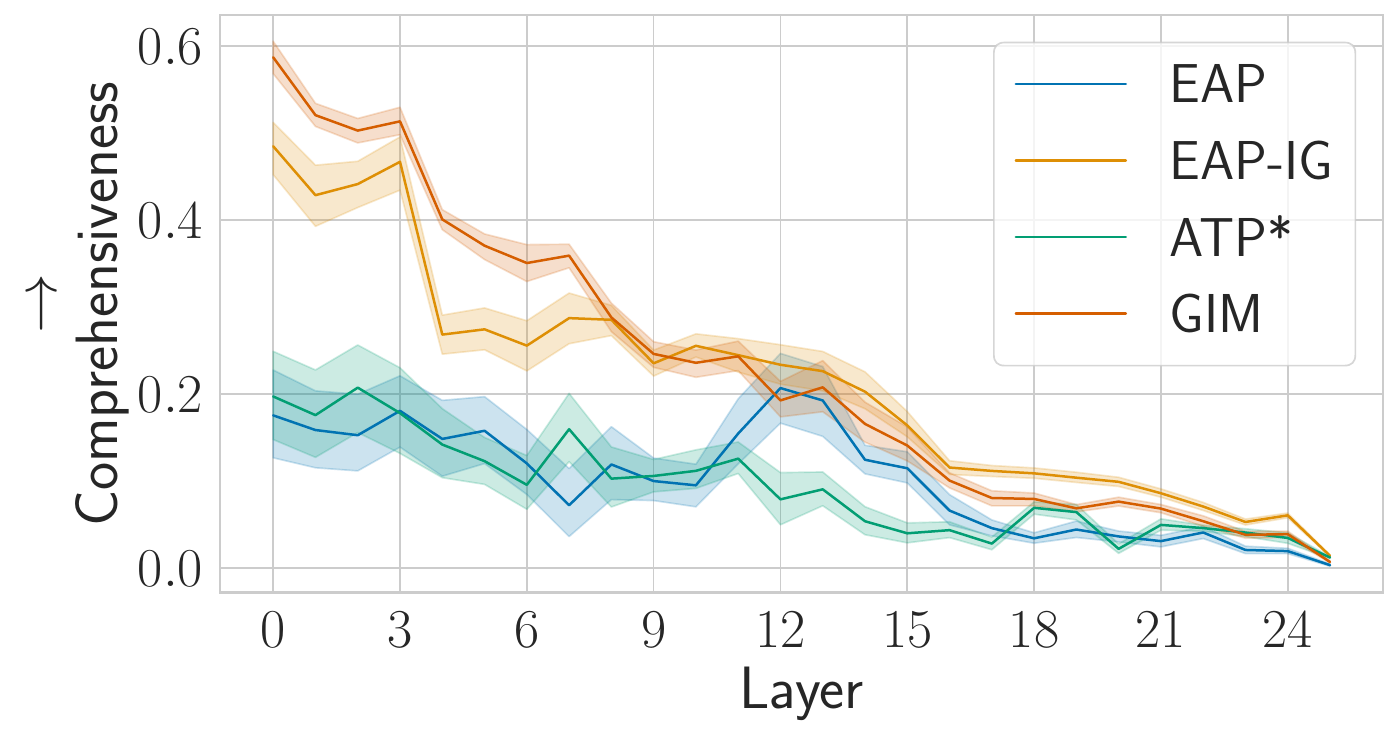}
         \includegraphics[width=\textwidth]{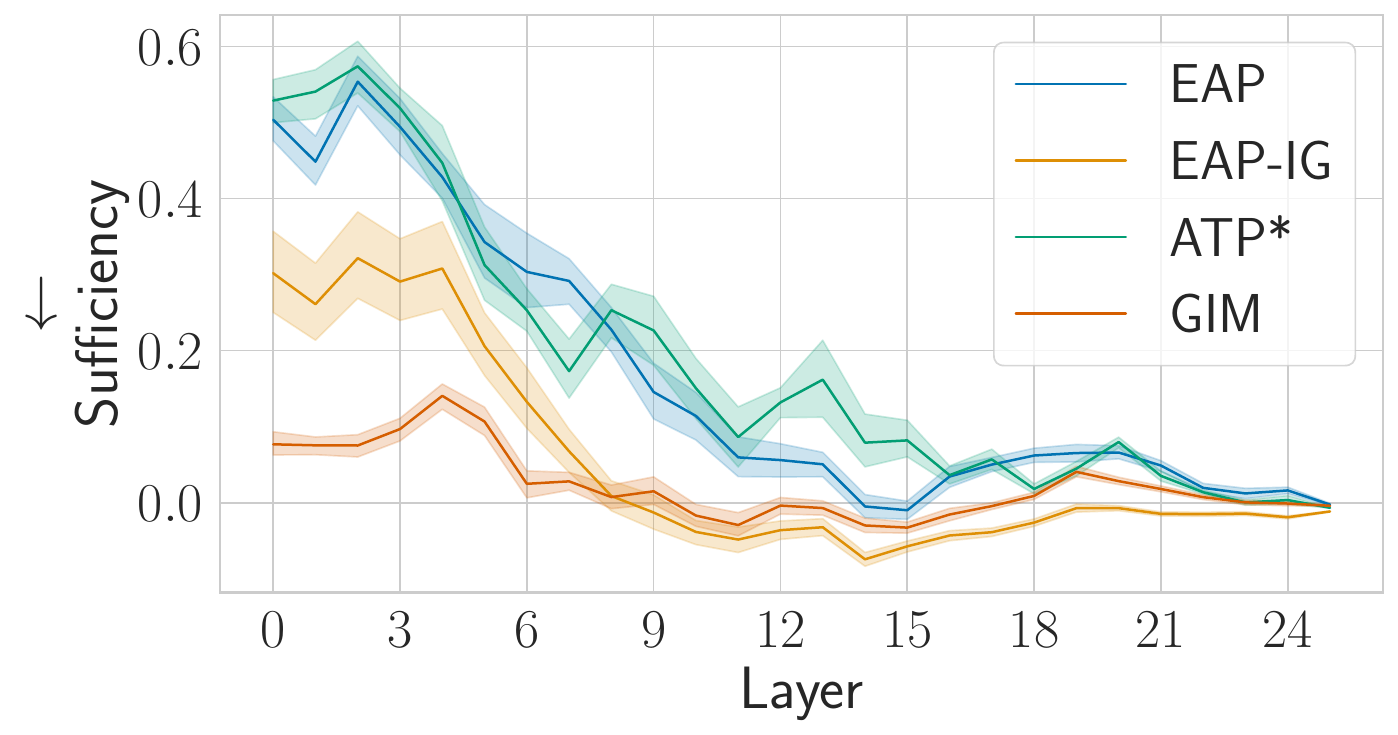}
         \caption{BoolQ}
     \end{subfigure}
     \hfill
     \begin{subfigure}[b]{0.25\textwidth}
         \centering
         \includegraphics[width=\textwidth]{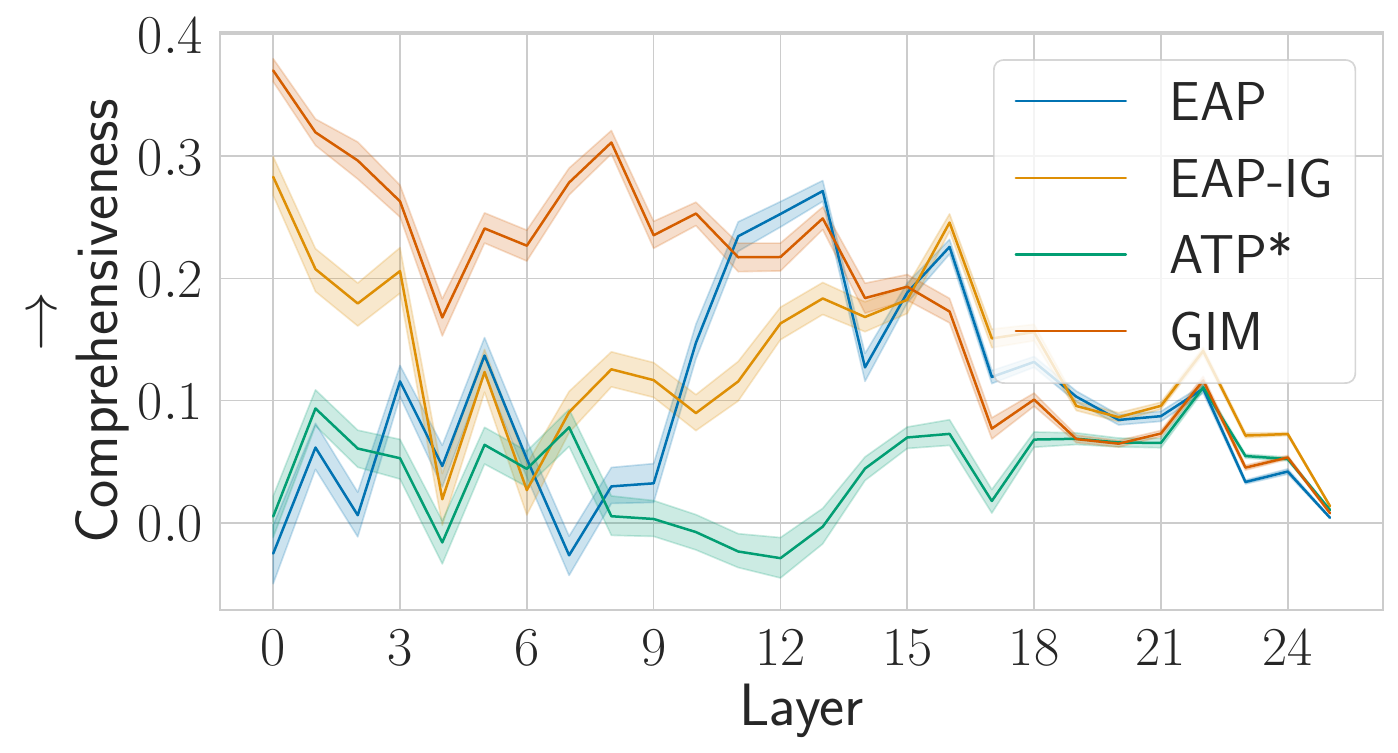}
         \includegraphics[width=\textwidth]{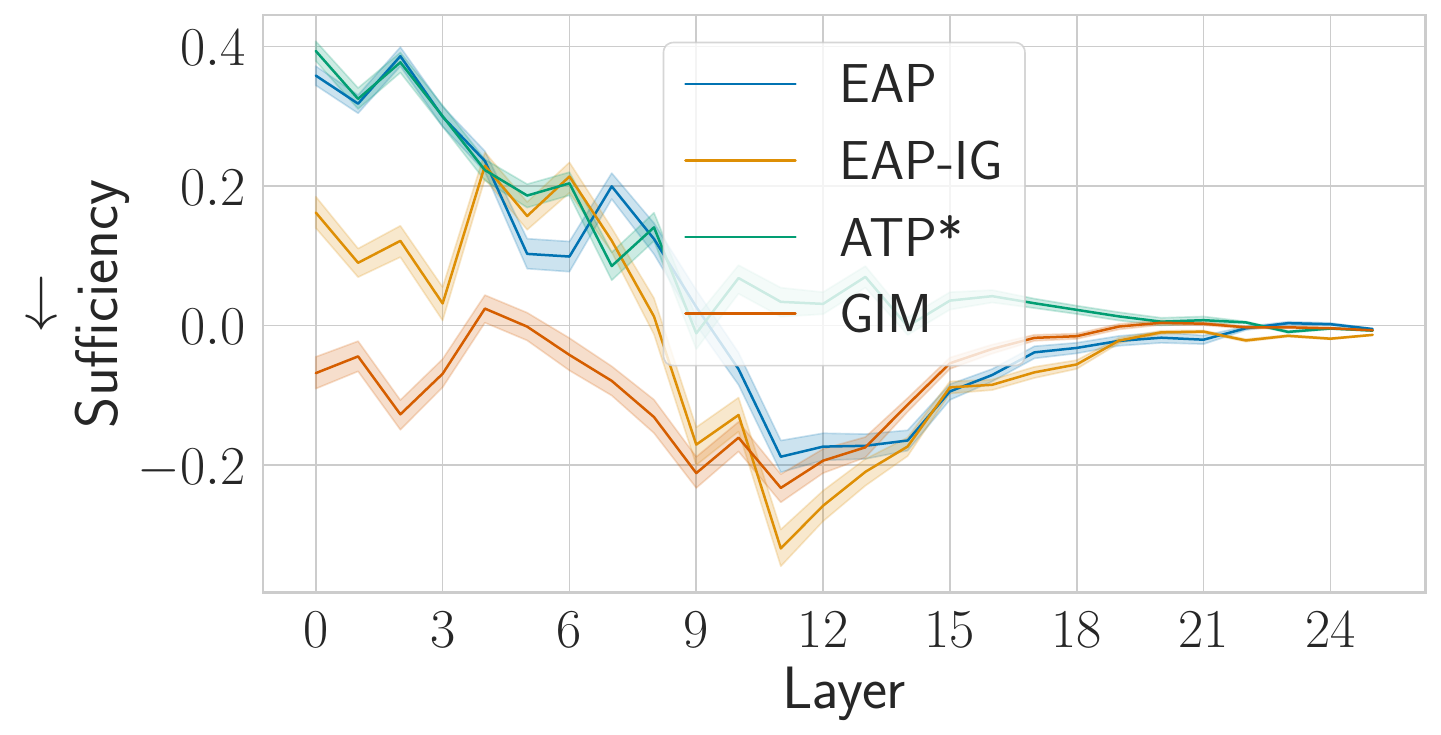}
         \caption{Fever}
     \end{subfigure}
     \hfill
     \begin{subfigure}[b]{0.25\textwidth}
         \centering
         \includegraphics[width=\textwidth]{figures/layer_aopc/movie/llama3-3b-it/layer_aopc_comprehensiveness.pdf}
         \includegraphics[width=\textwidth]{figures/layer_aopc/movie/llama3-3b-it/layer_aopc_sufficiency.pdf}
         
         \caption{Movie review}
    \end{subfigure}
    \hfill
    \begin{subfigure}[b]{0.25\textwidth}
         \centering
         \includegraphics[width=\textwidth]{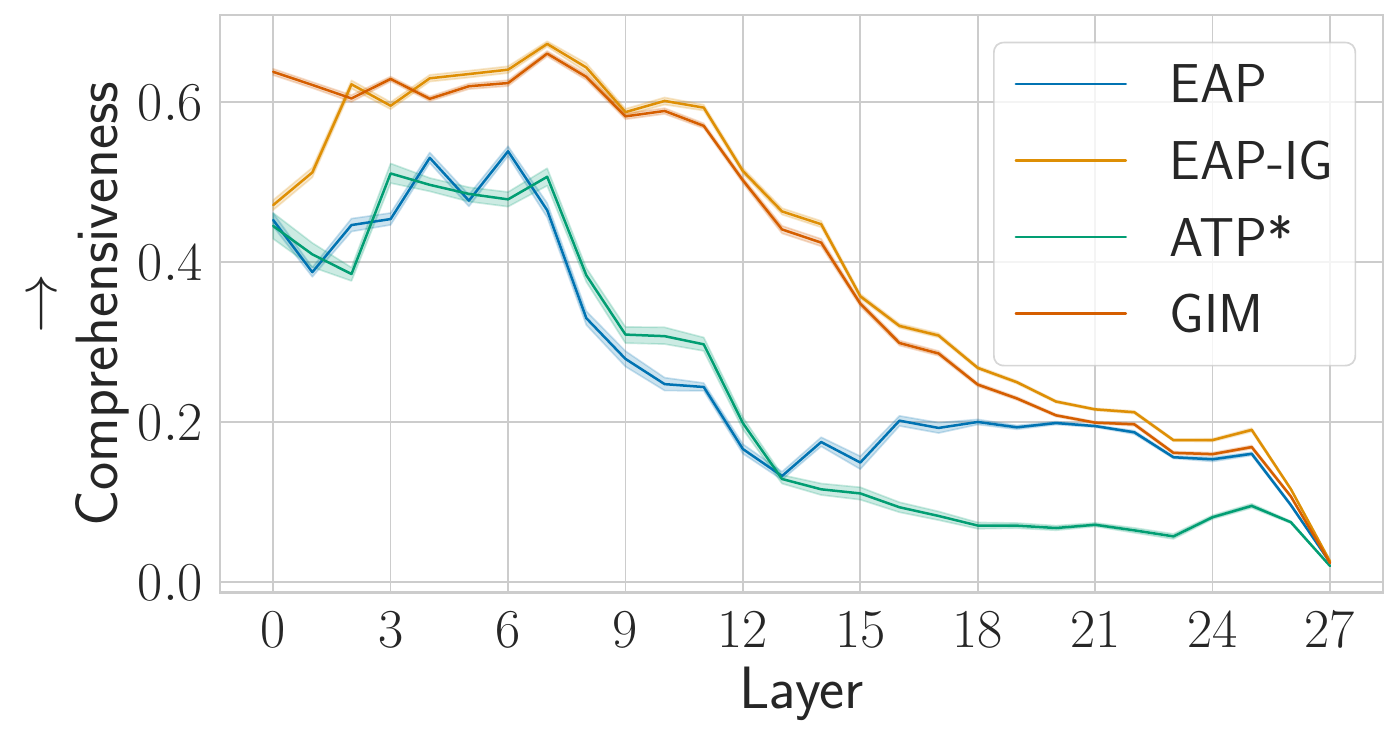}
         \includegraphics[width=\textwidth]{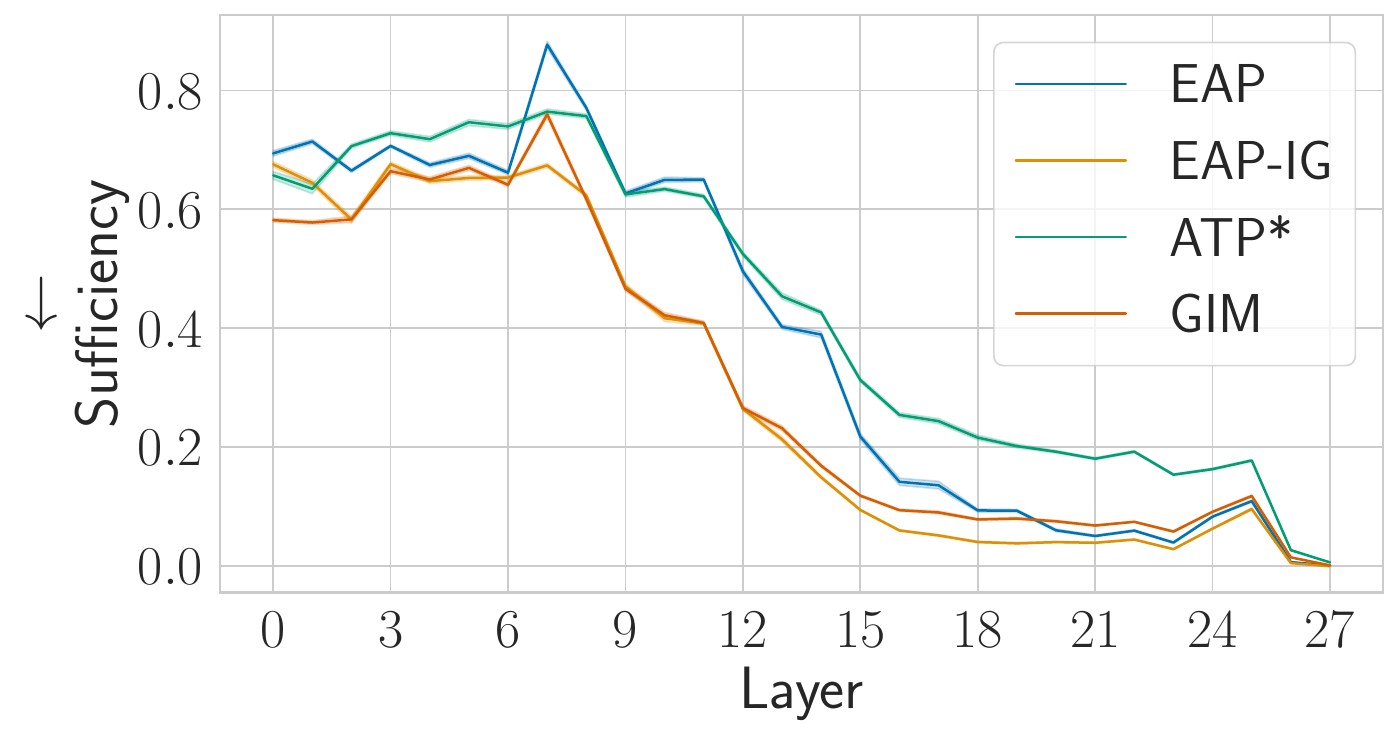}
         \caption{Twitter}
     \end{subfigure}
     \hfill
     \begin{subfigure}[b]{0.25\textwidth}
         \centering
         \includegraphics[width=\textwidth]{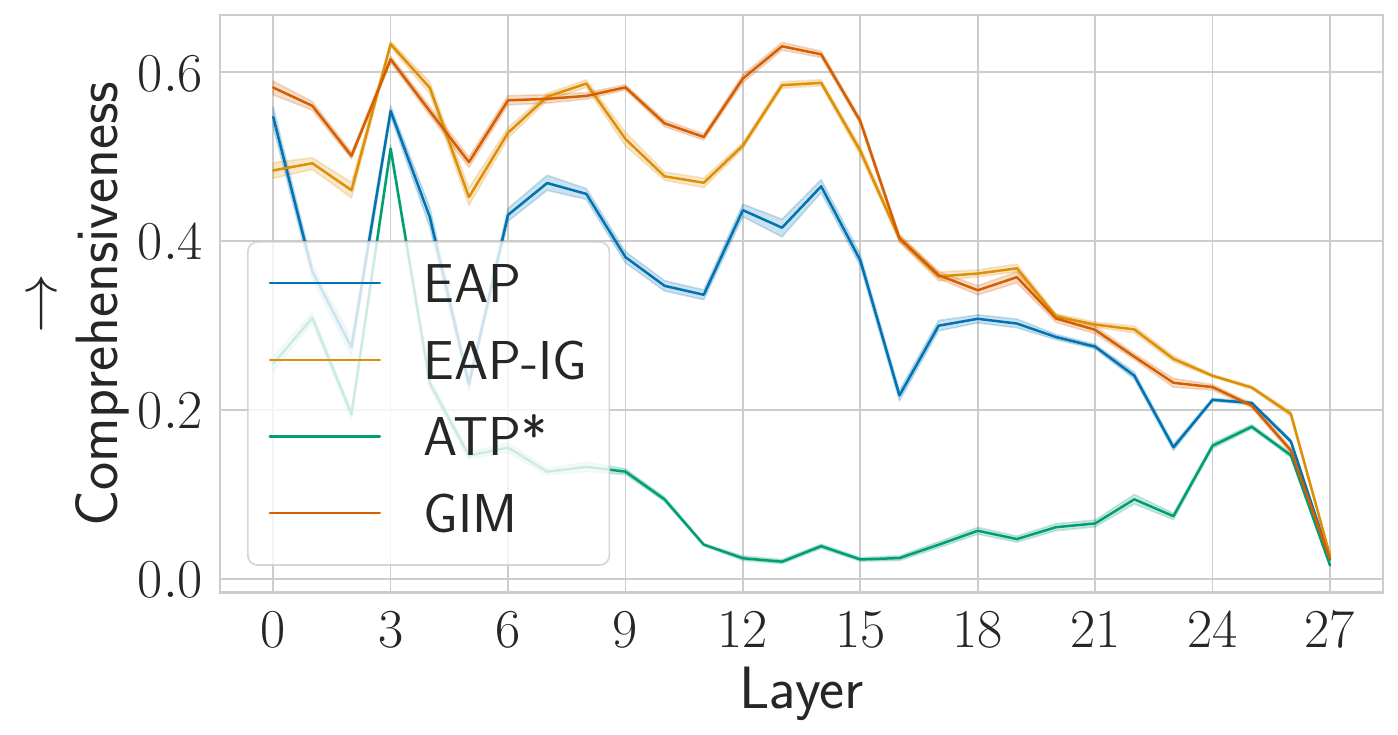}
         \includegraphics[width=\textwidth]{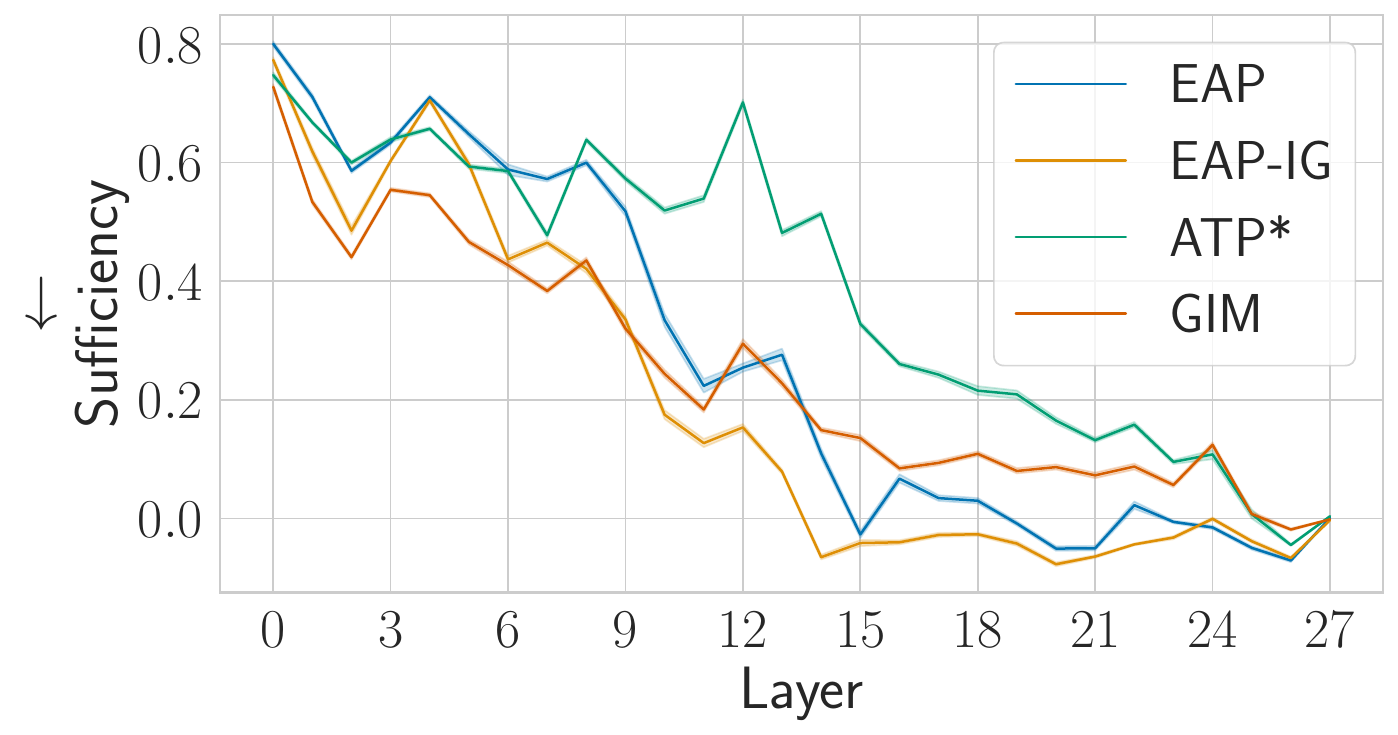}
         \caption{Hatexplain}
     \end{subfigure}
     \hfill
     \begin{subfigure}[b]{0.25\textwidth}
         \centering
         \includegraphics[width=\textwidth]{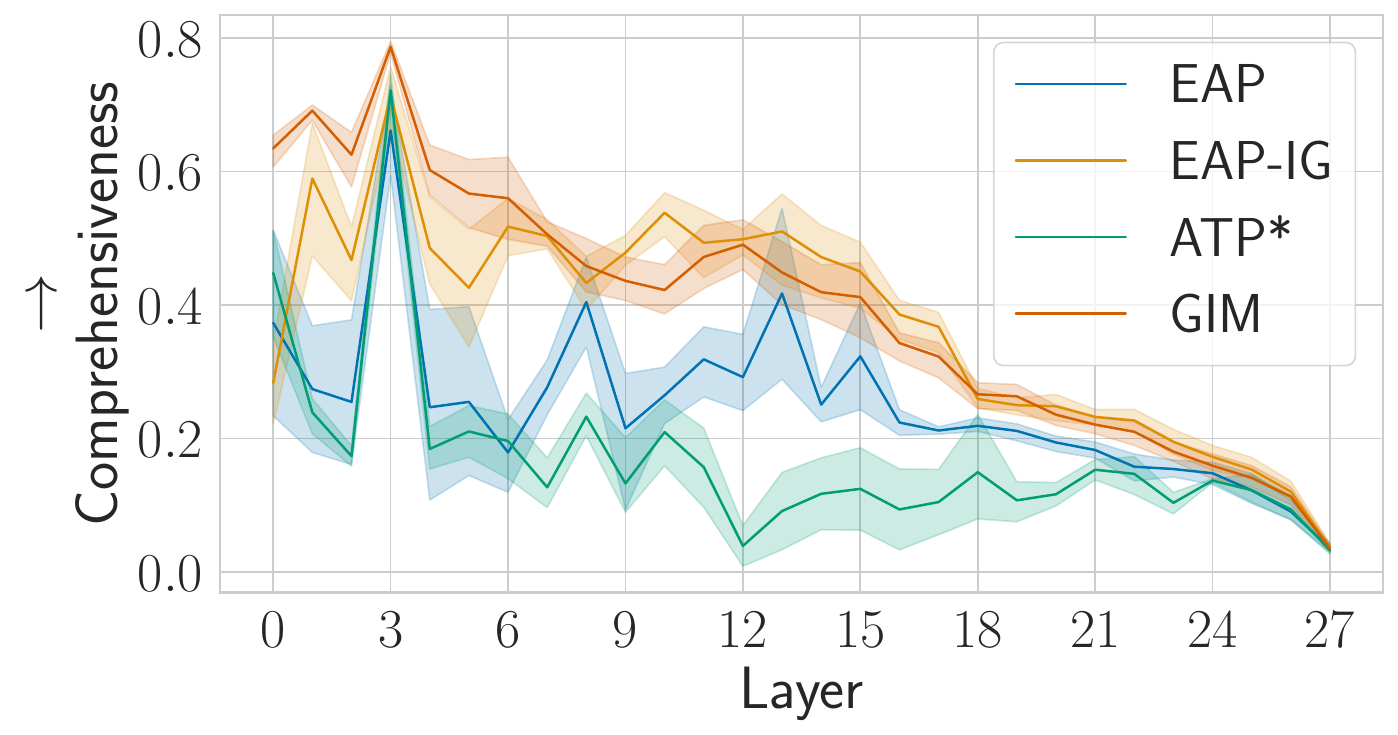}
         \includegraphics[width=\textwidth]{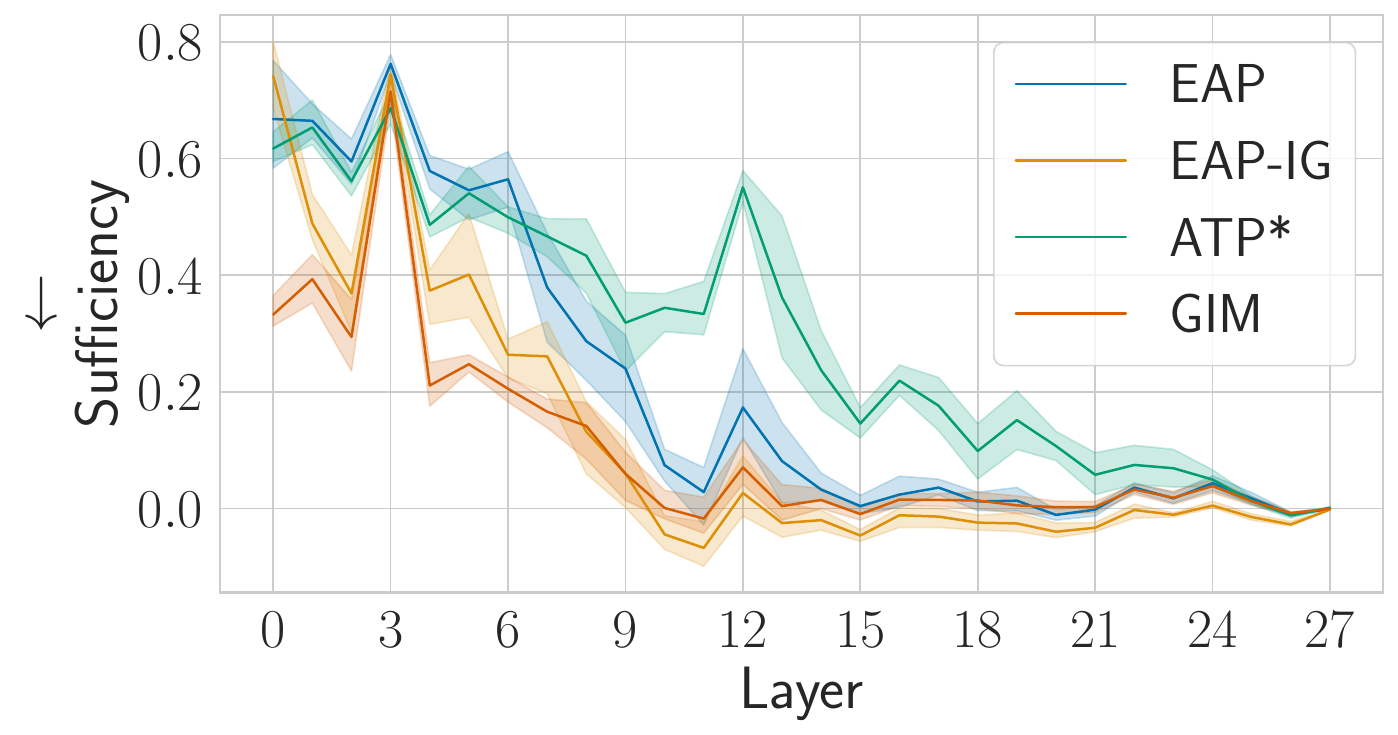}
         
         \caption{Scifact}
     \end{subfigure}
        \caption{Faithfulness per layer for \textbf{LLAMA-3.2 3B}. The top row depicts comprehensiveness per layer, where higher is better. The bottom row depicts sufficiency, where lower is better.}
        \label{fig:layer_aopc_llama3}
\end{figure*}

\Cref{fig:layer_aopc_llama1_2}, \Cref{fig:layer_aopc_gemma2}, and \Cref{fig:layer_aopc_llama3} shows comprehensiveness and sufficiency metrics across layers for LLAMA-3.2 1B, Gemma-2 2B, and LLAMA-3.2 3B across all datasets. GIM significantly outperforms other circuit localization methods in the early layers, while EAP-IG shows slight advantages in the middle and late layers. \citet{csordasLanguageModelsUse2025a} showed that LLMs primarily use the attention heads in the early layers. Since most of our modifications target the attention heads, this could explain the pattern. Across all methods, the scores worsen in the top layers of the LLM. This could either be because they become less faithful or because the upper and lower limits of comprehensiveness and sufficiency change across layers~\citep{edinNormalizedAOPCFixing2025}.

\begin{figure*}[h]
     \centering
     \begin{subfigure}[b]{0.25\textwidth}
         \centering
         \includegraphics[width=\textwidth]{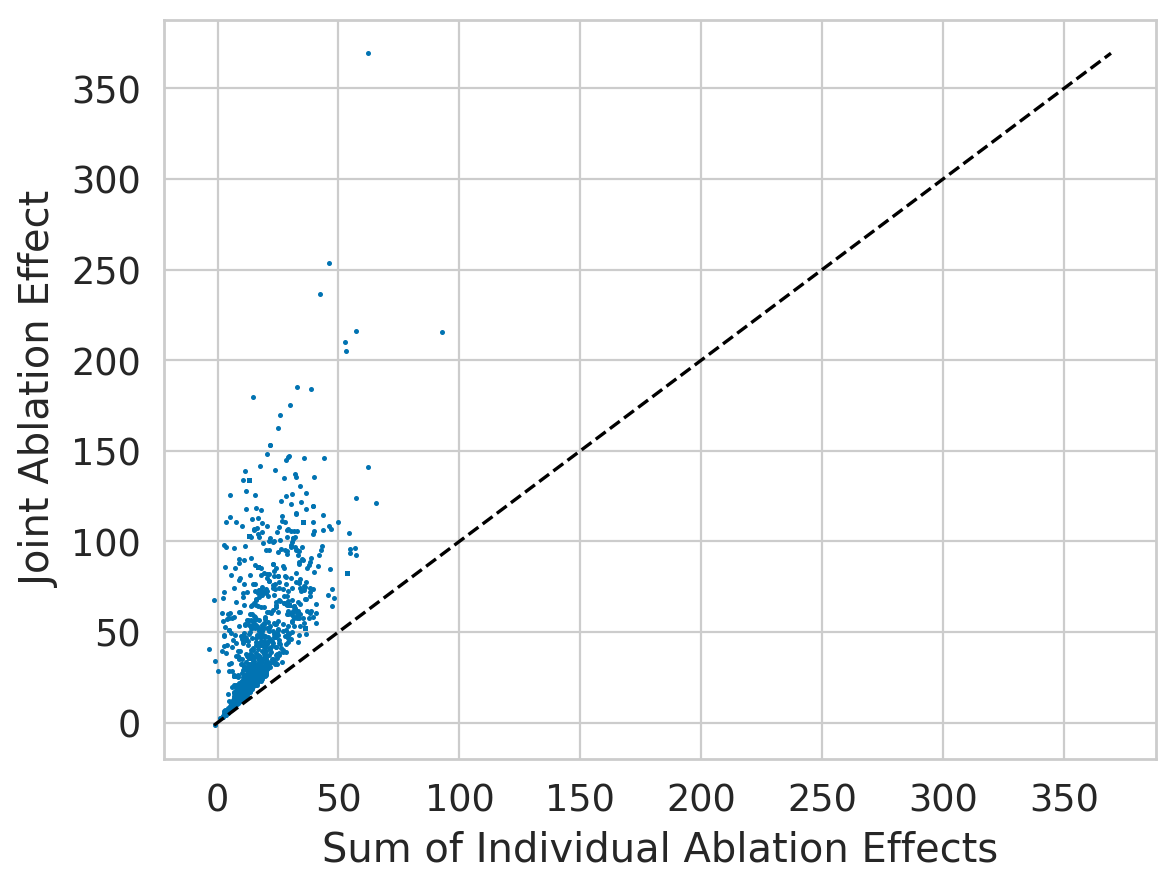}
         \includegraphics[width=\textwidth]{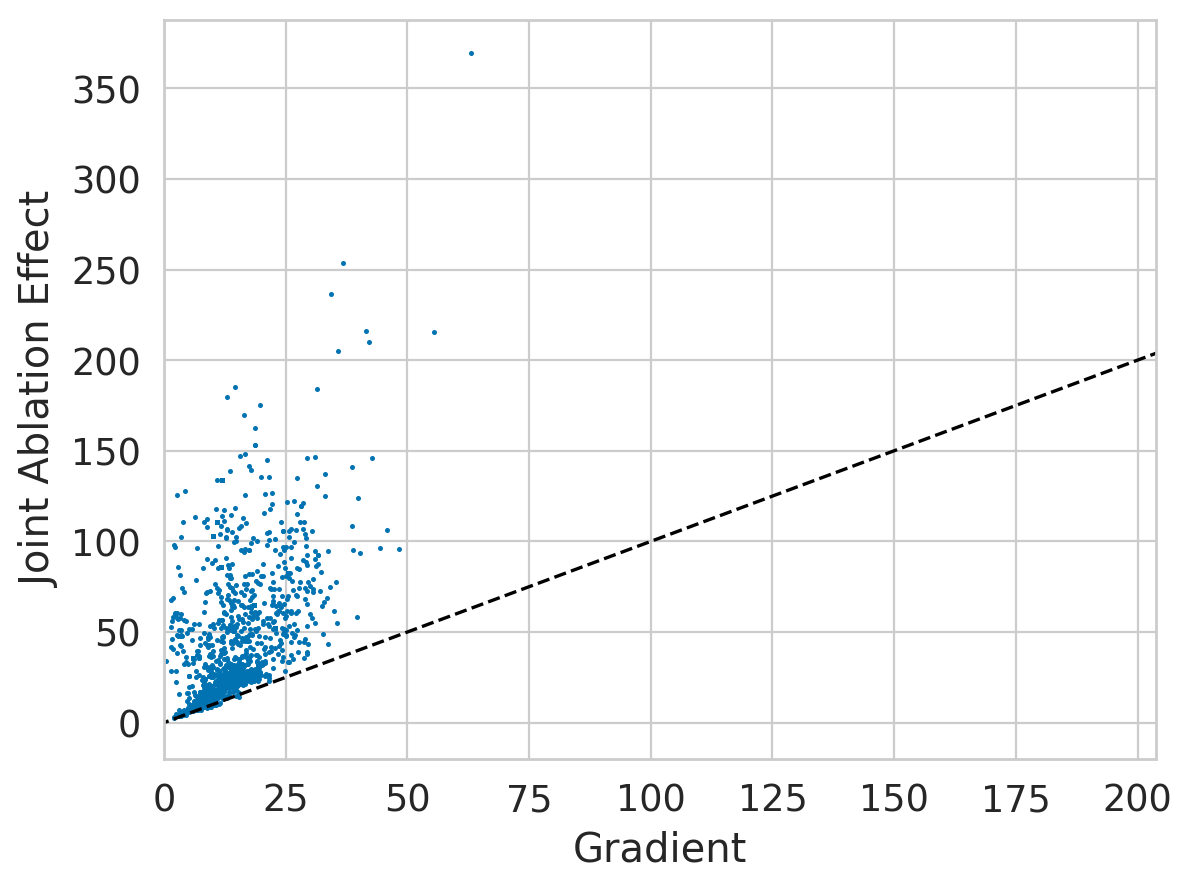}
         \includegraphics[width=\textwidth]{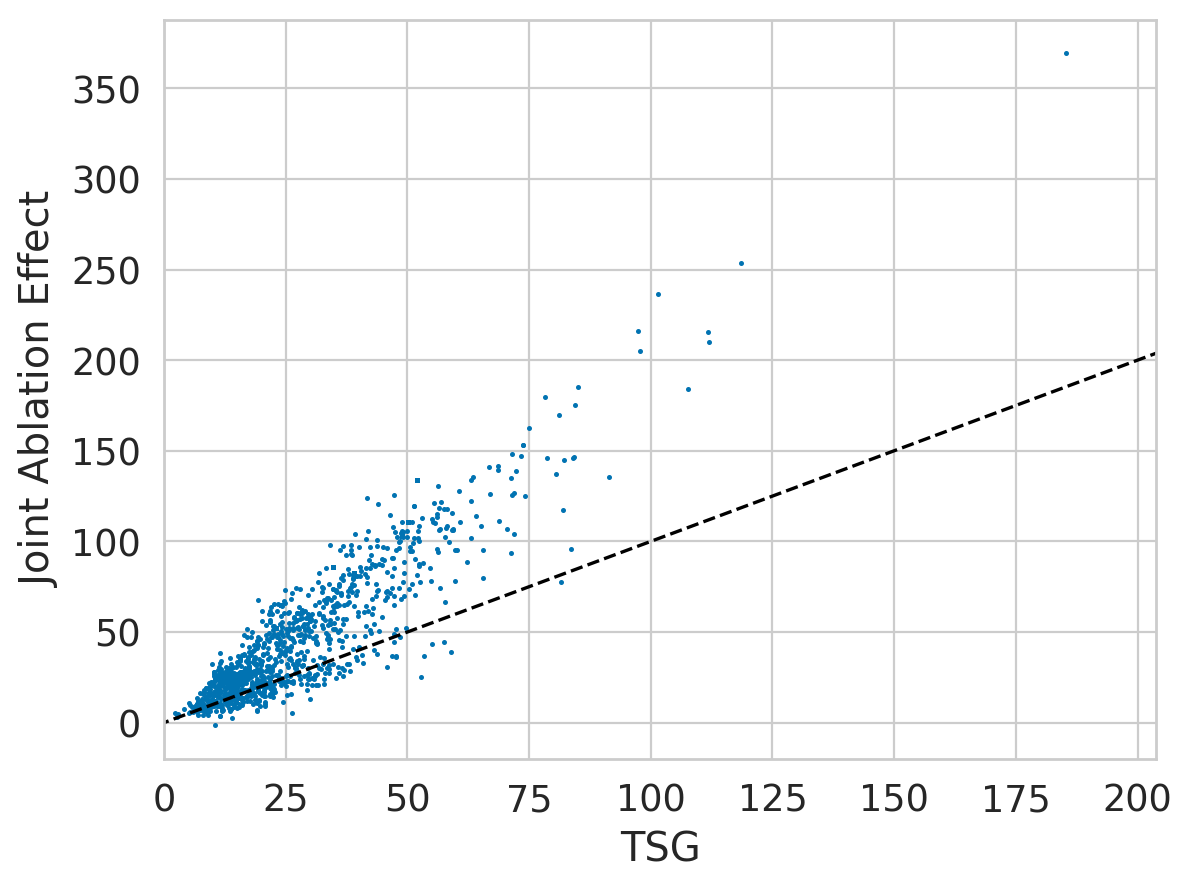}
         \caption{Twitter}
     \end{subfigure}
     \hfill
     \begin{subfigure}[b]{0.25\textwidth}
         \centering
         \includegraphics[width=\textwidth]{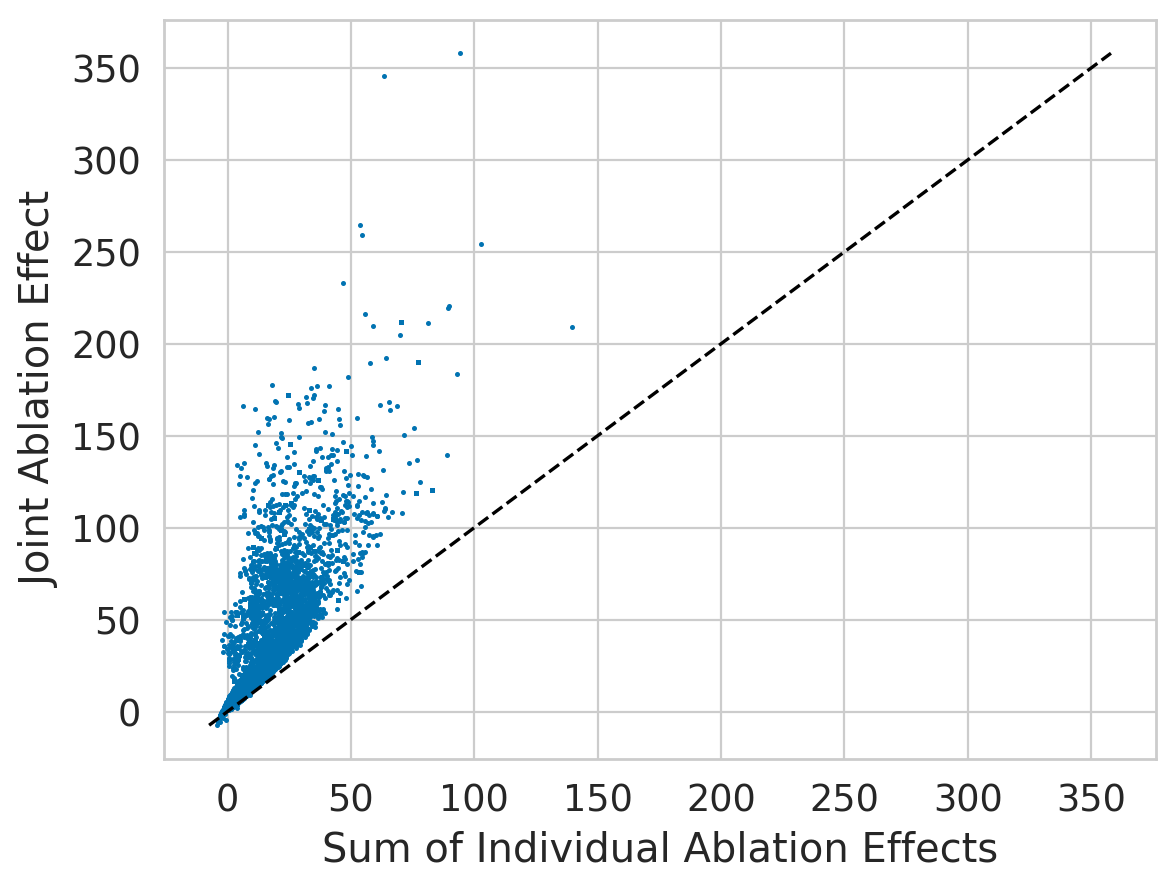}
         \includegraphics[width=\textwidth]{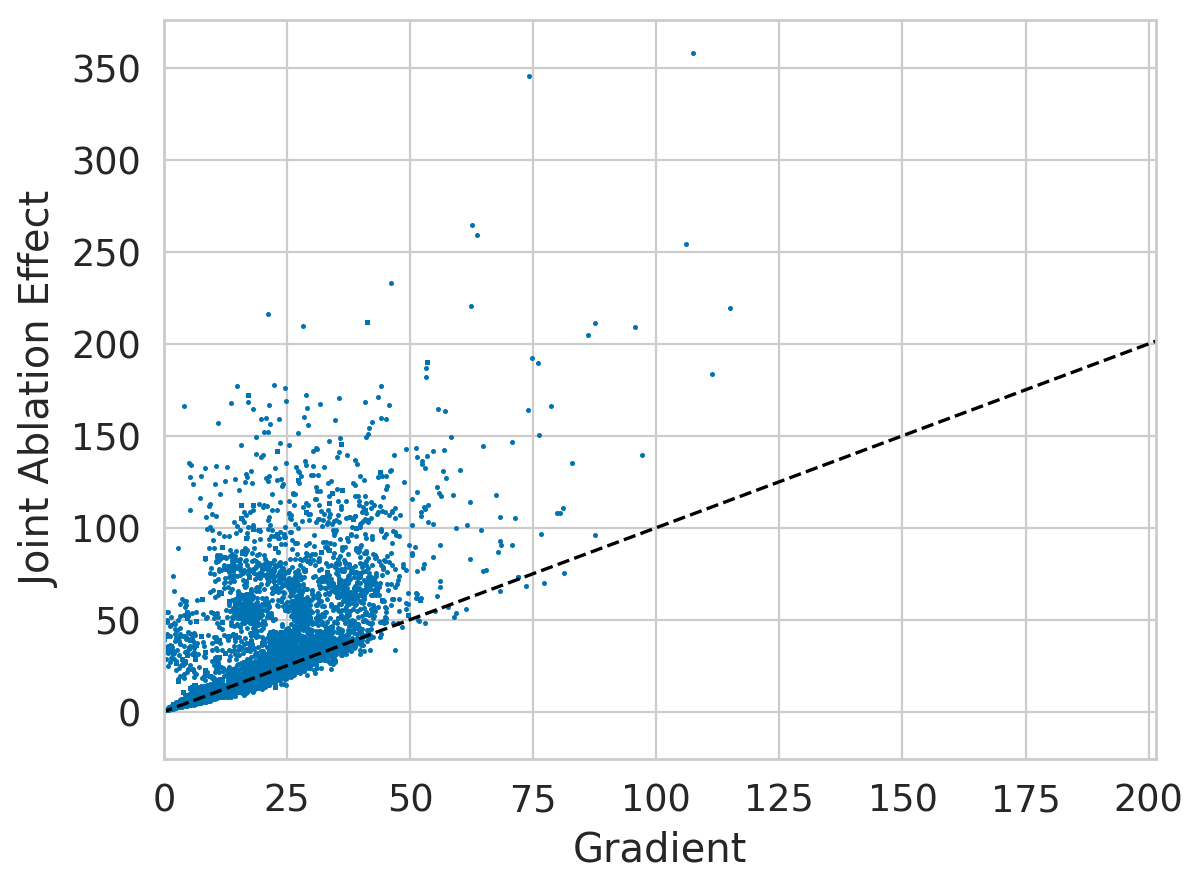}
         \includegraphics[width=\textwidth]{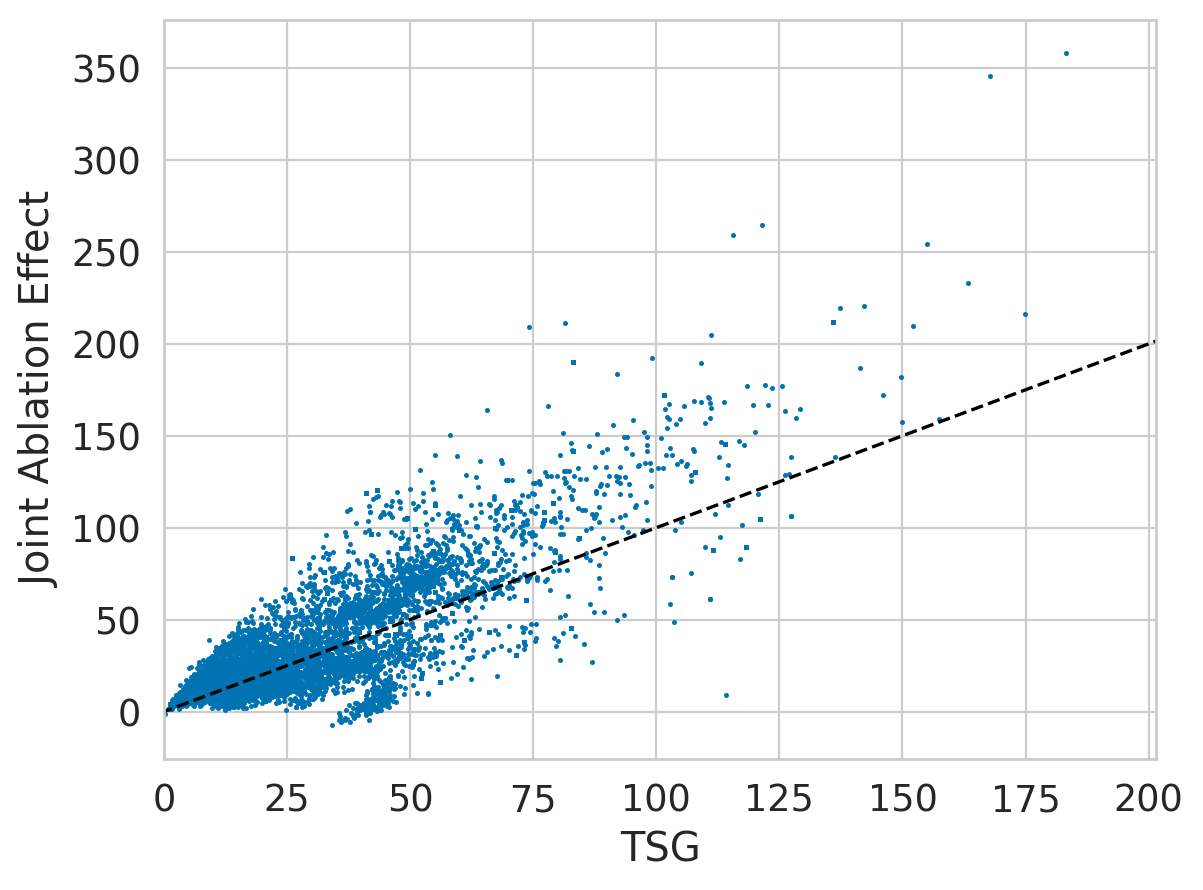}
         \caption{Hatexplain}
     \end{subfigure}
     \hfill
     \begin{subfigure}[b]{0.25\textwidth}
         \centering
         \includegraphics[width=\textwidth]{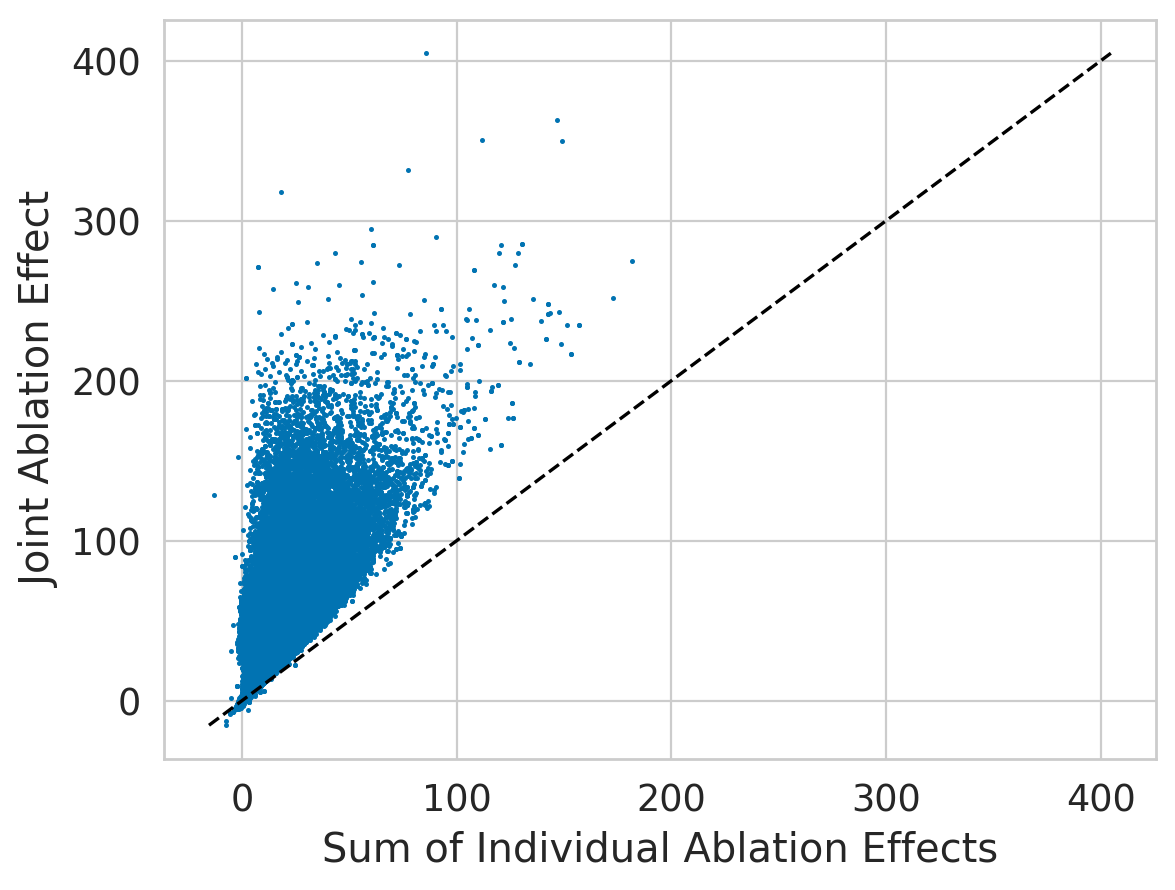}
         \includegraphics[width=\textwidth]{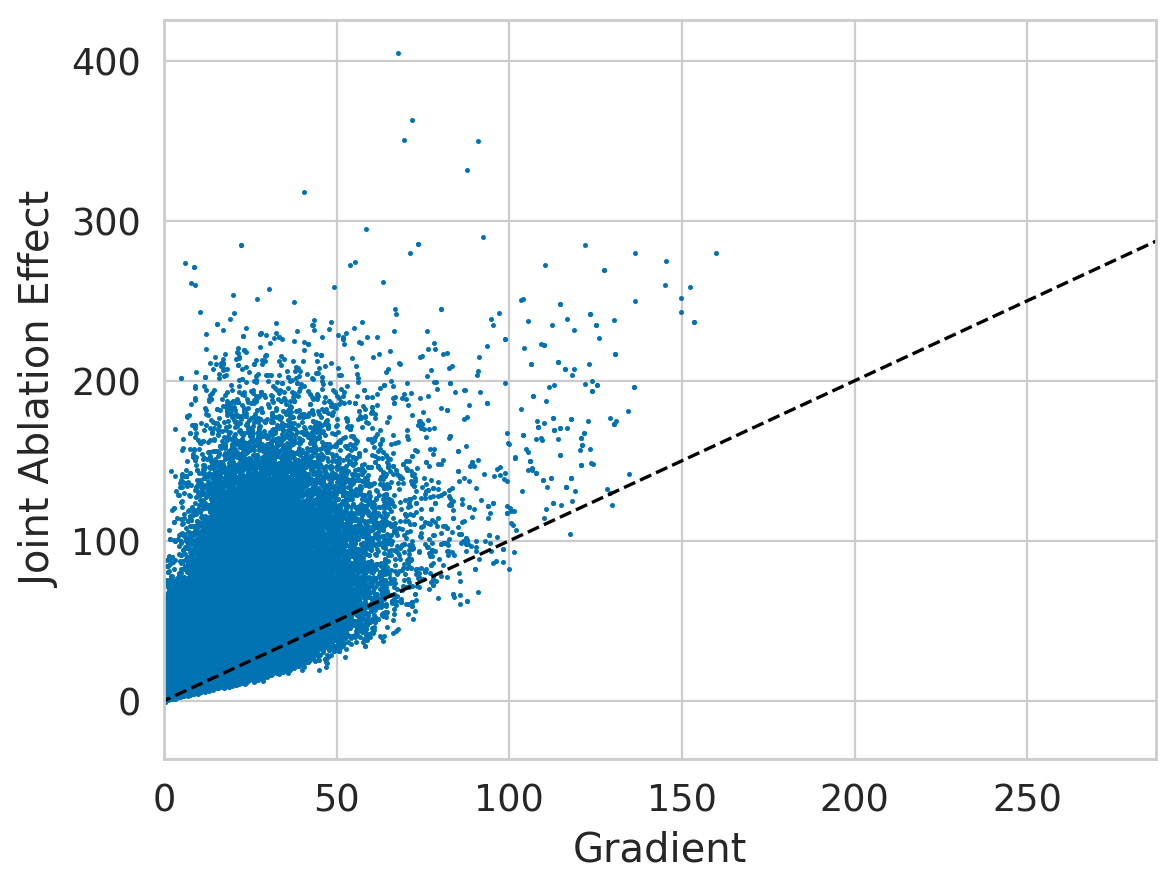}
         \includegraphics[width=\textwidth]{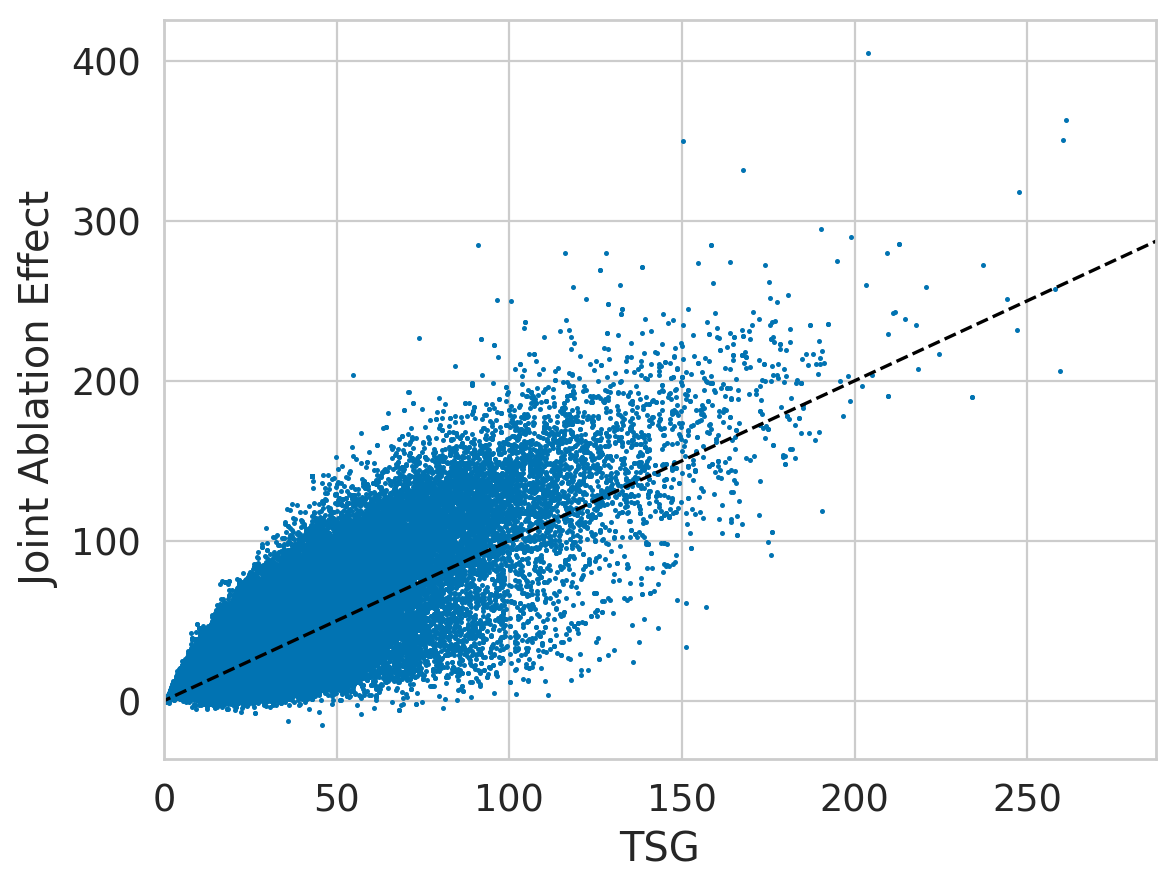}
         
         \caption{Scifact}
     \end{subfigure}
     \hfill
     \begin{subfigure}[b]{0.25\textwidth}
         \centering
         \includegraphics[width=\textwidth]{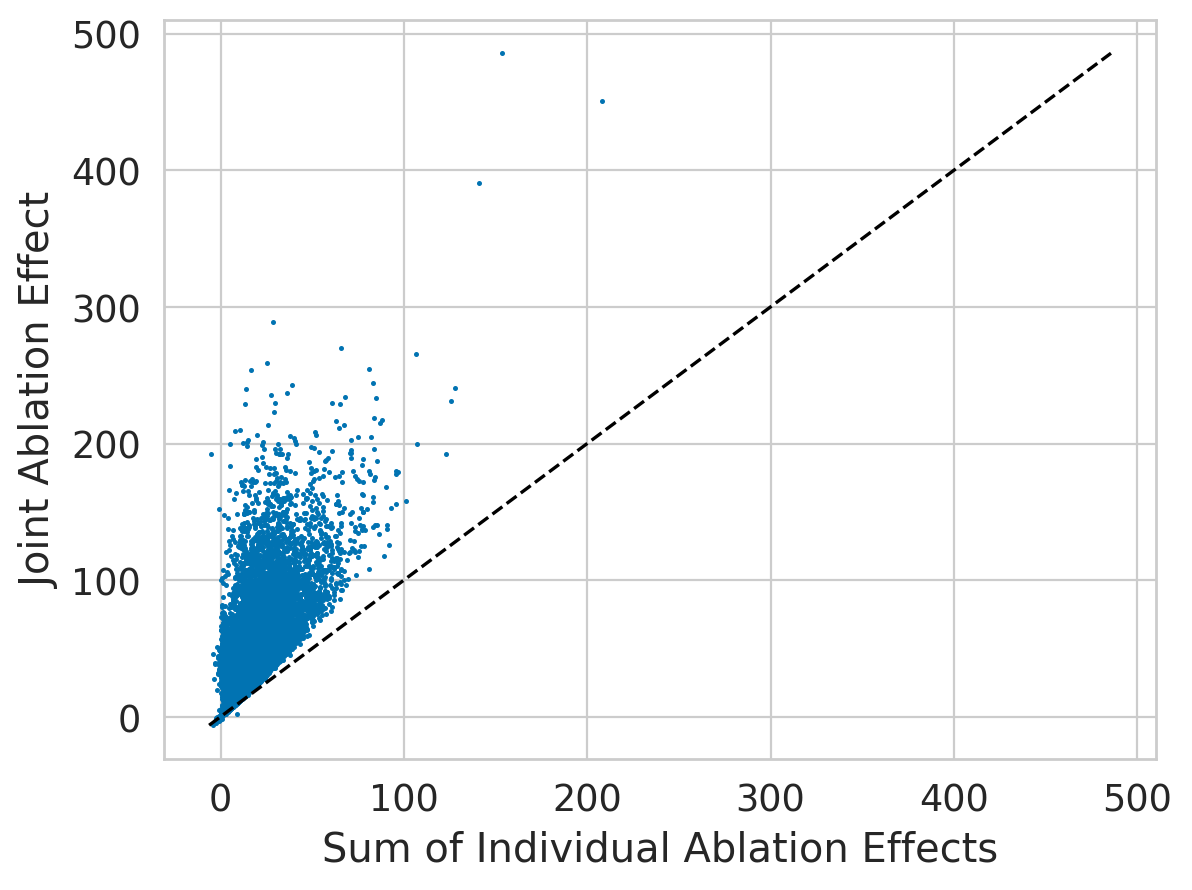}
         \includegraphics[width=\textwidth]{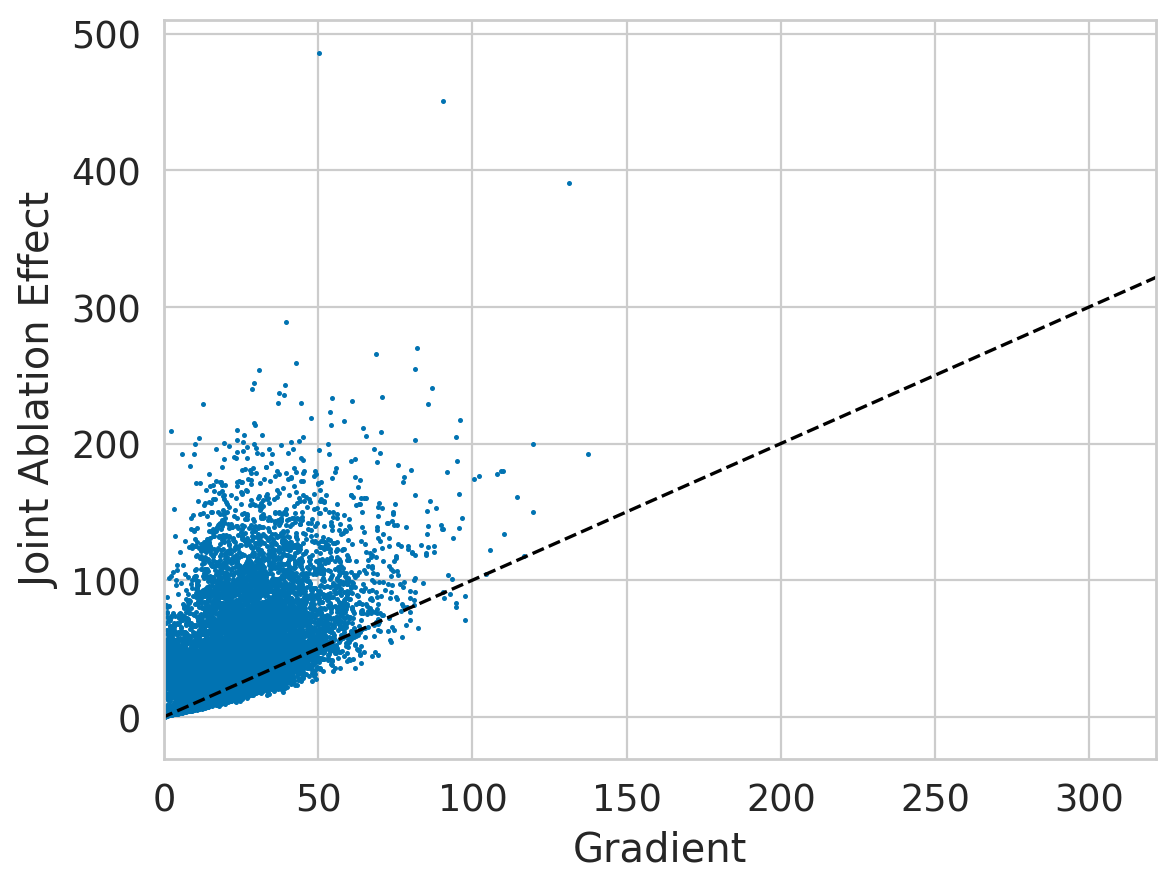}
         \includegraphics[width=\textwidth]{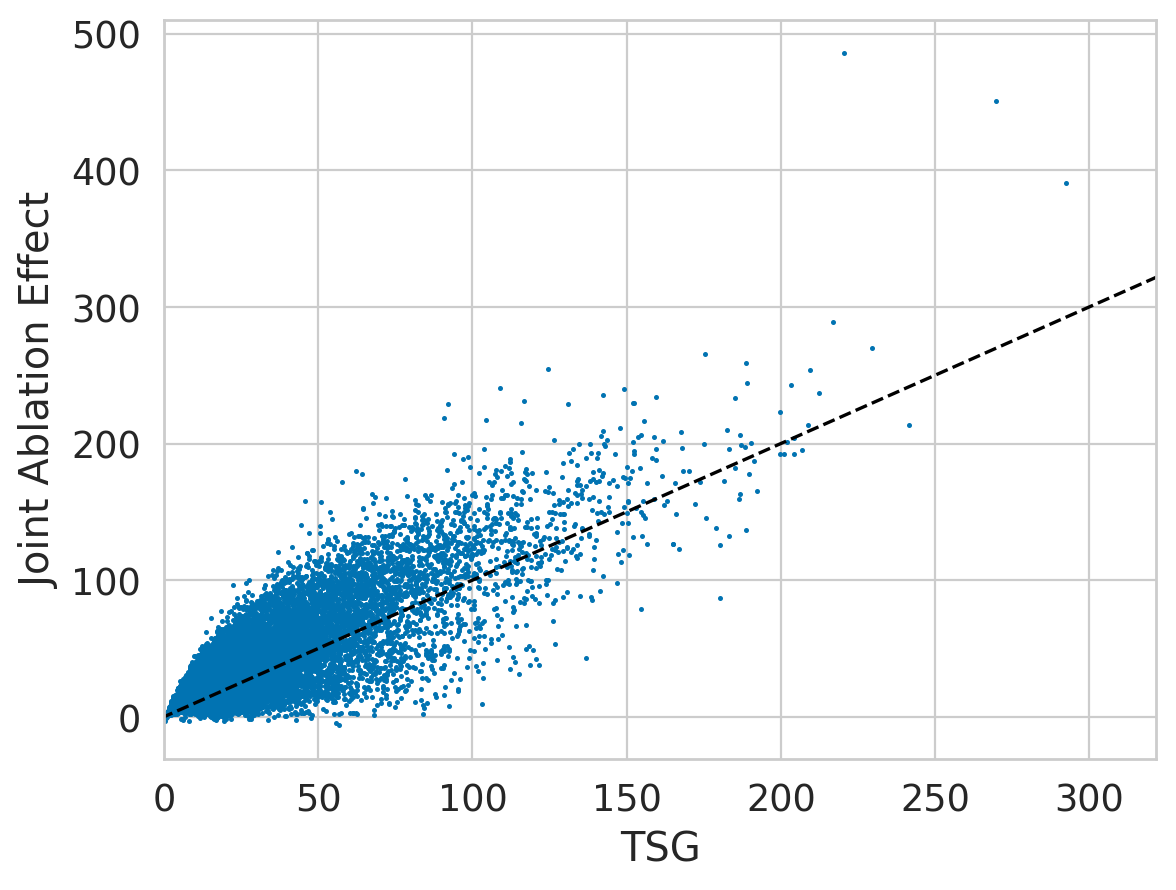}
         \caption{BoolQ}
     \end{subfigure}
     \hfill
     \begin{subfigure}[b]{0.25\textwidth}
         \centering
         \includegraphics[width=\textwidth]{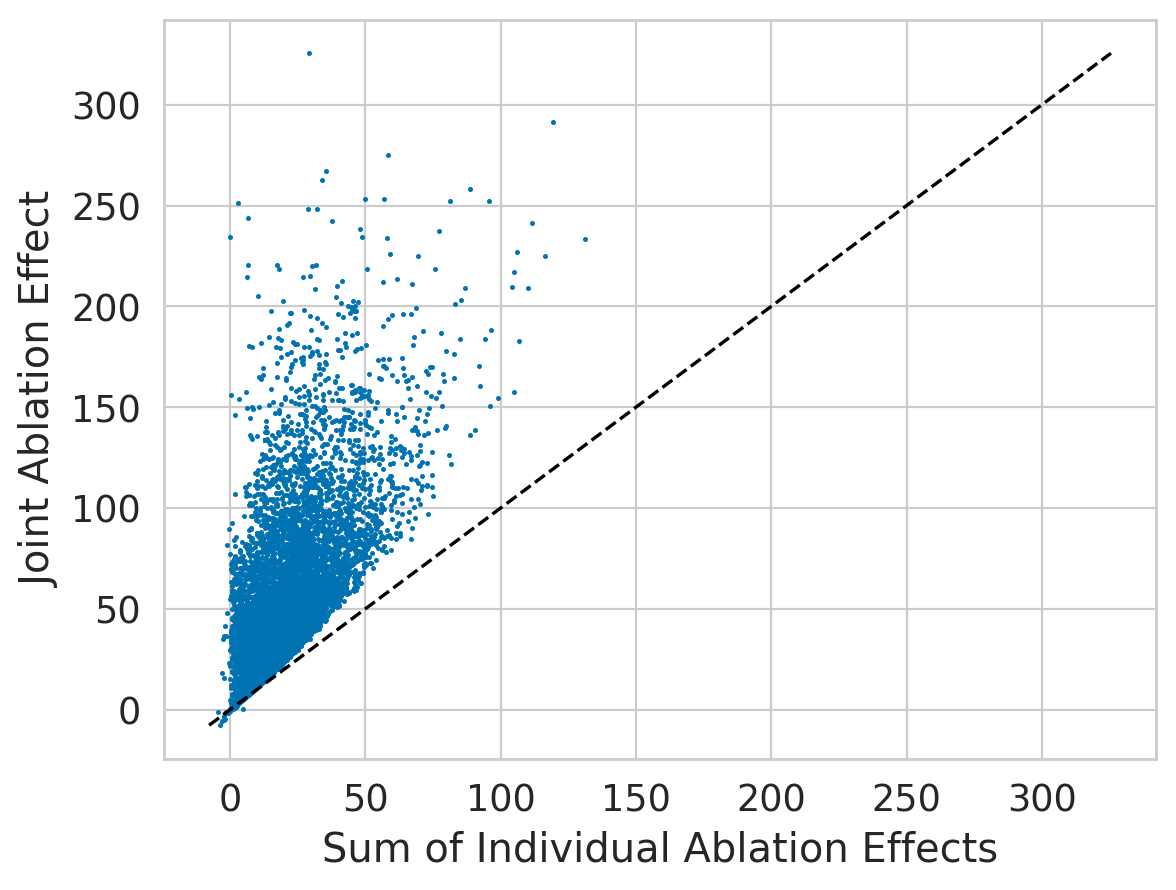}
         \includegraphics[width=\textwidth]{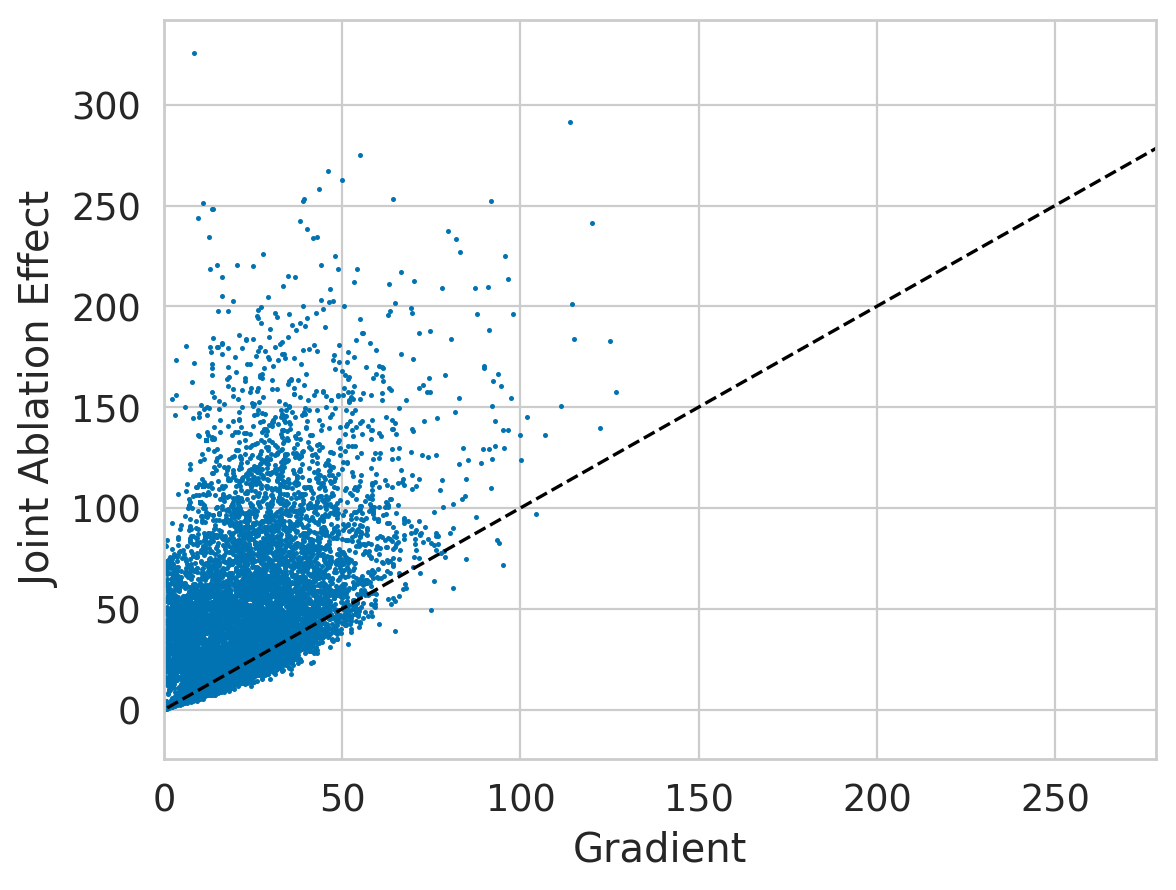}
         \includegraphics[width=\textwidth]{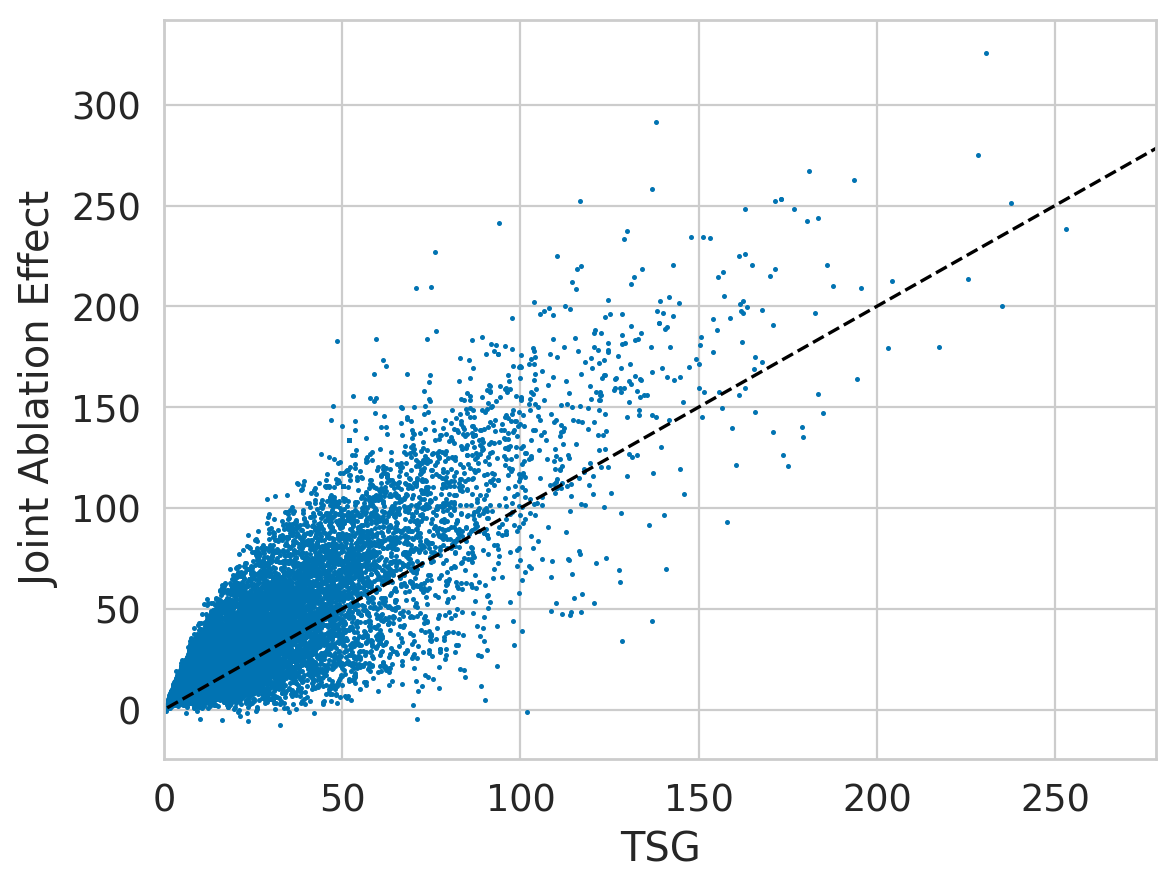}
         \caption{Movie}
     \end{subfigure}
     \hfill
     \begin{subfigure}[b]{0.25\textwidth}
         \centering
         \includegraphics[width=\textwidth]{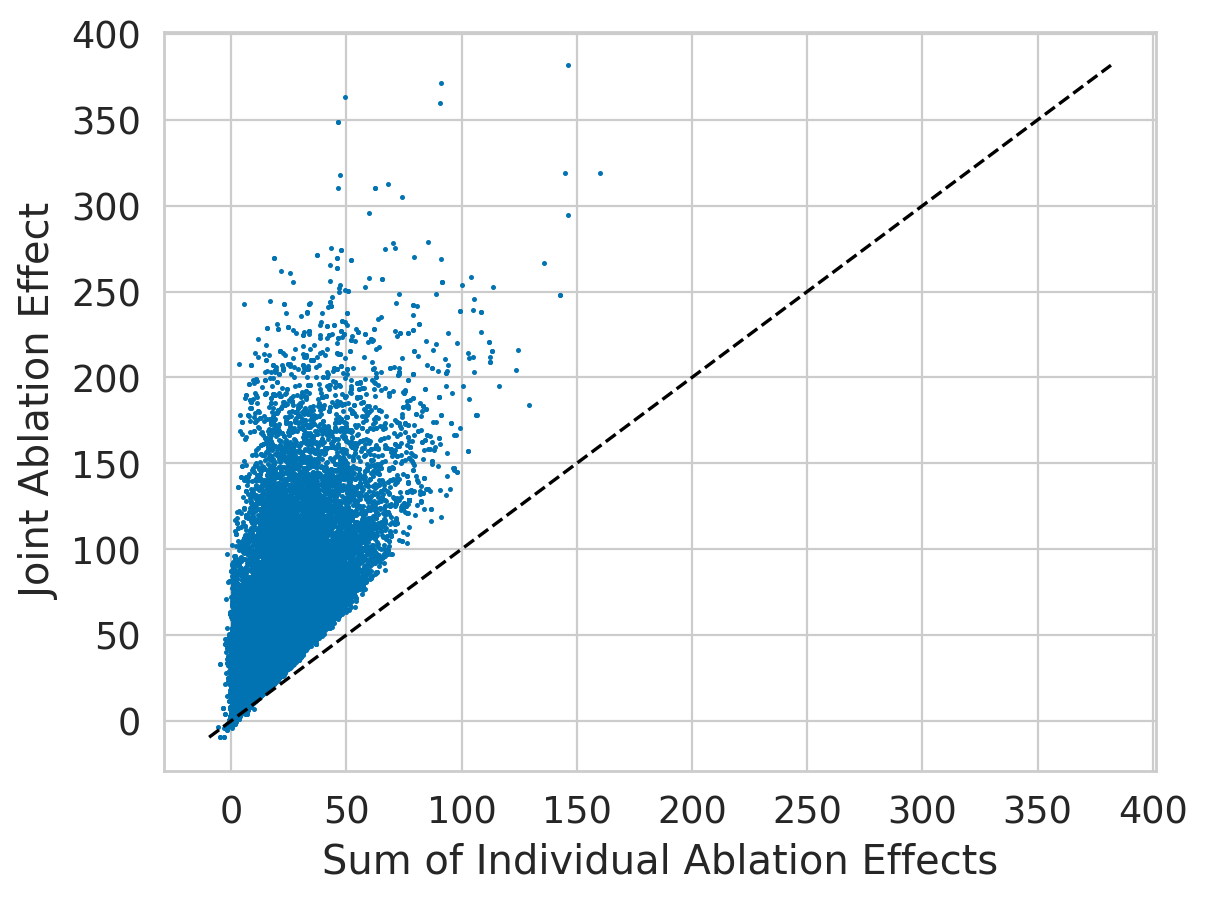}
         \includegraphics[width=\textwidth]{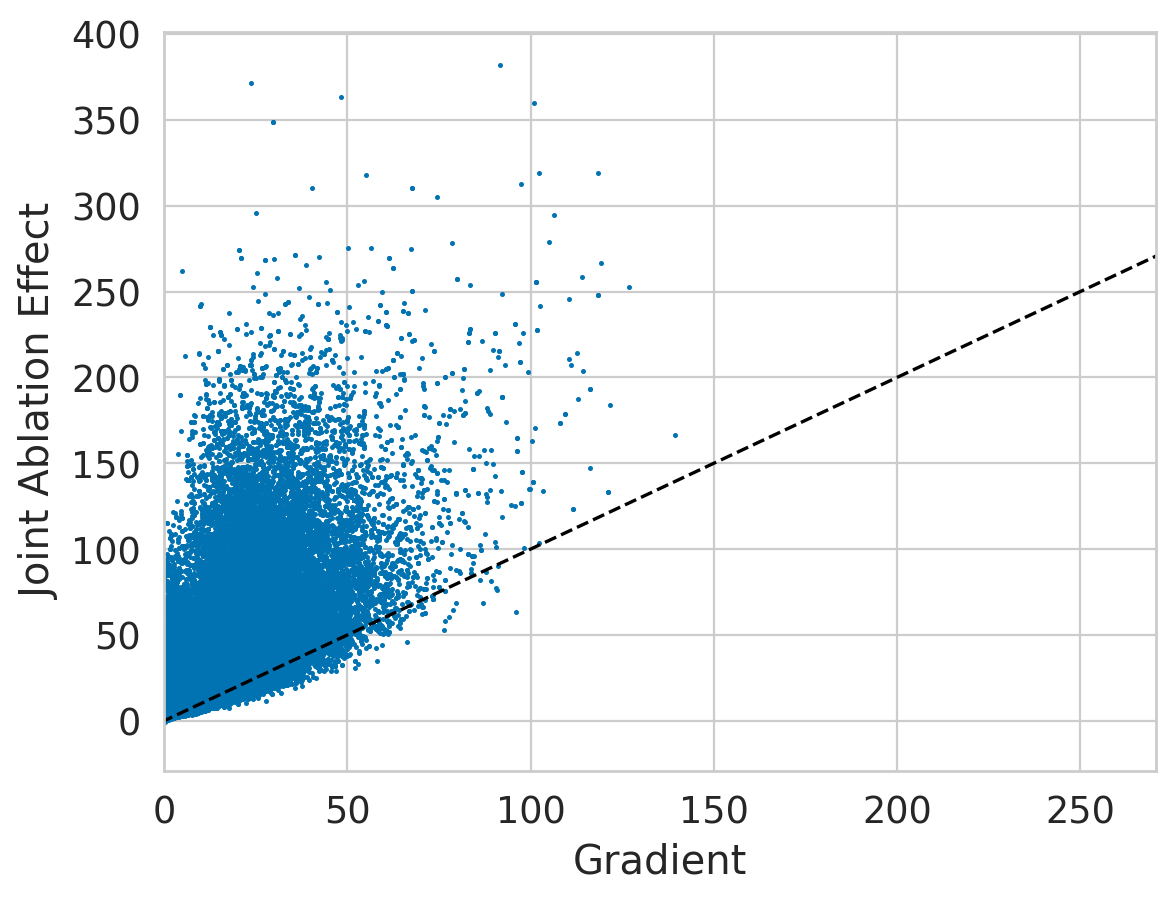}
         \includegraphics[width=\textwidth]{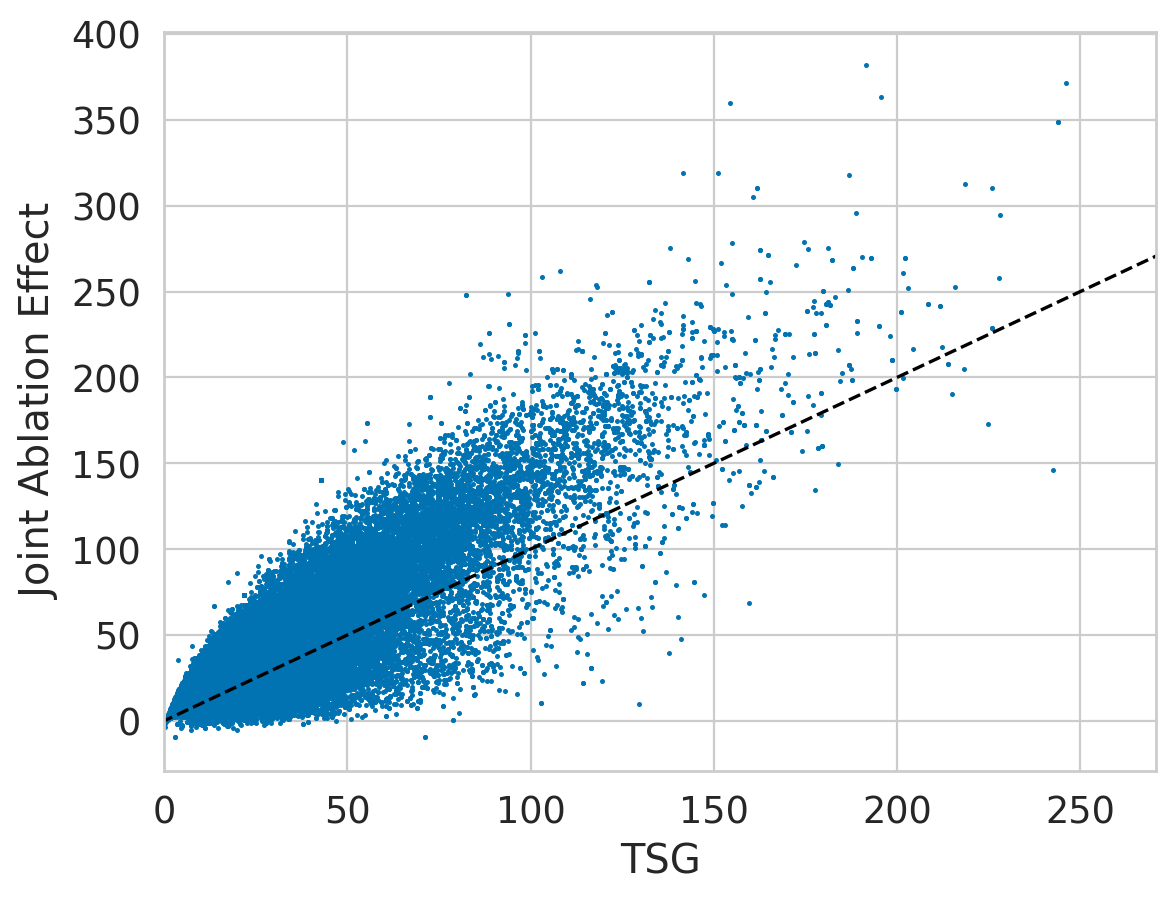}
         \caption{Fever}
     \end{subfigure}
        \caption{Self-repair for \textbf{Gemma-2 2B} and how TSG increases the attributions for the attention scores with the strongest self-repair effects}
        \label{fig:self-repair-gemma2}
\end{figure*}

\begin{figure*}[h]
     \centering
     \begin{subfigure}[b]{0.25\textwidth}
         \centering
         \includegraphics[width=\textwidth]{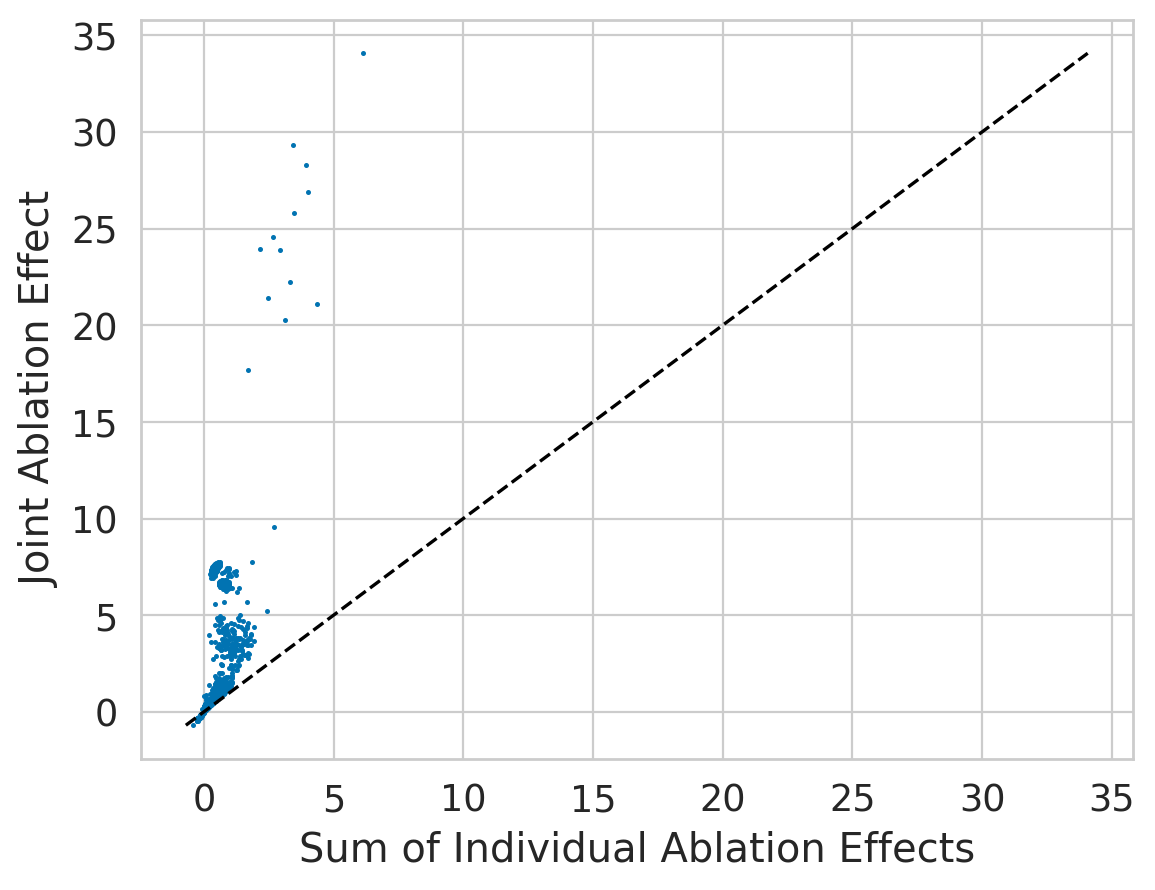}
         \includegraphics[width=\textwidth]{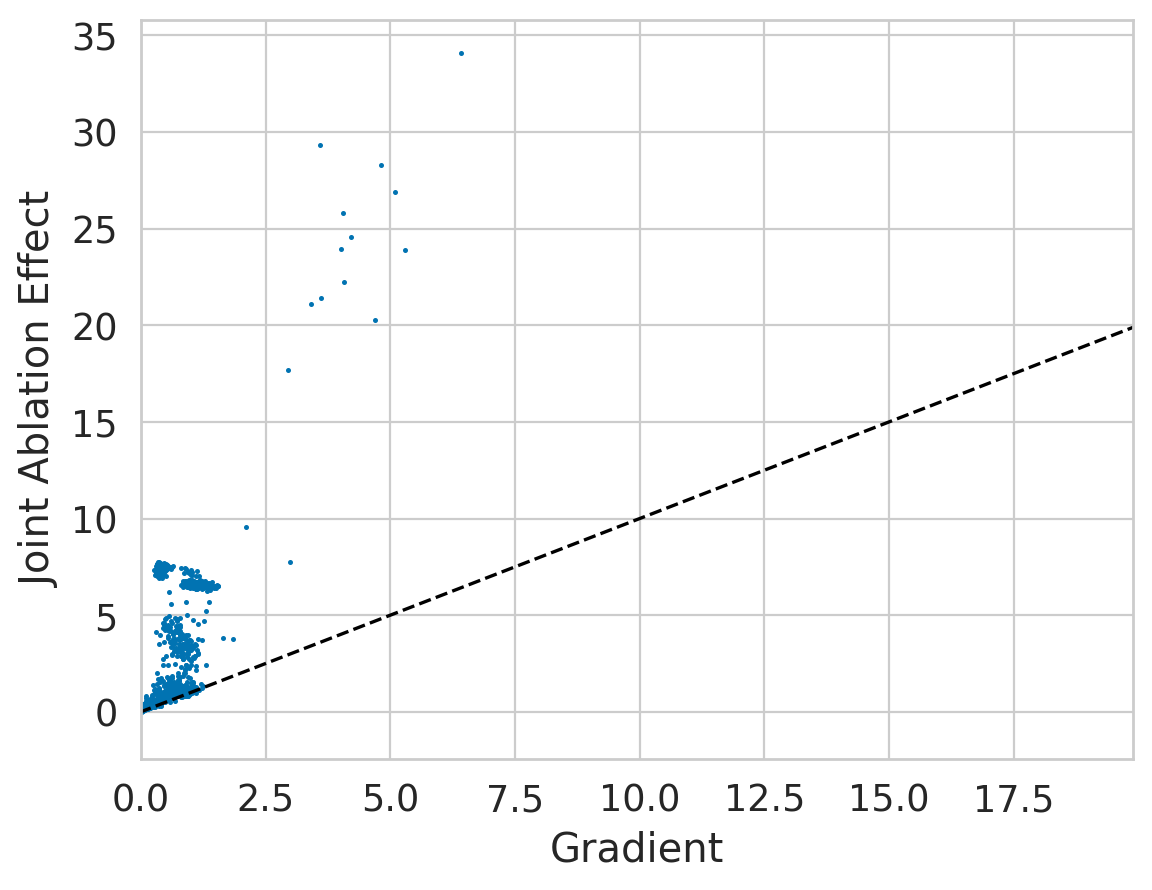}
         \includegraphics[width=\textwidth]{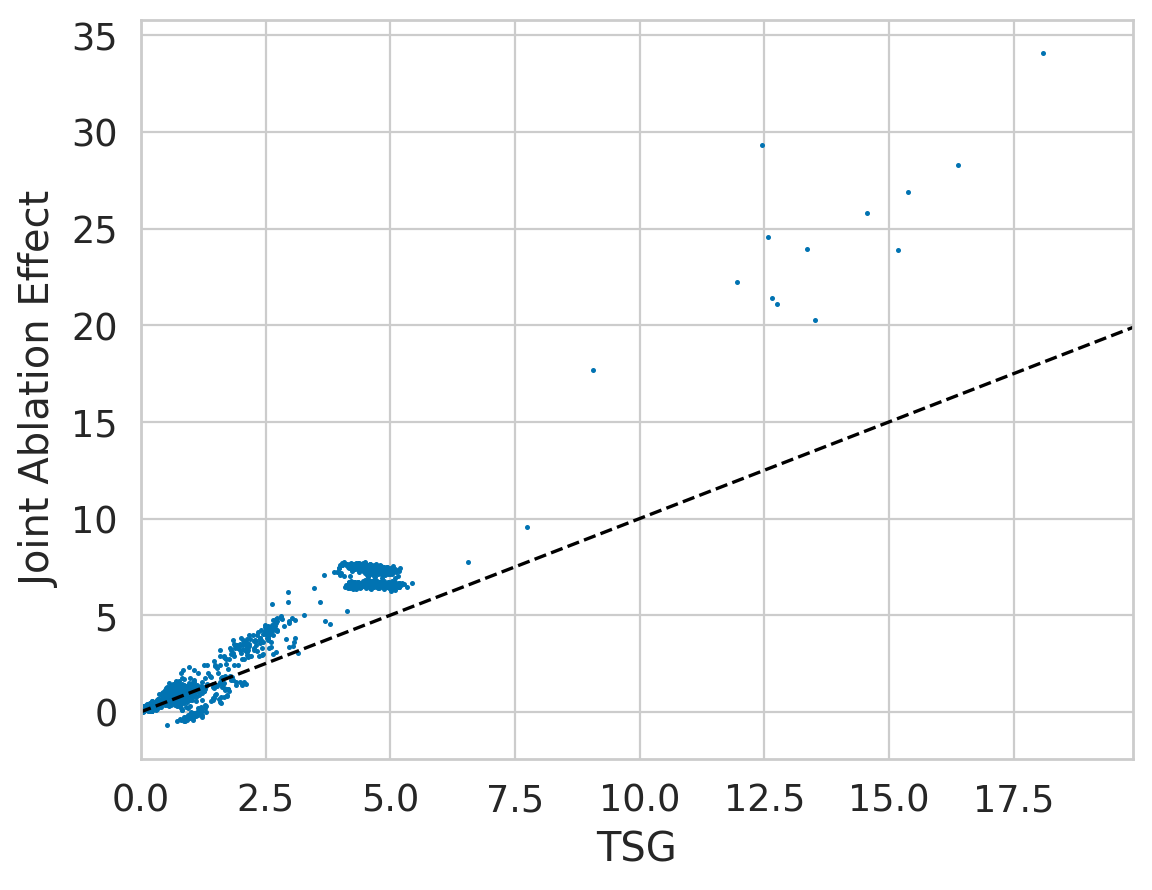}
         \caption{Twitter}
     \end{subfigure}
     \hfill
     \begin{subfigure}[b]{0.25\textwidth}
         \centering
         \includegraphics[width=\textwidth]{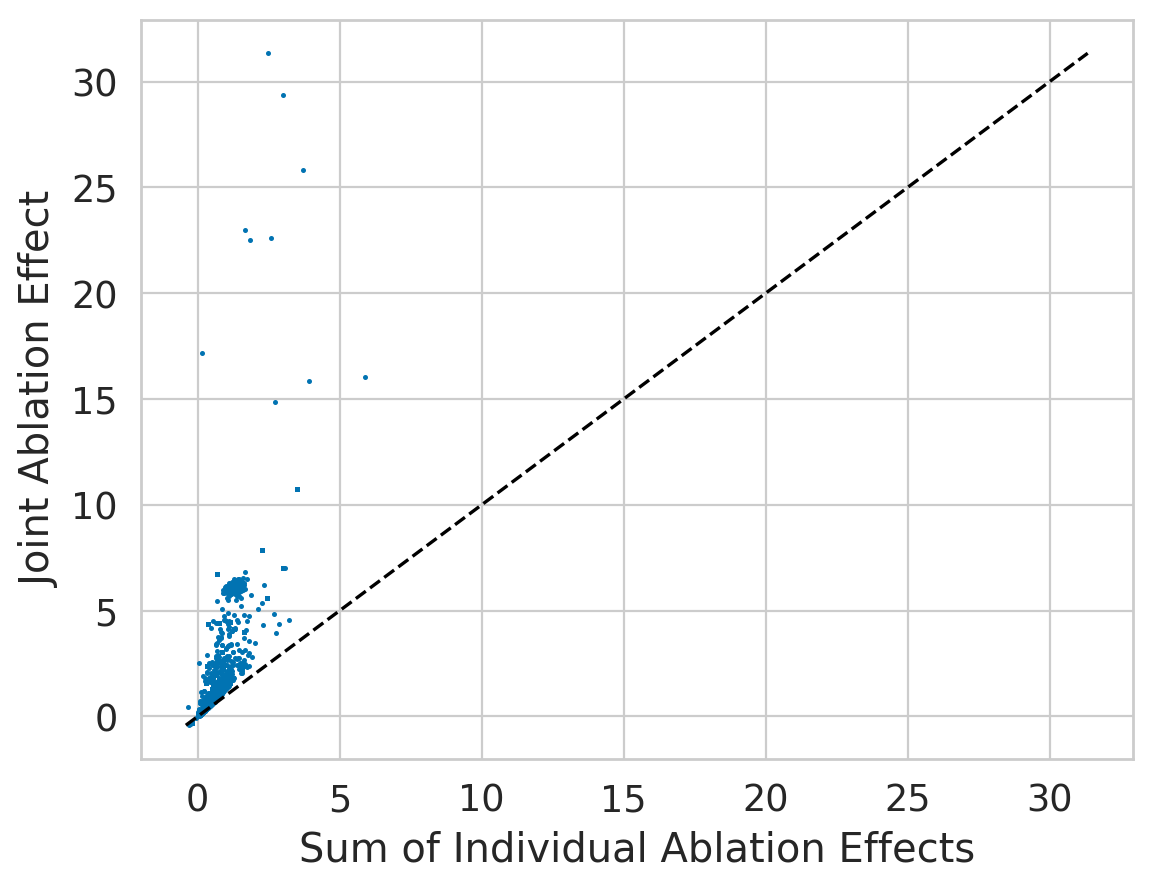}
         \includegraphics[width=\textwidth]{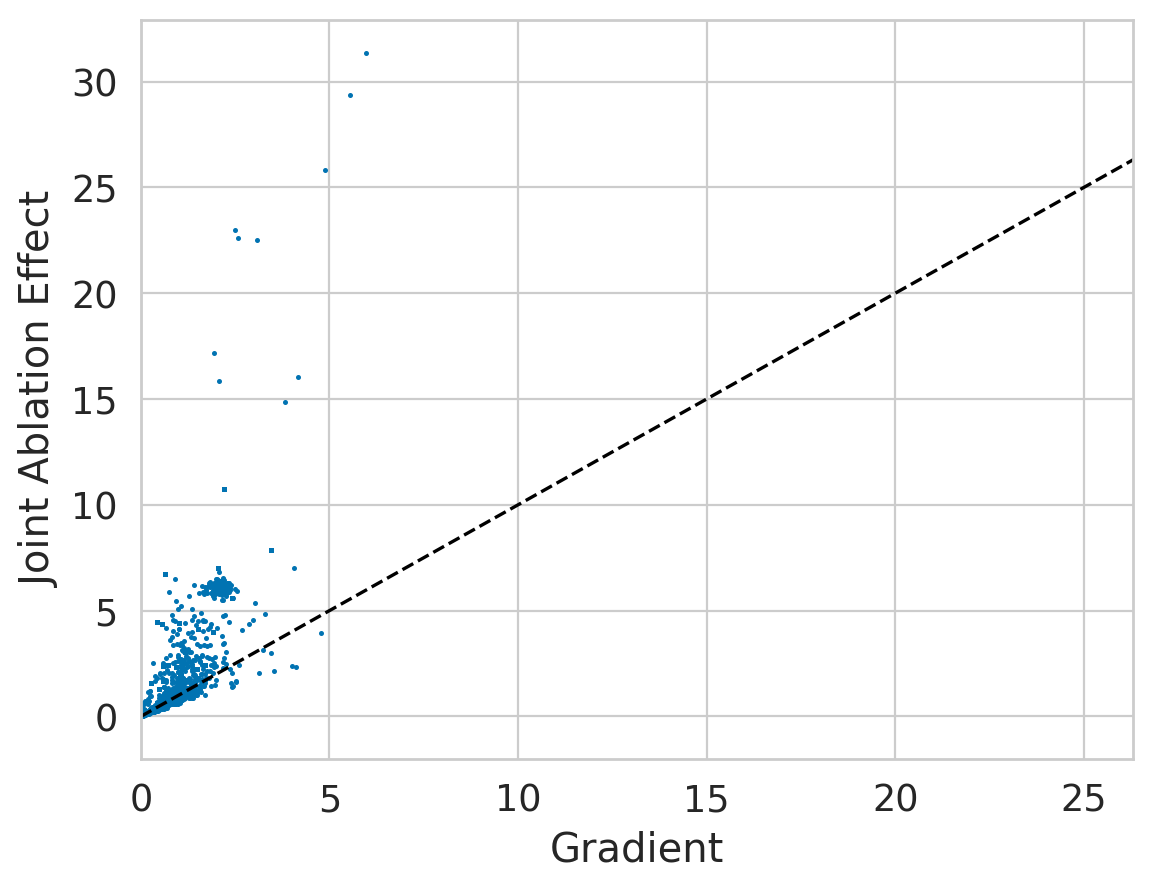}
         \includegraphics[width=\textwidth]{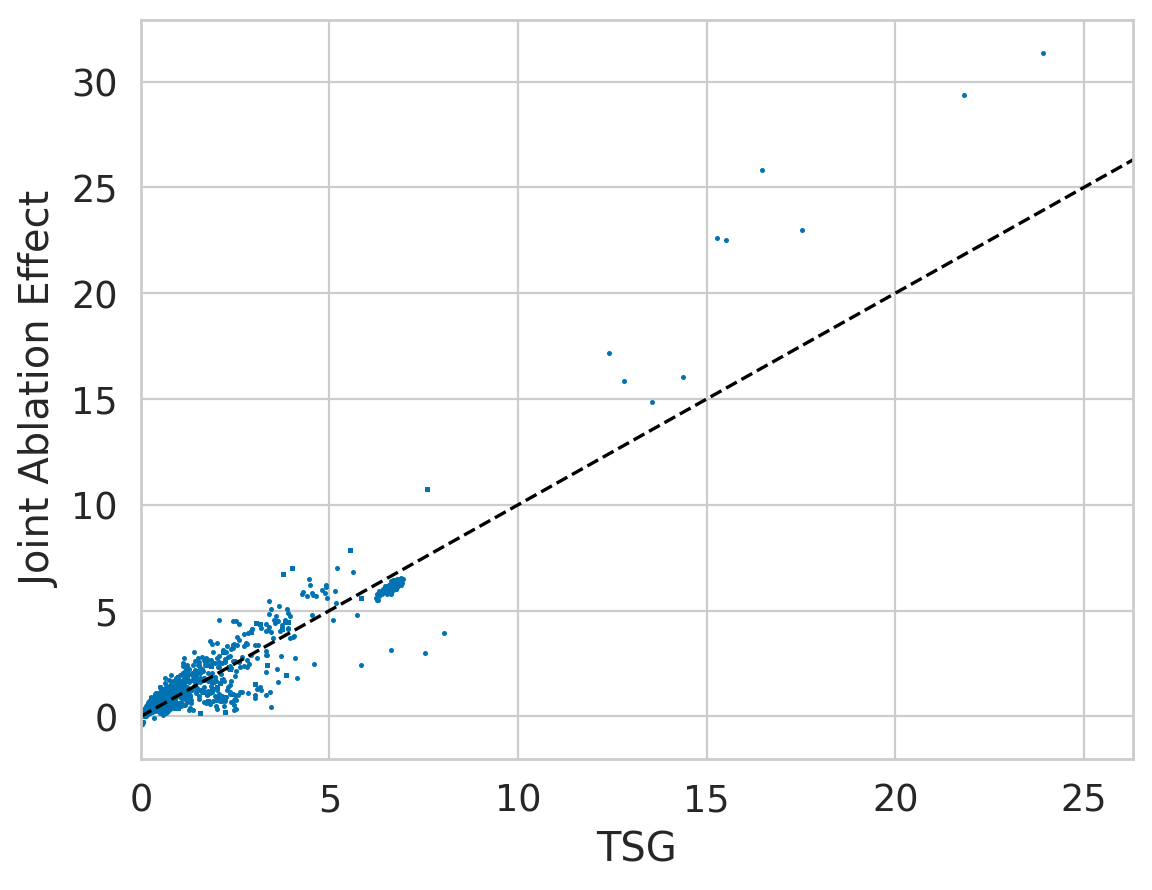}
         \caption{Hatexplain}
     \end{subfigure}
     \hfill
     \begin{subfigure}[b]{0.25\textwidth}
         \centering
         \includegraphics[width=\textwidth]{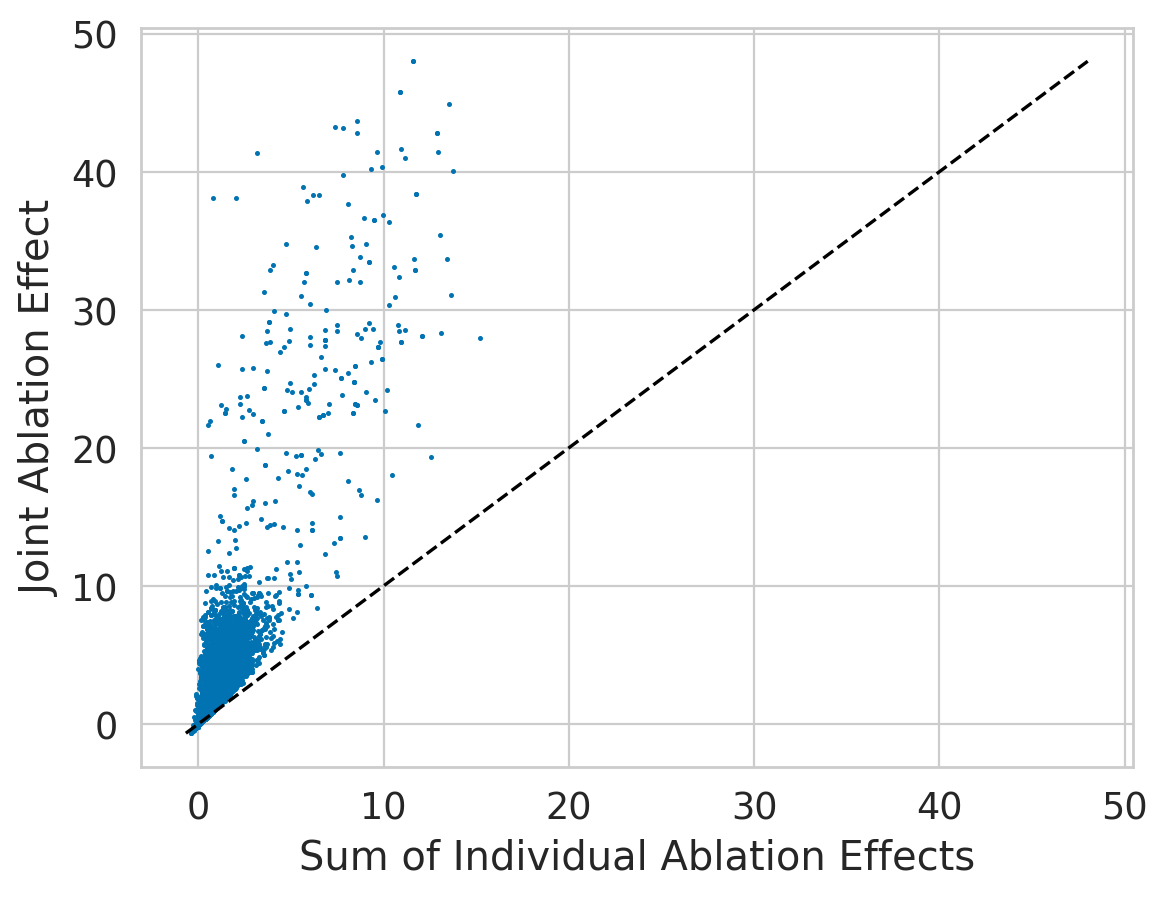}
         \includegraphics[width=\textwidth]{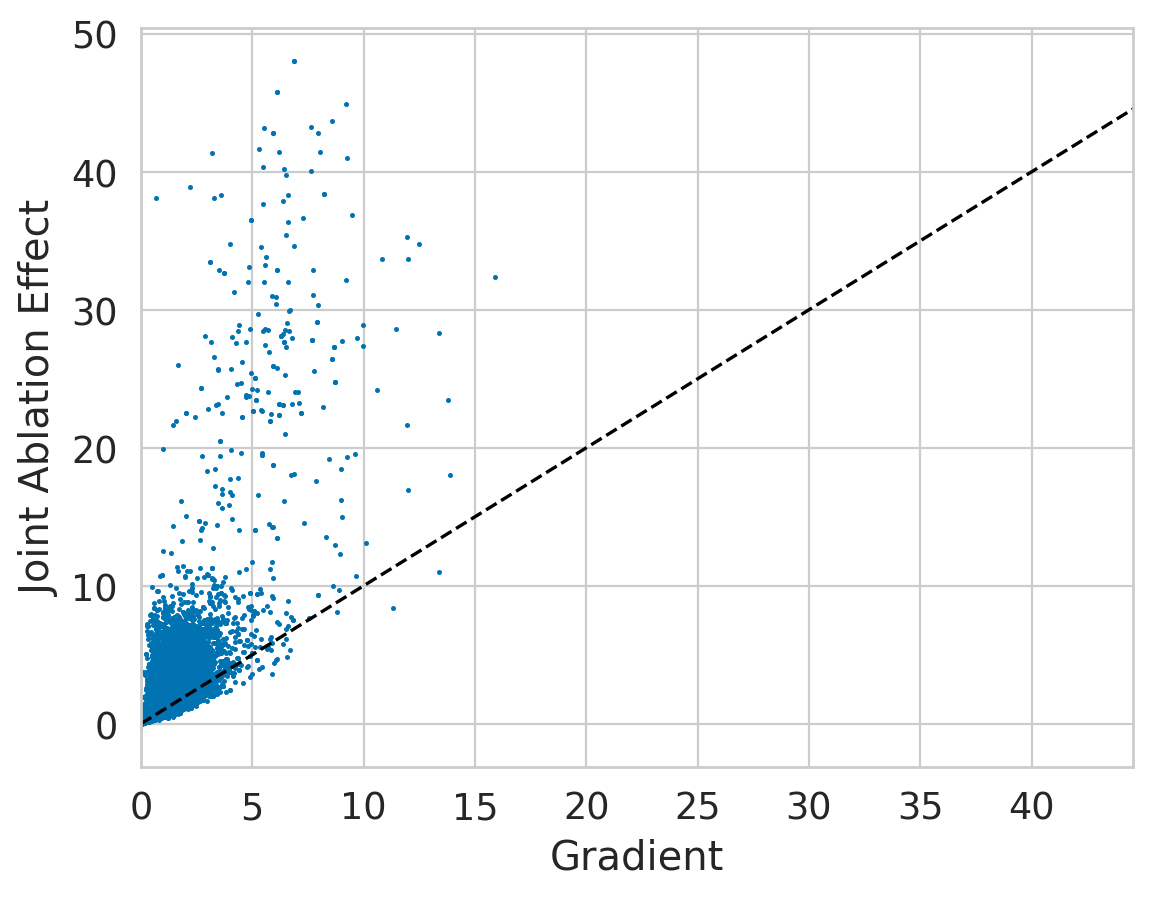}
         \includegraphics[width=\textwidth]{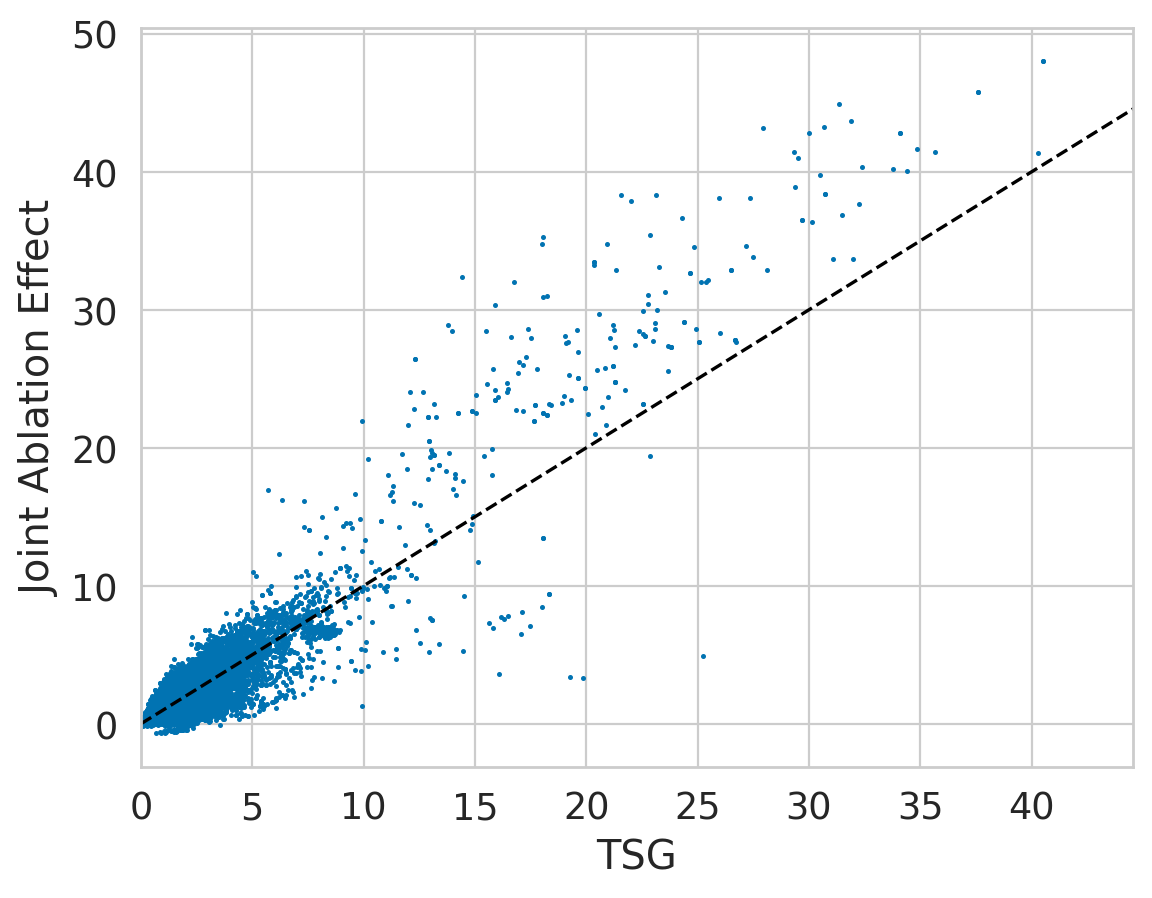}
         
         \caption{Scifact}
     \end{subfigure}
     \hfill
     \begin{subfigure}[b]{0.25\textwidth}
         \centering
         \includegraphics[width=\textwidth]{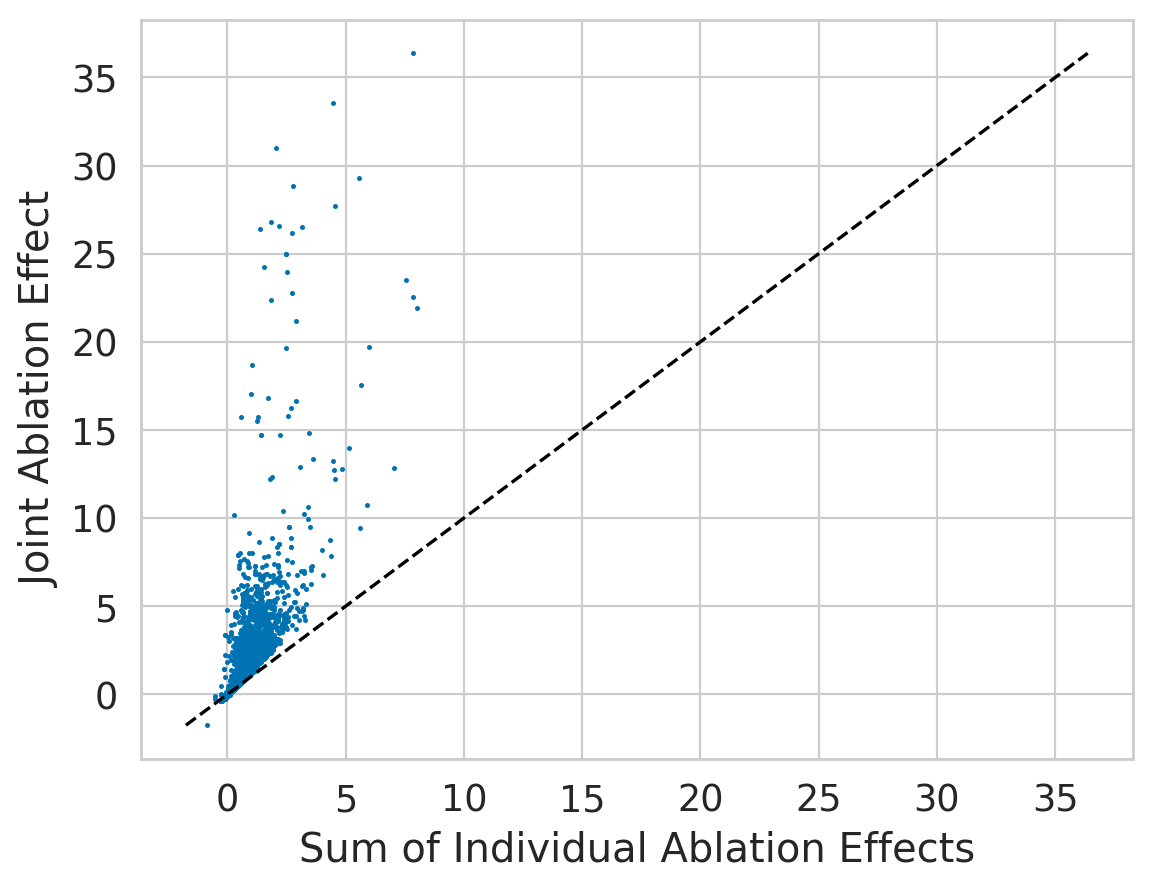}
         \includegraphics[width=\textwidth]{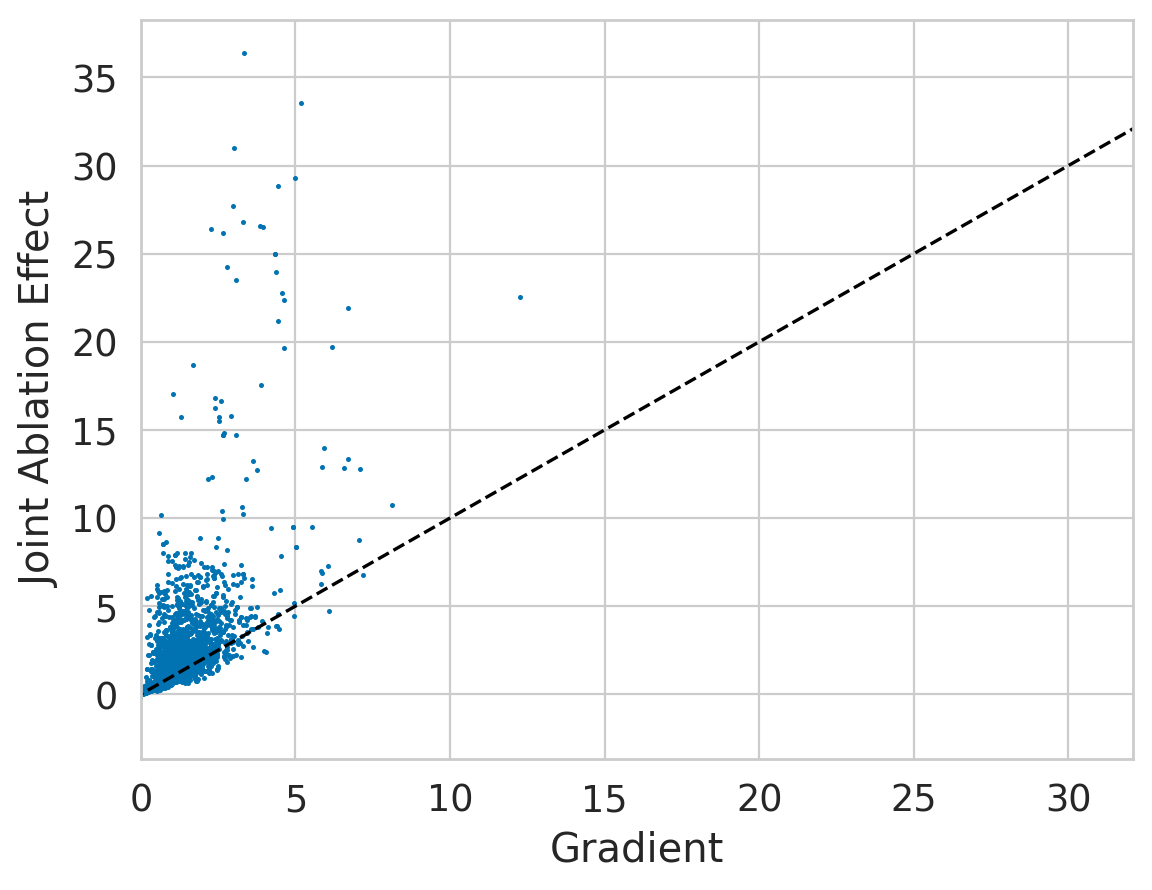}
         \includegraphics[width=\textwidth]{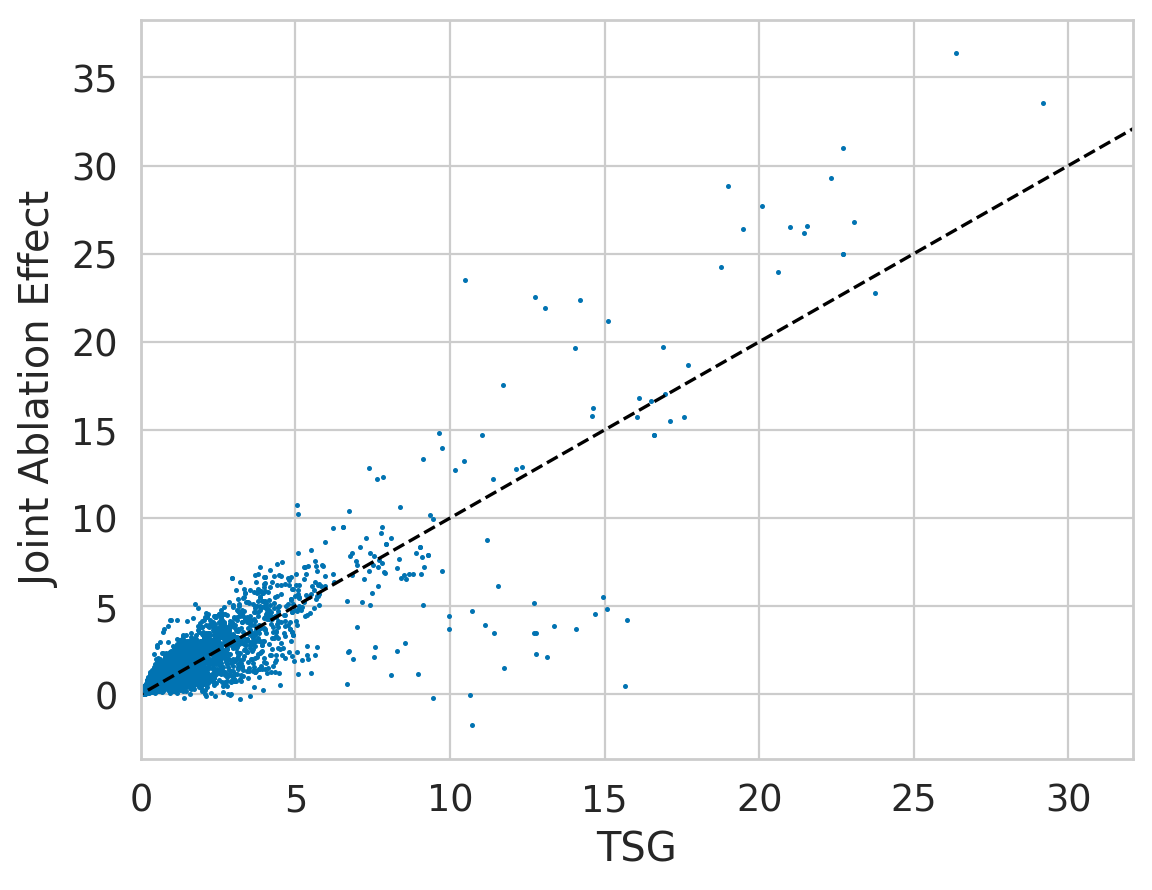}
         \caption{BoolQ}
     \end{subfigure}
     \hfill
     \begin{subfigure}[b]{0.25\textwidth}
         \centering
         \includegraphics[width=\textwidth]{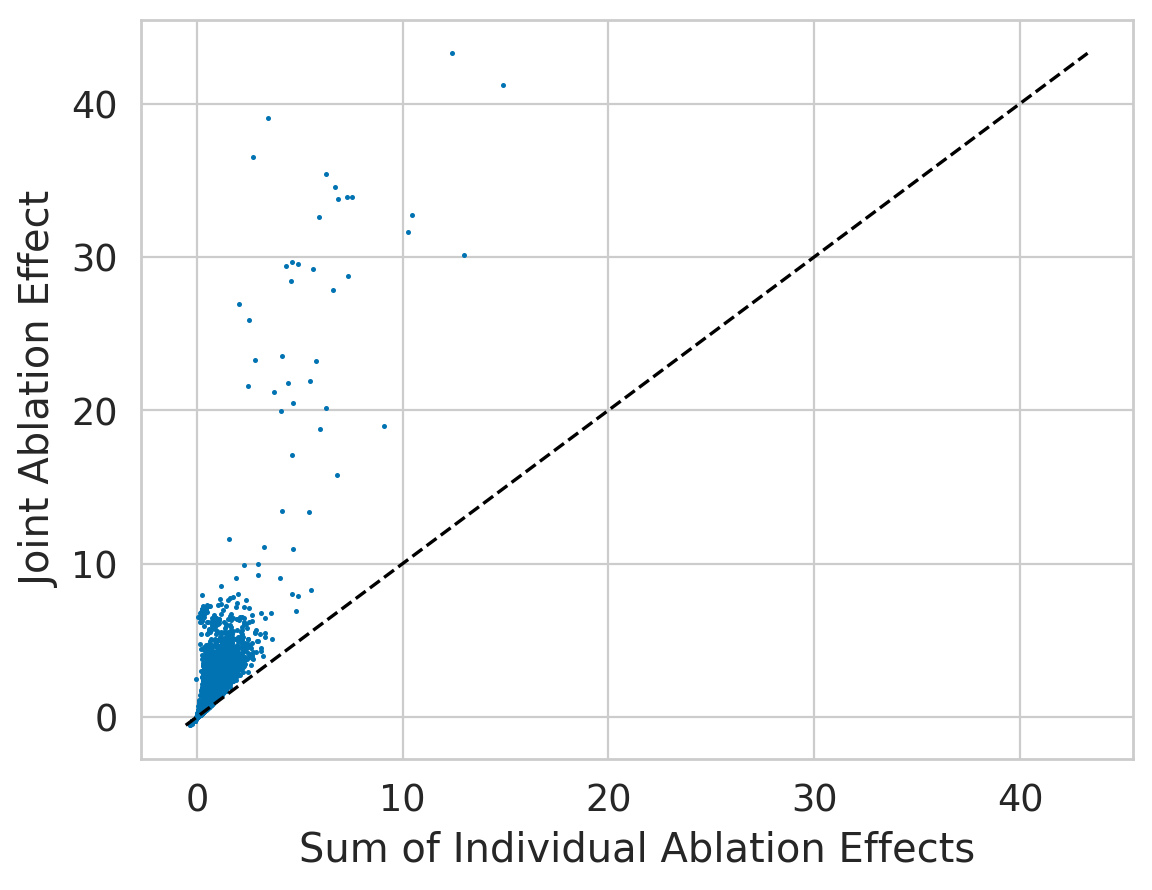}
         \includegraphics[width=\textwidth]{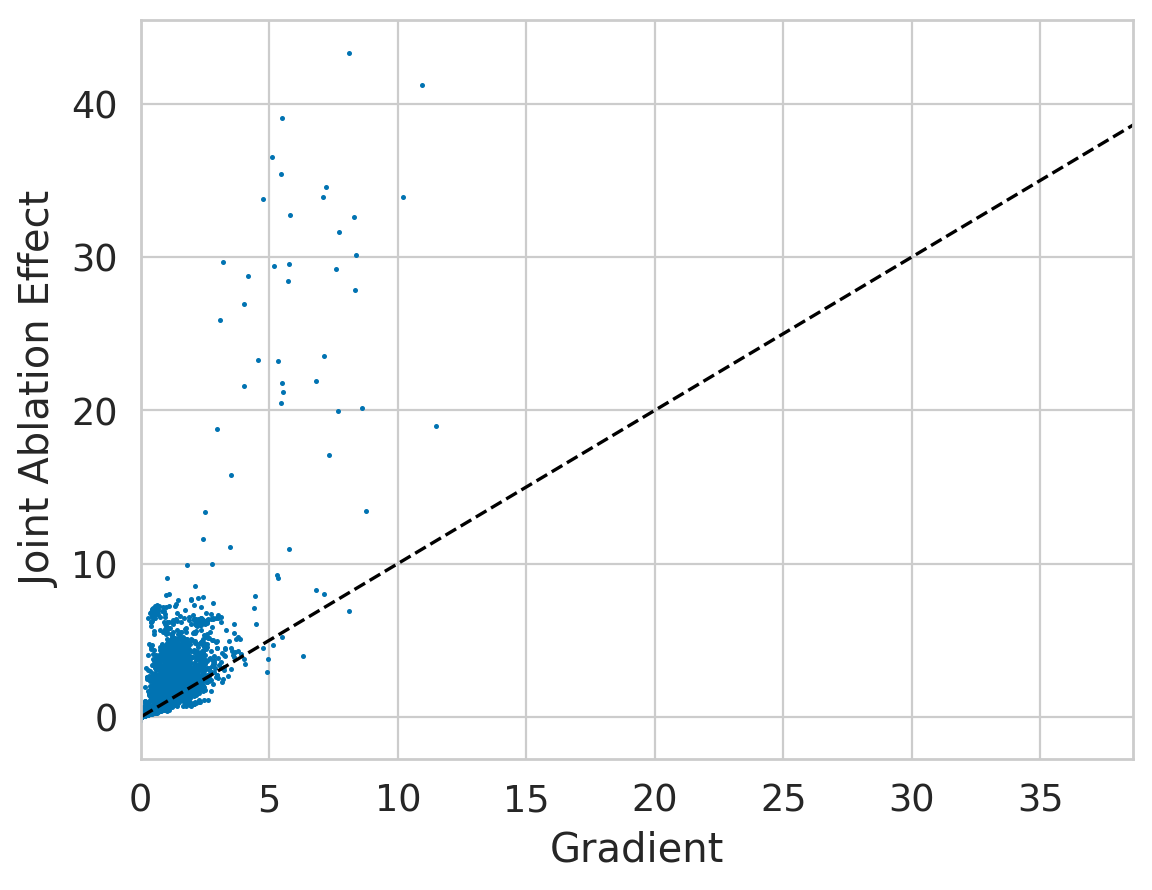}
         \includegraphics[width=\textwidth]{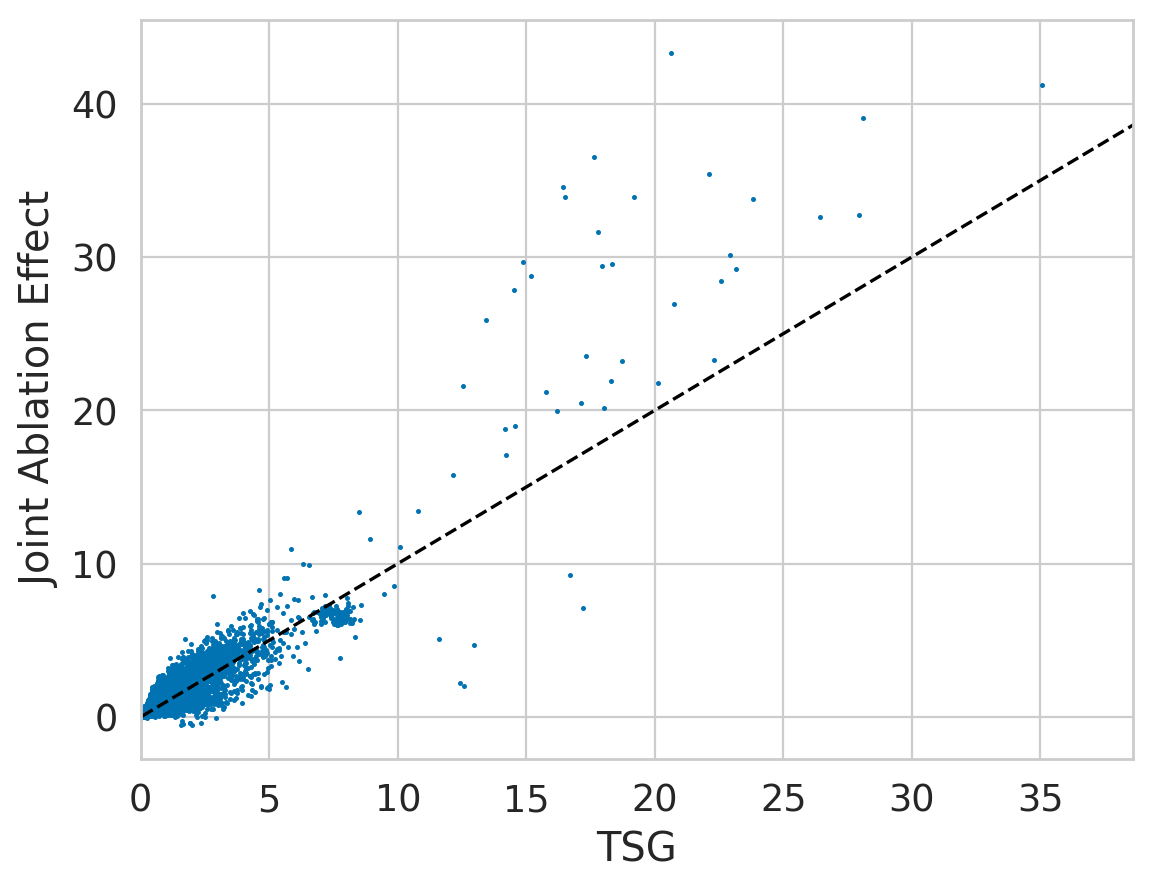}
         \caption{Movie}
     \end{subfigure}
     \hfill
     \begin{subfigure}[b]{0.25\textwidth}
         \centering
         \includegraphics[width=\textwidth]{figures/self_repair_0.1/fever/llama3-1b-it/ablation_effect.png}
         \includegraphics[width=\textwidth]{figures/self_repair_0.1/fever/llama3-1b-it/grad_vs_joint.png}
         \includegraphics[width=\textwidth]{figures/self_repair_0.1/fever/llama3-1b-it/tsg_vs_joint.png}
         \caption{Fever}
     \end{subfigure}
        \caption{Self-repair for \textbf{LLAMA-3.2 1B} and how TSG increases the attributions for the attention scores with the strongest self-repair effects}
        \label{fig:self-repair-llama1}
\end{figure*}

\begin{figure*}[h]
     \centering
     \begin{subfigure}[b]{0.25\textwidth}
         \centering
         \includegraphics[width=\textwidth]{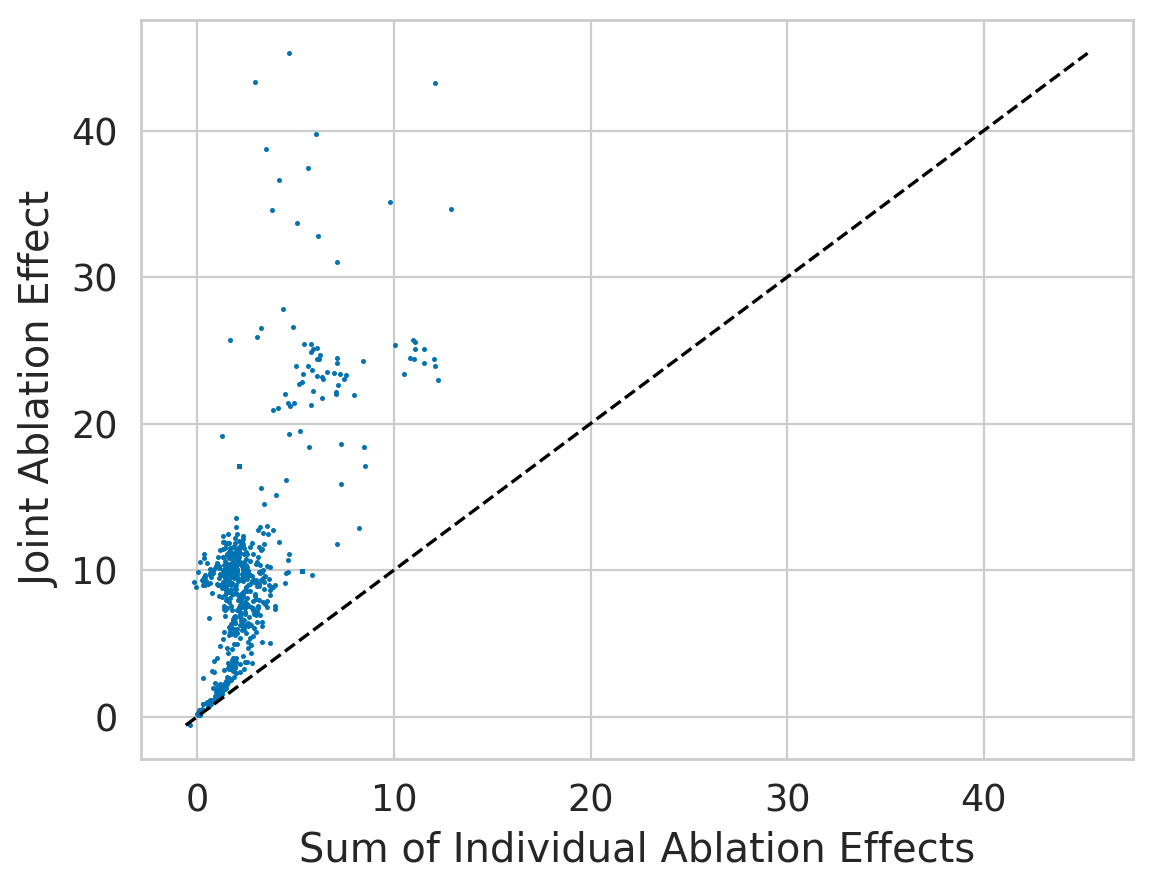}
         \includegraphics[width=\textwidth]{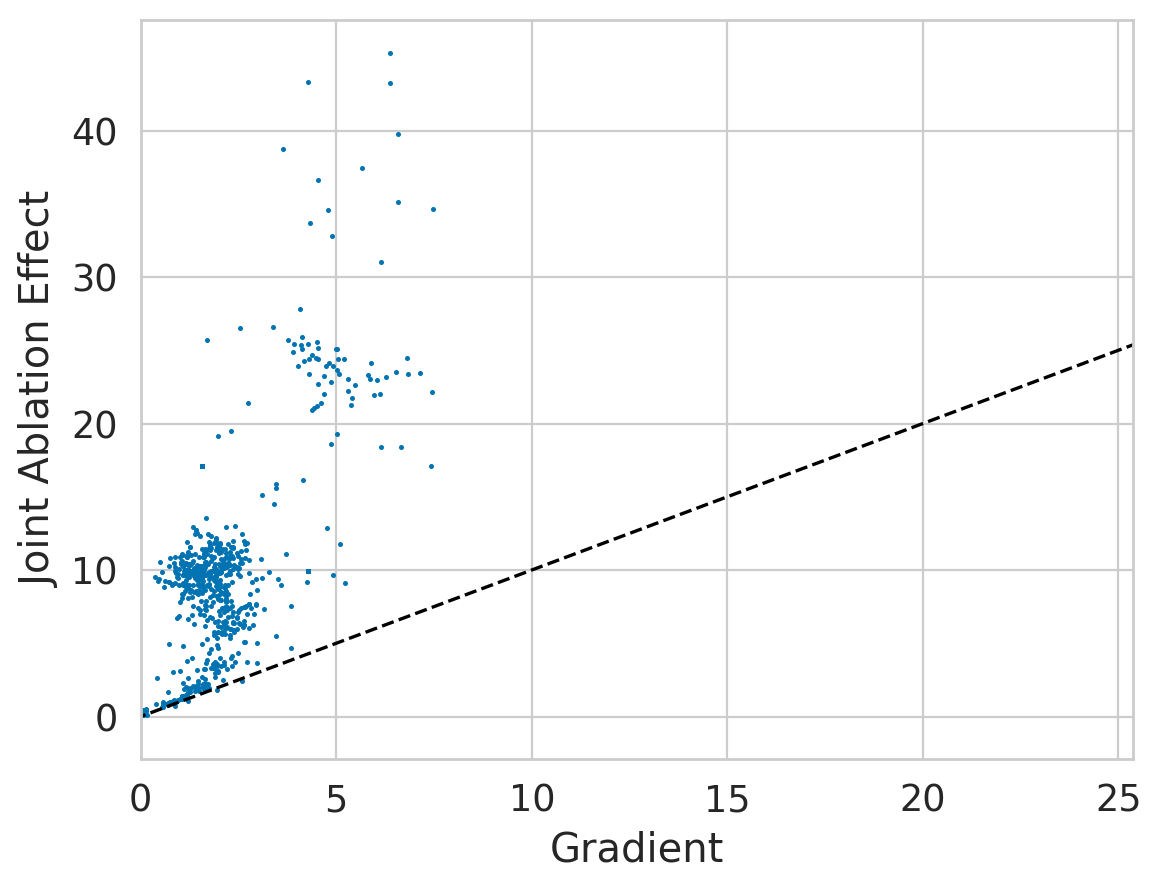}
         \includegraphics[width=\textwidth]{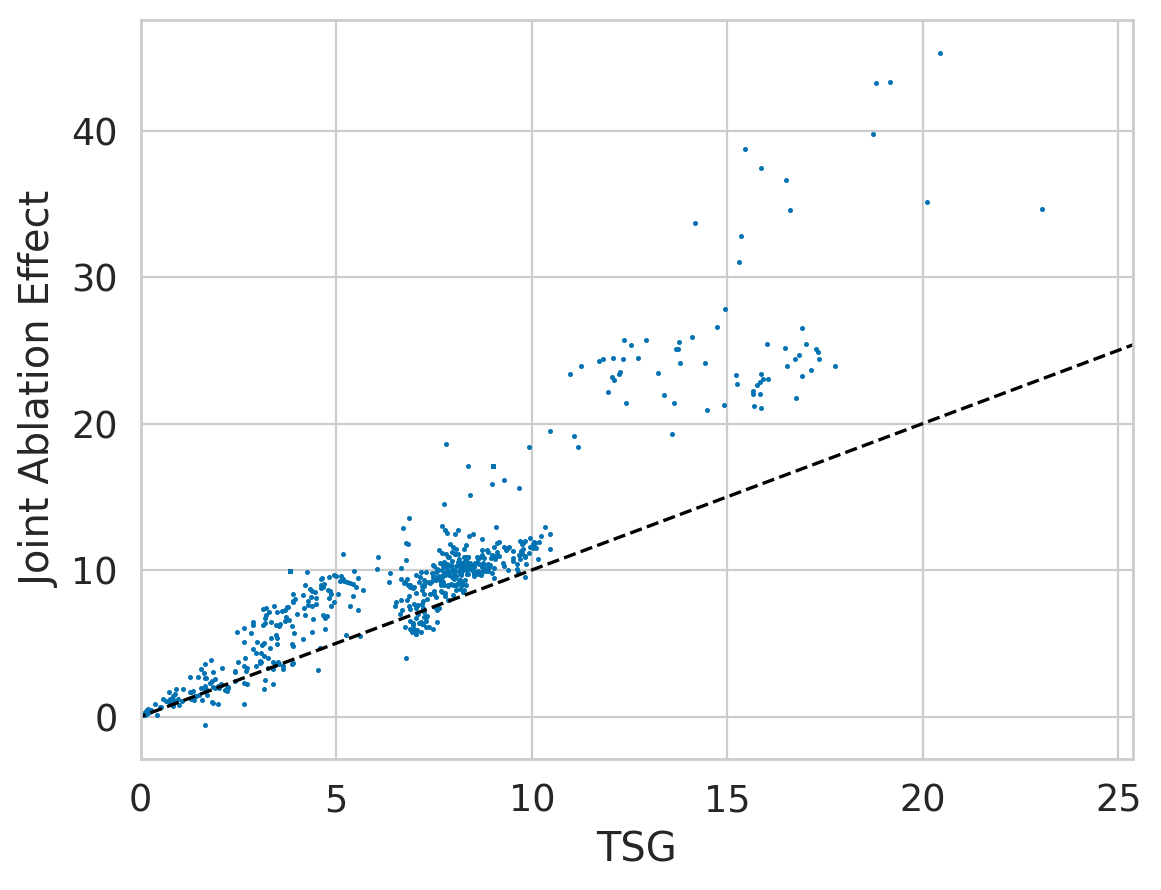}
         \caption{Twitter}
     \end{subfigure}
     \hfill
     \begin{subfigure}[b]{0.25\textwidth}
         \centering
         \includegraphics[width=\textwidth]{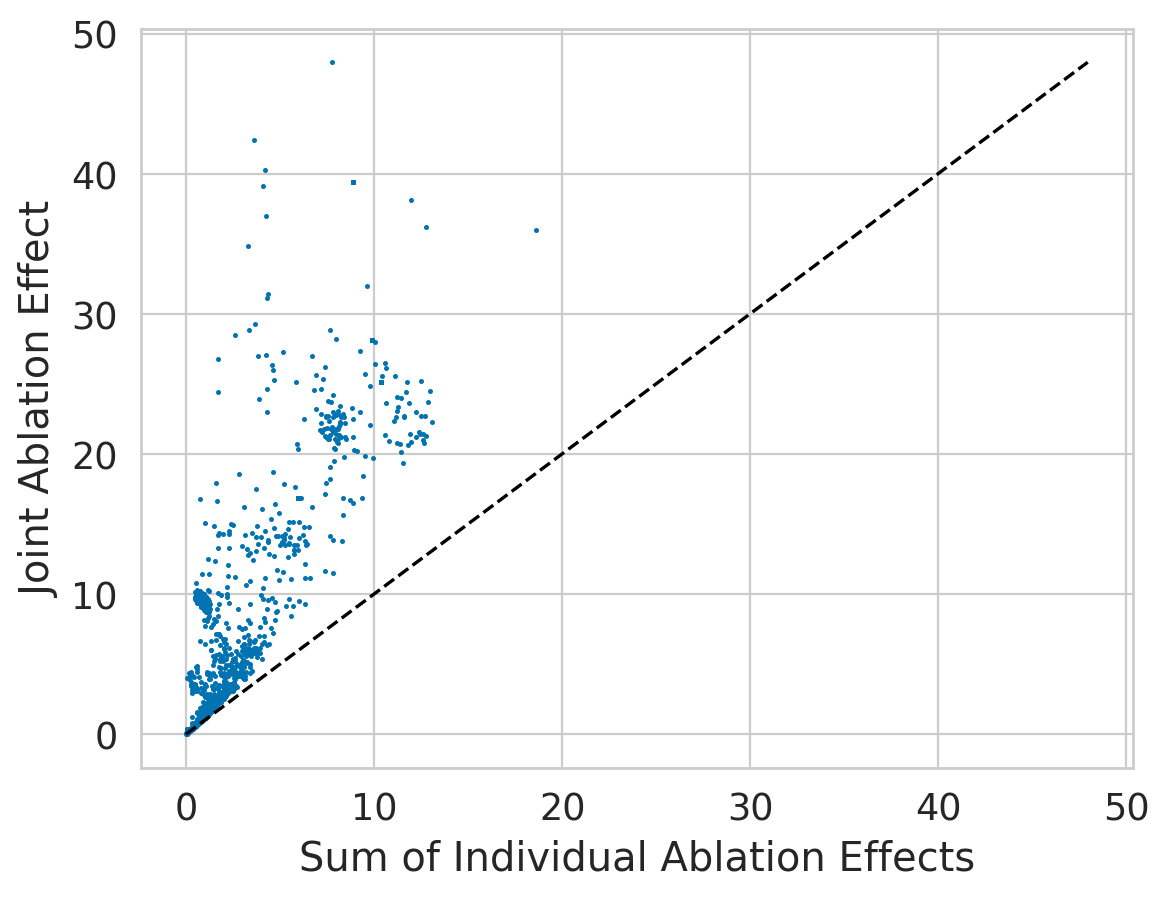}
         \includegraphics[width=\textwidth]{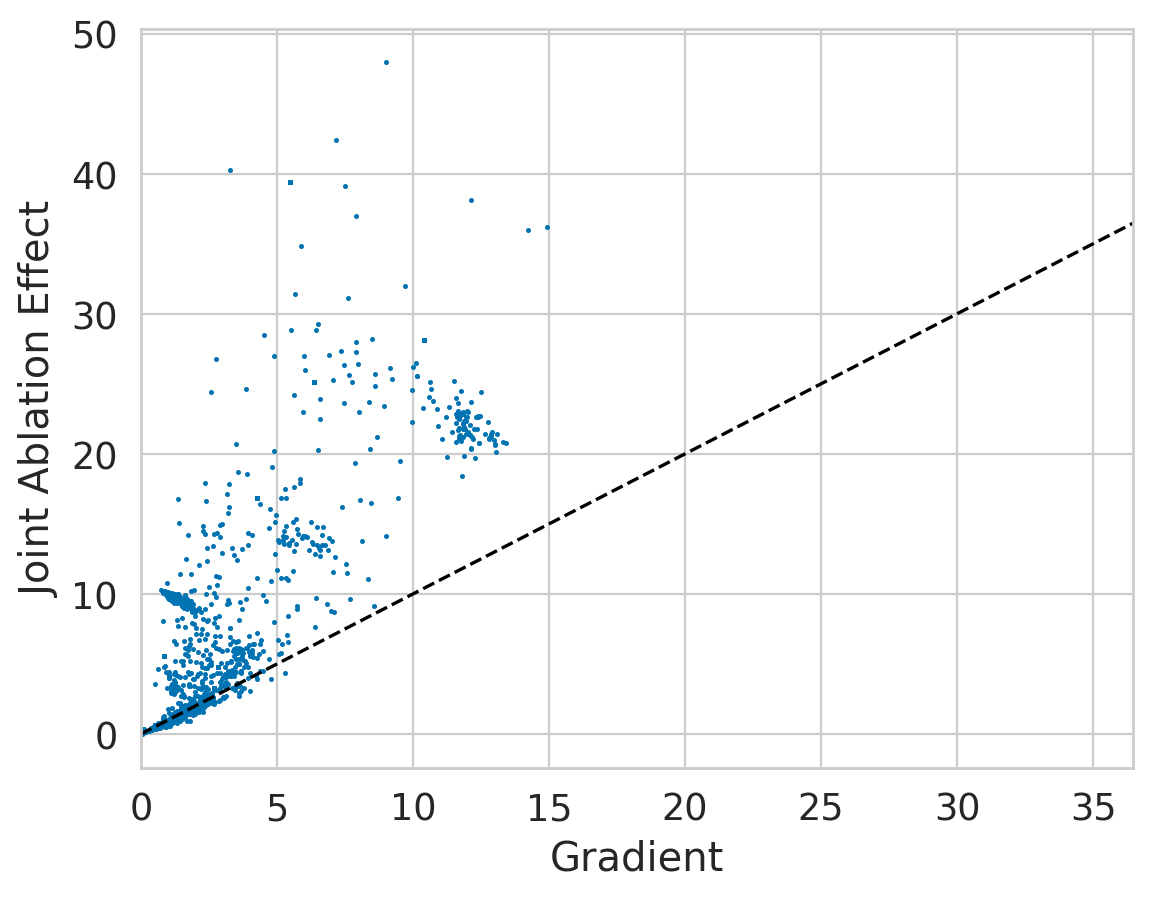}
         \includegraphics[width=\textwidth]{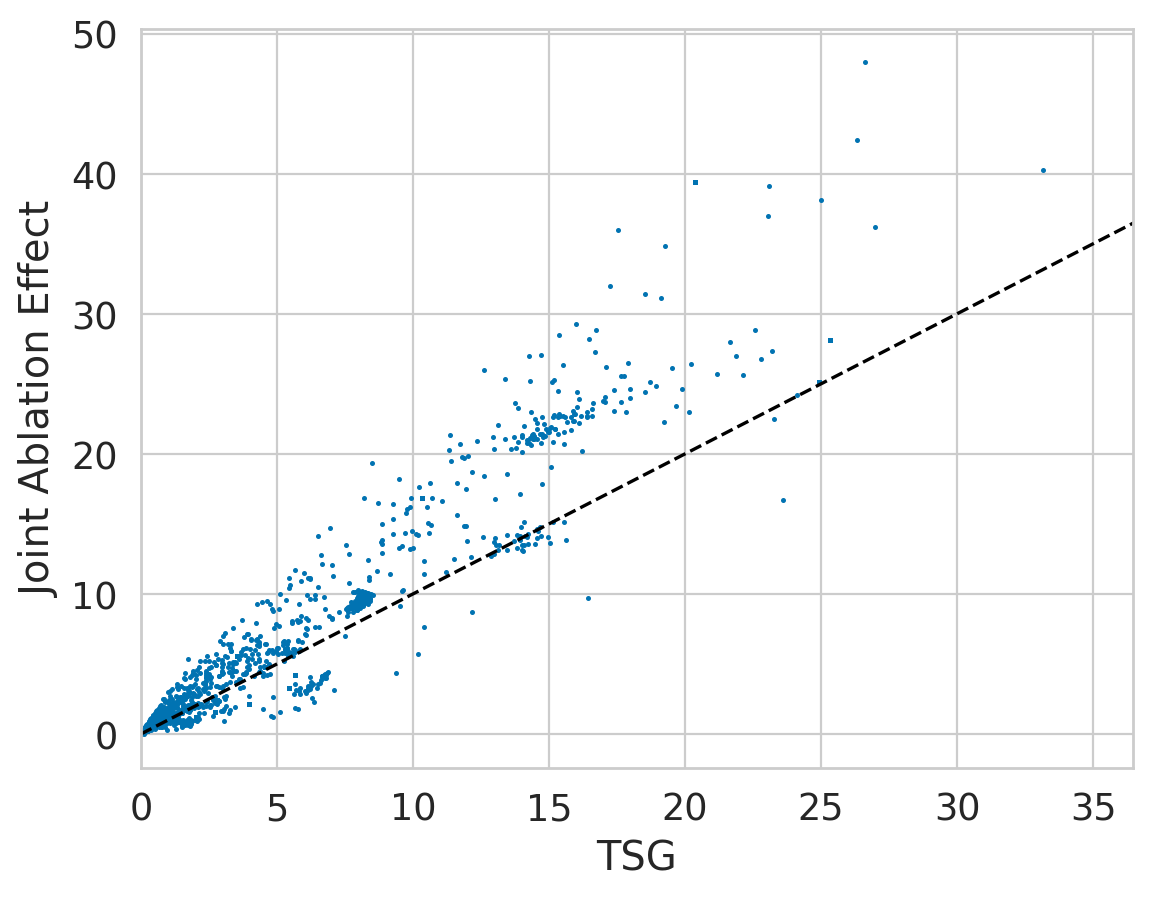}
         \caption{Hatexplain}
     \end{subfigure}
     \hfill
     \begin{subfigure}[b]{0.25\textwidth}
         \centering
         \includegraphics[width=\textwidth]{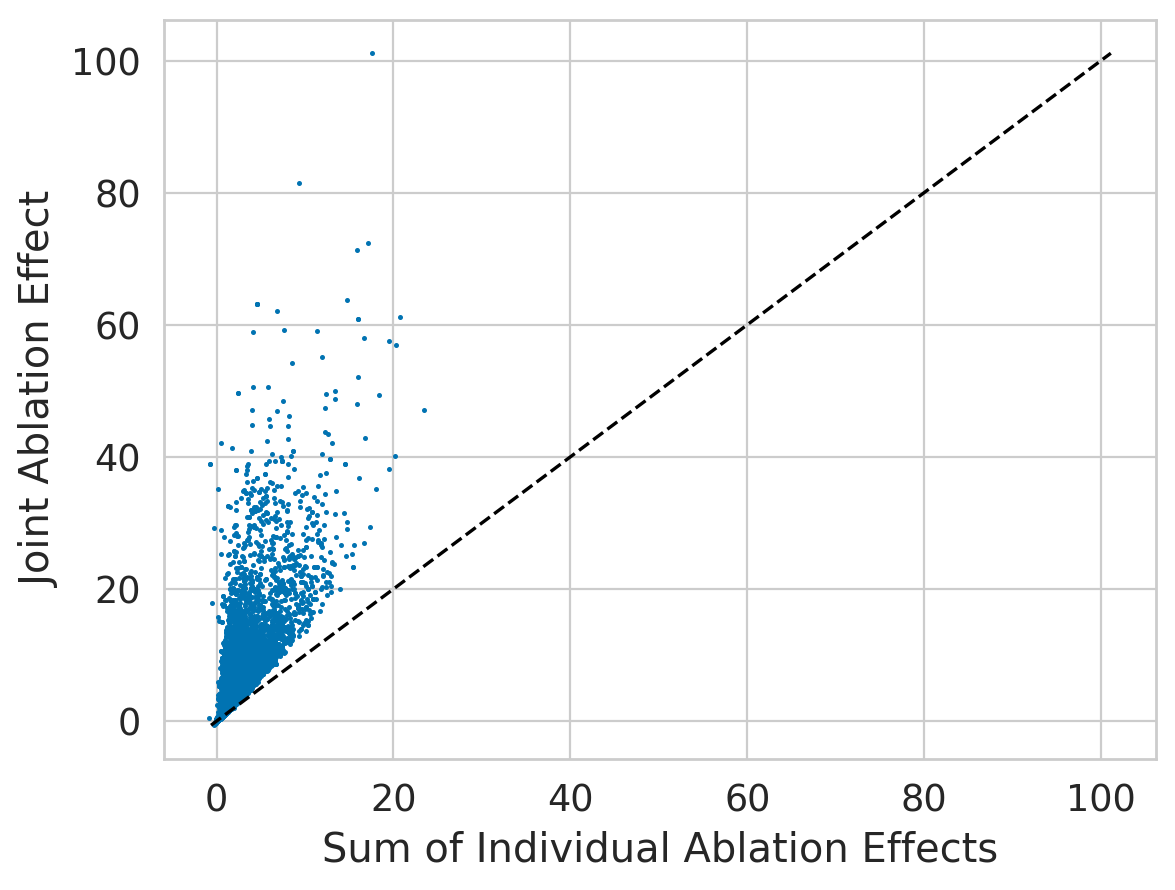}
         \includegraphics[width=\textwidth]{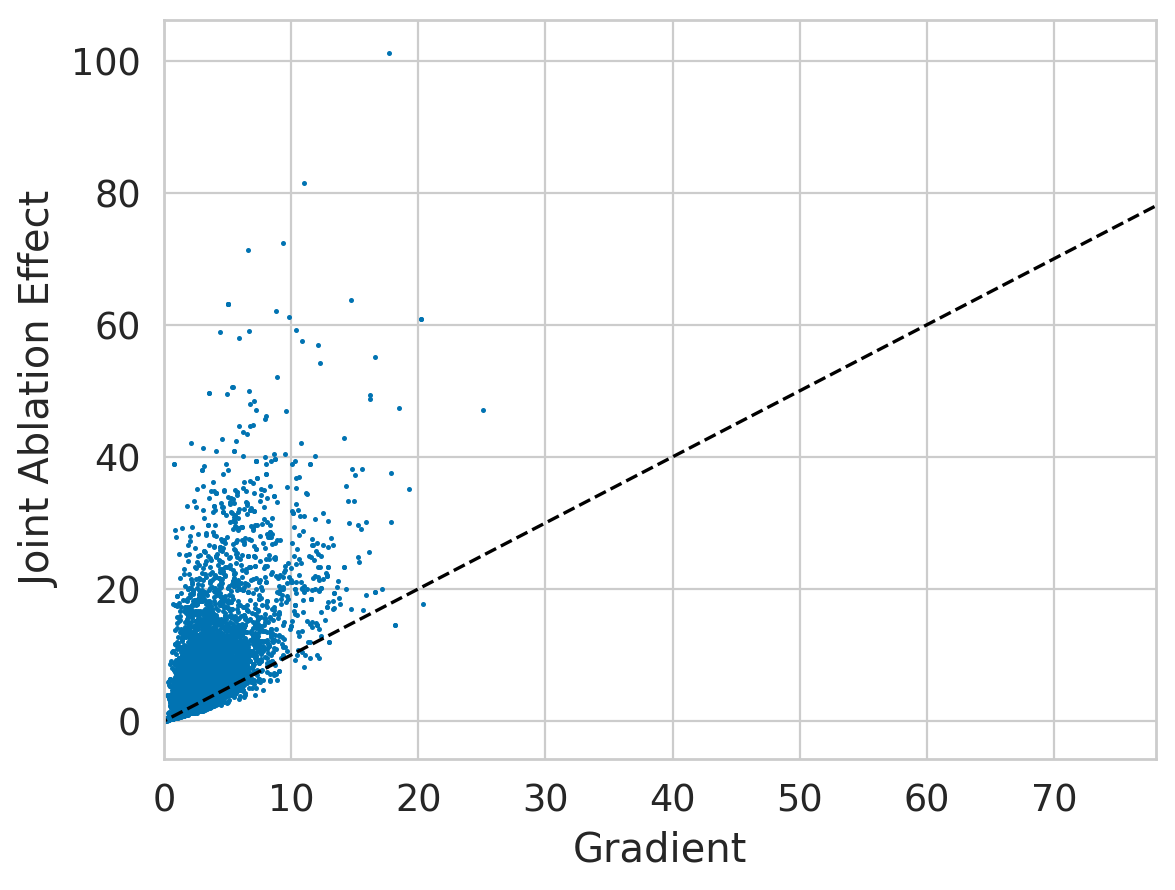}
         \includegraphics[width=\textwidth]{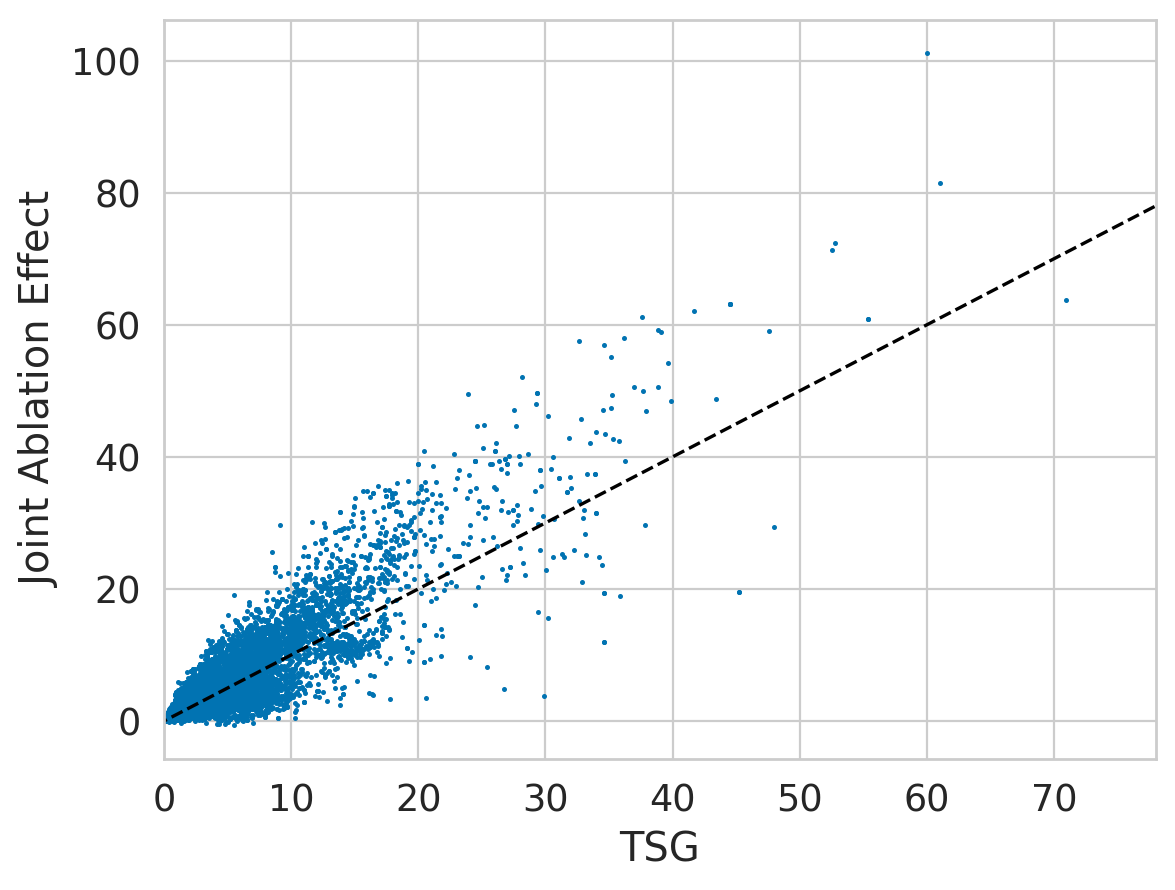}
         
         \caption{Scifact}
     \end{subfigure}
     \hfill
     \begin{subfigure}[b]{0.25\textwidth}
         \centering
         \includegraphics[width=\textwidth]{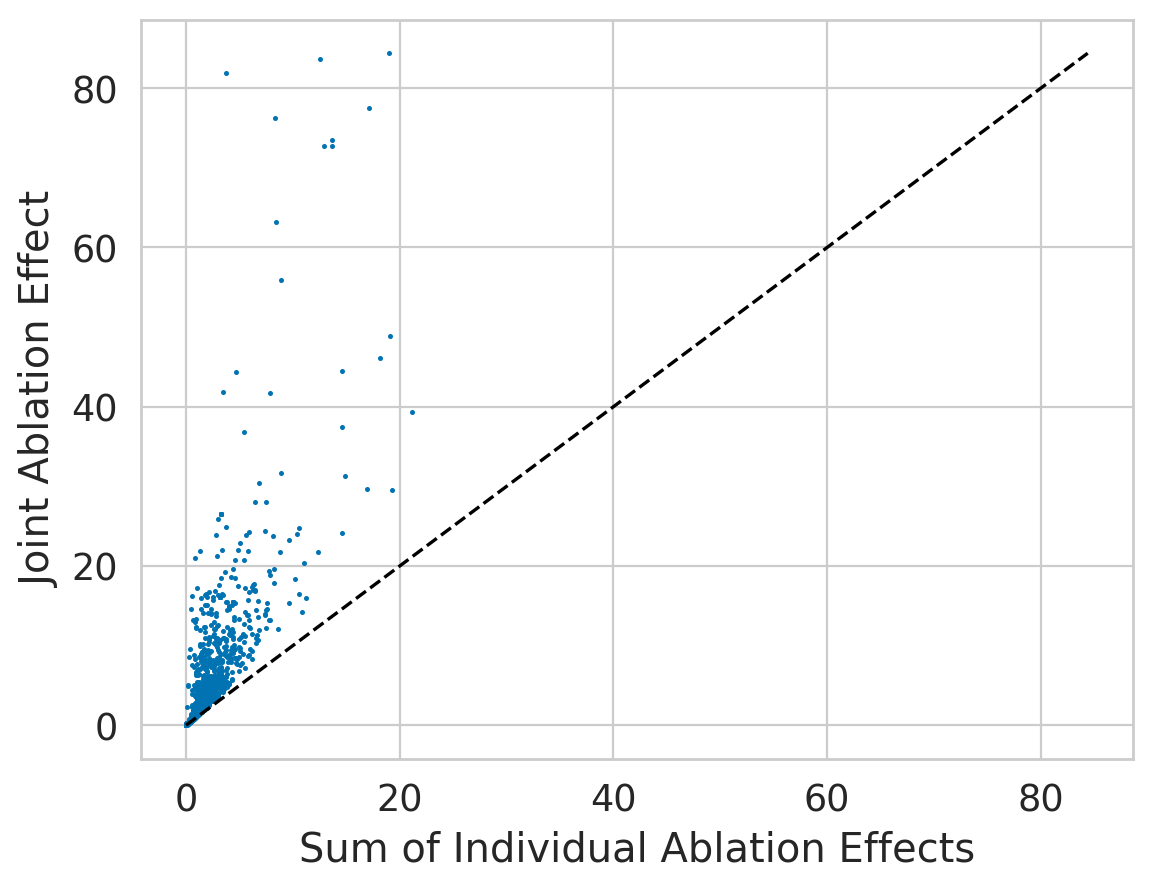}
         \includegraphics[width=\textwidth]{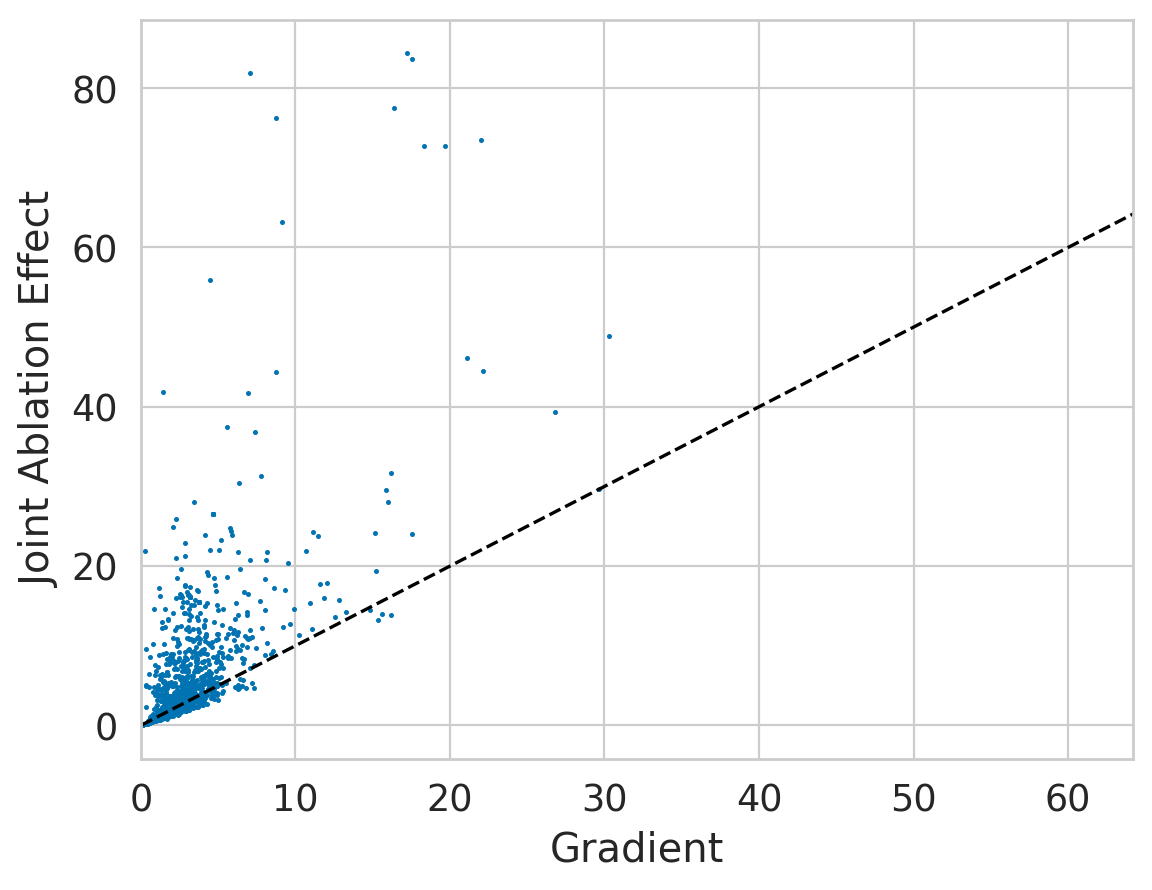}
         \includegraphics[width=\textwidth]{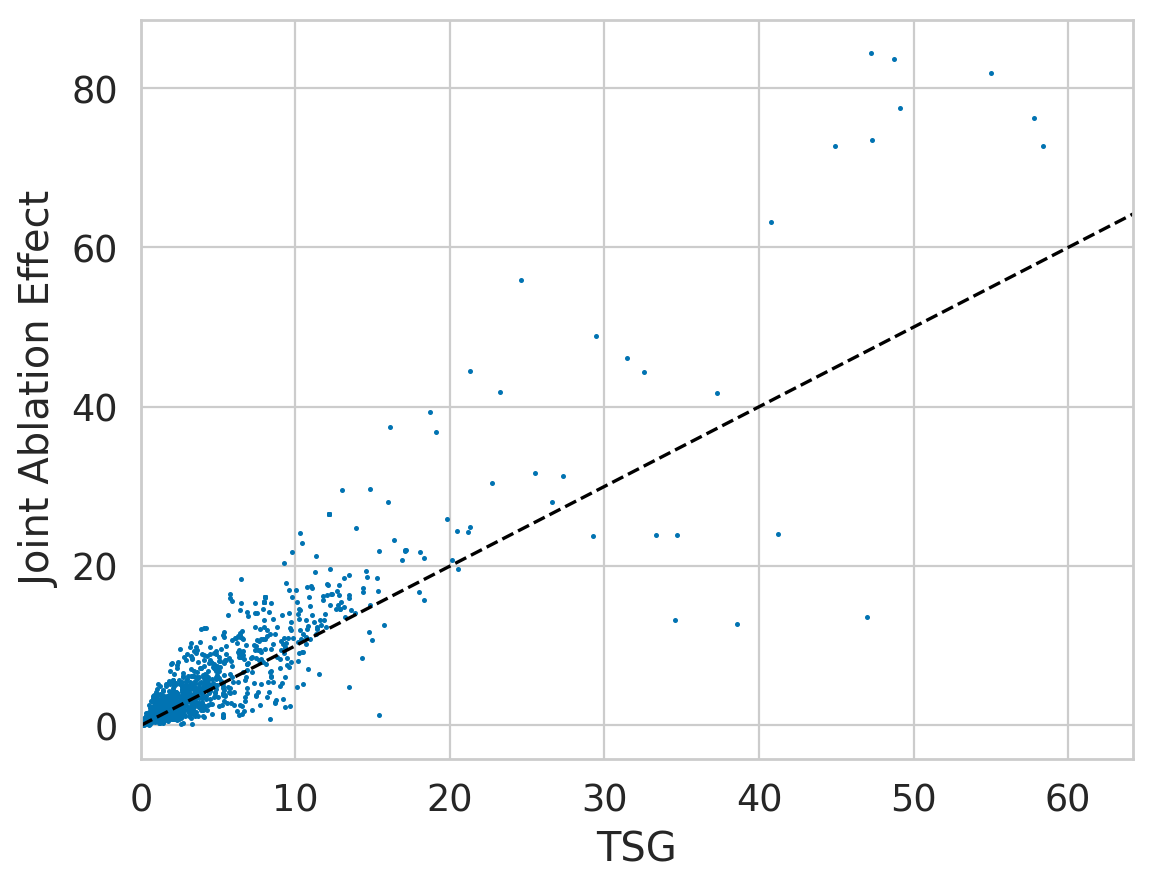}
         \caption{BoolQ}
     \end{subfigure}
     \hfill
     \begin{subfigure}[b]{0.25\textwidth}
         \centering
         \includegraphics[width=\textwidth]{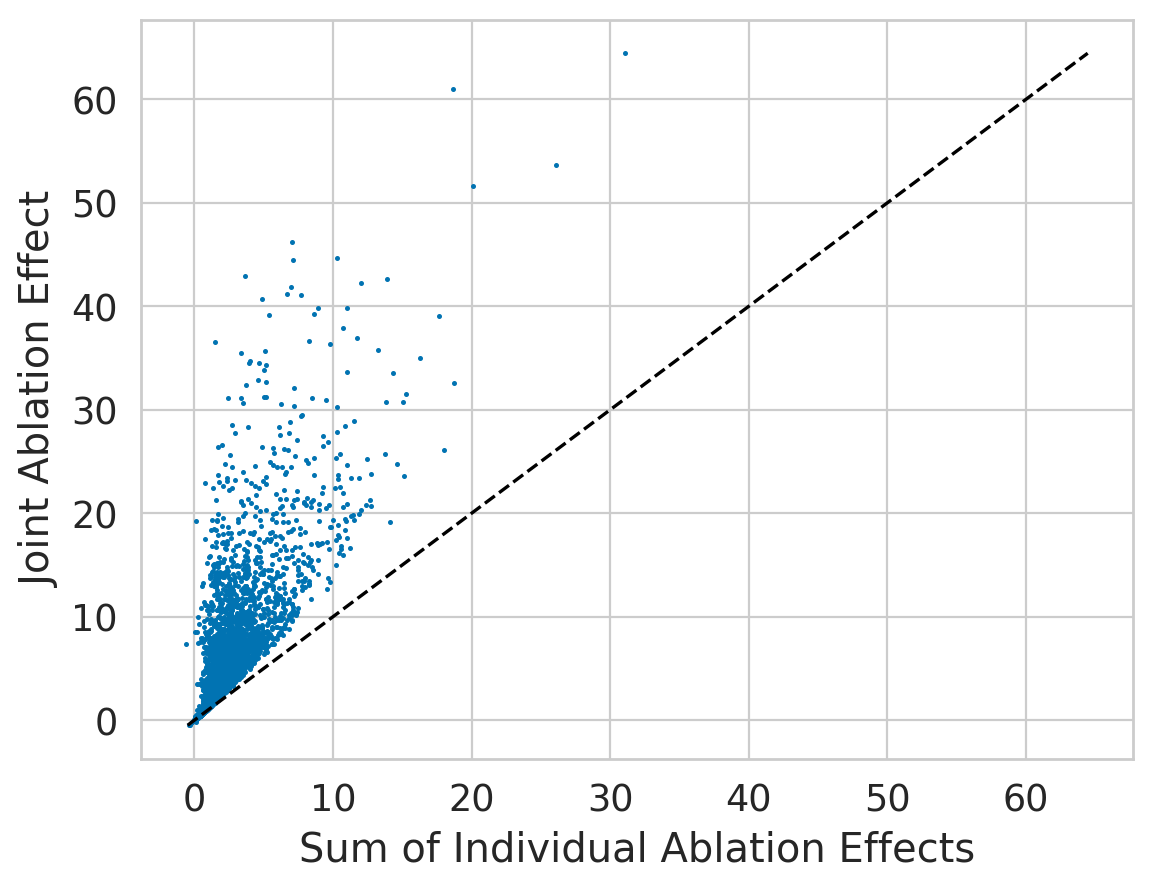}
         \includegraphics[width=\textwidth]{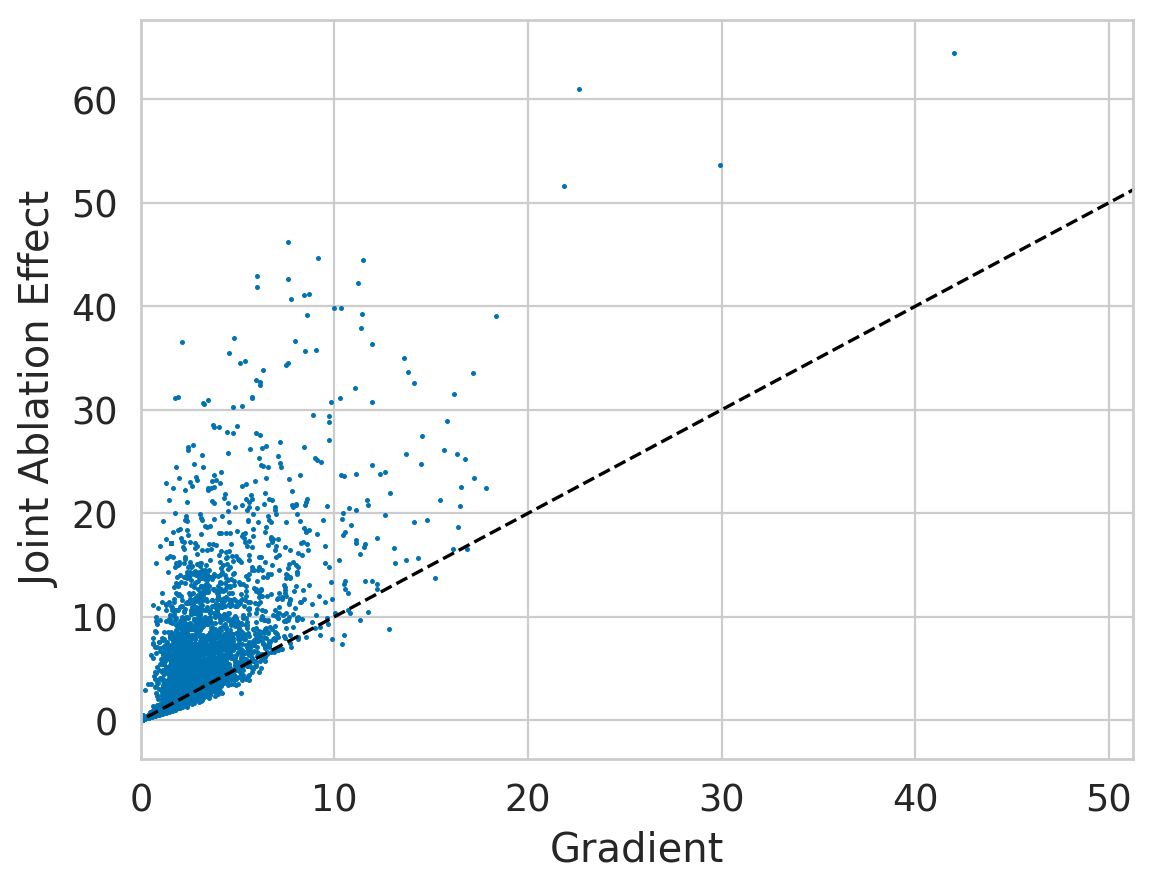}
         \includegraphics[width=\textwidth]{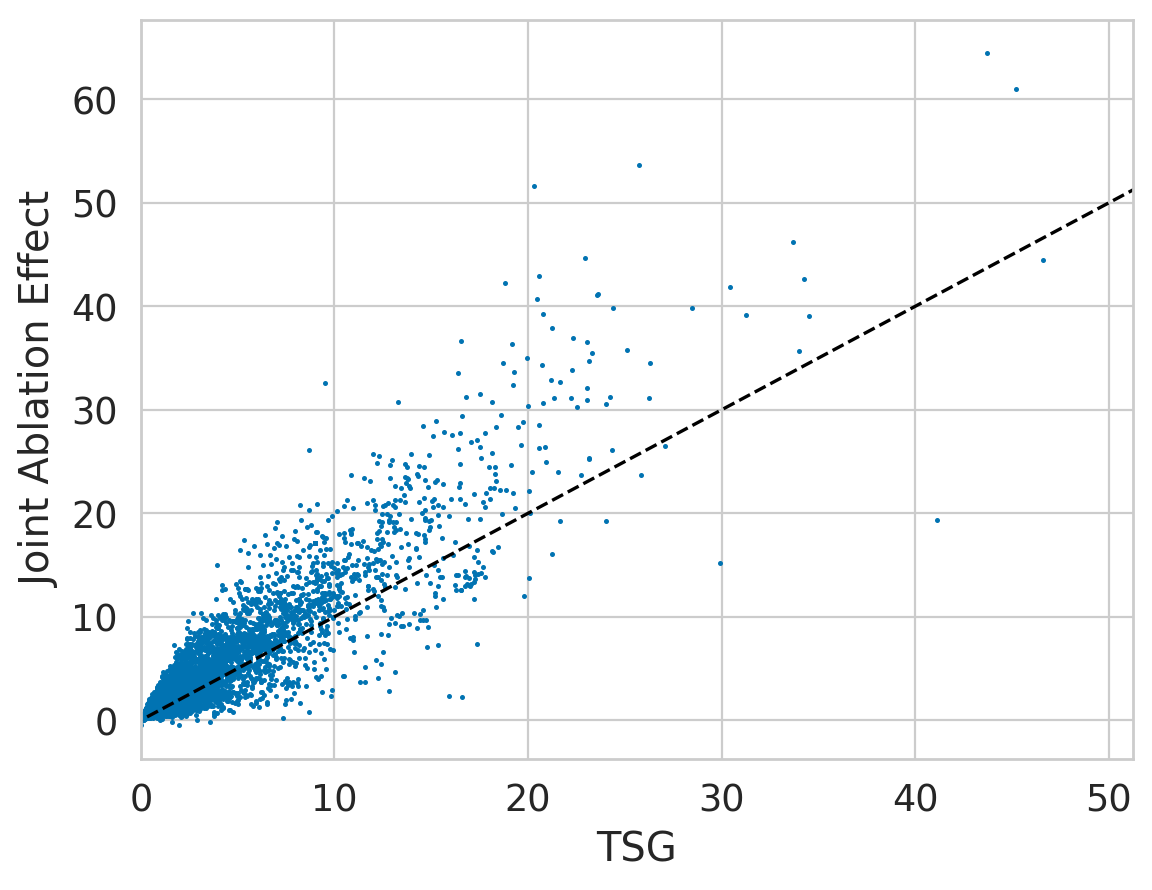}
         \caption{Movie}
     \end{subfigure}
     \hfill
     \begin{subfigure}[b]{0.25\textwidth}
         \centering
         \includegraphics[width=\textwidth]{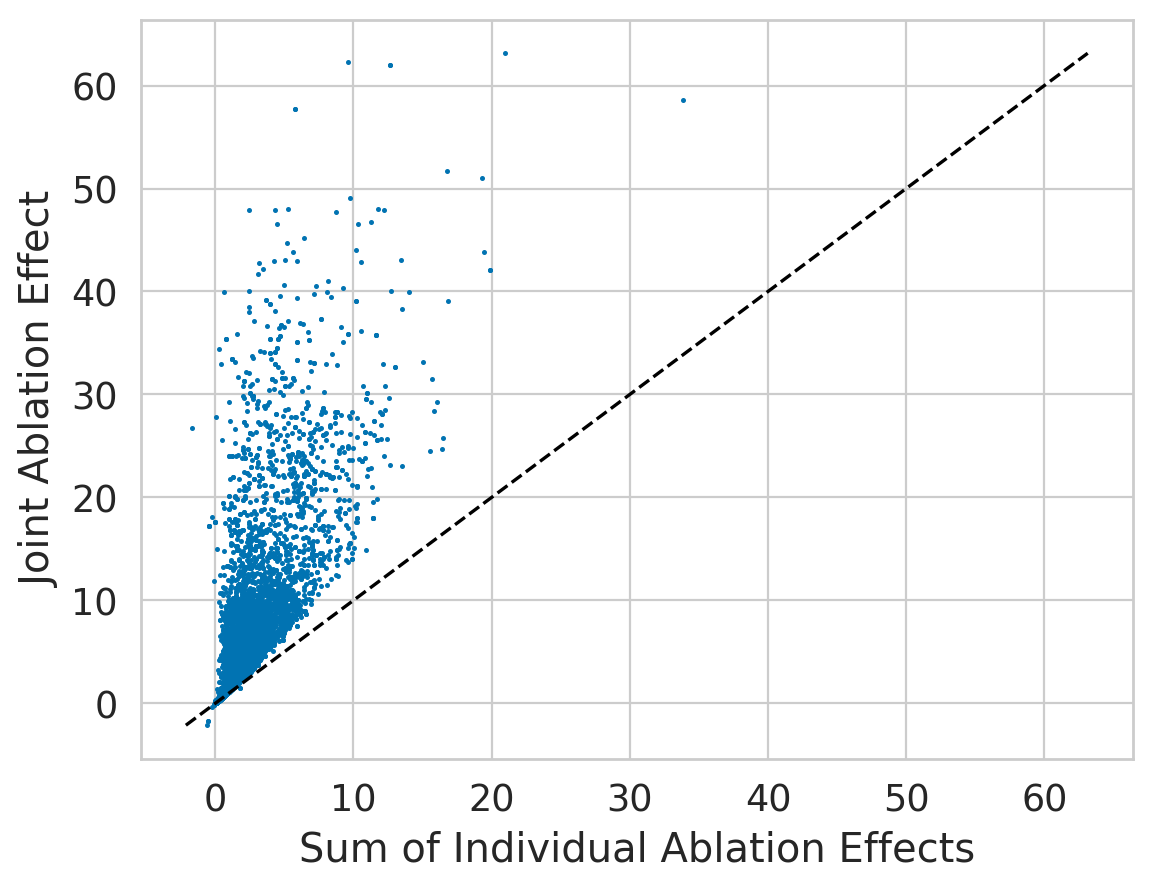}
         \includegraphics[width=\textwidth]{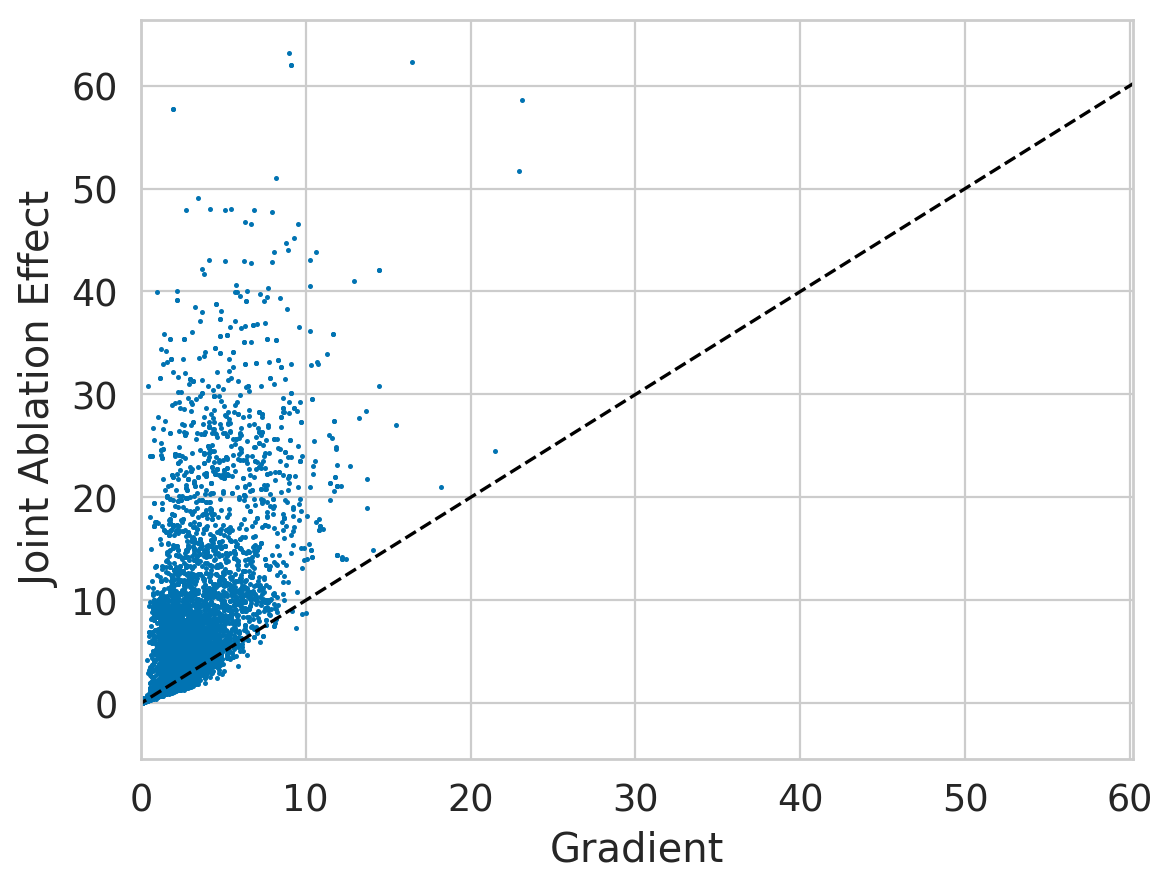}
         \includegraphics[width=\textwidth]{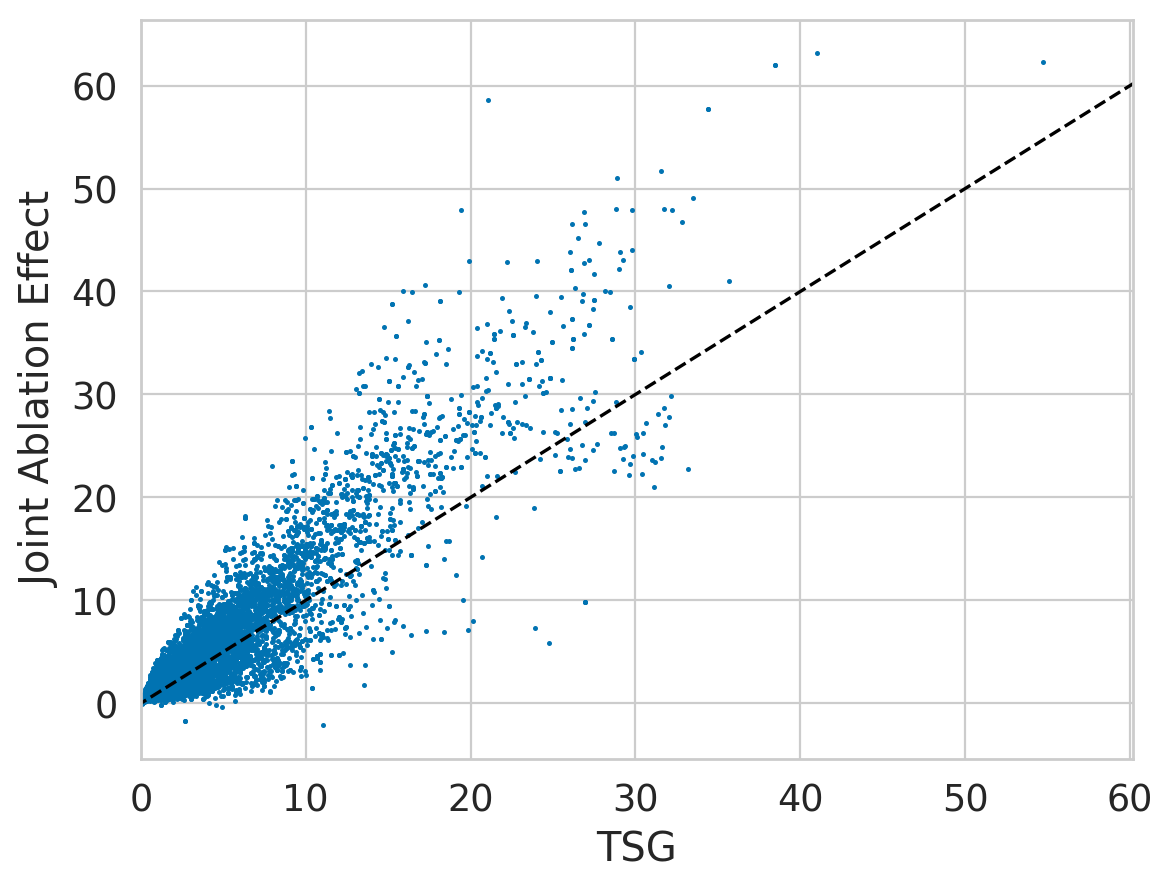}
         \caption{Fever}
     \end{subfigure}
        \caption{Self-repair for \textbf{LLAMA-3.2 3B} and how TSG increases the attributions for the attention scores with the strongest self-repair effects}
        \label{fig:self-repair-llama3}
\end{figure*}

\begin{figure*}[h]
     \centering
     \begin{subfigure}[b]{0.25\textwidth}
         \centering
         \includegraphics[width=\textwidth]{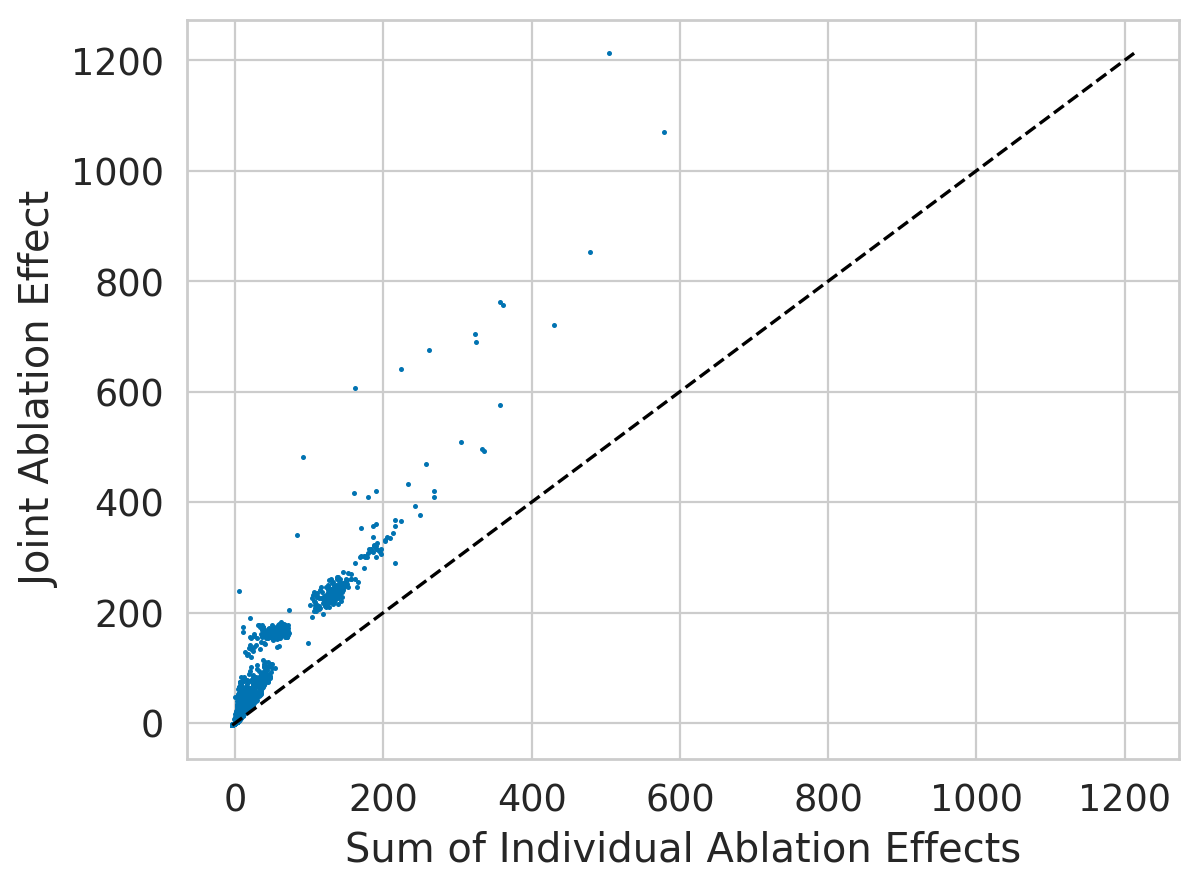}
         \includegraphics[width=\textwidth]{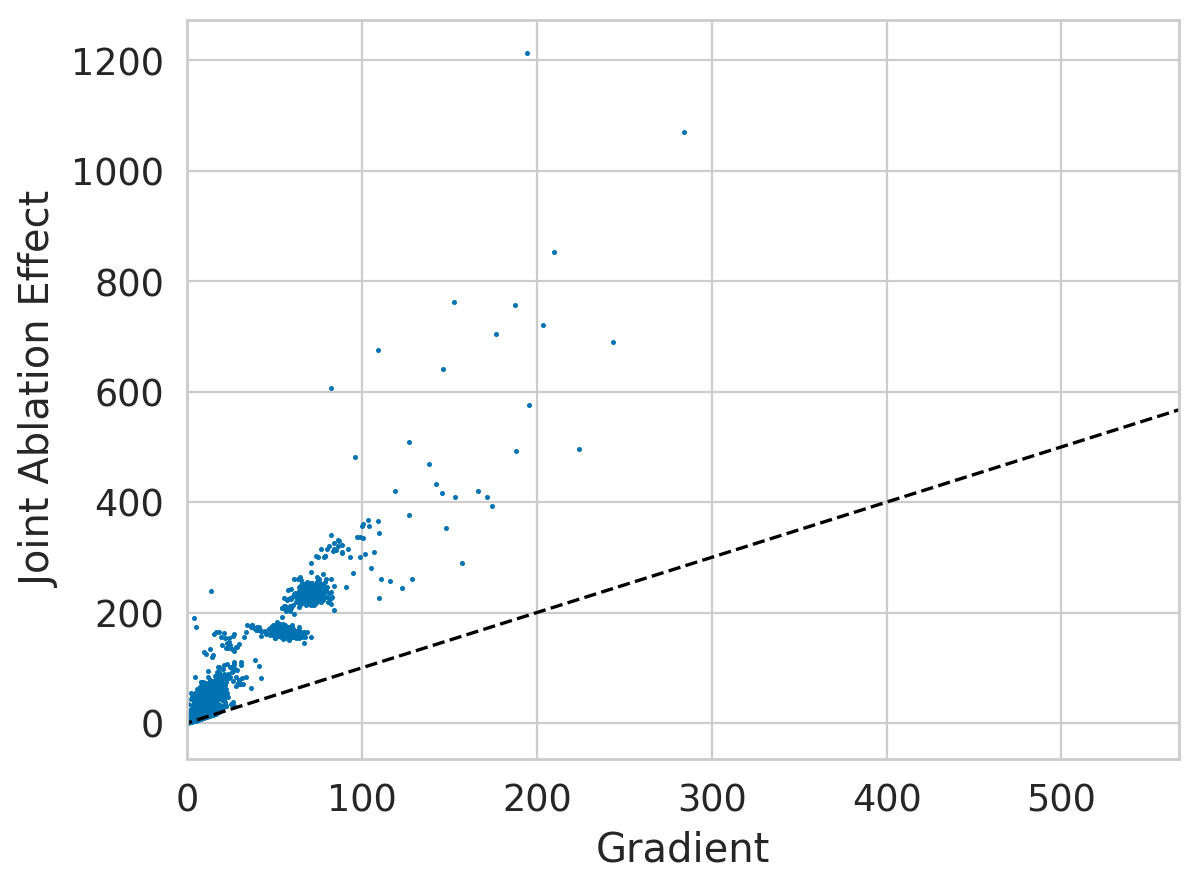}
         \includegraphics[width=\textwidth]{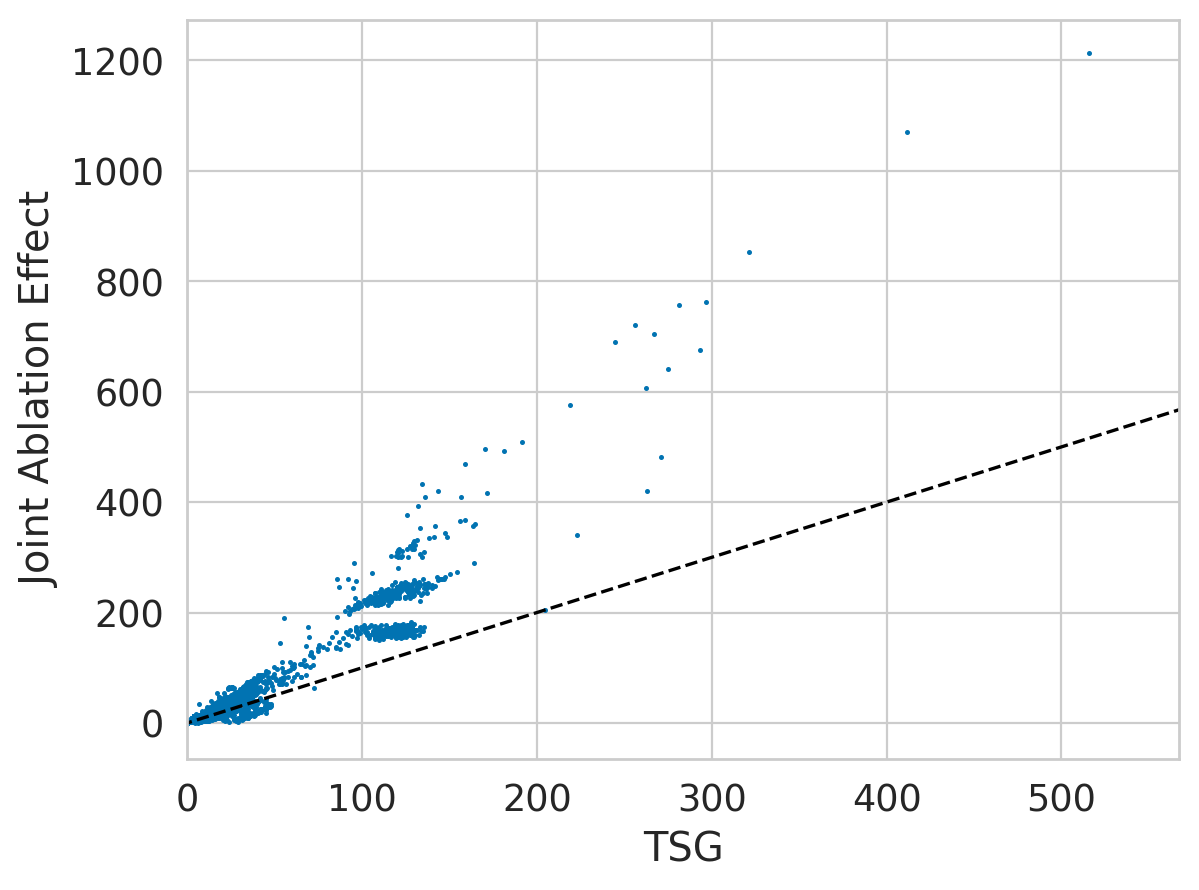}
         \caption{Twitter}
     \end{subfigure}
     \hfill
     \begin{subfigure}[b]{0.25\textwidth}
         \centering
         \includegraphics[width=\textwidth]{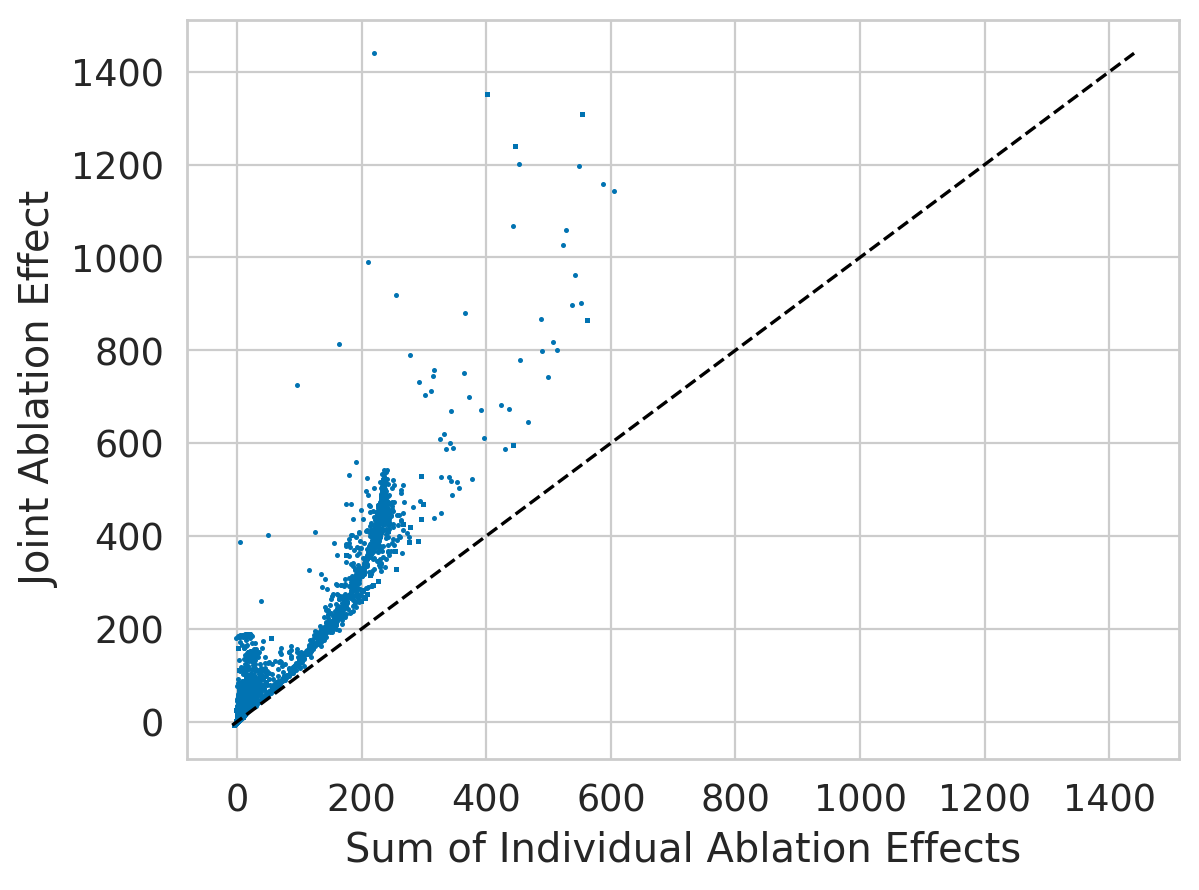}
         \includegraphics[width=\textwidth]{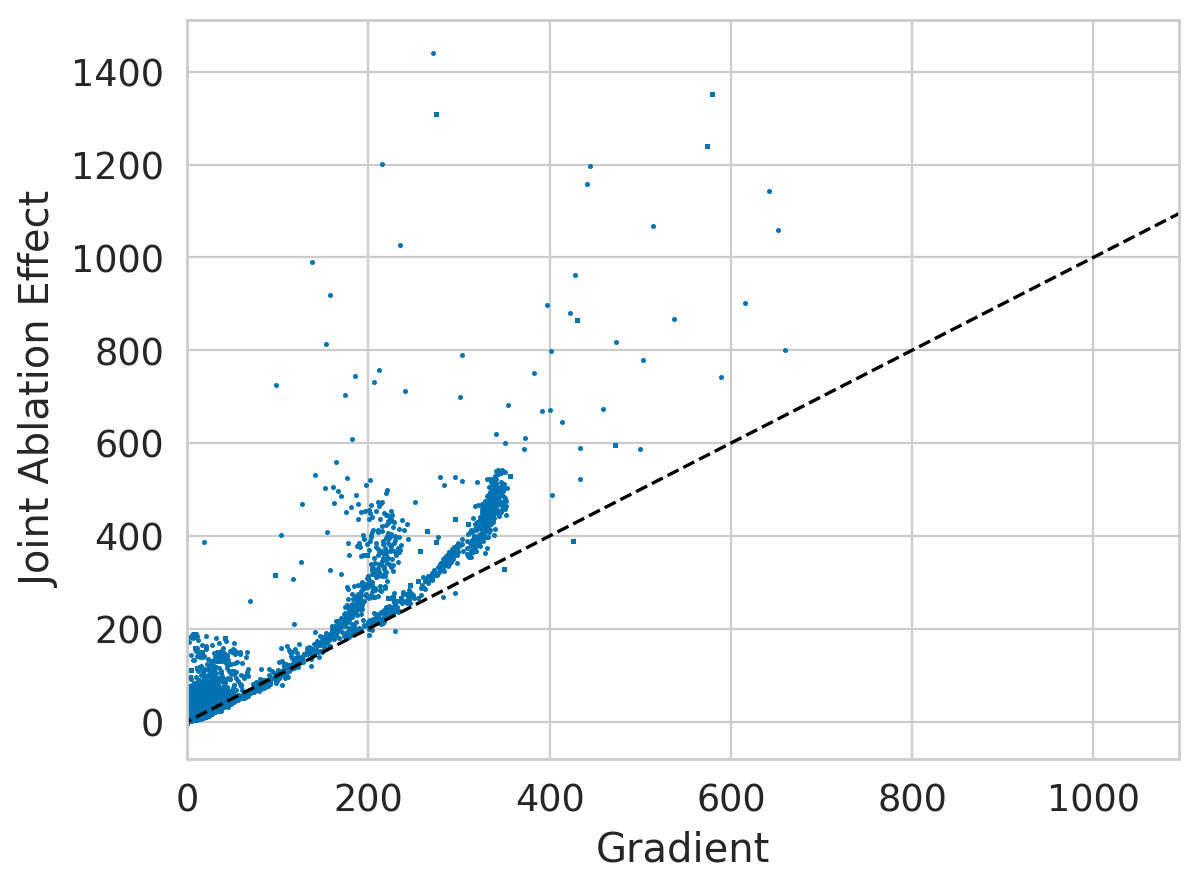}
         \includegraphics[width=\textwidth]{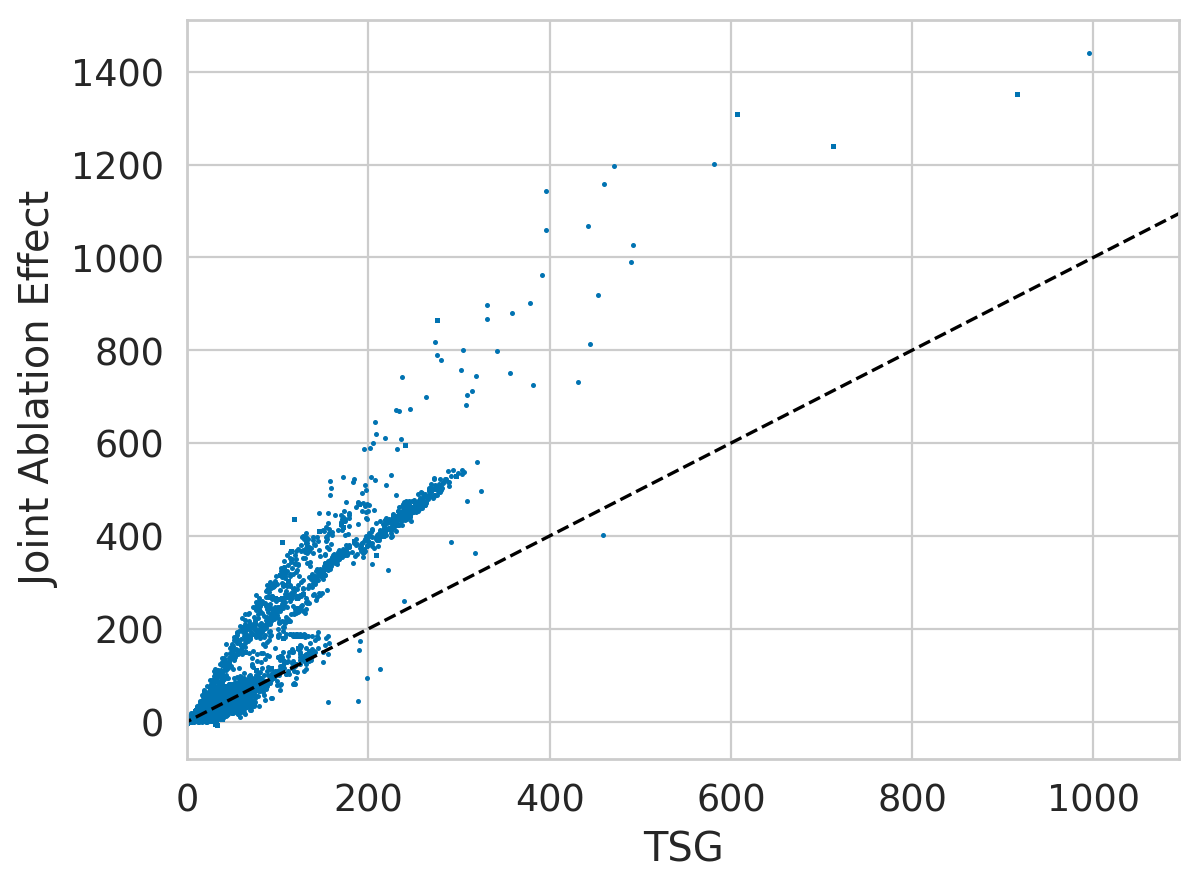}
         \caption{Hatexplain}
     \end{subfigure}
     \hfill
     \begin{subfigure}[b]{0.25\textwidth}
         \centering
         \includegraphics[width=\textwidth]{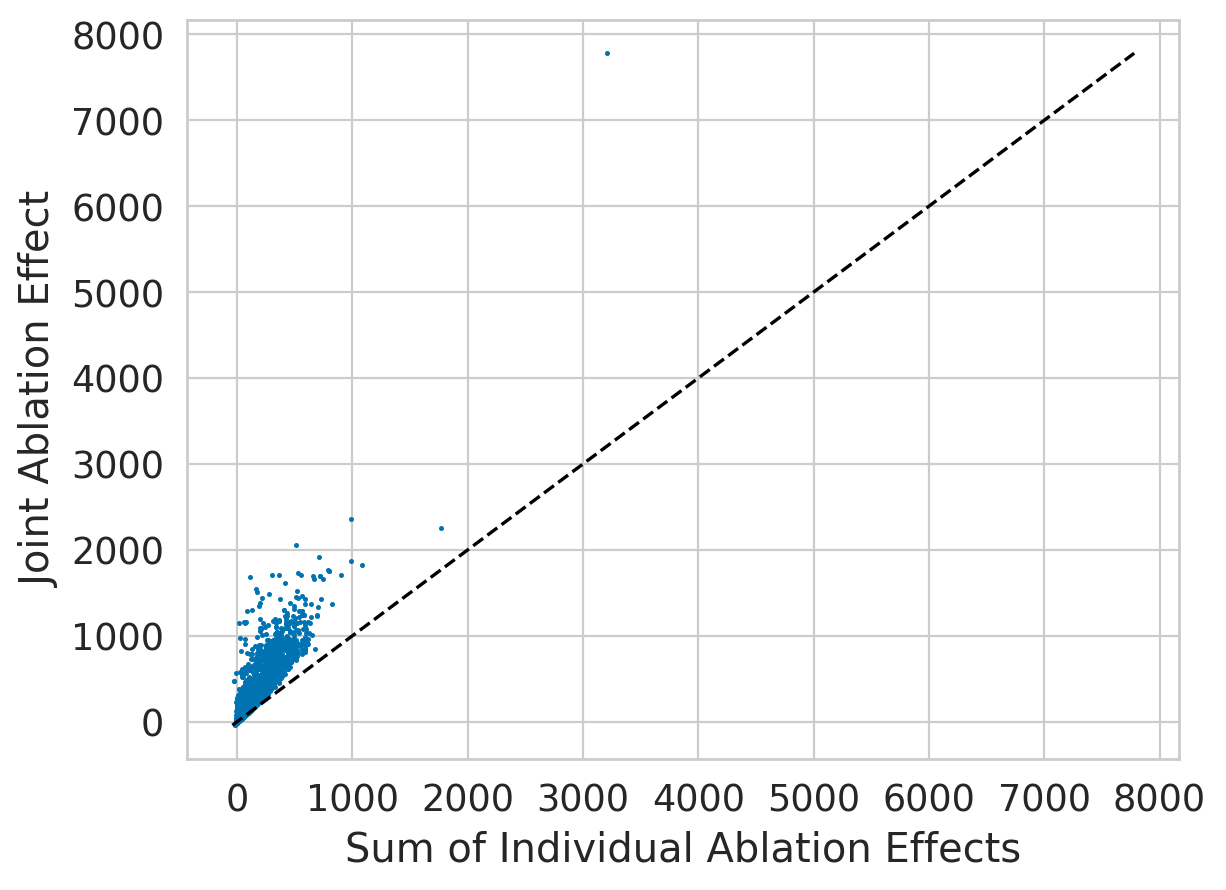}
         \includegraphics[width=\textwidth]{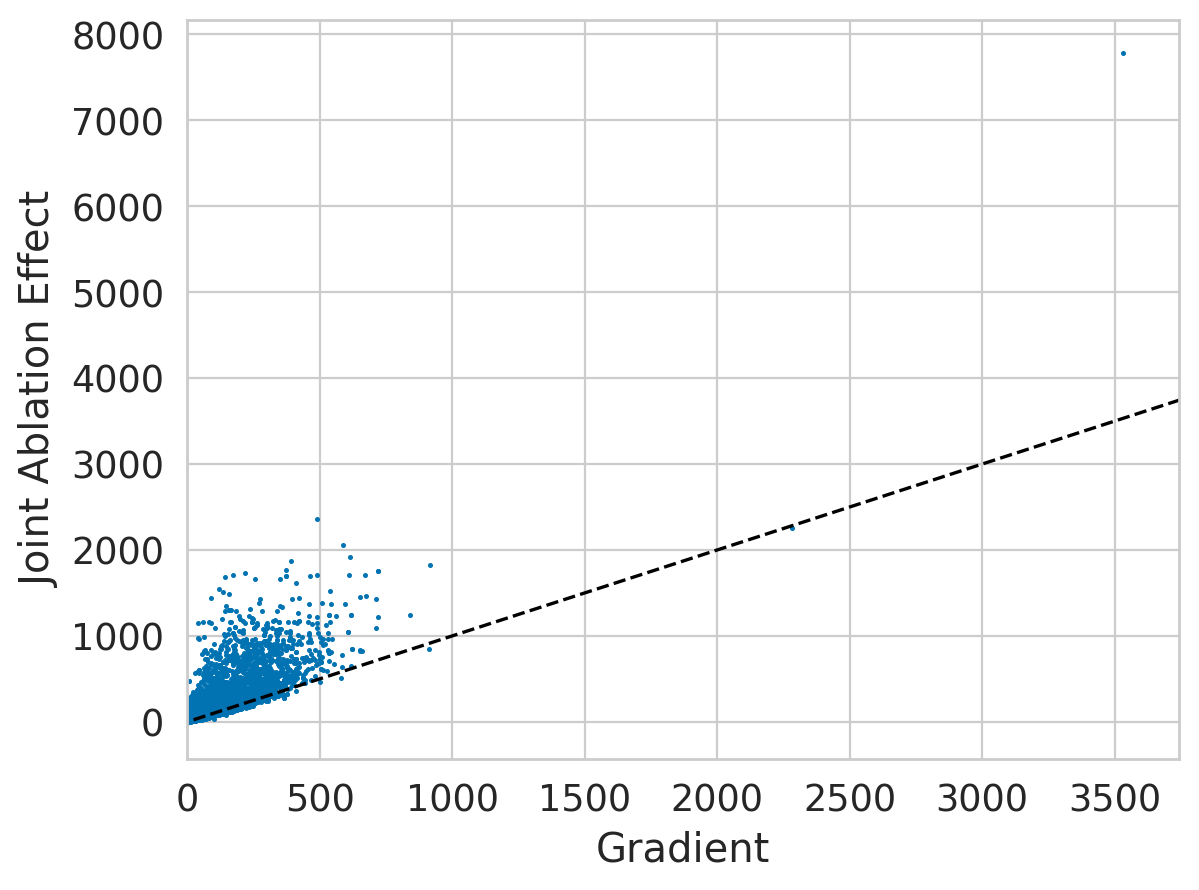}
         \includegraphics[width=\textwidth]{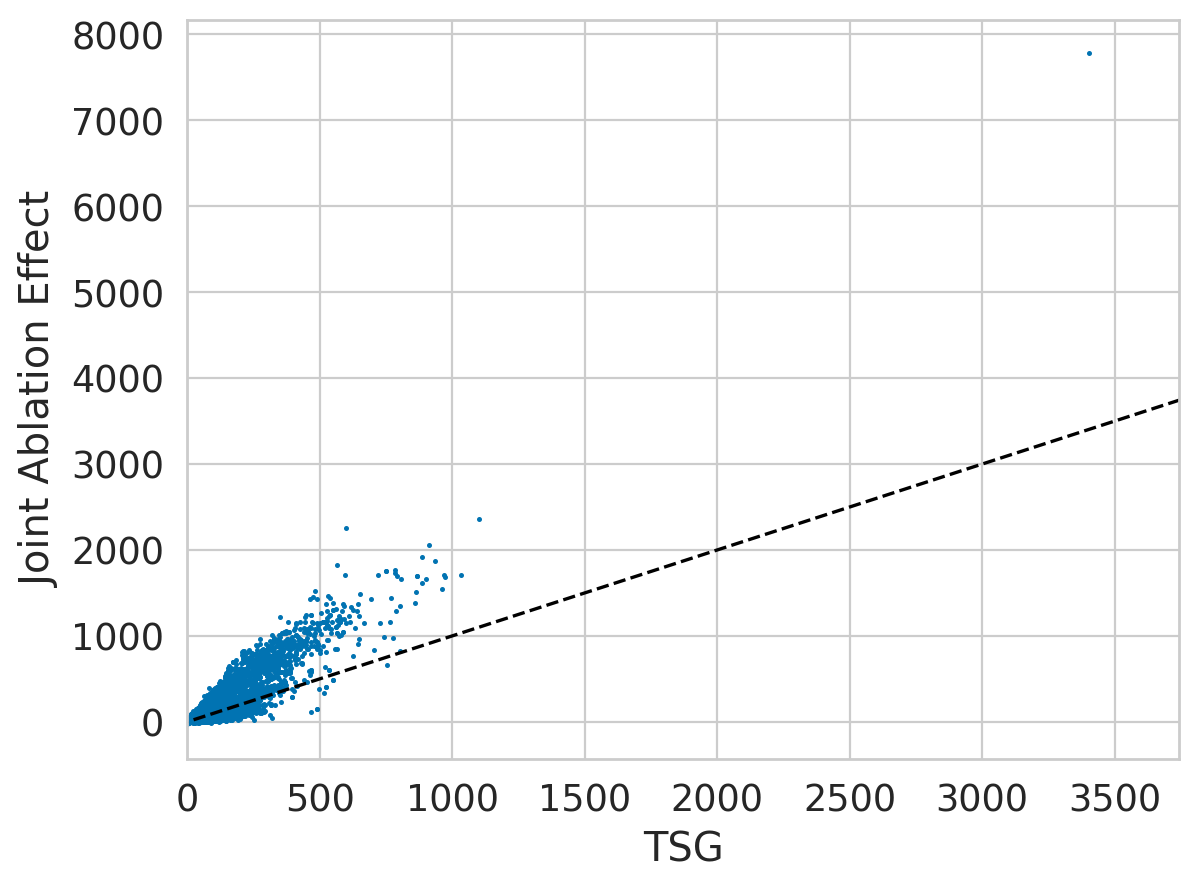}
         
         \caption{Scifact}
     \end{subfigure}
     \hfill
     \begin{subfigure}[b]{0.25\textwidth}
         \centering
         \includegraphics[width=\textwidth]{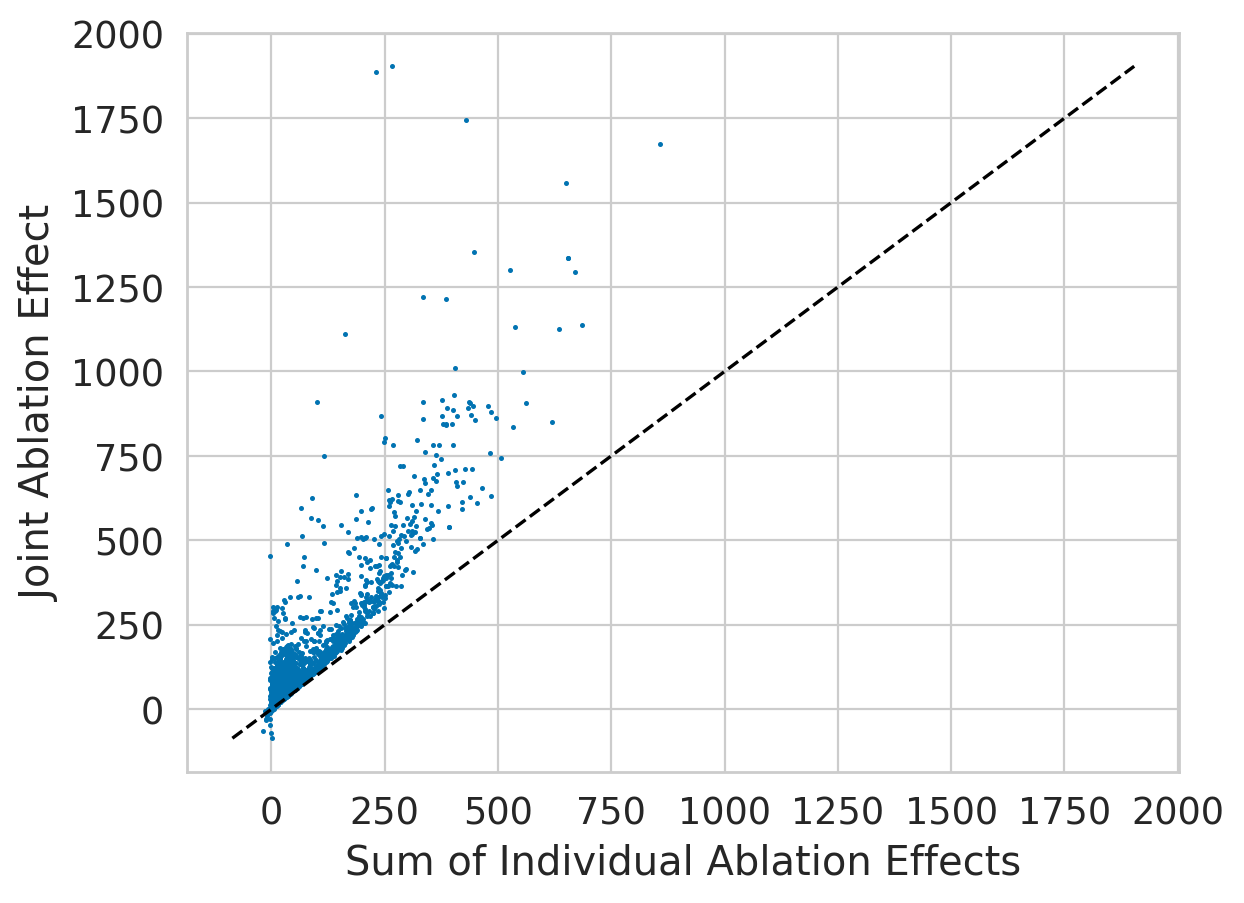}
         \includegraphics[width=\textwidth]{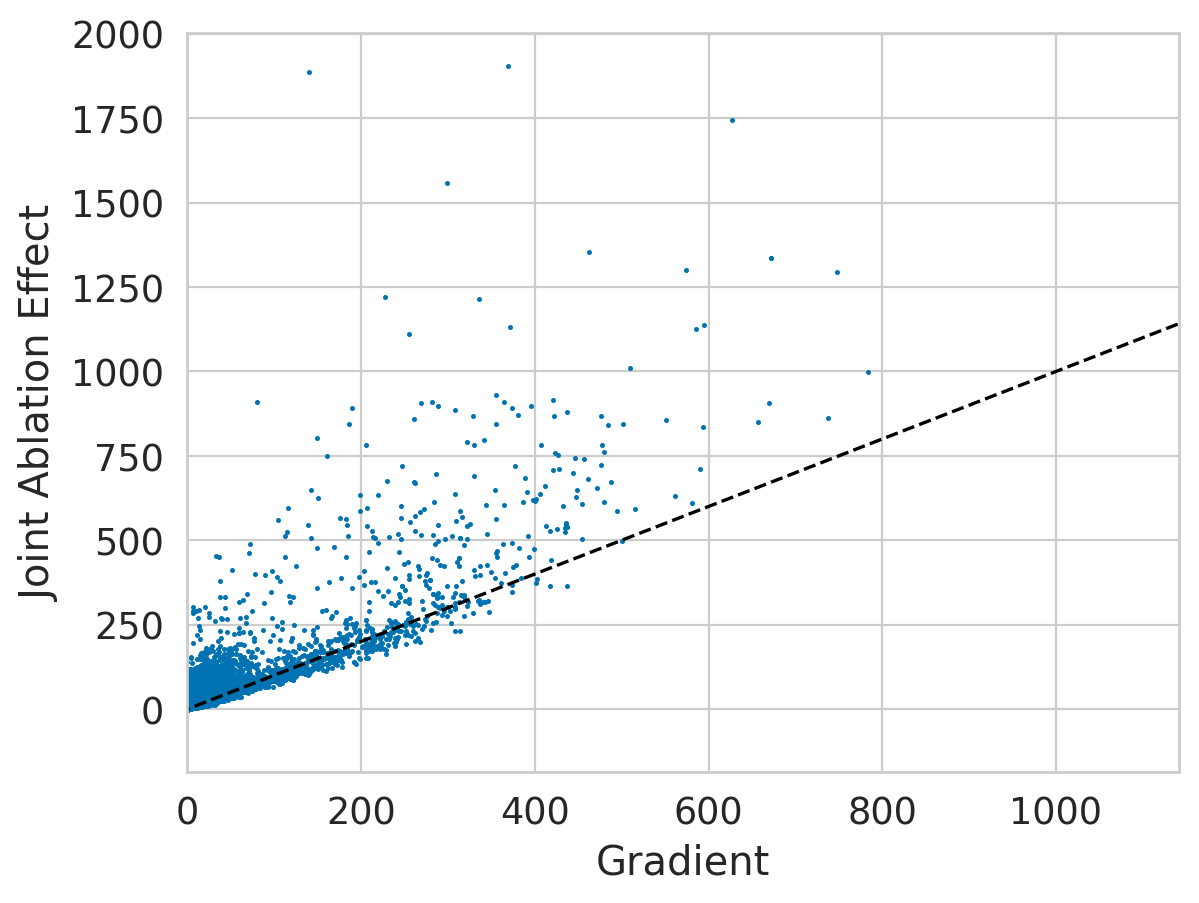}
         \includegraphics[width=\textwidth]{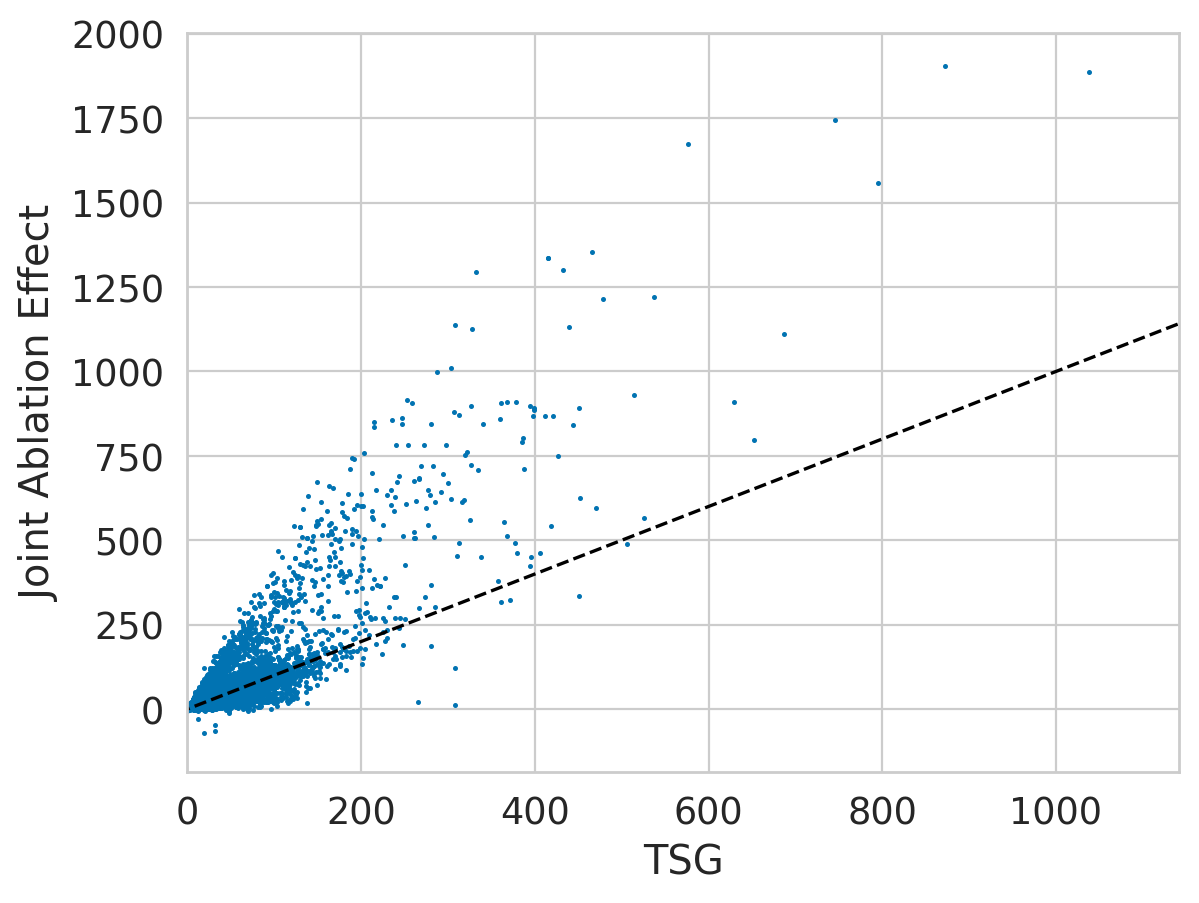}
         \caption{BoolQ}
     \end{subfigure}
     \hfill
     \begin{subfigure}[b]{0.25\textwidth}
         \centering
         \includegraphics[width=\textwidth]{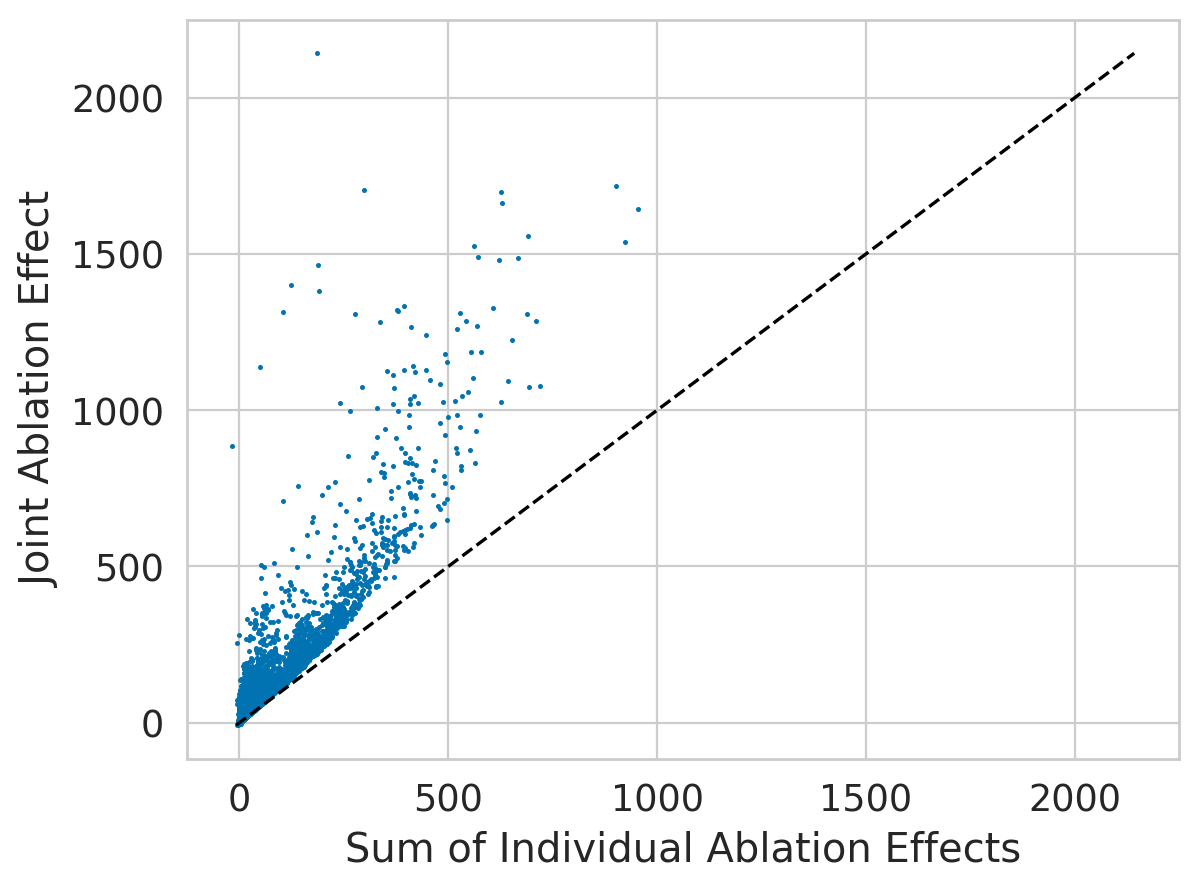}
         \includegraphics[width=\textwidth]{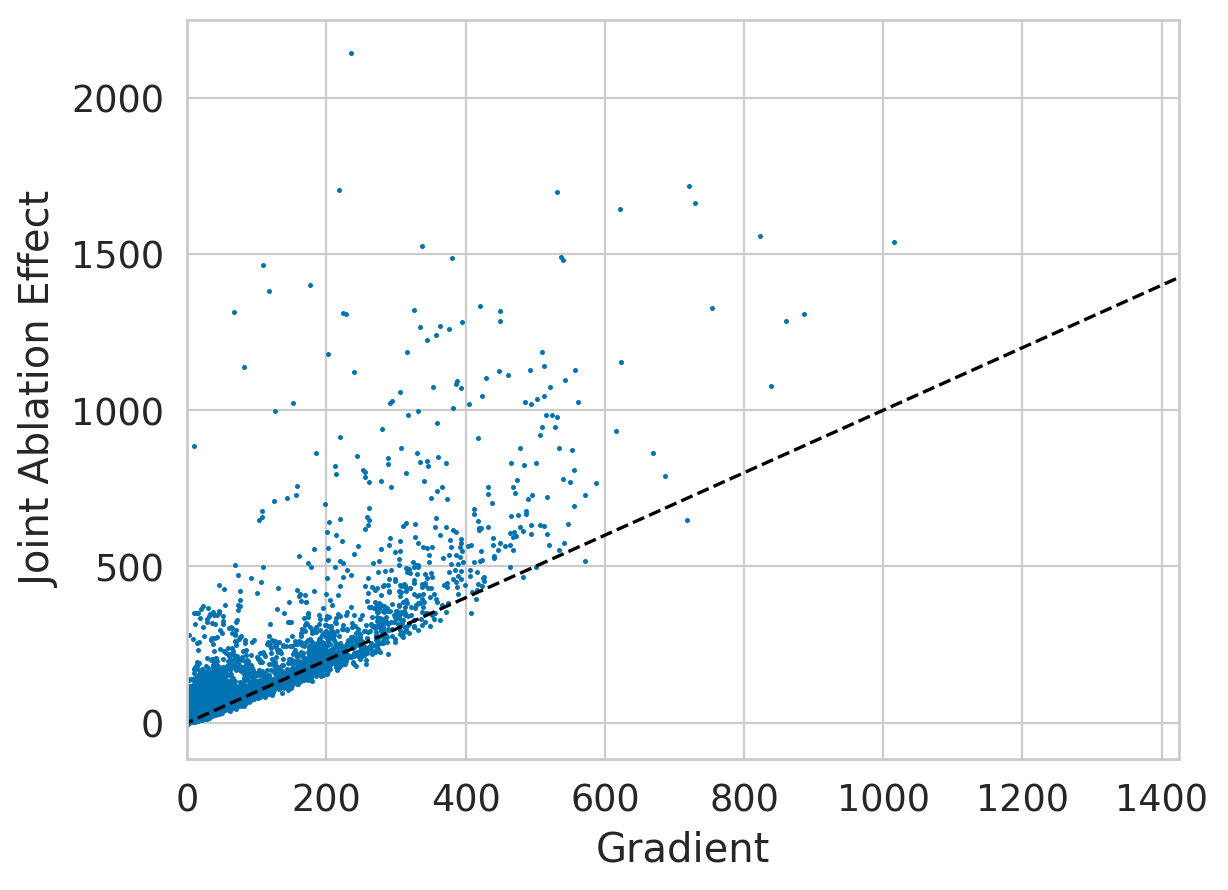}
         \includegraphics[width=\textwidth]{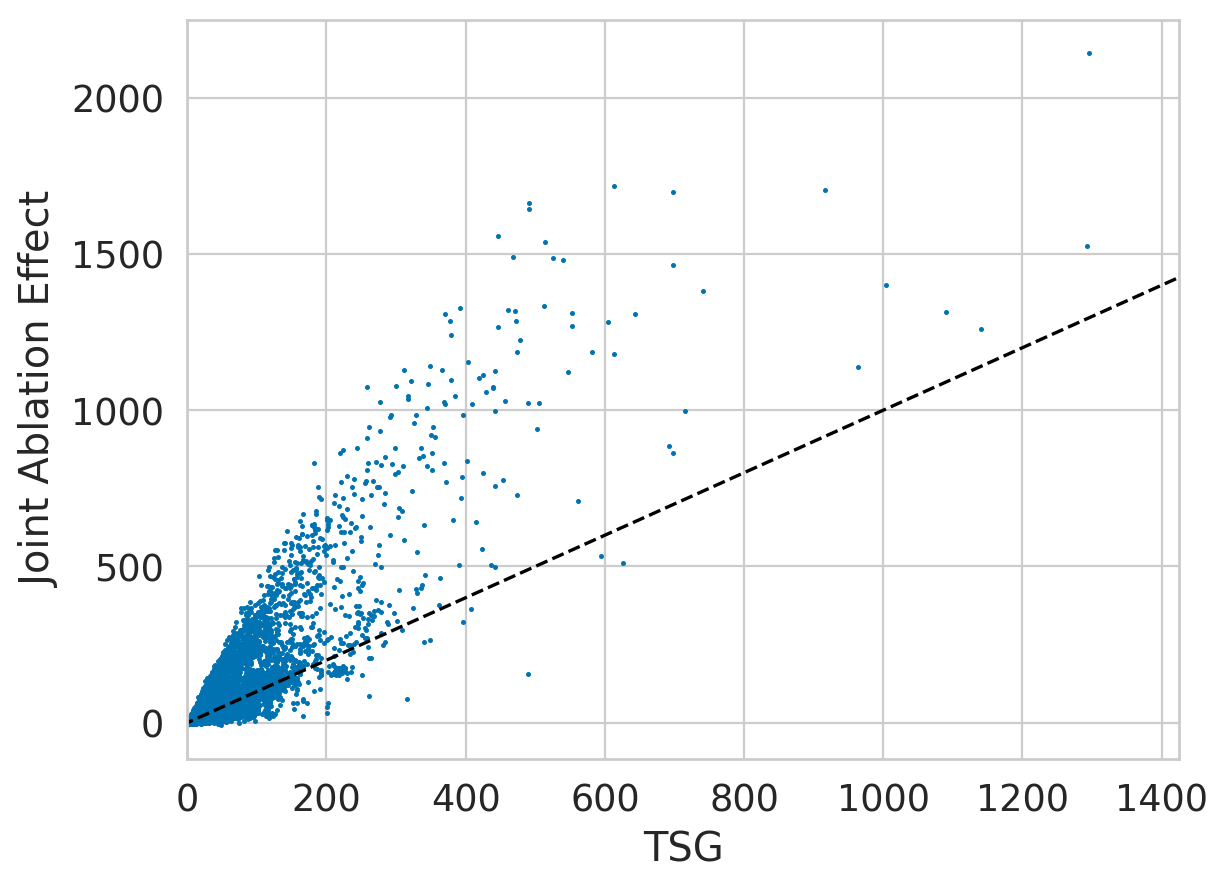}
         \caption{Movie}
     \end{subfigure}
     \hfill
     \begin{subfigure}[b]{0.25\textwidth}
         \centering
         \includegraphics[width=\textwidth]{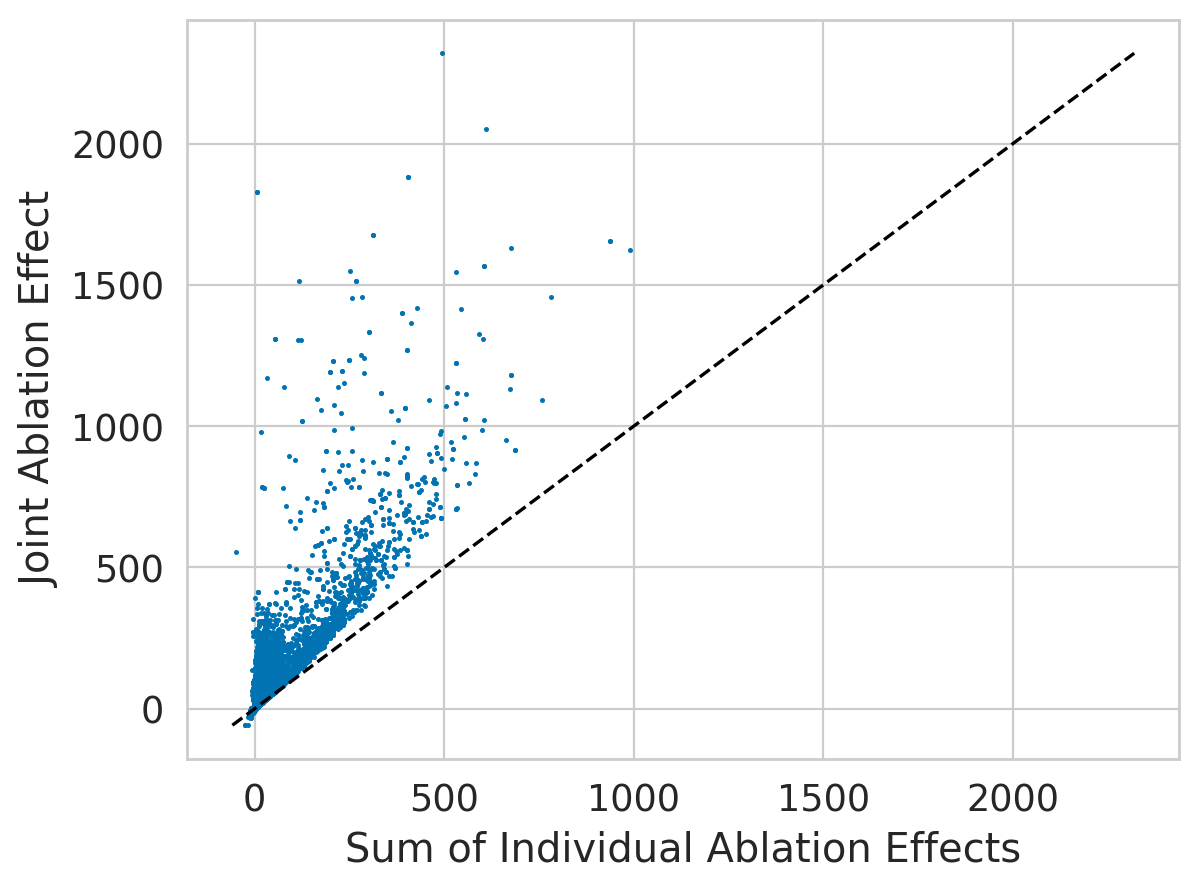}
         \includegraphics[width=\textwidth]{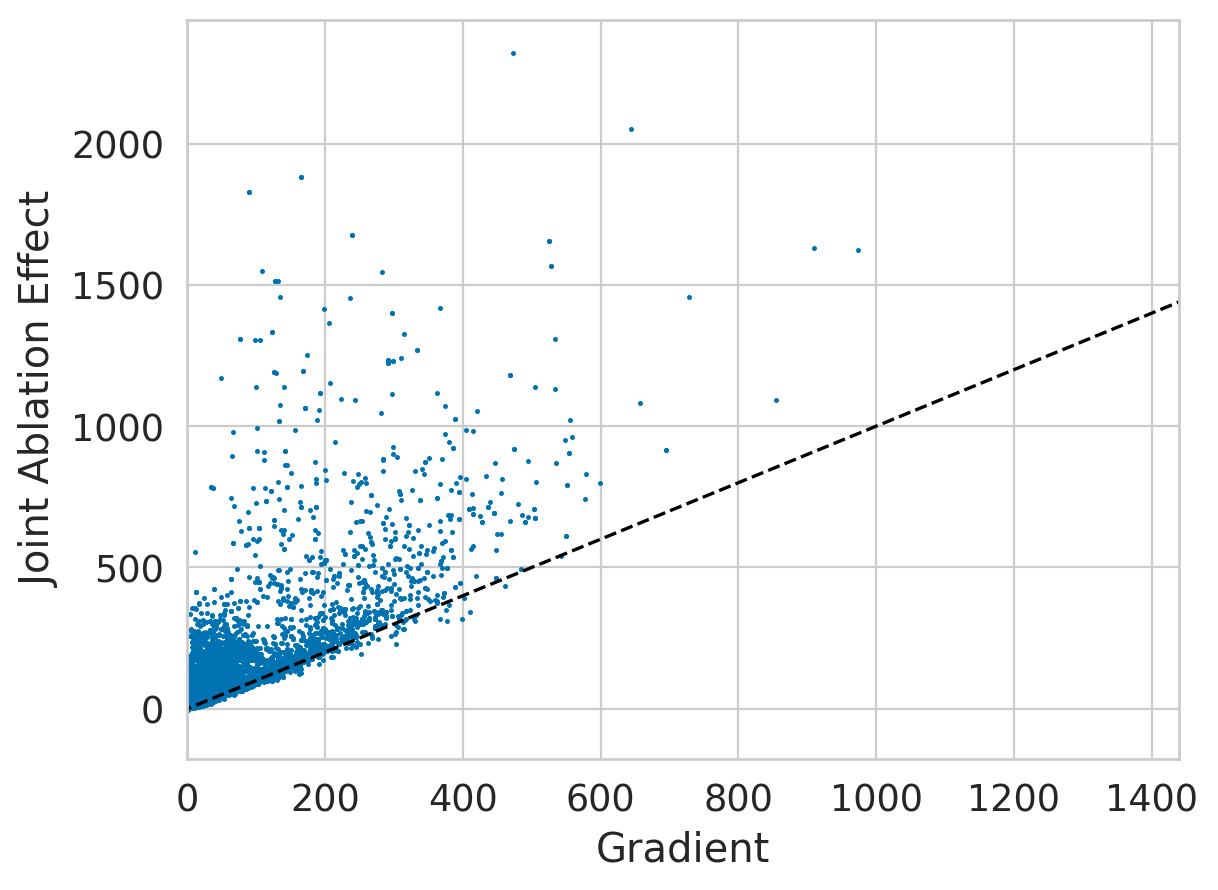}
         \includegraphics[width=\textwidth]{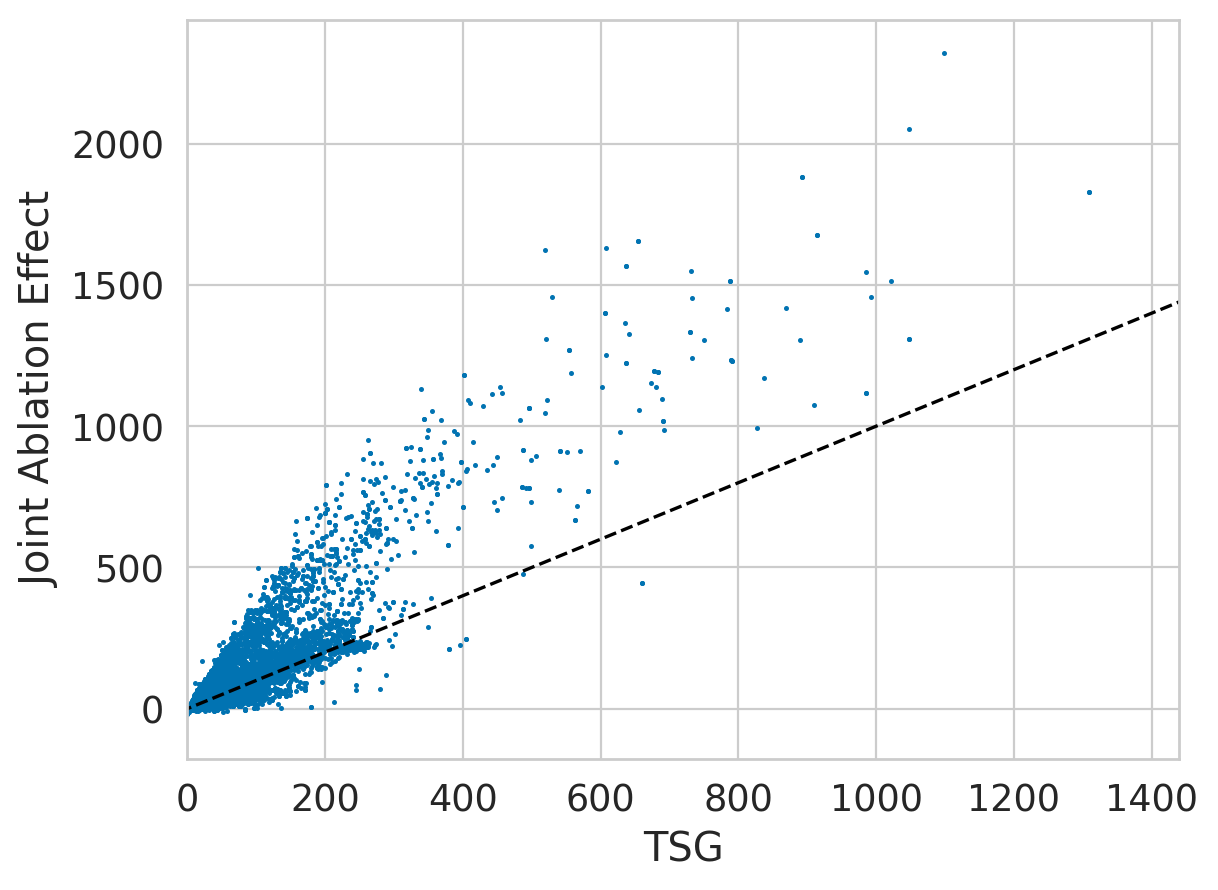}
         \caption{Fever}
     \end{subfigure}
        \caption{Self-repair for \textbf{Qwen-2.5 1.5B} and how TSG increases the attributions for the attention scores with the strongest self-repair effects}
        \label{fig:self-repair-qwen1.5}
\end{figure*}

\begin{figure*}[h]
     \centering
     \begin{subfigure}[b]{0.25\textwidth}
         \centering
         \includegraphics[width=\textwidth]{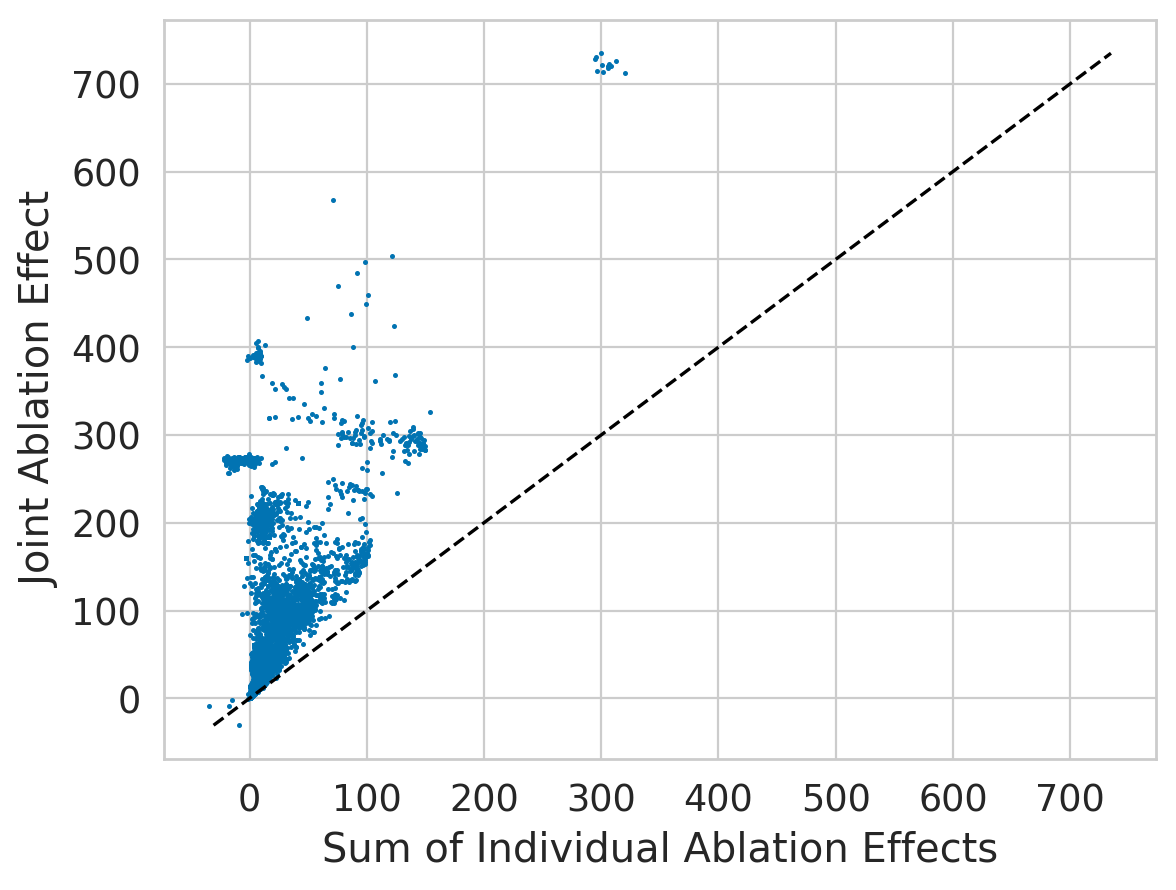}
         \includegraphics[width=\textwidth]{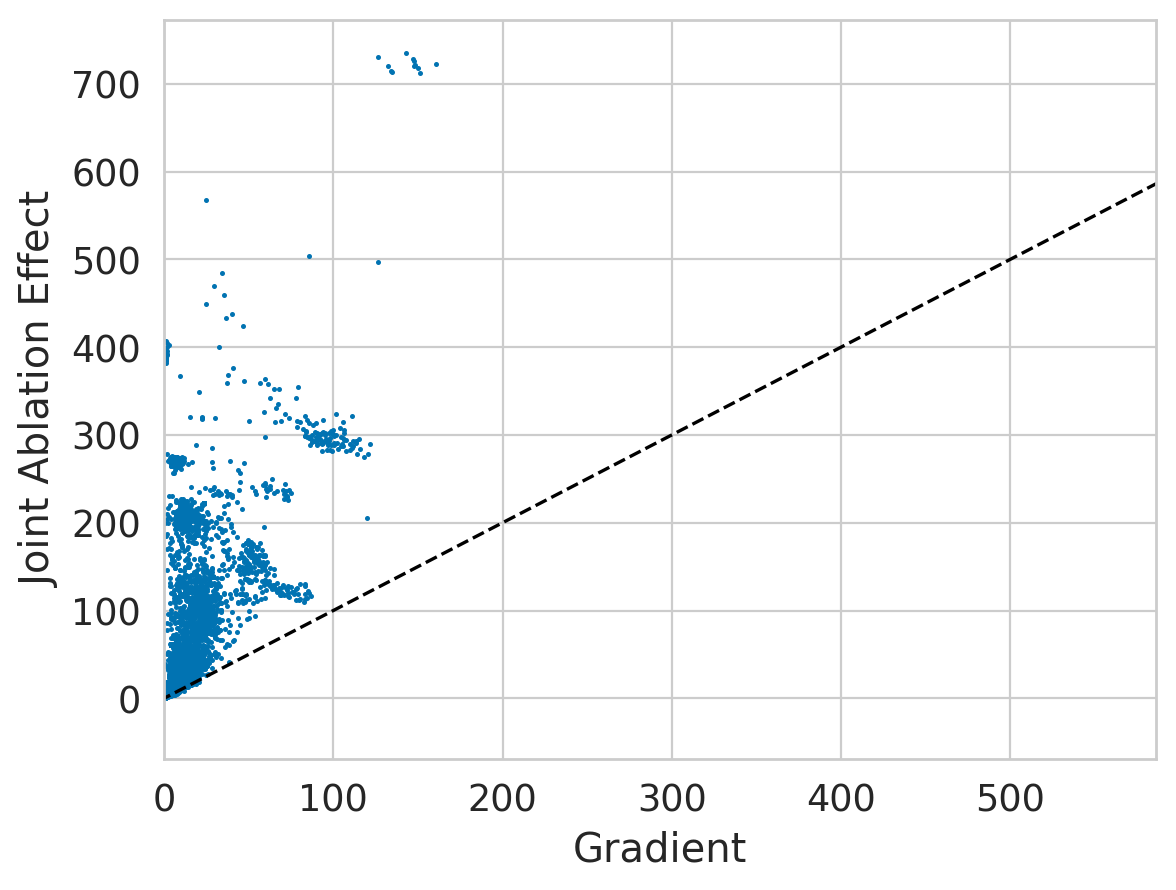}
         \includegraphics[width=\textwidth]{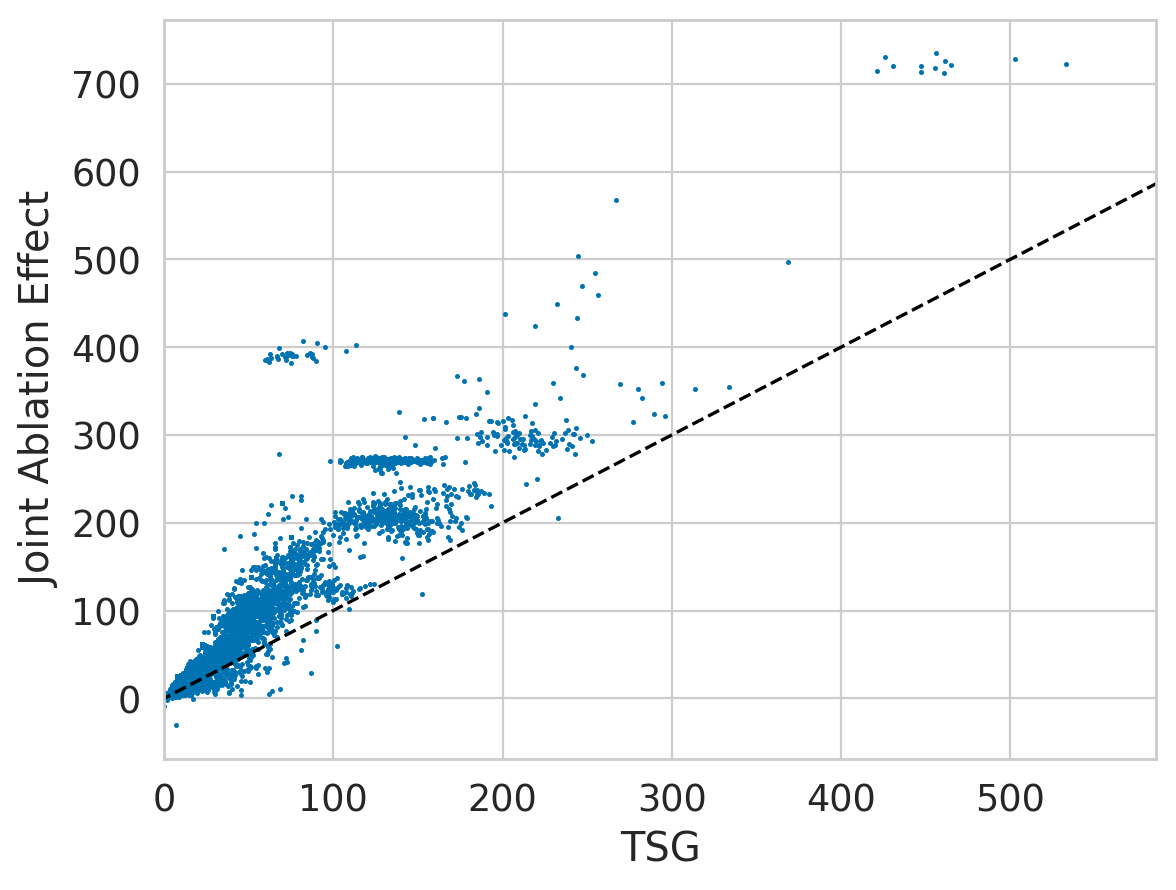}
         \caption{Twitter}
     \end{subfigure}
     \hfill
     \begin{subfigure}[b]{0.25\textwidth}
         \centering
         \includegraphics[width=\textwidth]{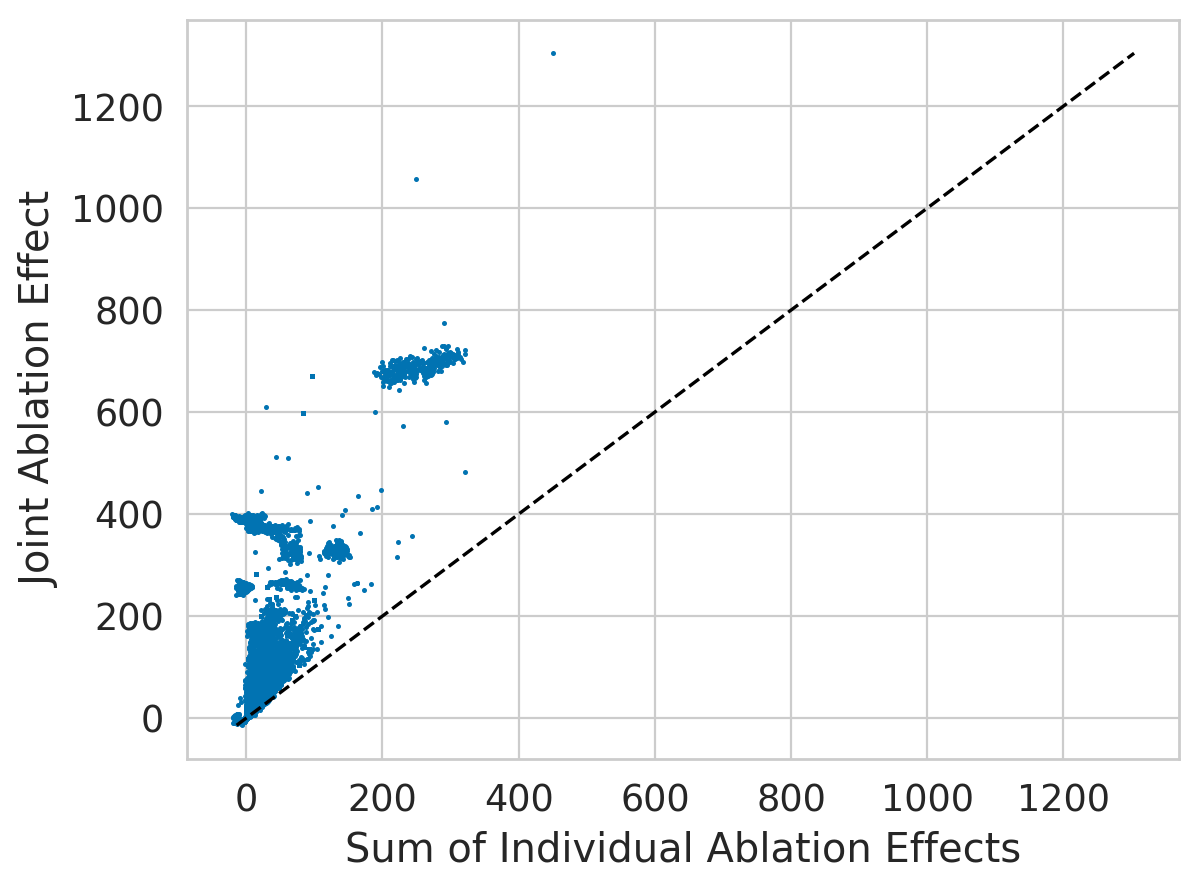}
         \includegraphics[width=\textwidth]{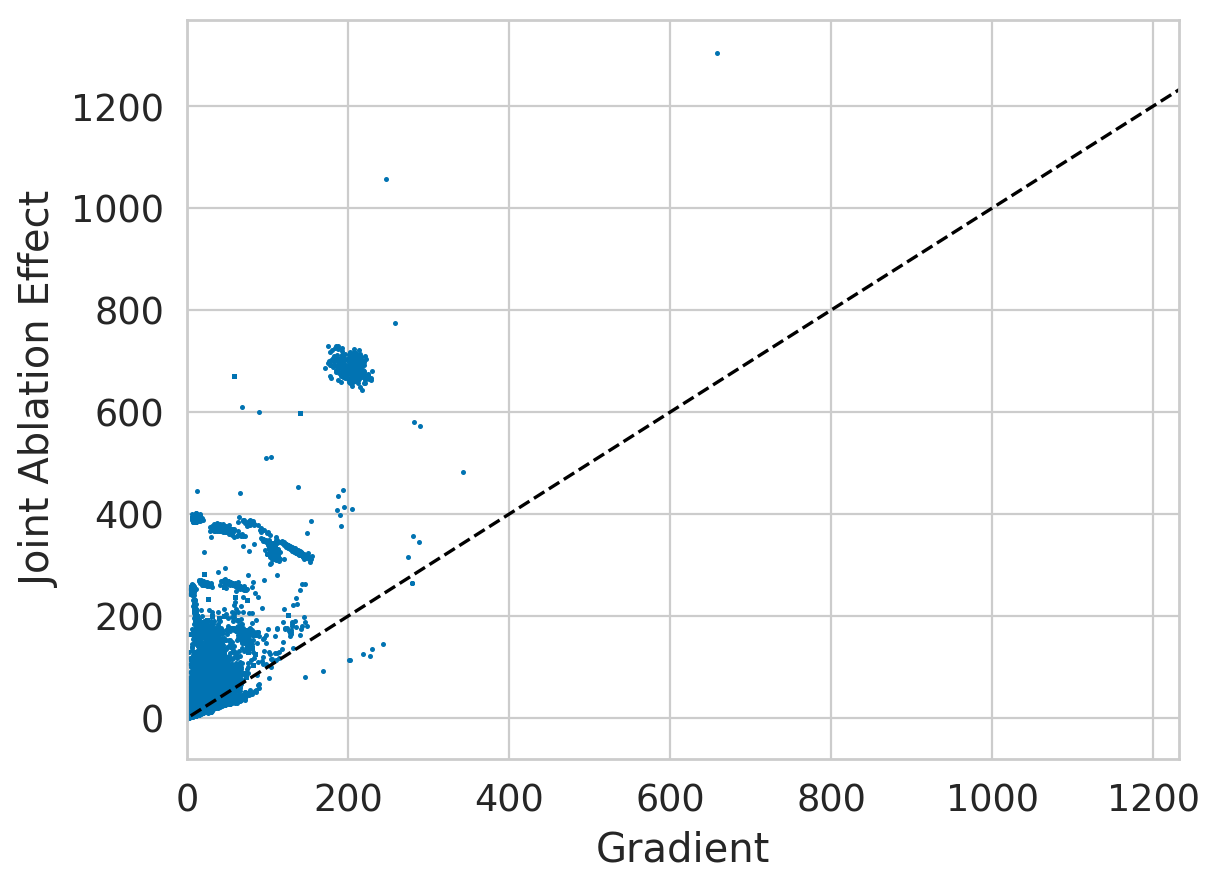}
         \includegraphics[width=\textwidth]{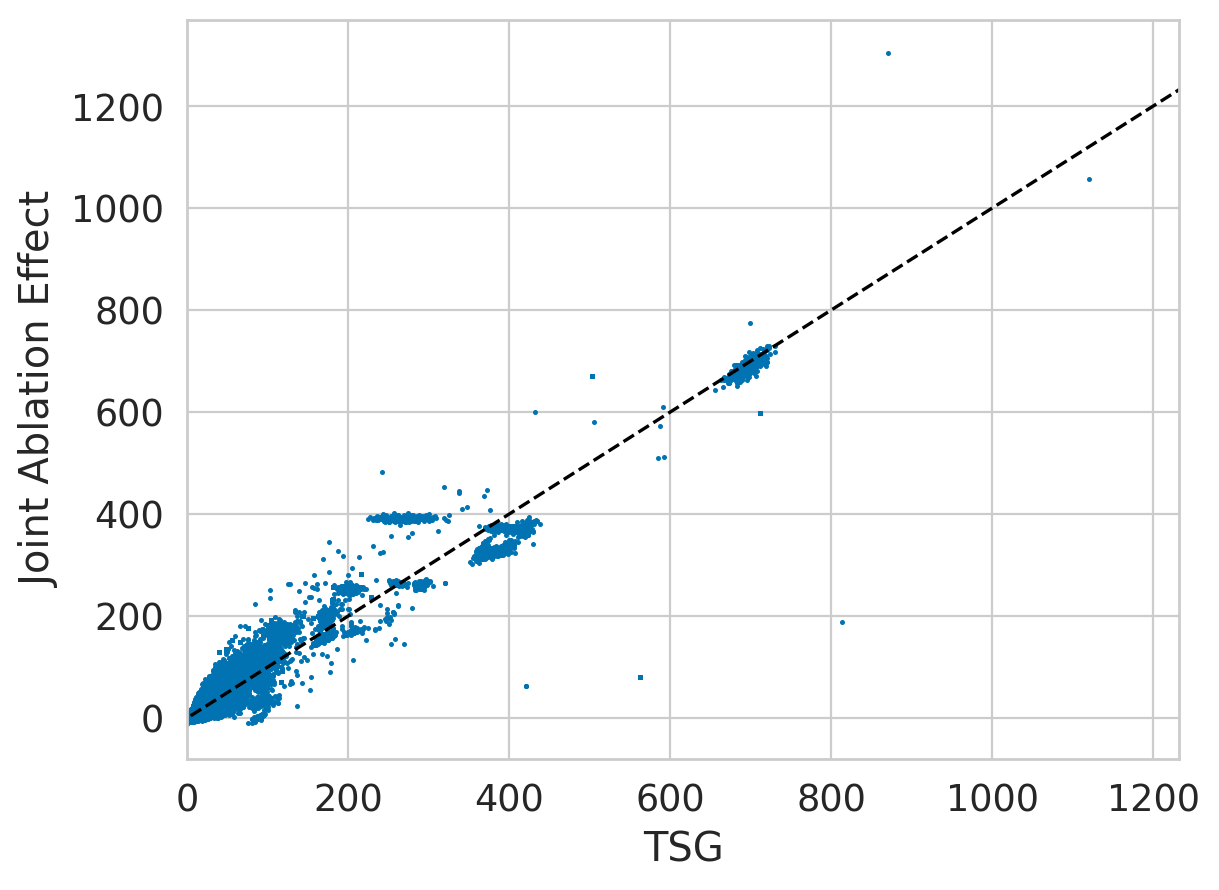}
         \caption{Hatexplain}
     \end{subfigure}
     \hfill
     \begin{subfigure}[b]{0.25\textwidth}
         \centering
         \includegraphics[width=\textwidth]{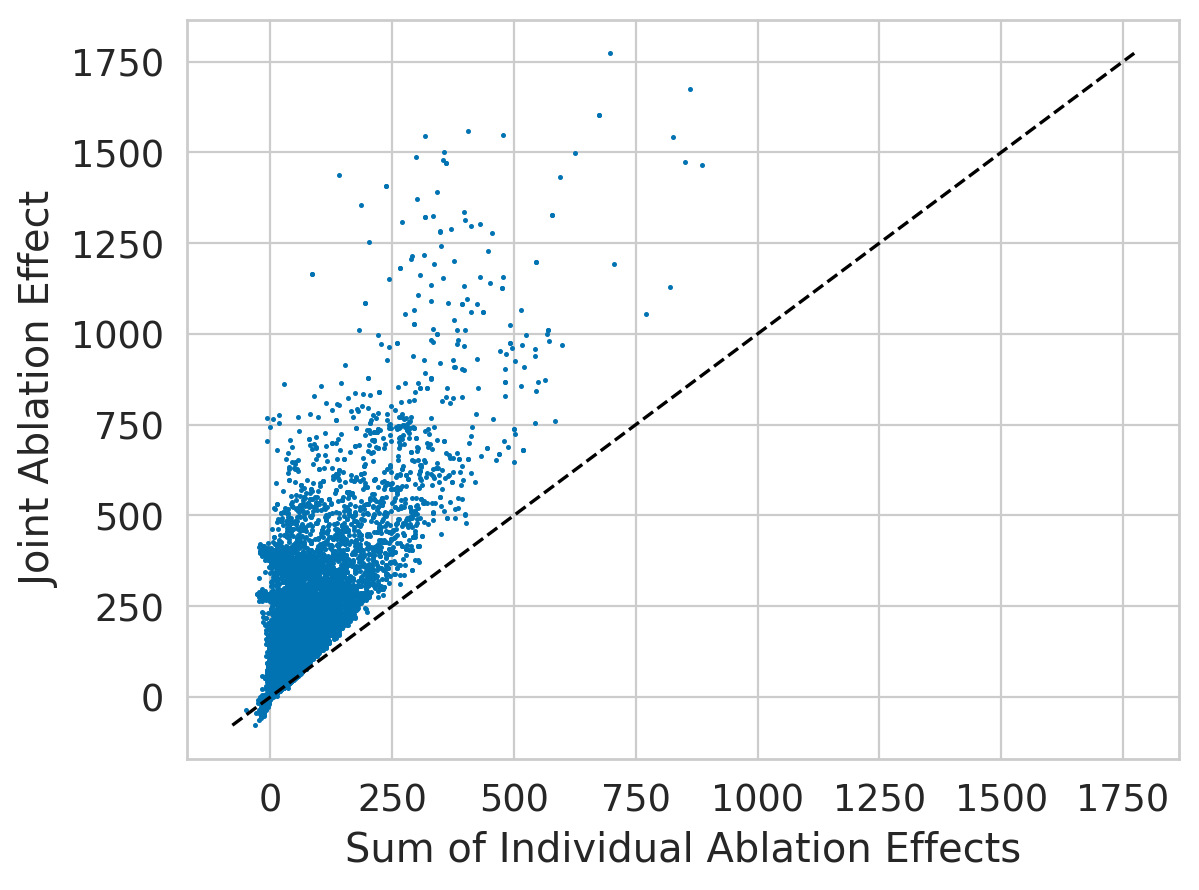}
         \includegraphics[width=\textwidth]{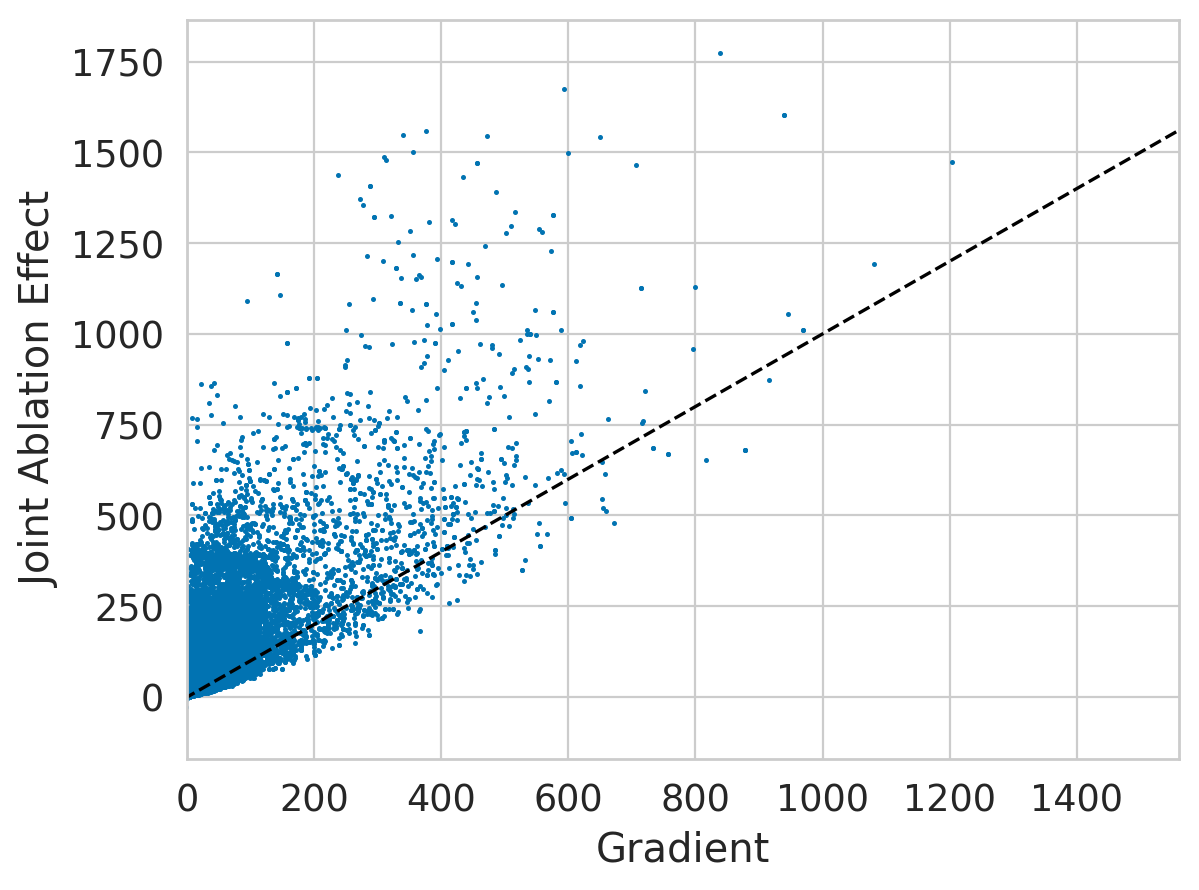}
         \includegraphics[width=\textwidth]{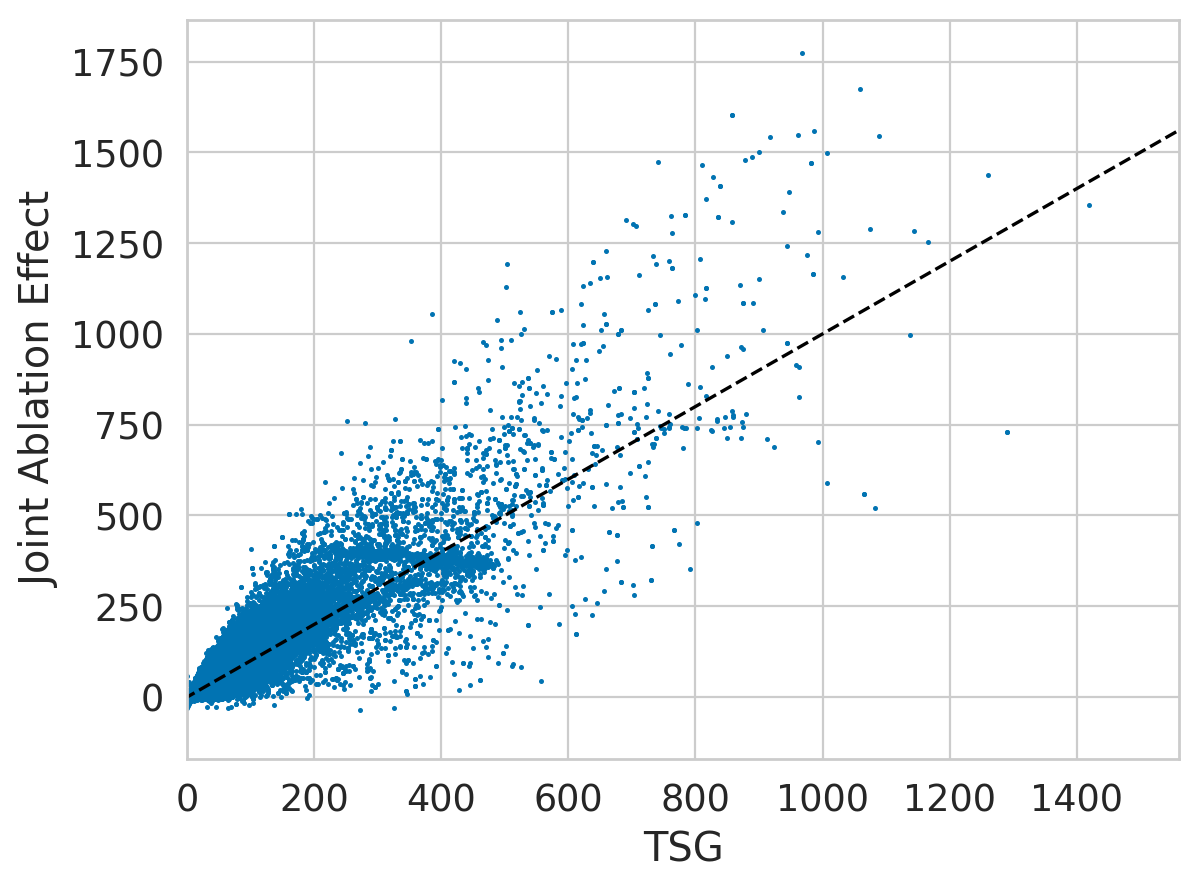}
         
         \caption{Scifact}
     \end{subfigure}
     \hfill
     \begin{subfigure}[b]{0.25\textwidth}
         \centering
         \includegraphics[width=\textwidth]{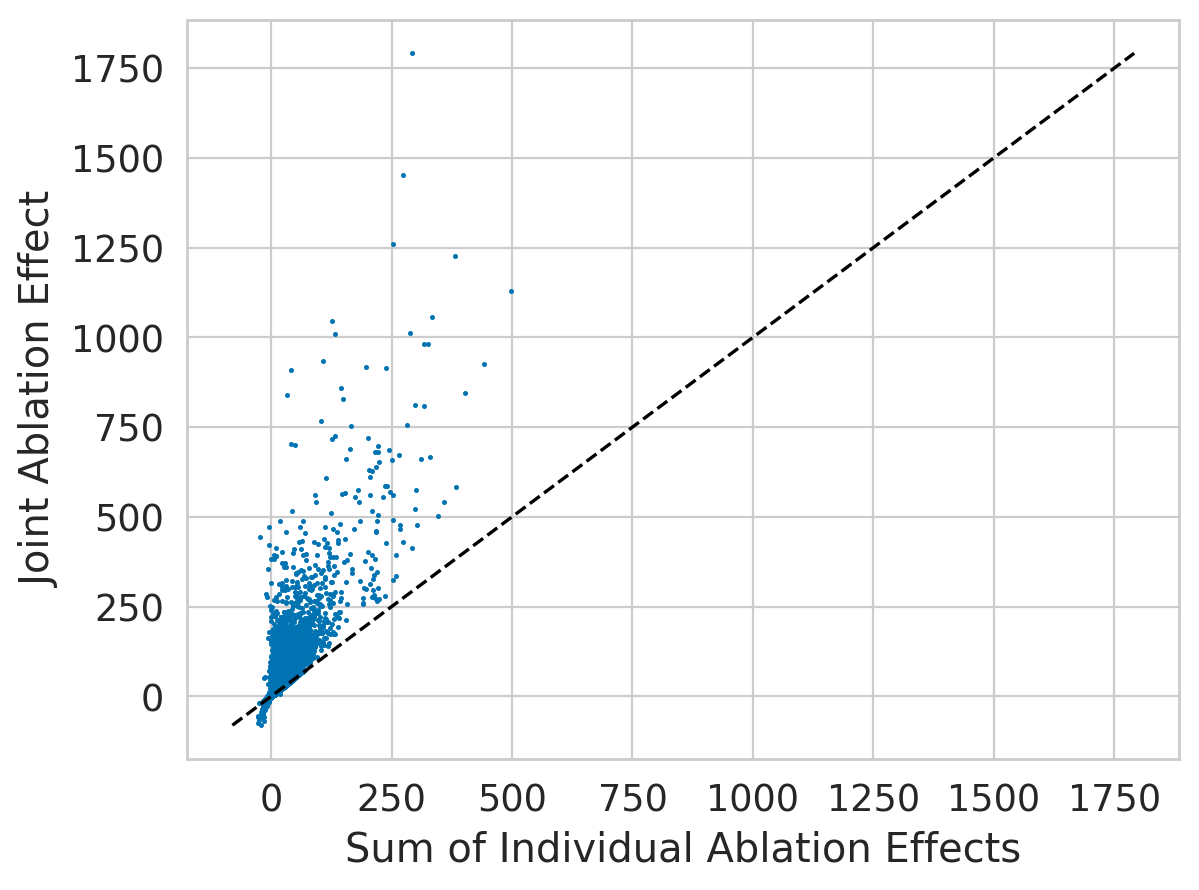}
         \includegraphics[width=\textwidth]{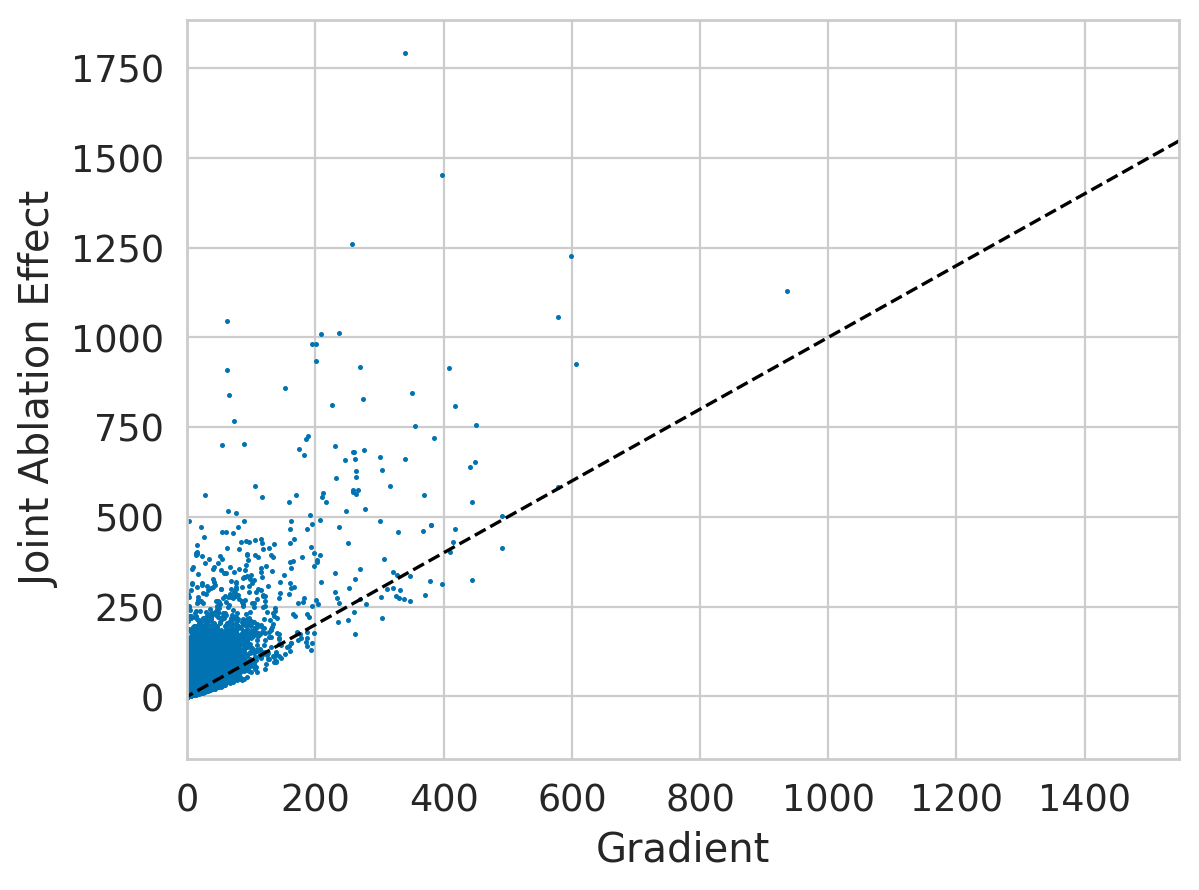}
         \includegraphics[width=\textwidth]{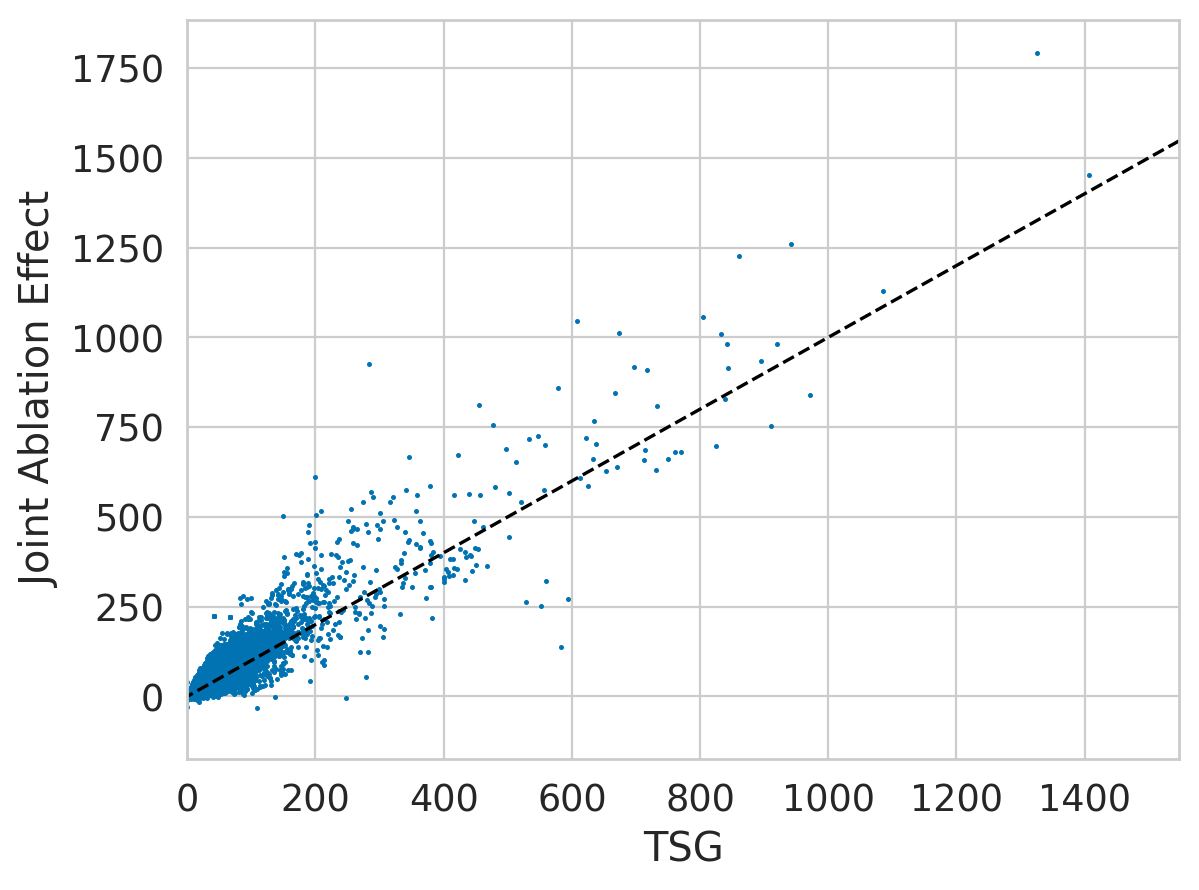}
         \caption{BoolQ}
     \end{subfigure}
     \hfill
     \begin{subfigure}[b]{0.25\textwidth}
         \centering
         \includegraphics[width=\textwidth]{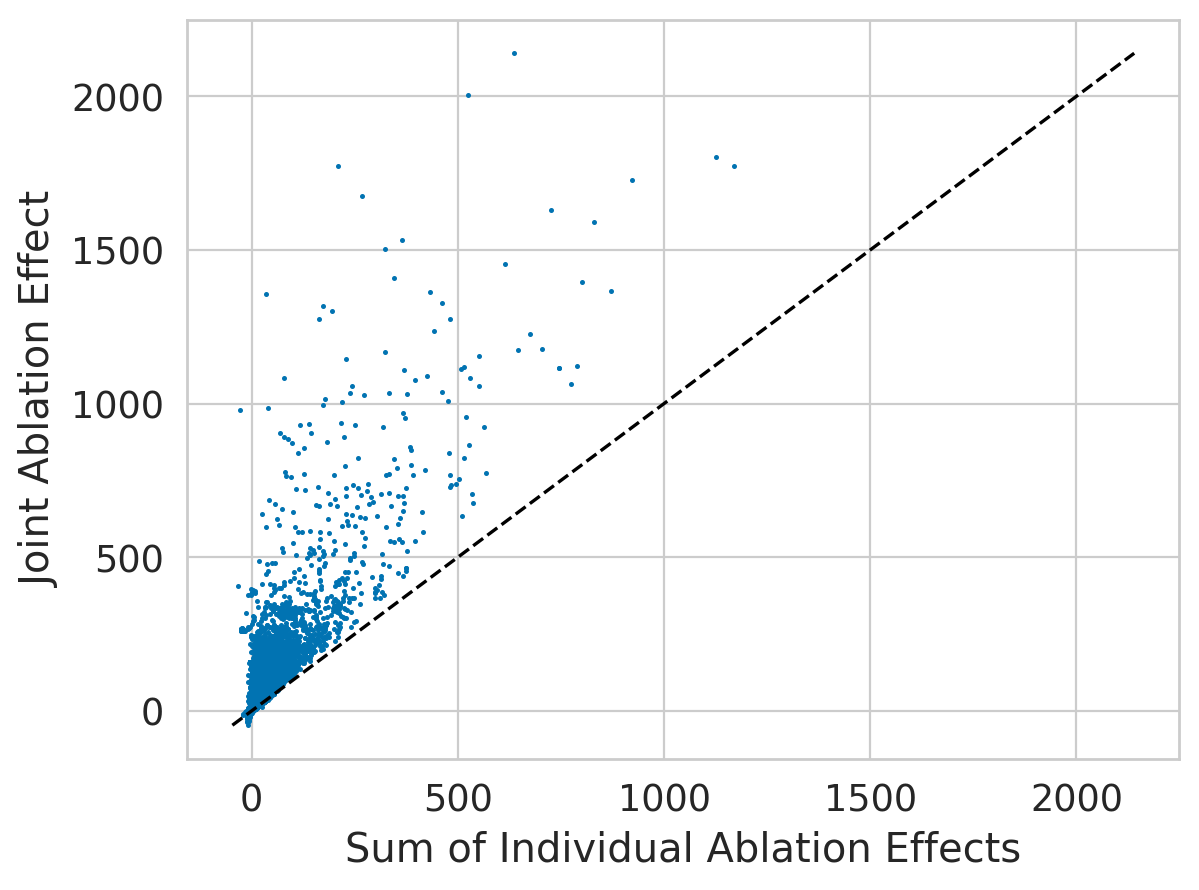}
         \includegraphics[width=\textwidth]{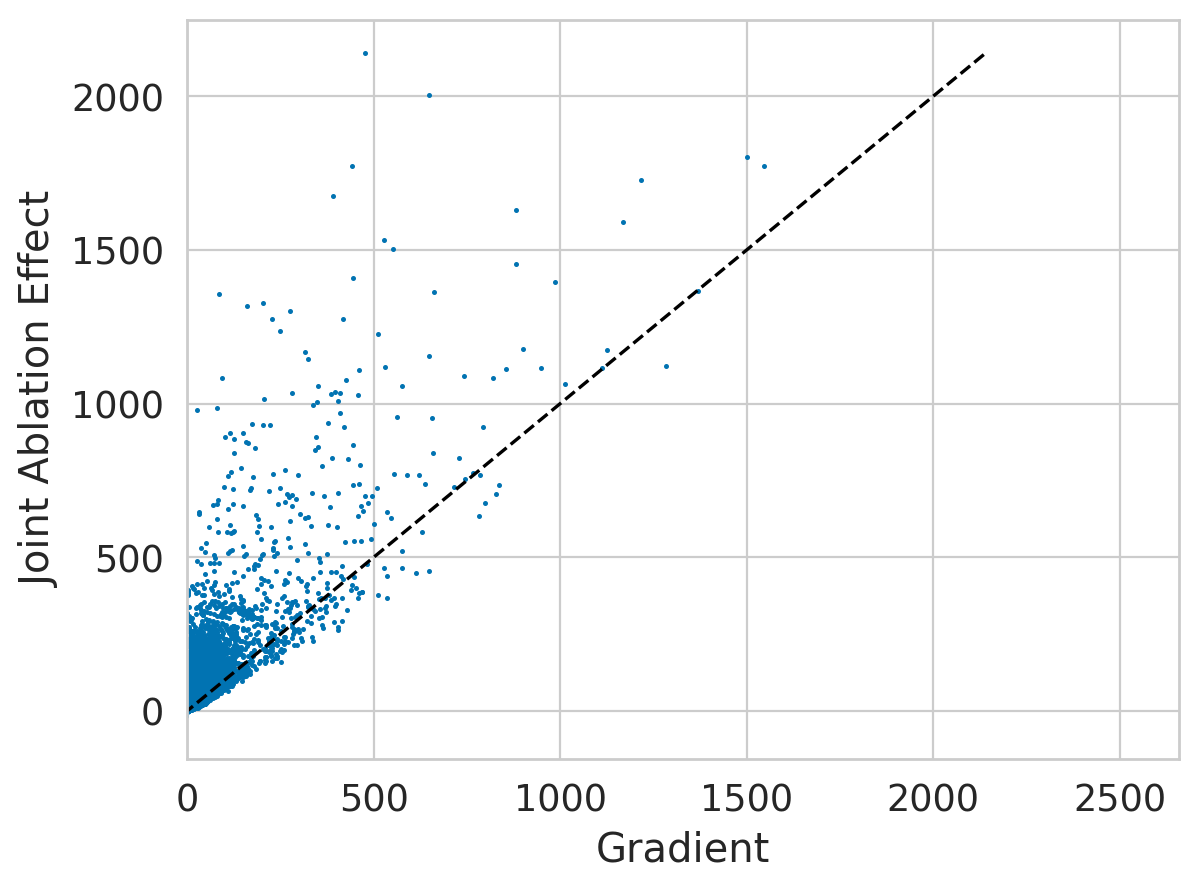}
         \includegraphics[width=\textwidth]{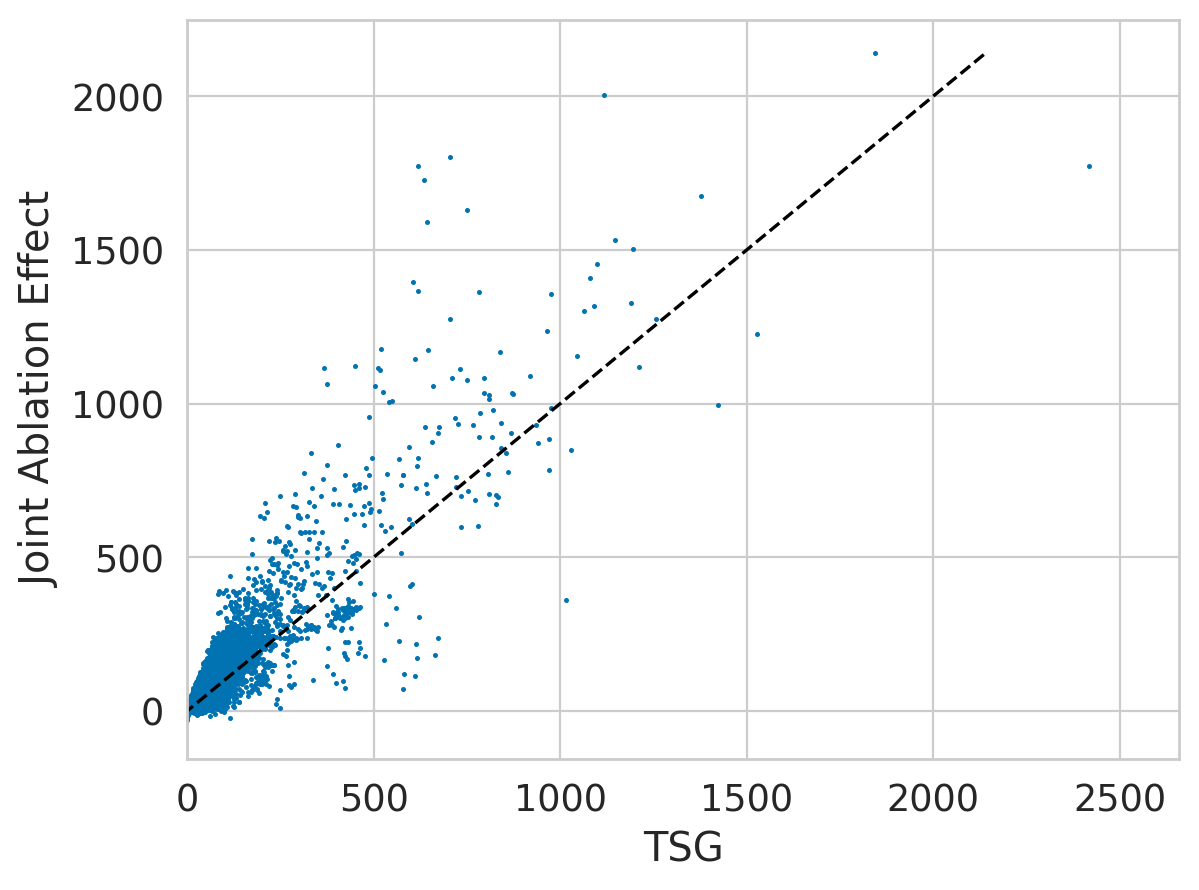}
         \caption{Movie}
     \end{subfigure}
     \hfill
     \begin{subfigure}[b]{0.25\textwidth}
         \centering
         \includegraphics[width=\textwidth]{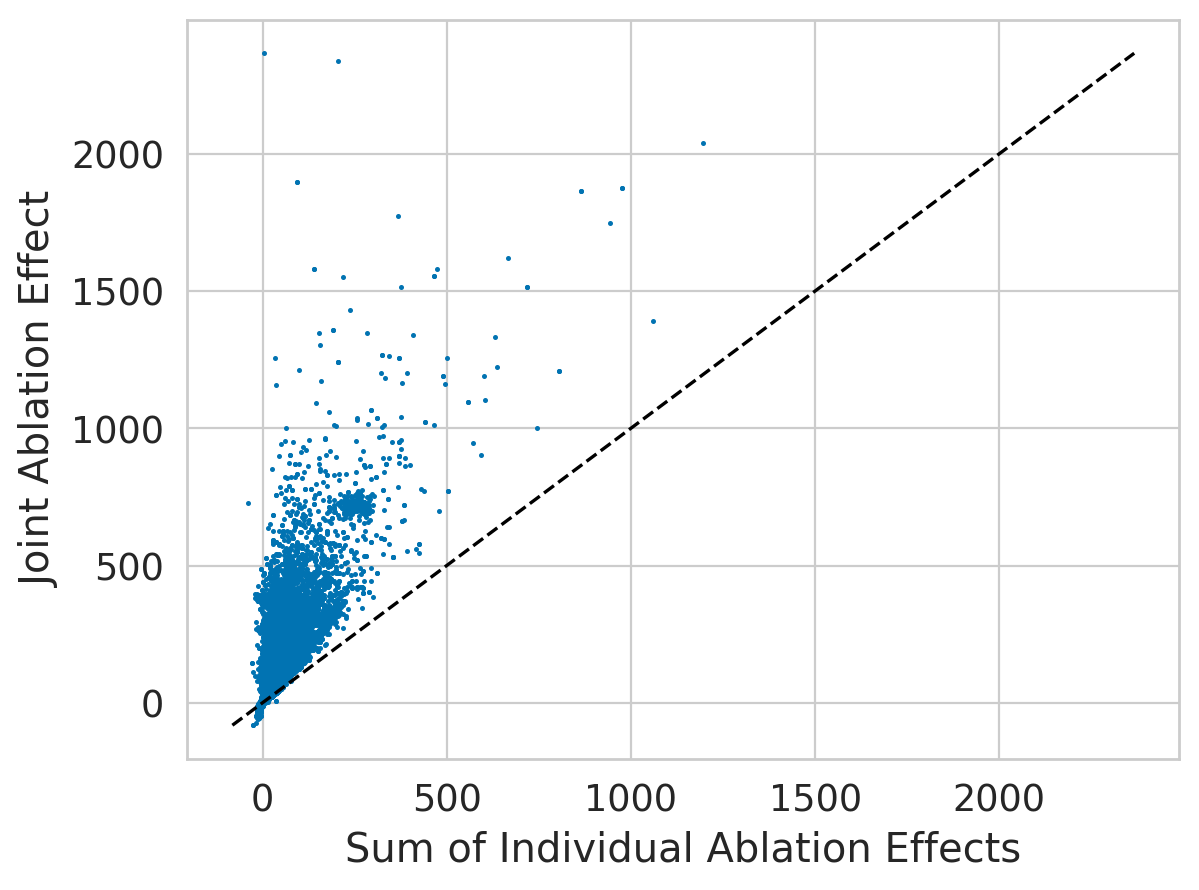}
         \includegraphics[width=\textwidth]{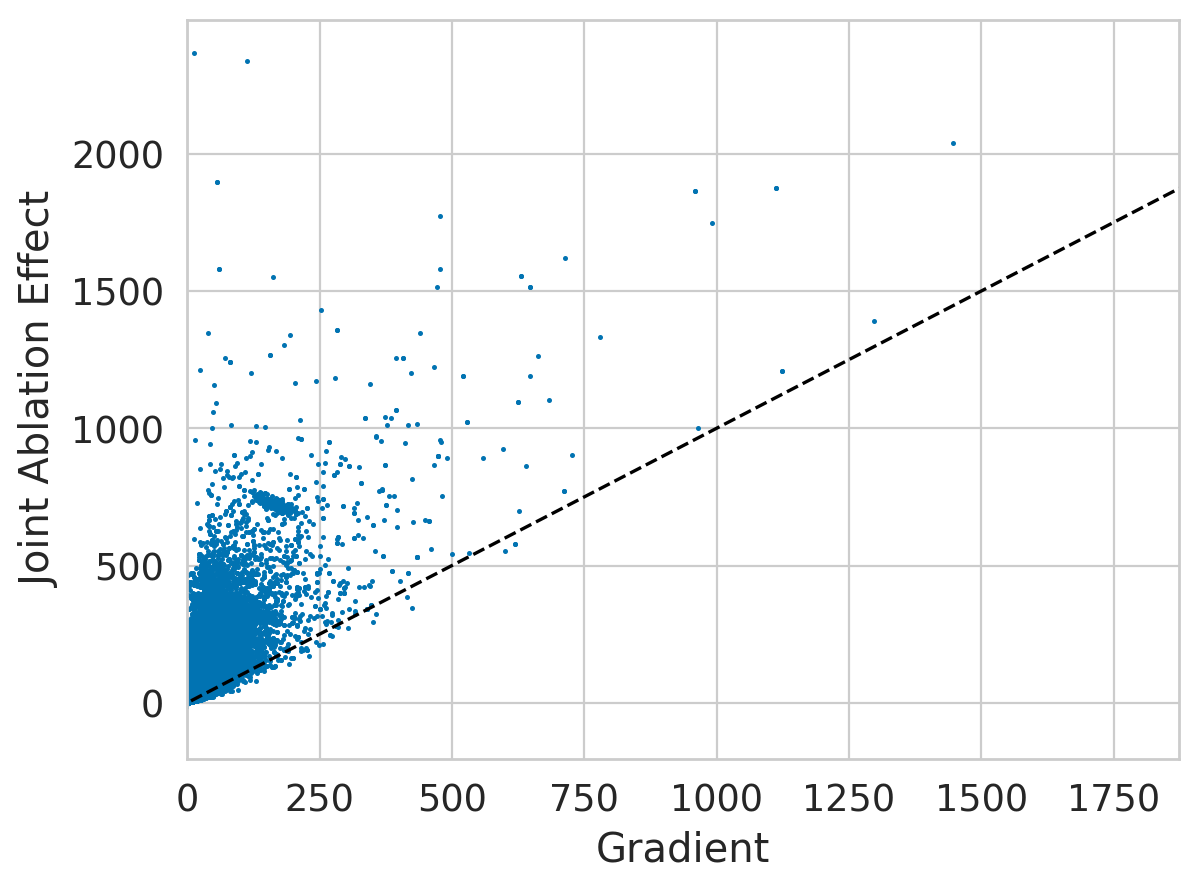}
         \includegraphics[width=\textwidth]{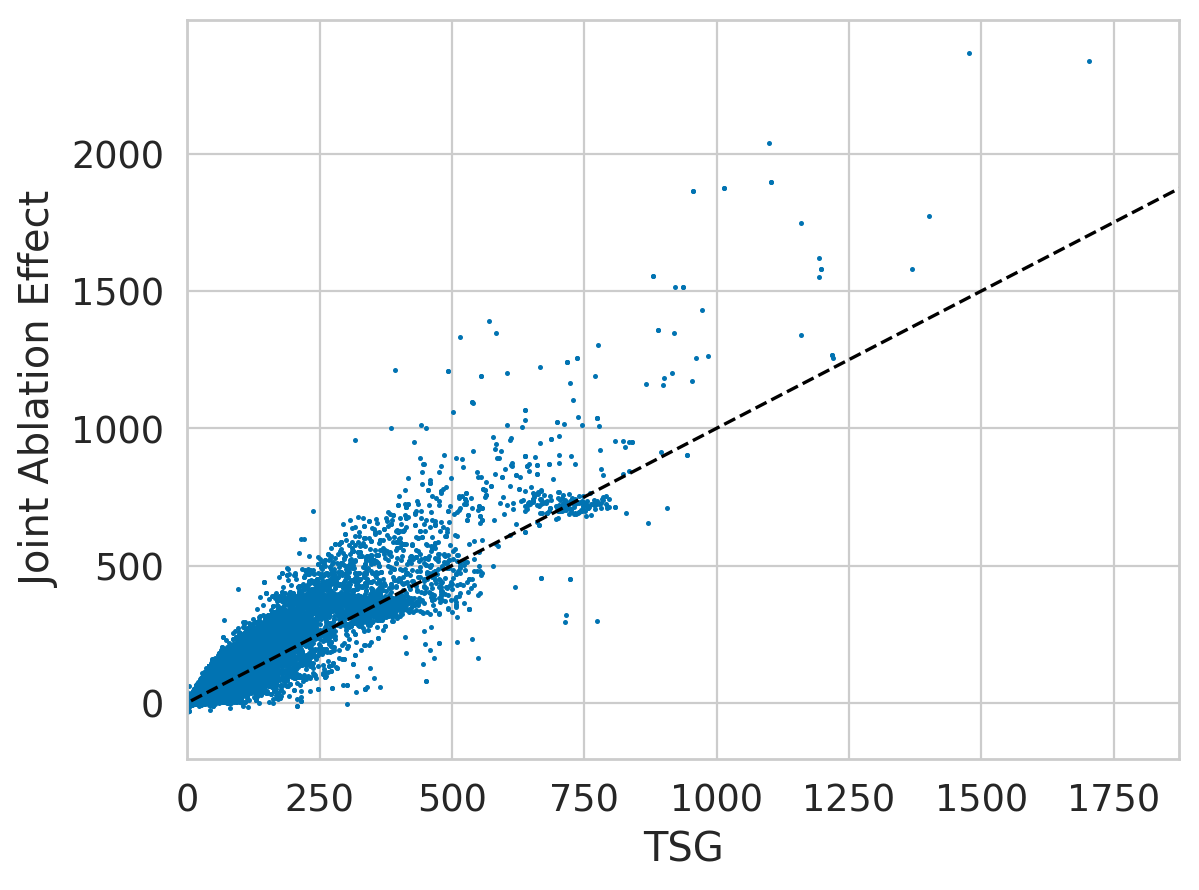}
         \caption{Fever}
     \end{subfigure}
        \caption{Self-repair for \textbf{Qwen-2.5 3B} and how TSG increases the attributions for the attention scores with the strongest self-repair effects}
        \label{fig:self-repair-qwen3}
\end{figure*}

\subsection{Additional self-repair results} \label{app:self-repair}
In \Cref{fig:self-repair-gemma2}, \Cref{fig:self-repair-llama1}, \Cref{fig:self-repair-llama3}, and \Cref{fig:self-repair-qwen1.5}, we present additional empirical evidence for self-repair. We compare the joint ablation effect with the gradients, TSG, and the sum of individual ablation effects. The figures show similar results to those in the main paper. For models such as Qwen, the gradients are generally very large, which results in many points being plotted.

\begin{figure*}[h]
     \centering
     \begin{subfigure}[b]{0.25\textwidth}
         \centering
         \includegraphics[width=\textwidth]{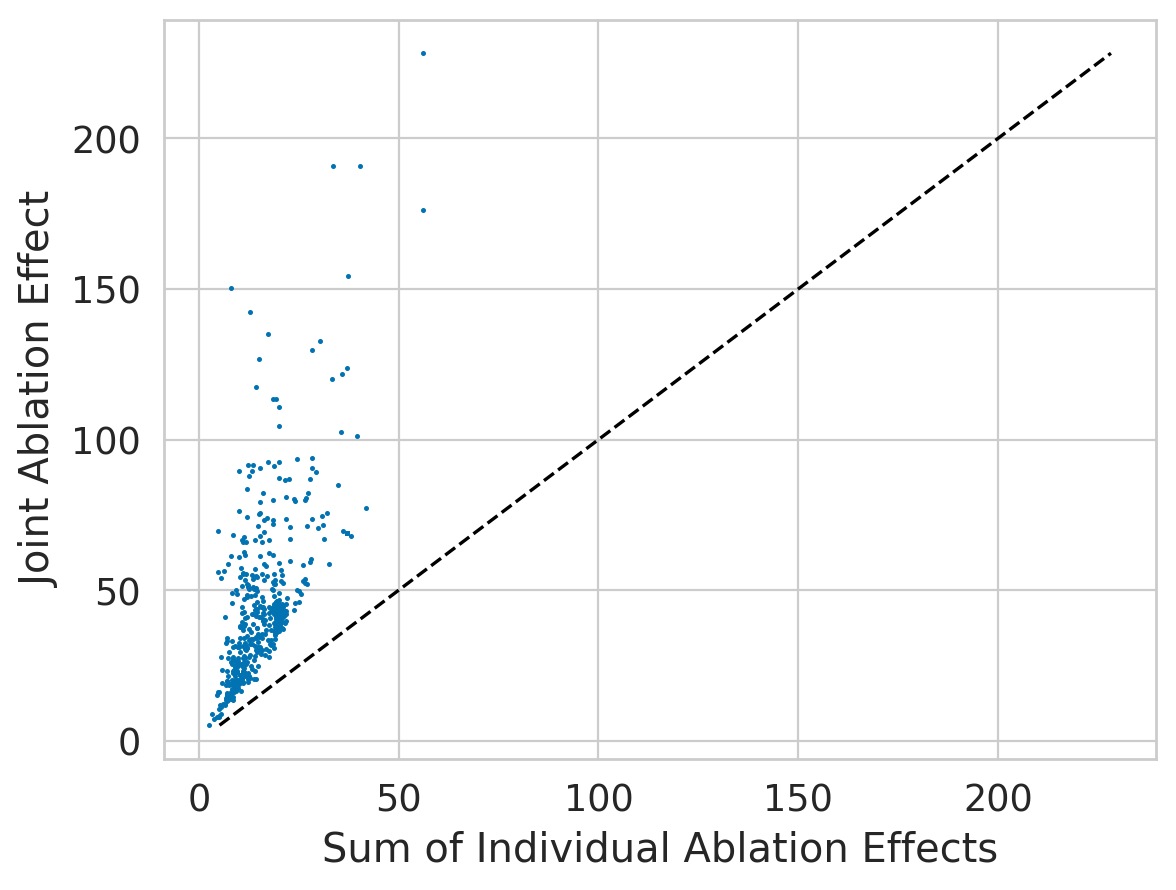}
         \includegraphics[width=\textwidth]{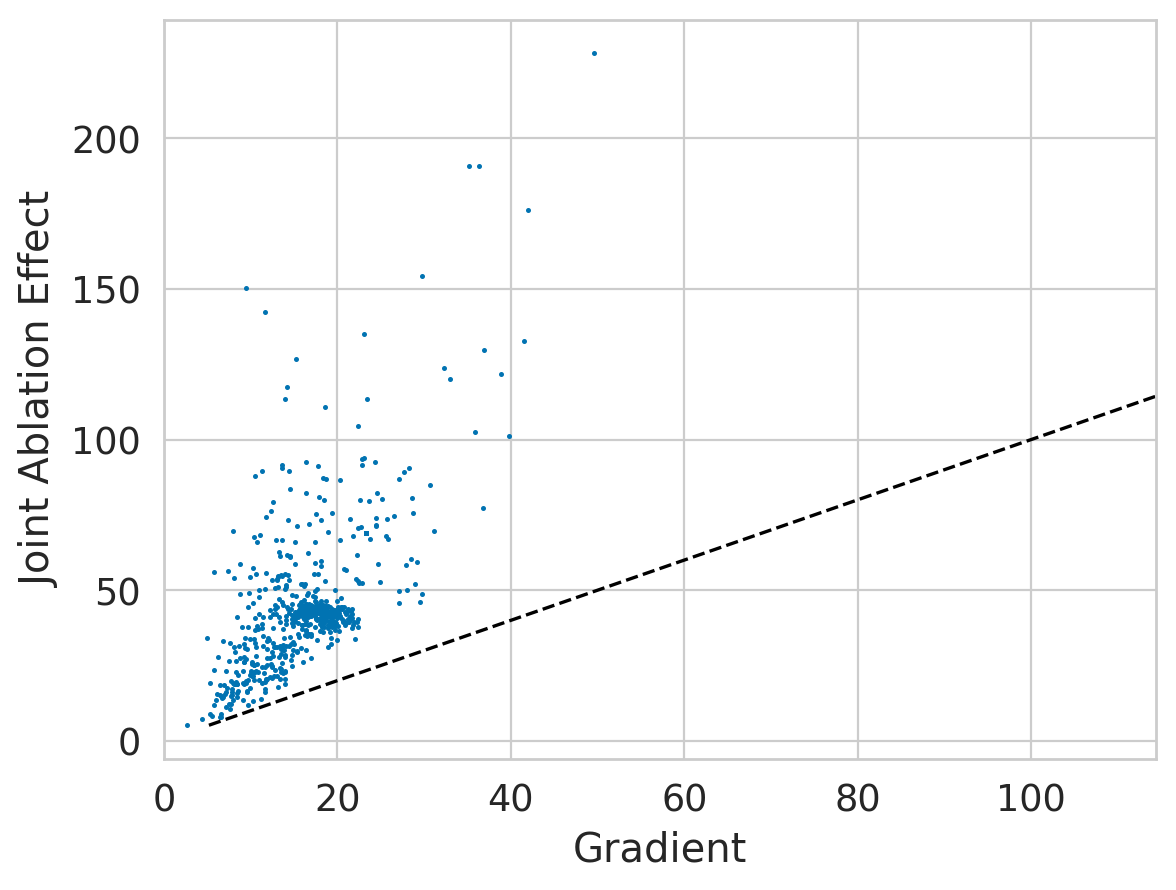}
         \includegraphics[width=\textwidth]{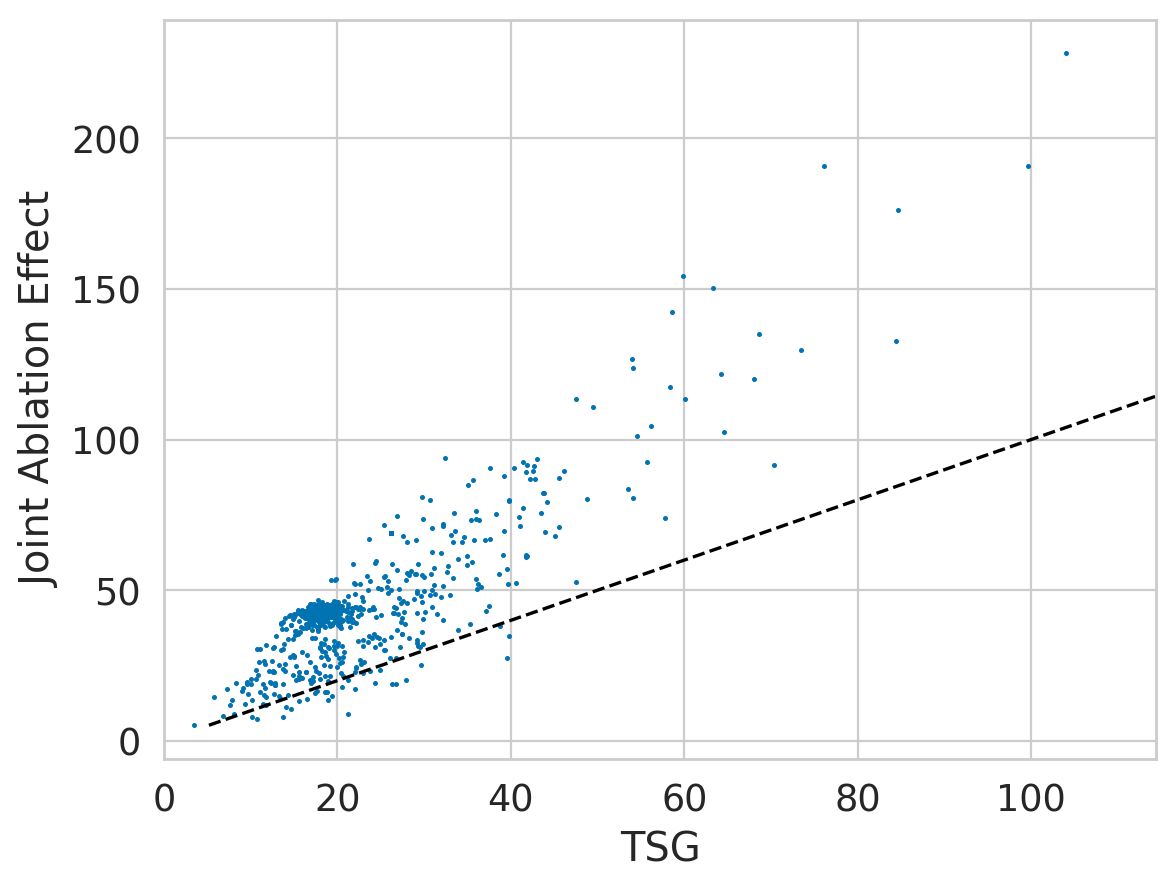}
         \caption{Twitter}
     \end{subfigure}
     \hfill
     \begin{subfigure}[b]{0.25\textwidth}
         \centering
         \includegraphics[width=\textwidth]{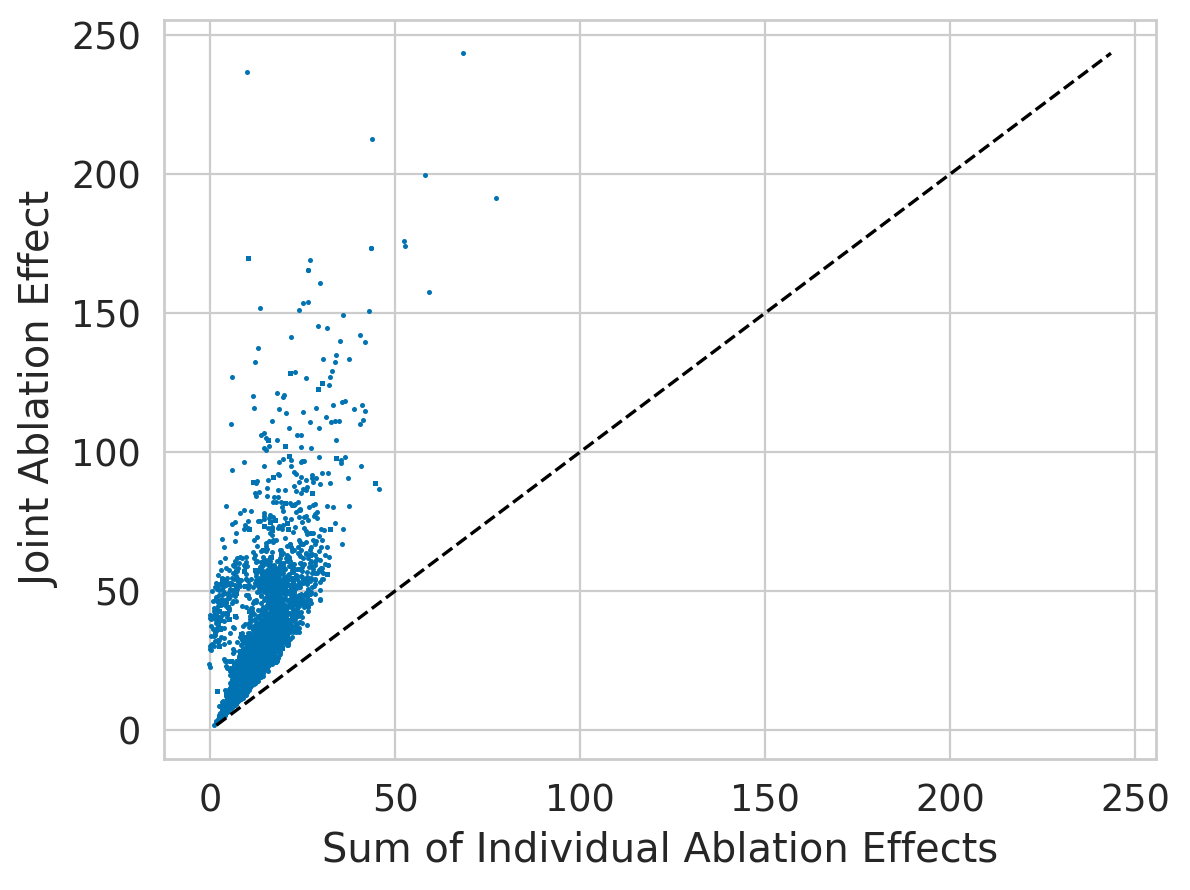}
         \includegraphics[width=\textwidth]{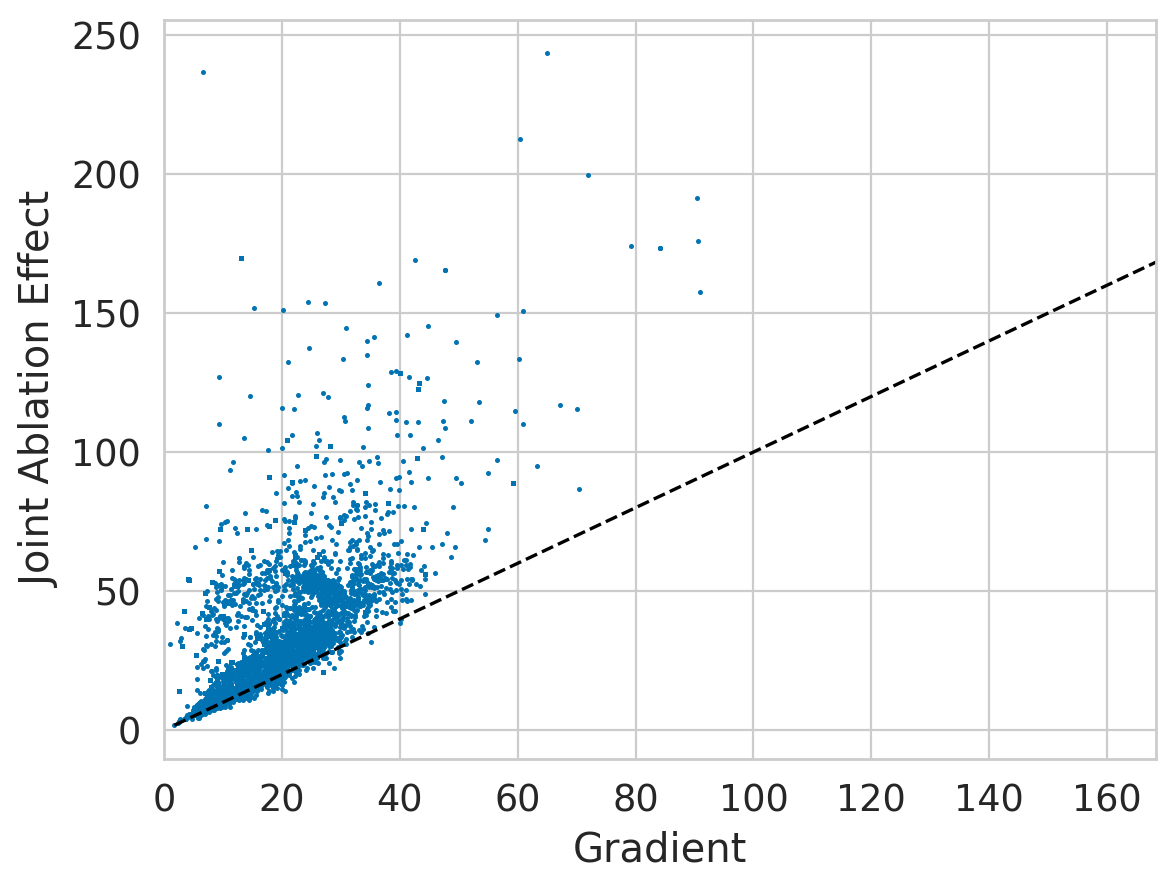}
         \includegraphics[width=\textwidth]{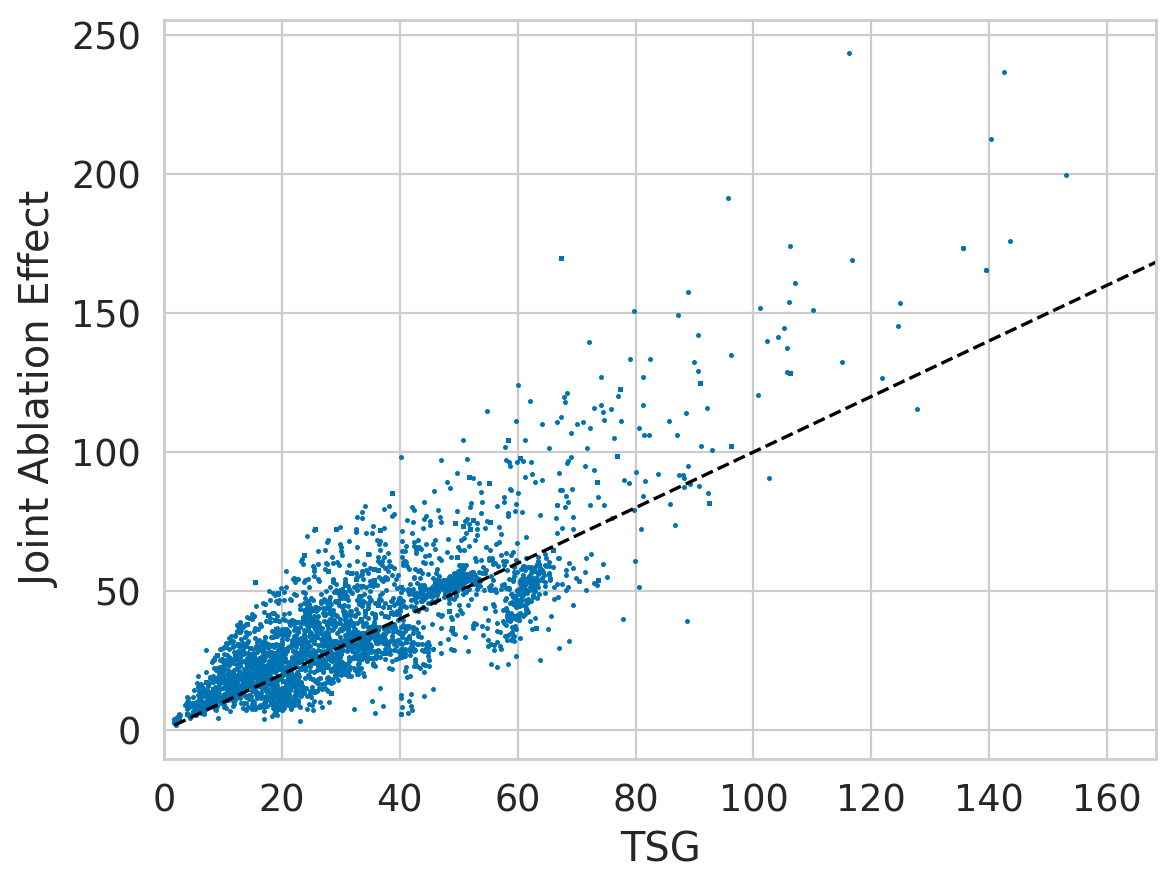}
         \caption{Hatexplain}
     \end{subfigure}
     \hfill
     \begin{subfigure}[b]{0.25\textwidth}
         \centering
         \includegraphics[width=\textwidth]{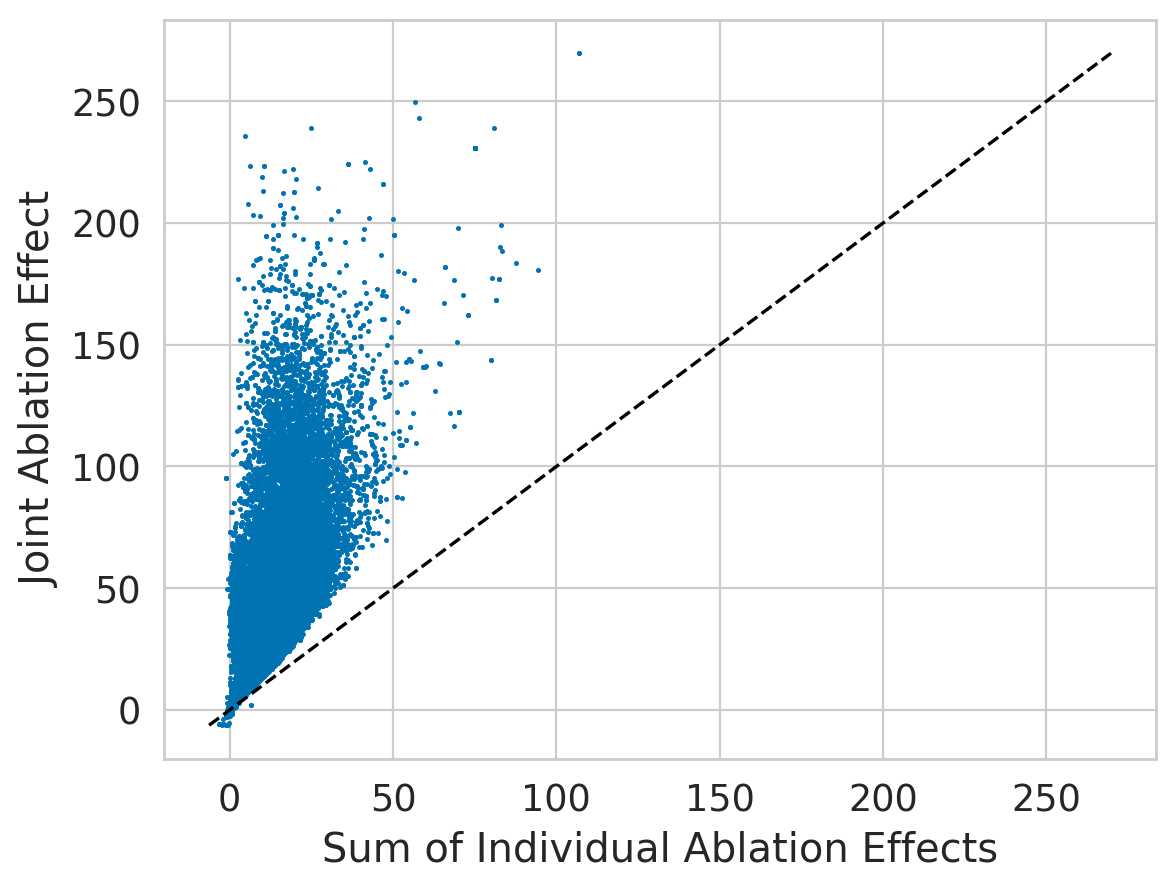}
         \includegraphics[width=\textwidth]{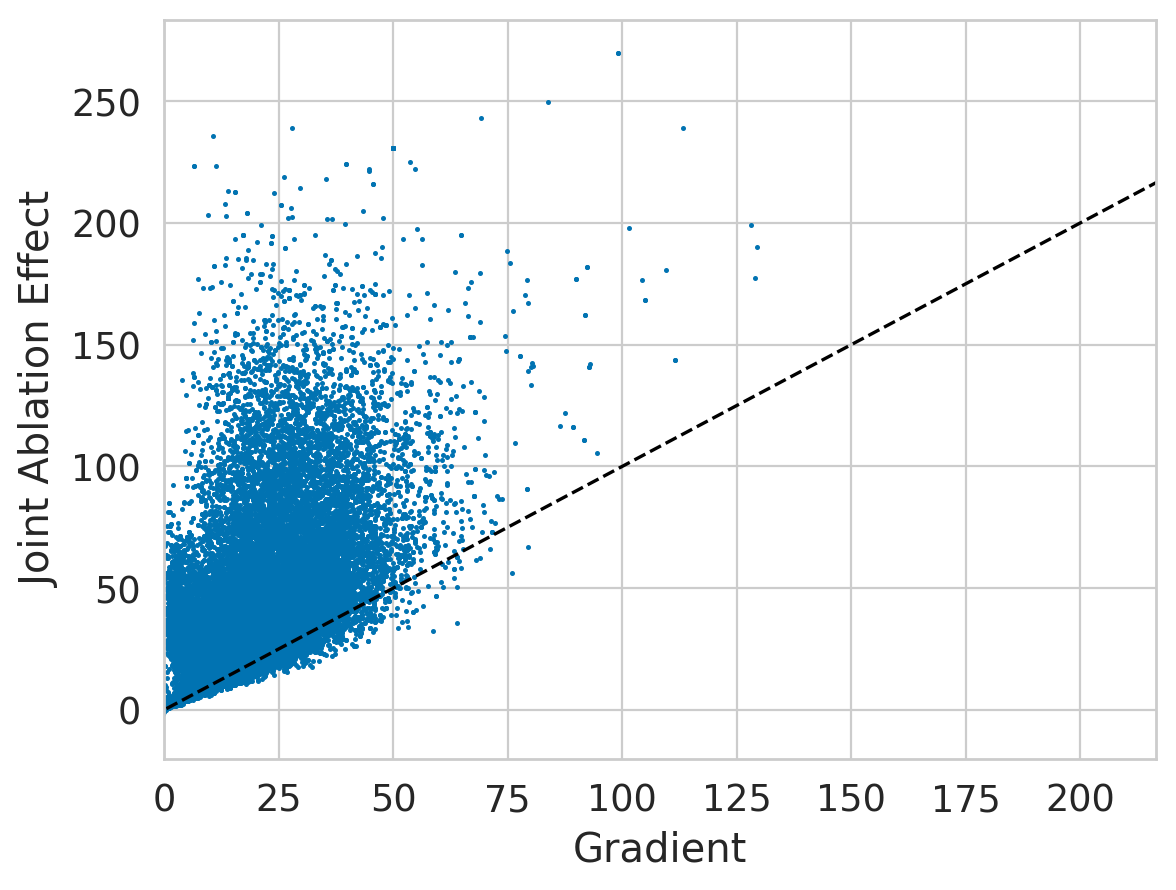}
         \includegraphics[width=\textwidth]{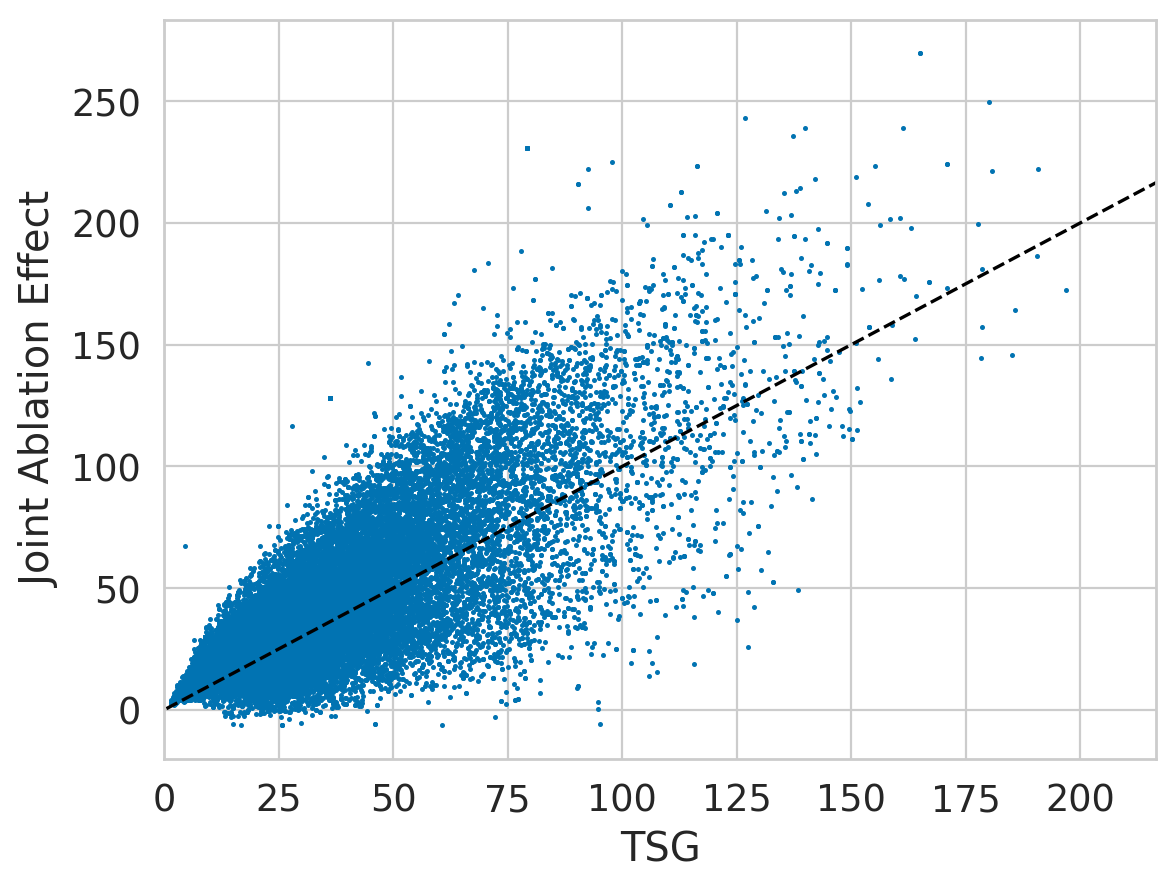}
         
         \caption{Scifact}
     \end{subfigure}
     \hfill
     \begin{subfigure}[b]{0.25\textwidth}
         \centering
         \includegraphics[width=\textwidth]{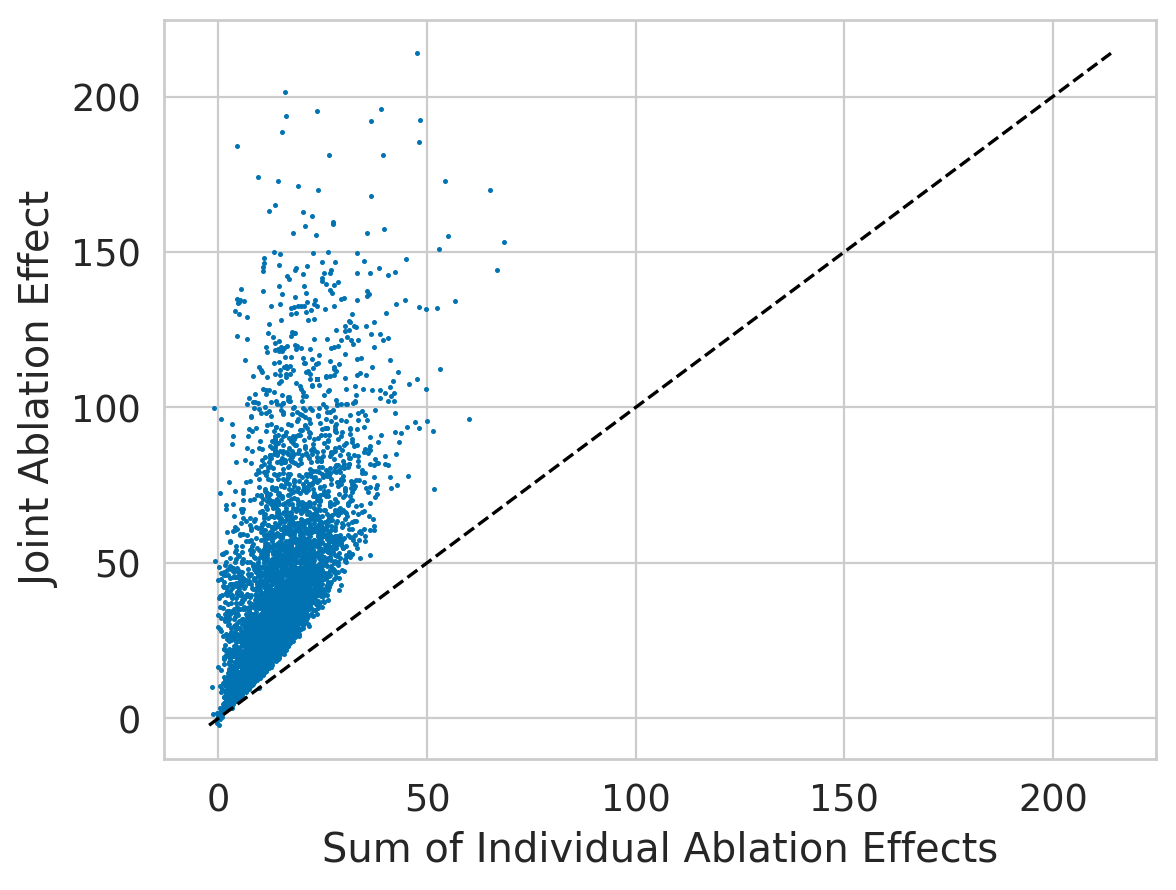}
         \includegraphics[width=\textwidth]{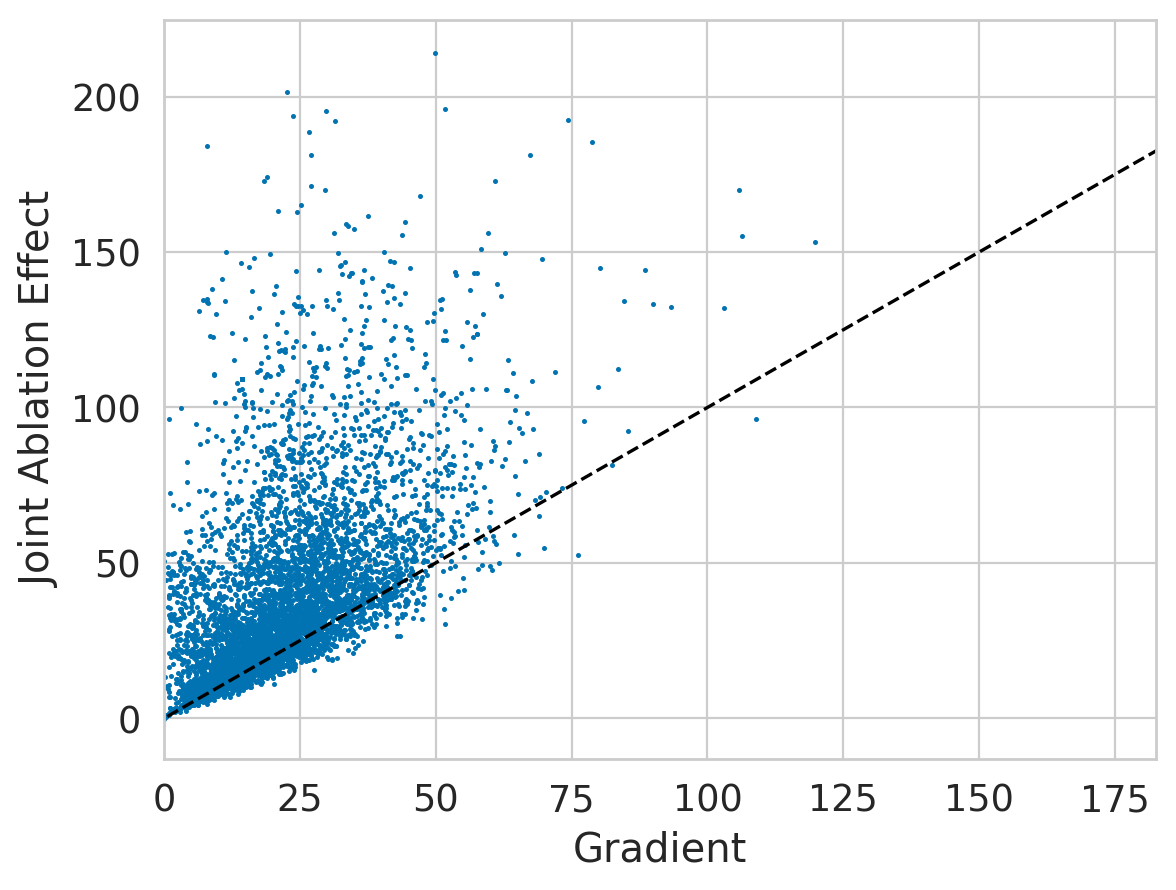}
         \includegraphics[width=\textwidth]{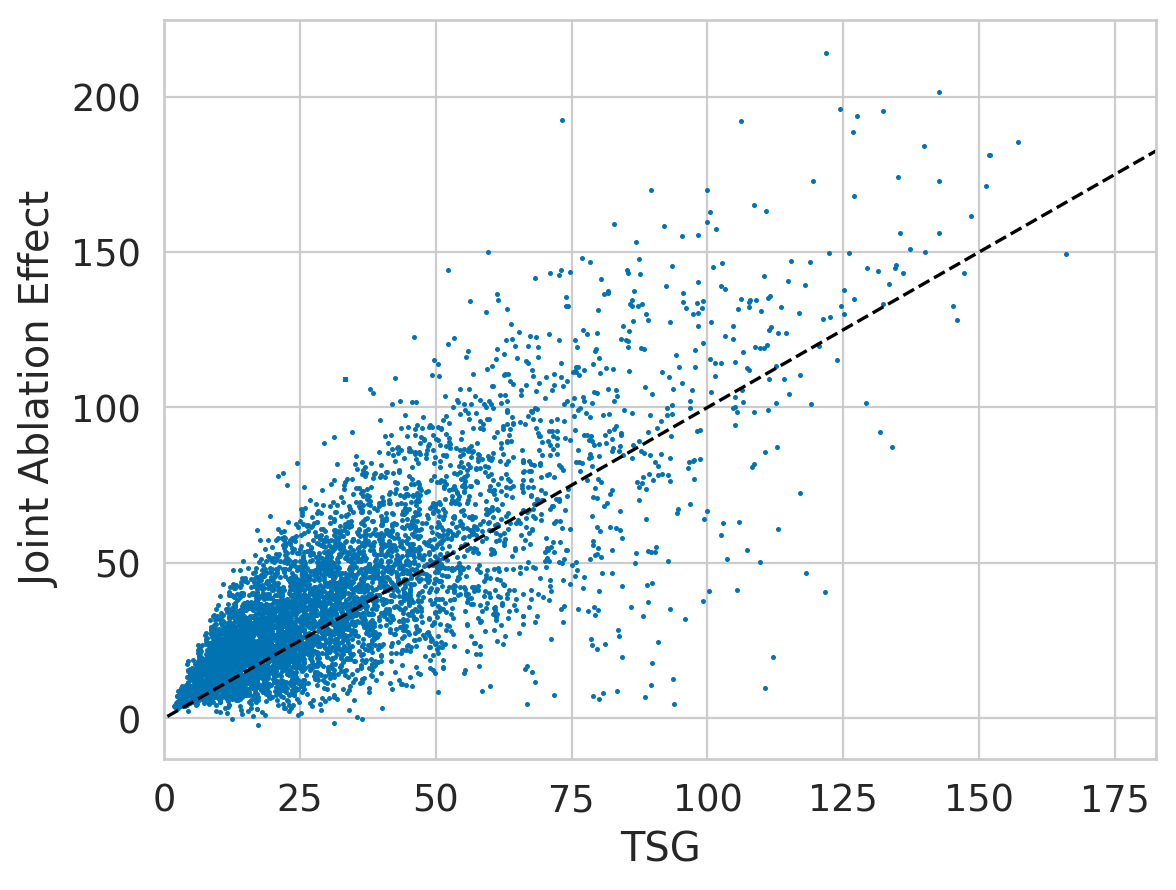}
         \caption{BoolQ}
     \end{subfigure}
     \hfill
     \begin{subfigure}[b]{0.25\textwidth}
         \centering
         \includegraphics[width=\textwidth]{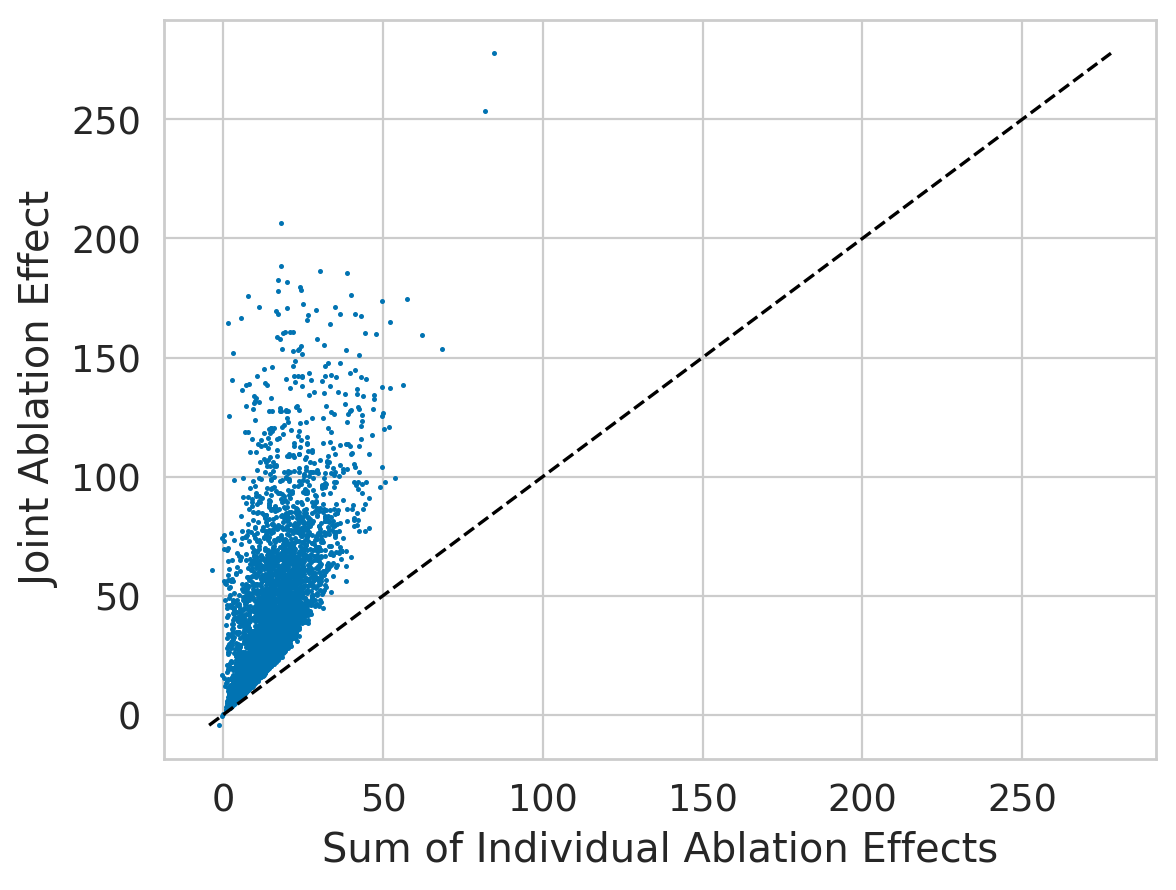}
         \includegraphics[width=\textwidth]{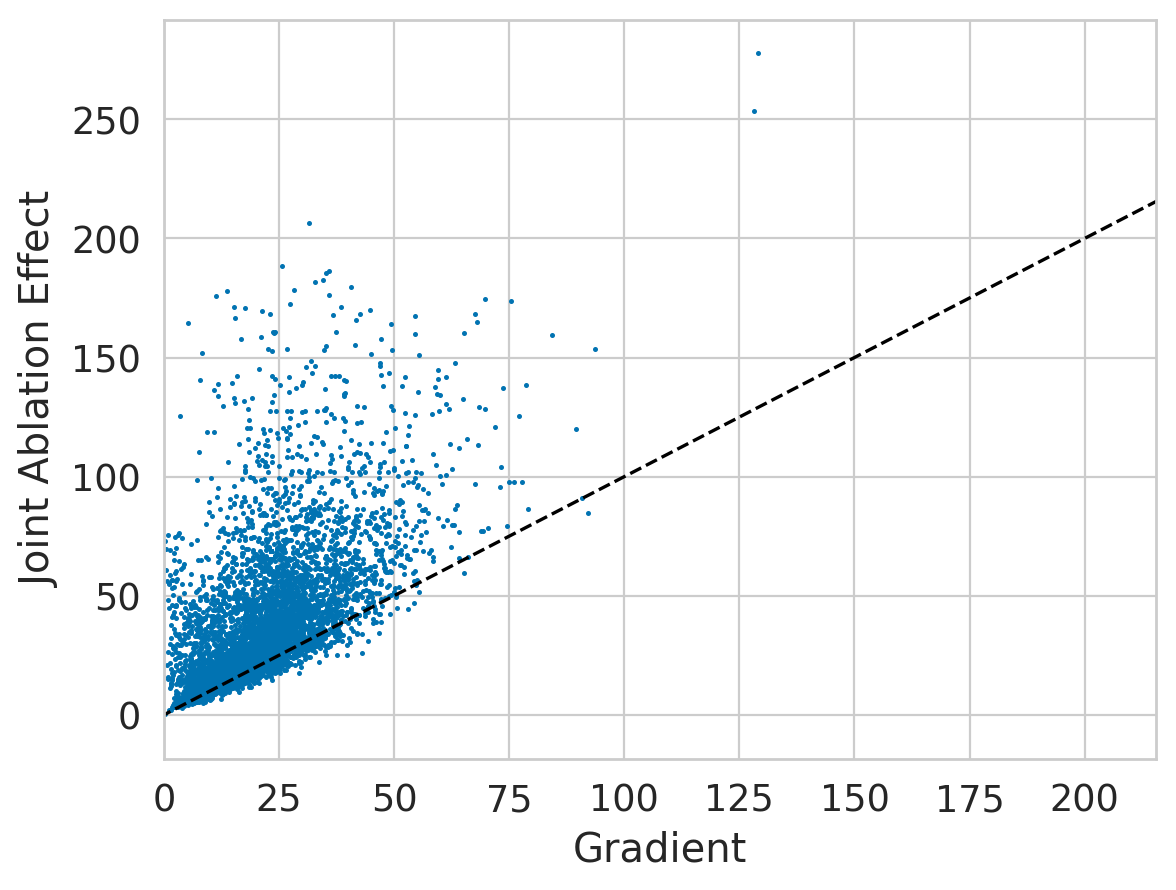}
         \includegraphics[width=\textwidth]{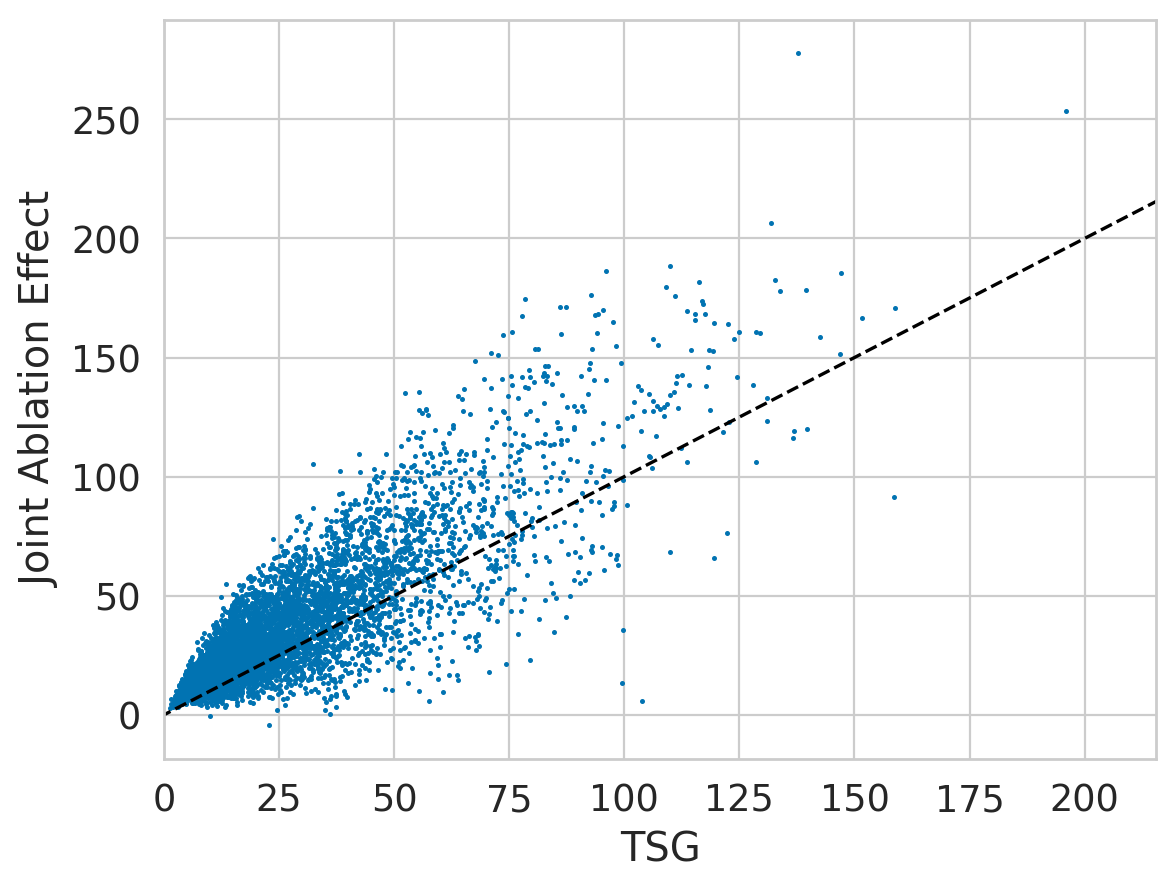}
         \caption{Movie}
     \end{subfigure}
     \hfill
     \begin{subfigure}[b]{0.25\textwidth}
         \centering
         \includegraphics[width=\textwidth]{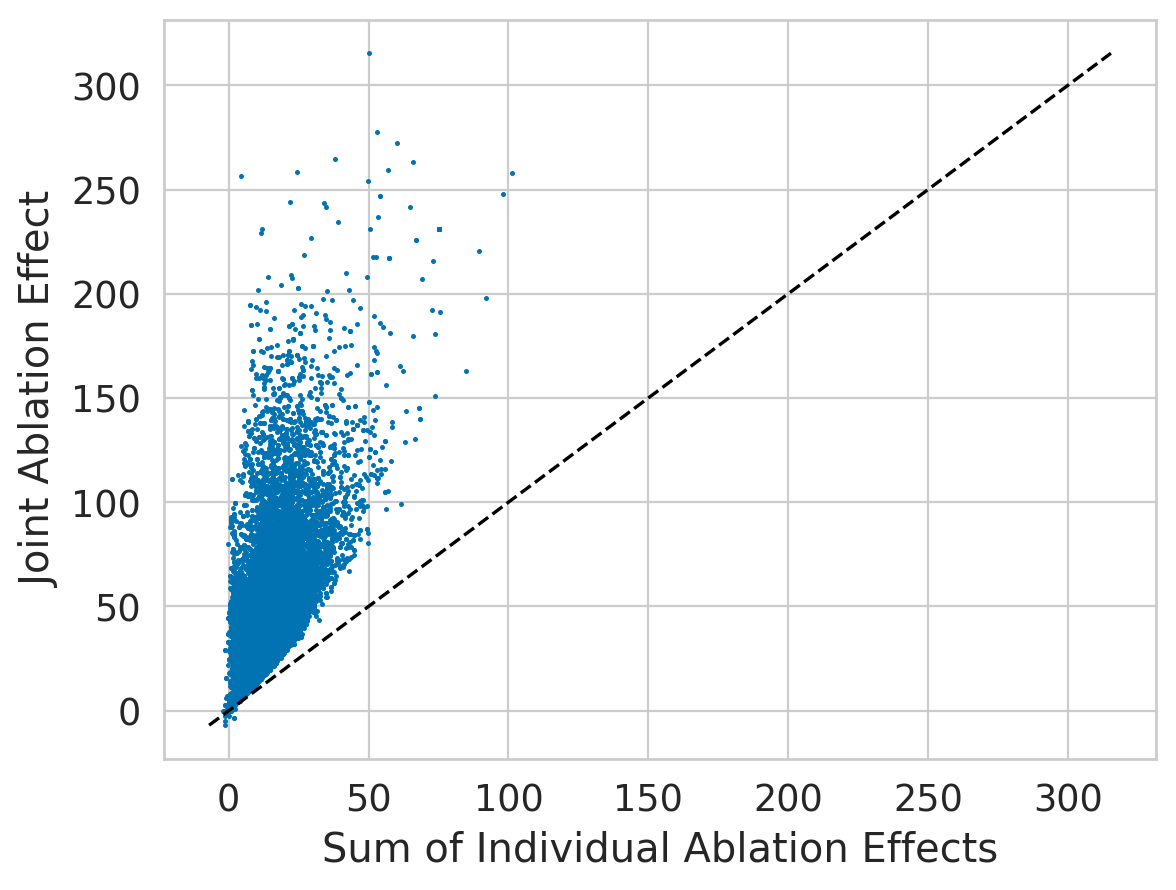}
         \includegraphics[width=\textwidth]{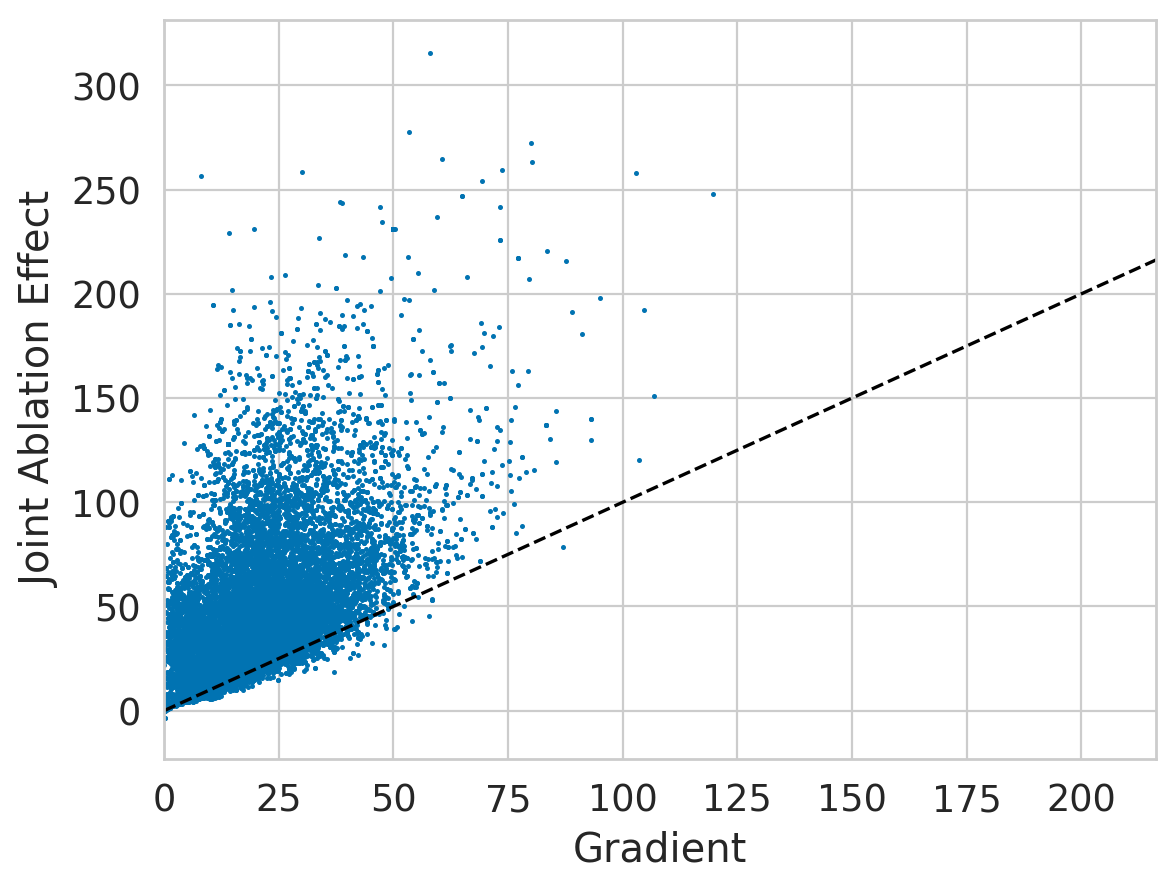}
         \includegraphics[width=\textwidth]{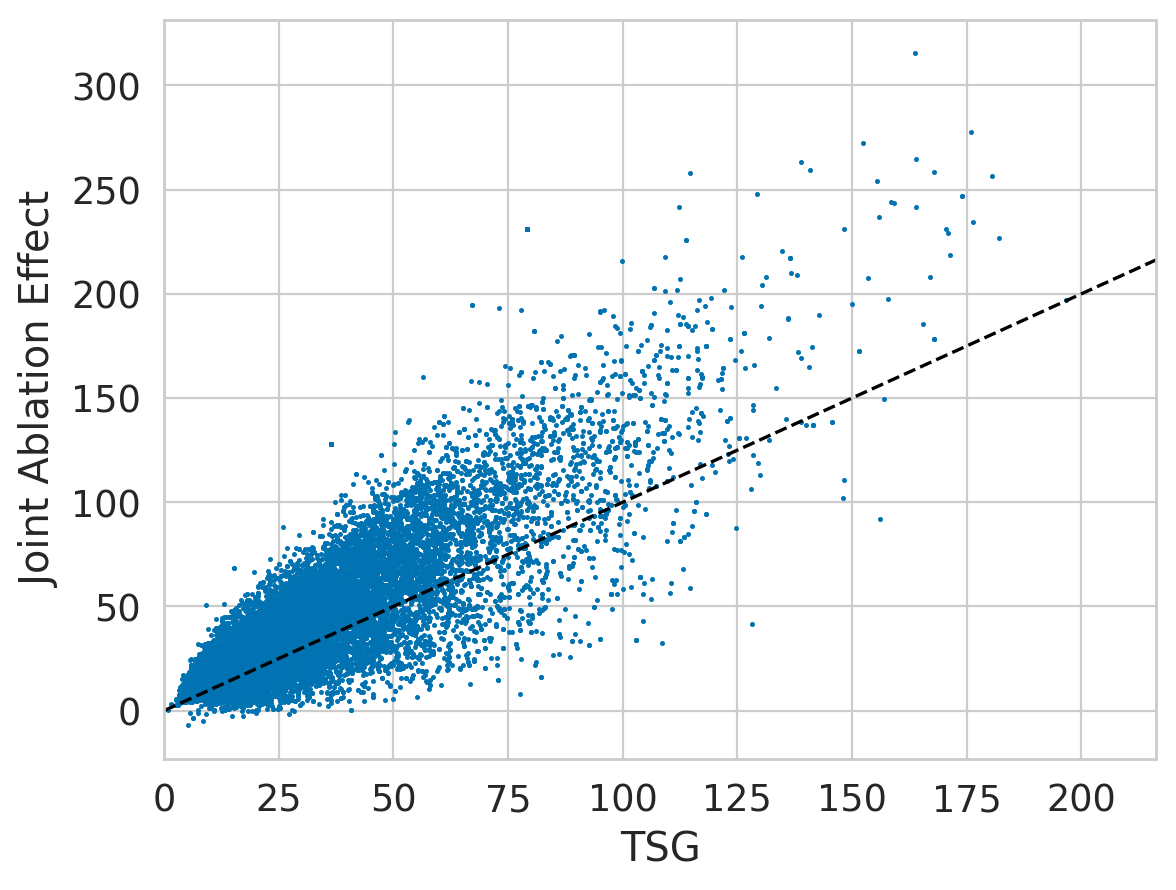}
         \caption{Fever}
     \end{subfigure}
        \caption{Self-repair in three attention scores for  \textbf{Gemma-2 2B} and how TSG increases the attributions for the attention scores with the strongest self-repair effects}
        \label{fig:self-repair-gemma2-three}
\end{figure*}

\begin{figure*}[h]
     \centering
     \begin{subfigure}[b]{0.25\textwidth}
         \centering
         \includegraphics[width=\textwidth]{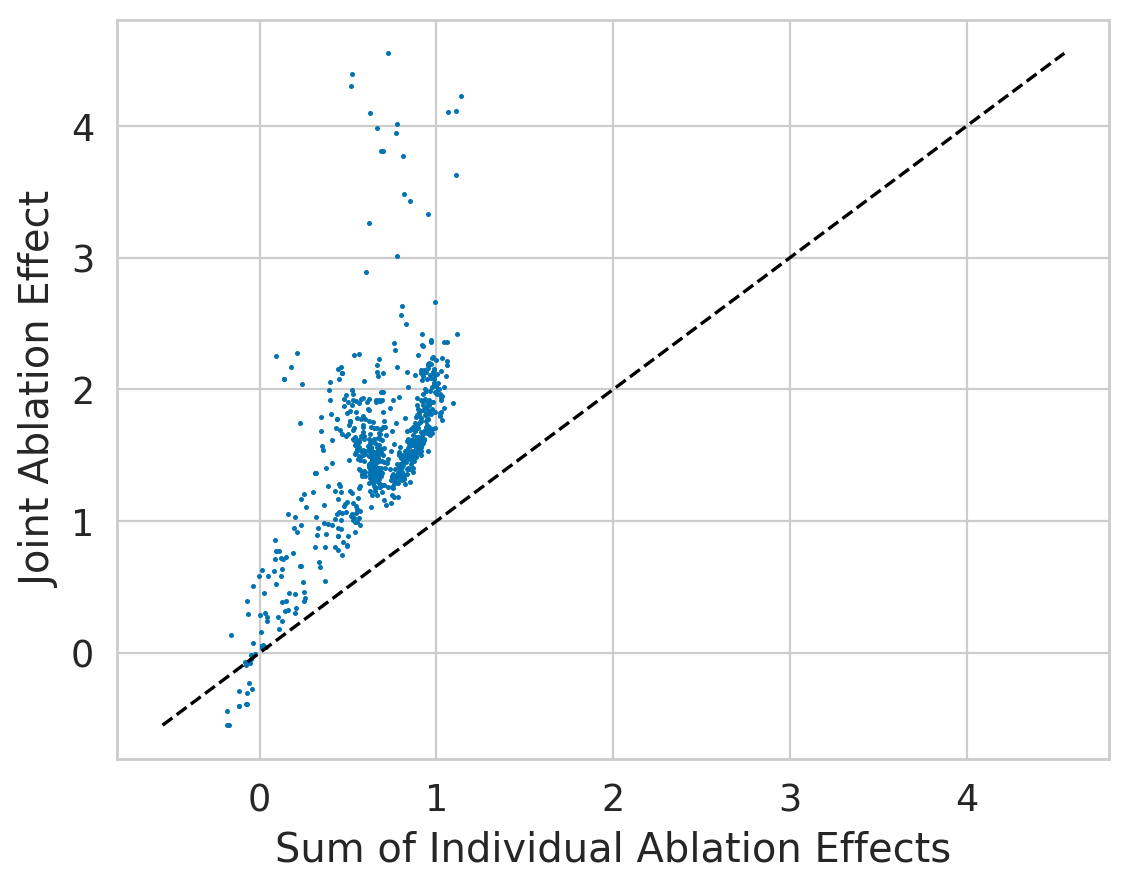}
         \includegraphics[width=\textwidth]{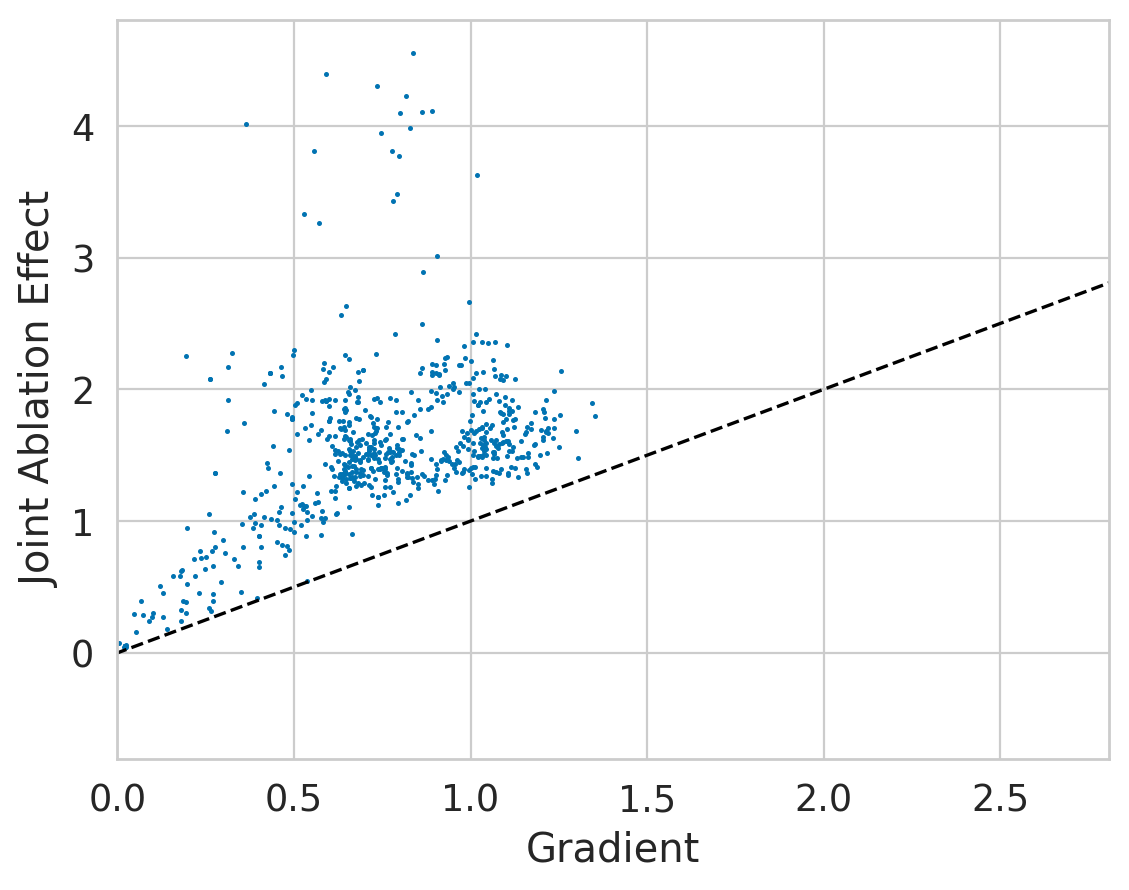}
         \includegraphics[width=\textwidth]{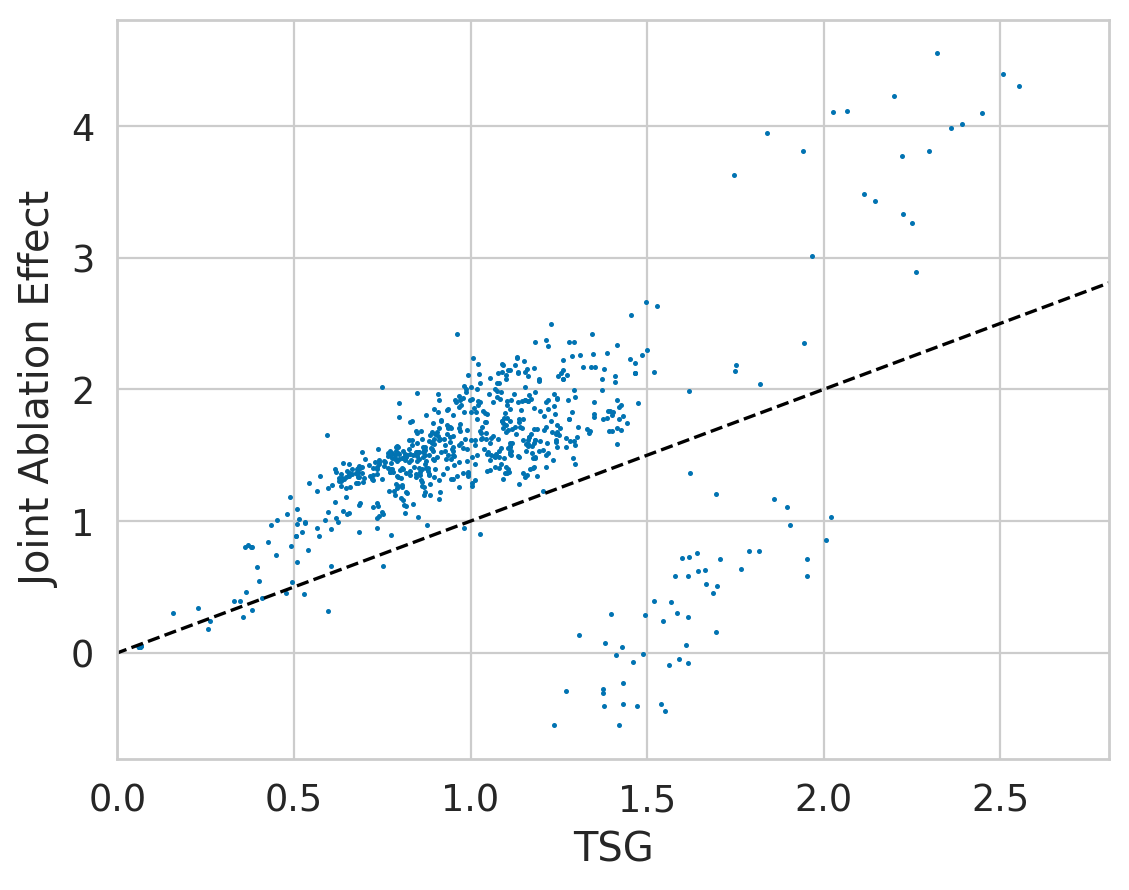}
         \caption{Twitter}
     \end{subfigure}
     \hfill
     \begin{subfigure}[b]{0.25\textwidth}
         \centering
         \includegraphics[width=\textwidth]{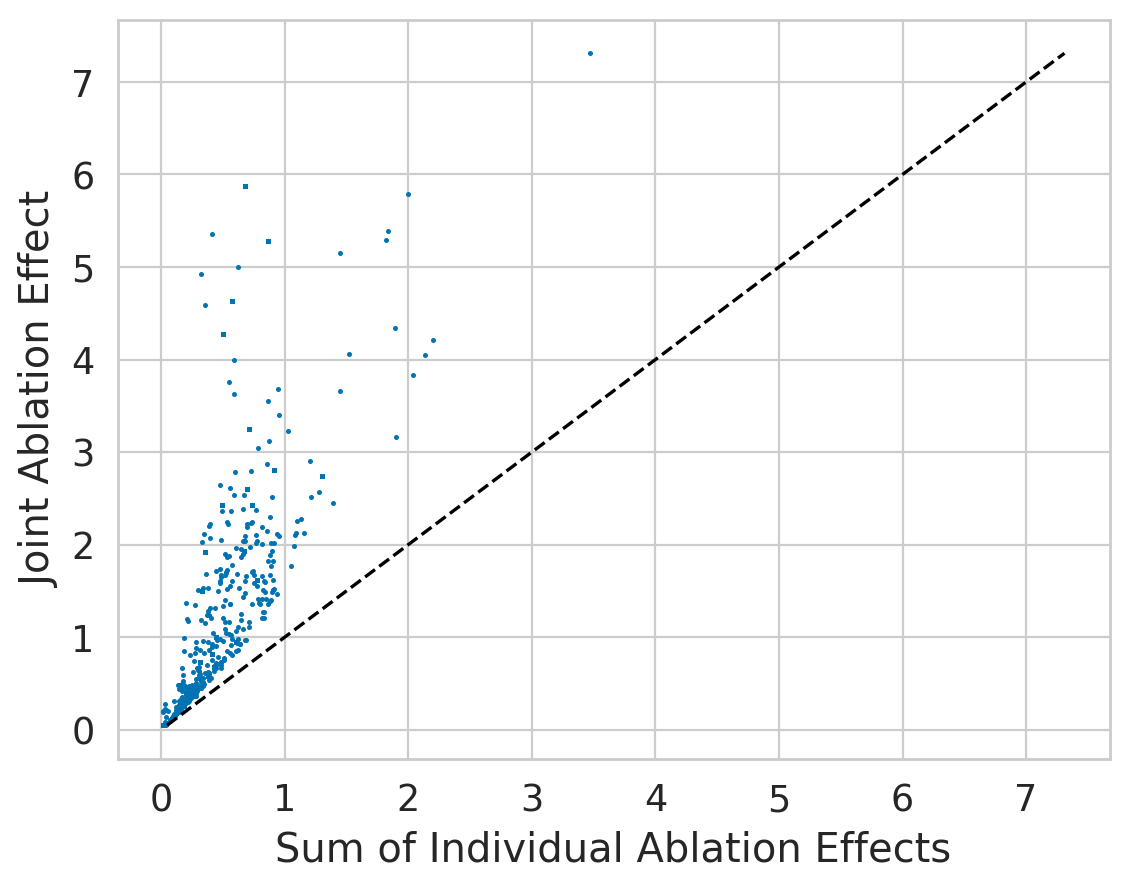}
         \includegraphics[width=\textwidth]{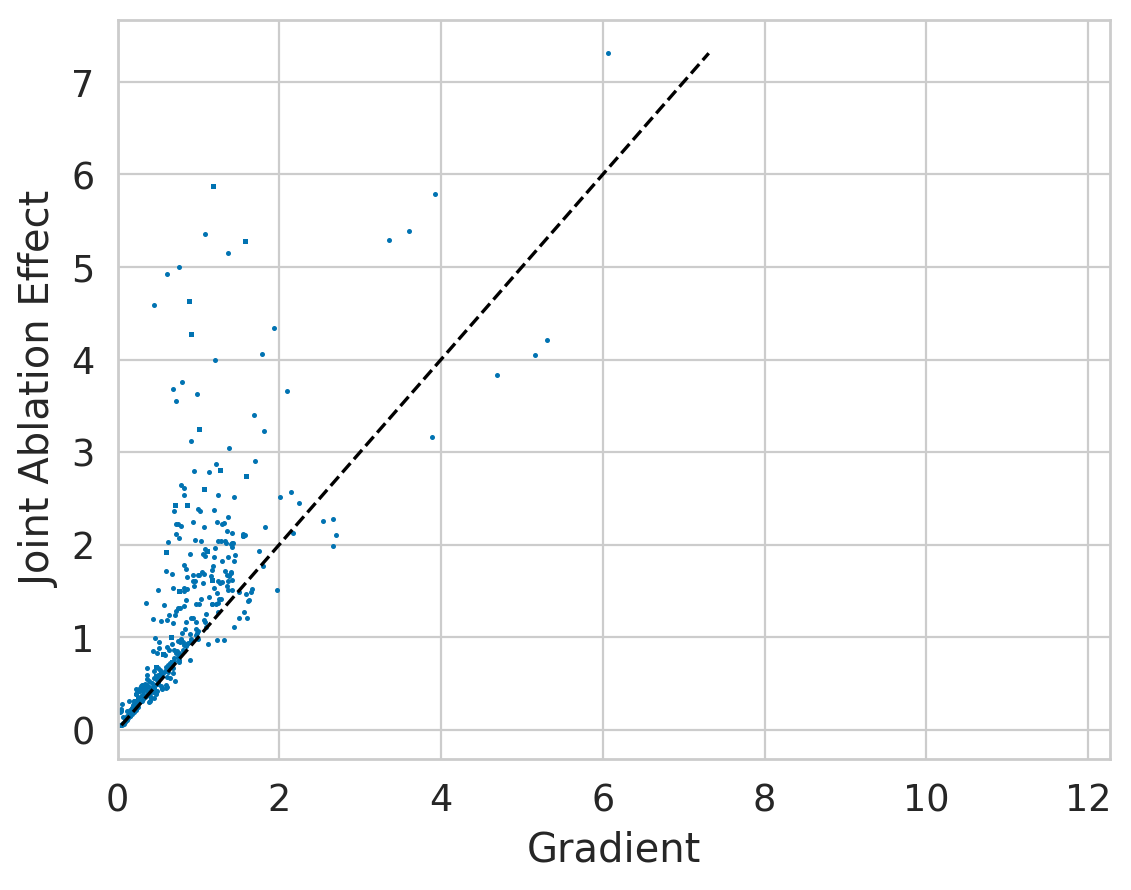}
         \includegraphics[width=\textwidth]{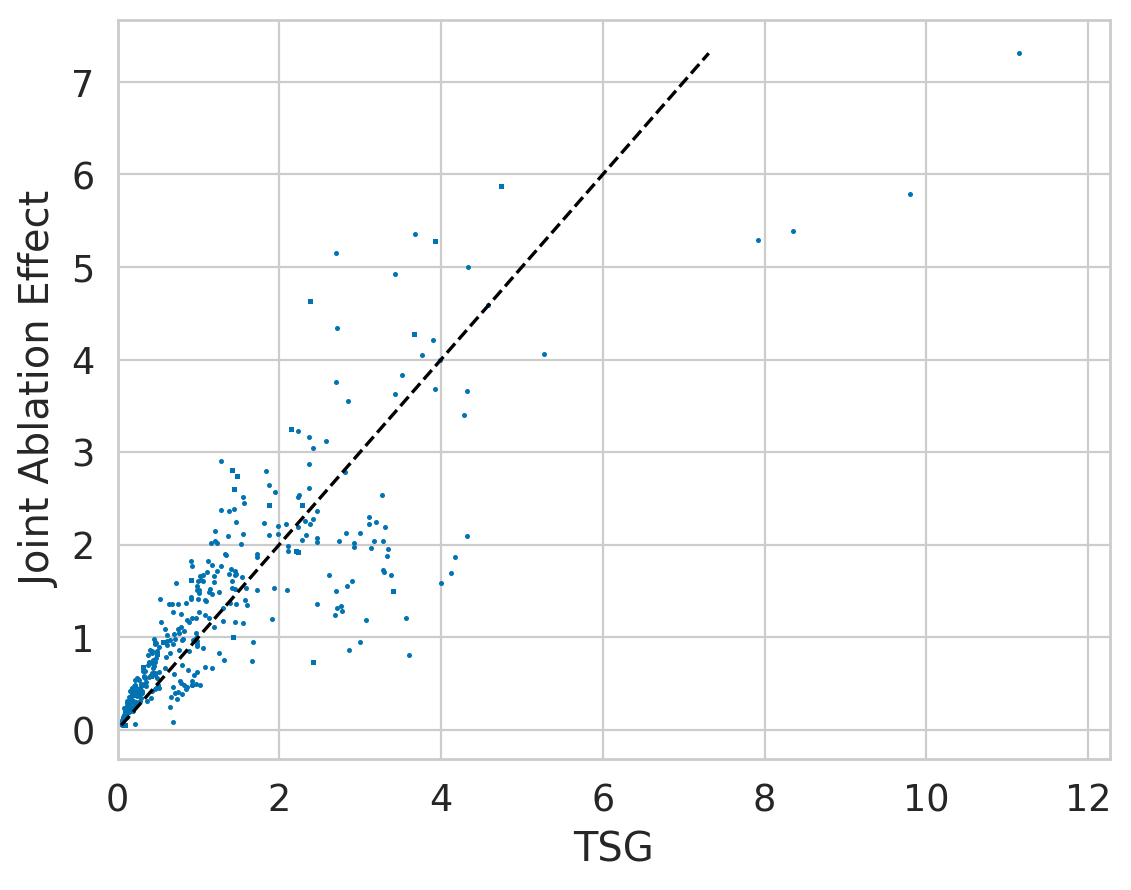}
         \caption{Hatexplain}
     \end{subfigure}
     \hfill
     \begin{subfigure}[b]{0.25\textwidth}
         \centering
         \includegraphics[width=\textwidth]{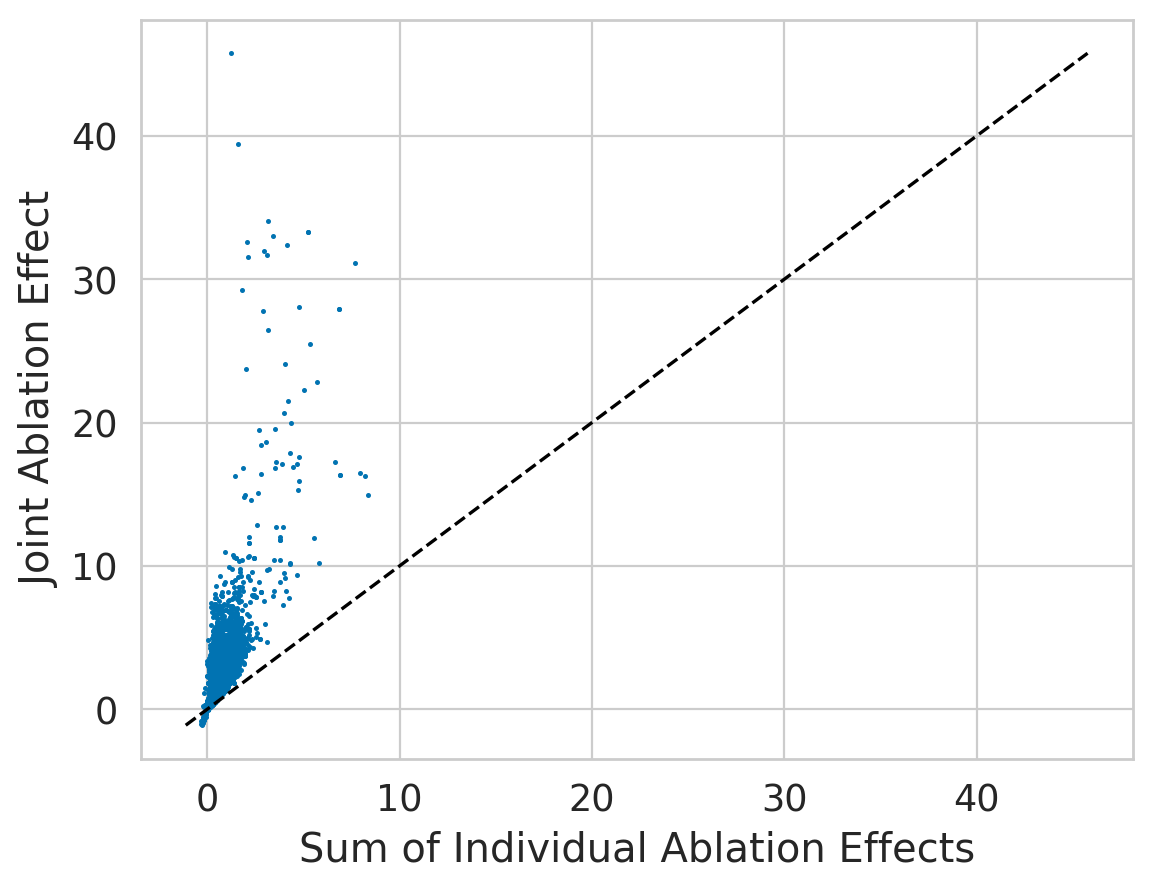}
         \includegraphics[width=\textwidth]{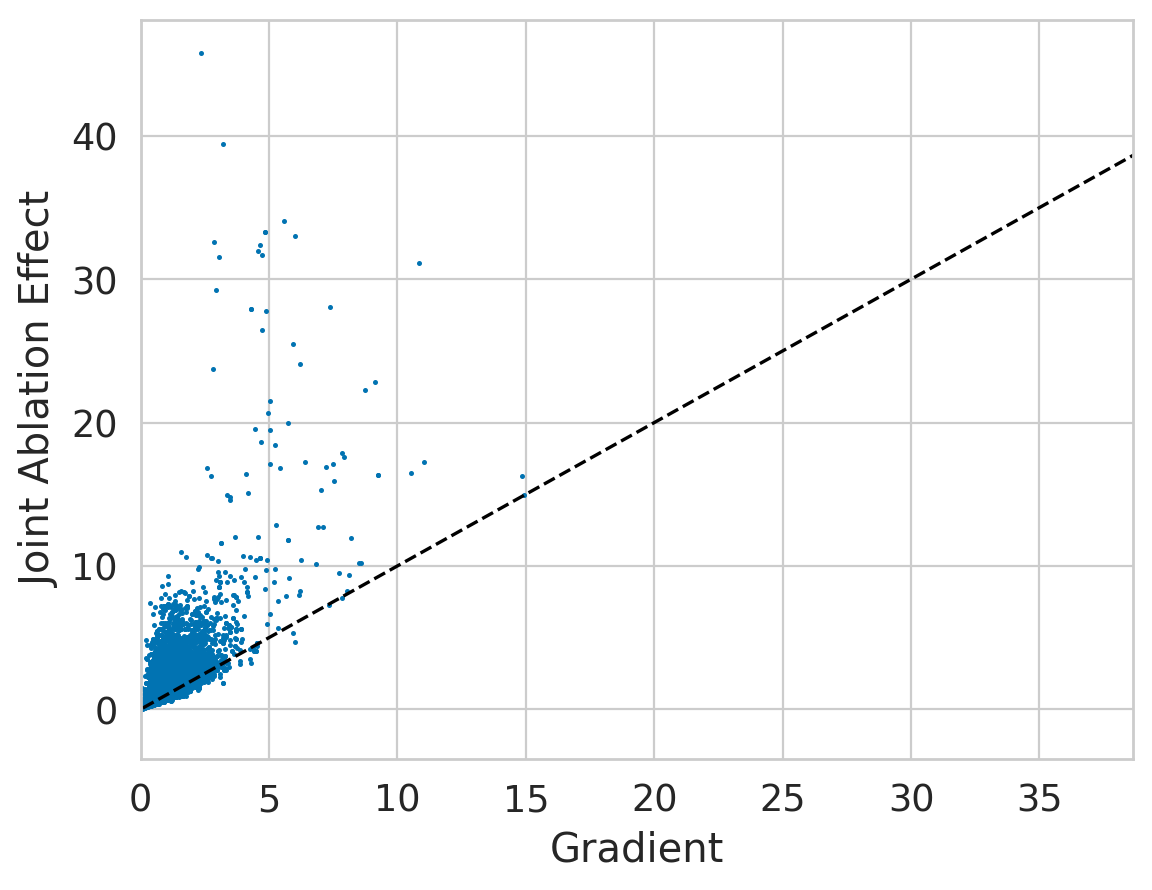}
         \includegraphics[width=\textwidth]{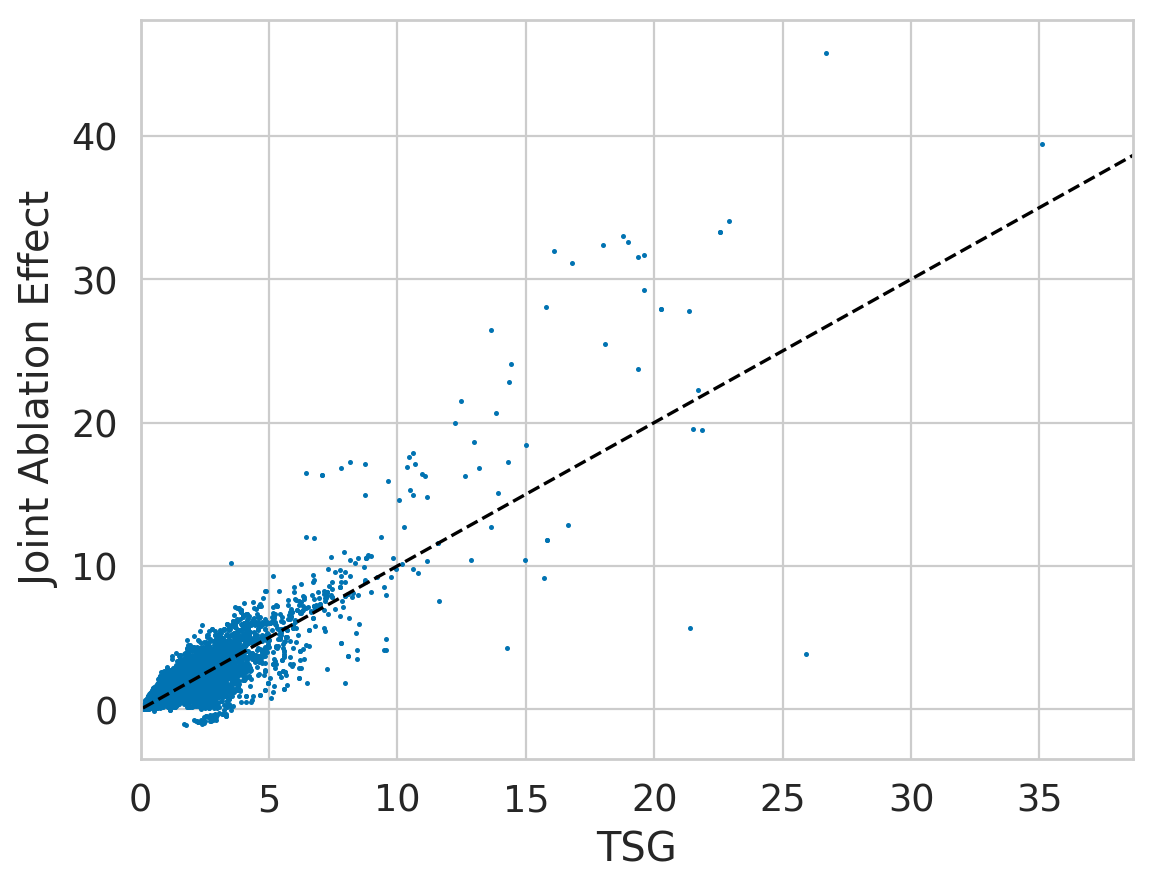}
         
         \caption{Scifact}
     \end{subfigure}
     \begin{subfigure}[b]{0.25\textwidth}
         \centering
         \includegraphics[width=\textwidth]{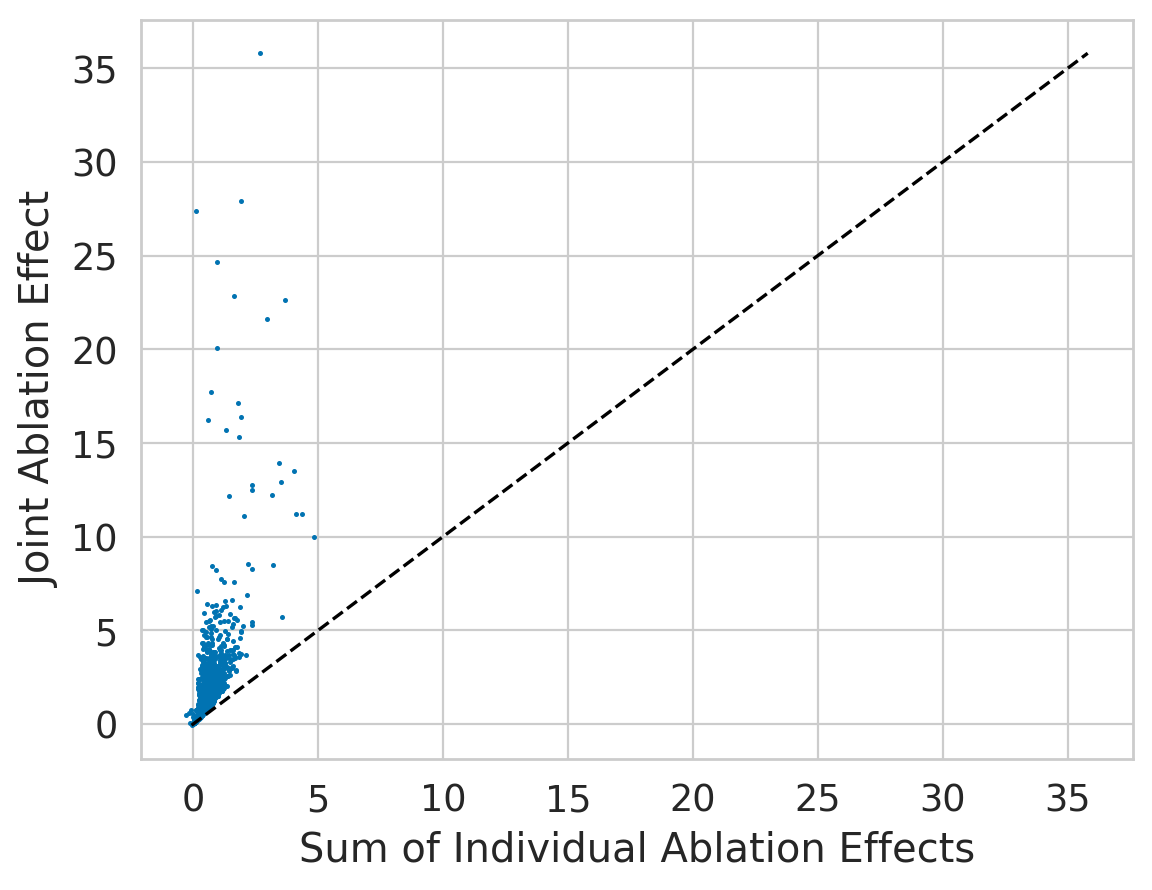}
         \includegraphics[width=\textwidth]{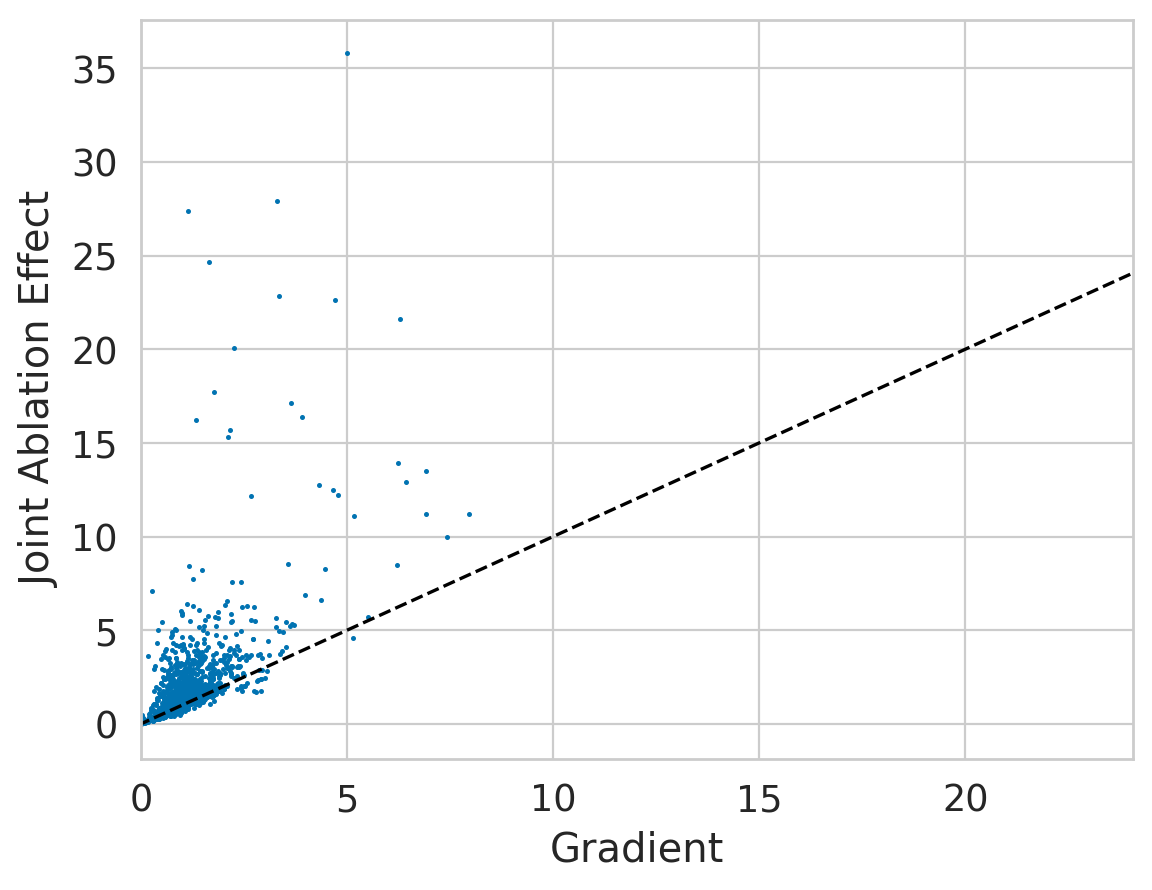}
         \includegraphics[width=\textwidth]{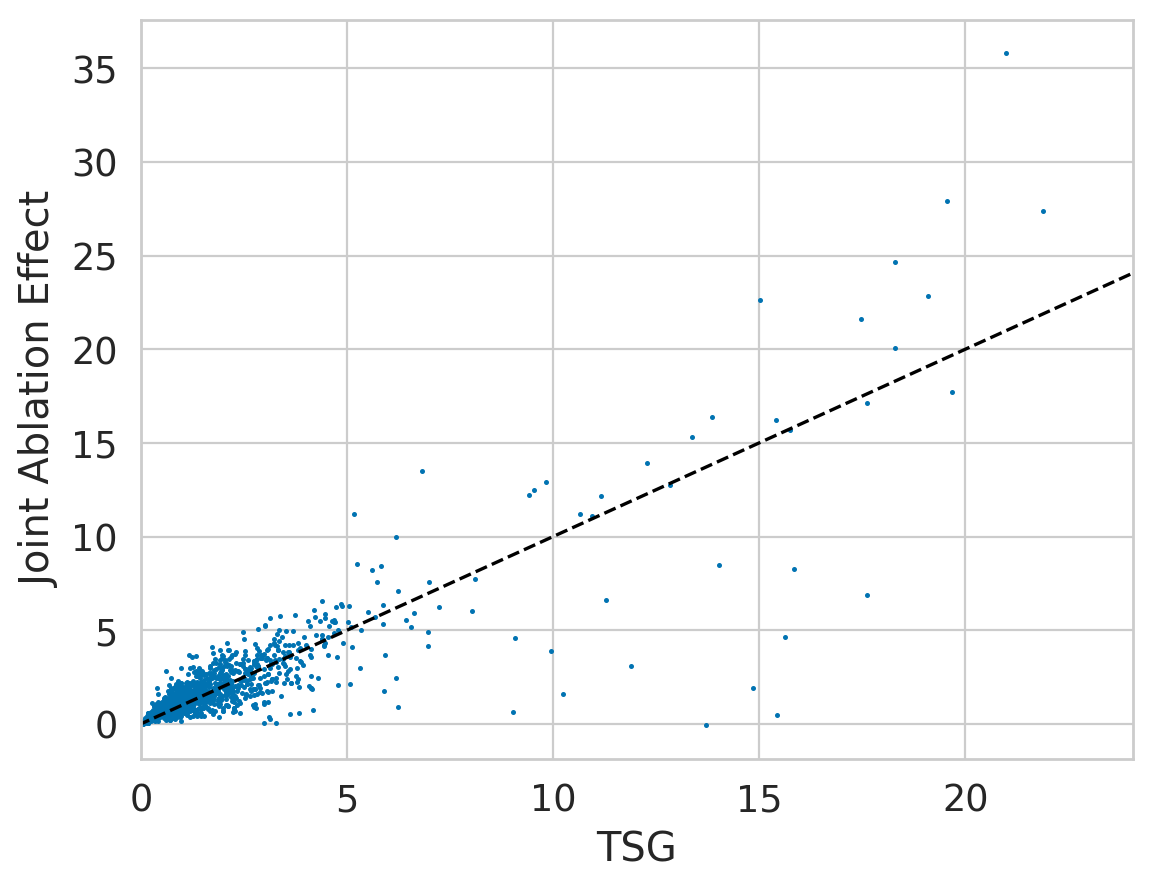}
         \caption{BoolQ}
     \end{subfigure}
     \hfill
     \begin{subfigure}[b]{0.25\textwidth}
         \centering
         \includegraphics[width=\textwidth]{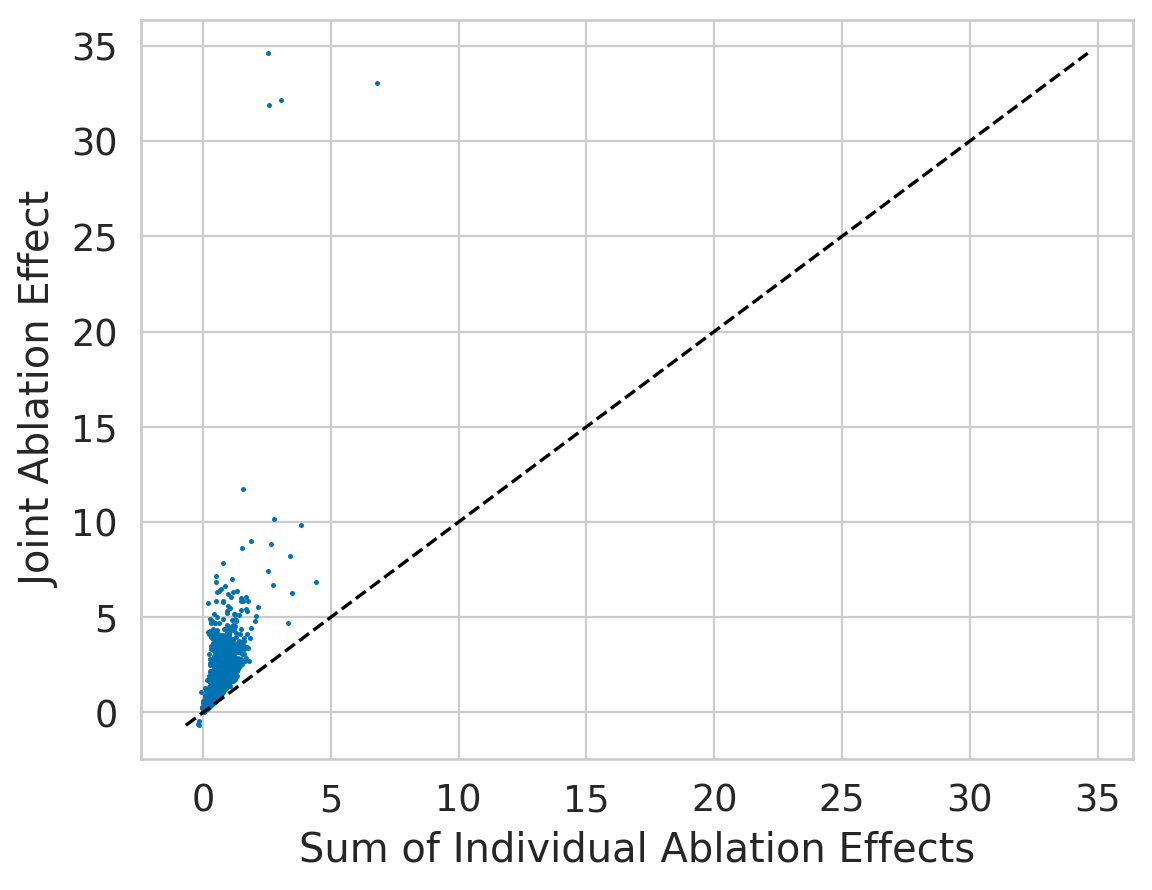}
         \includegraphics[width=\textwidth]{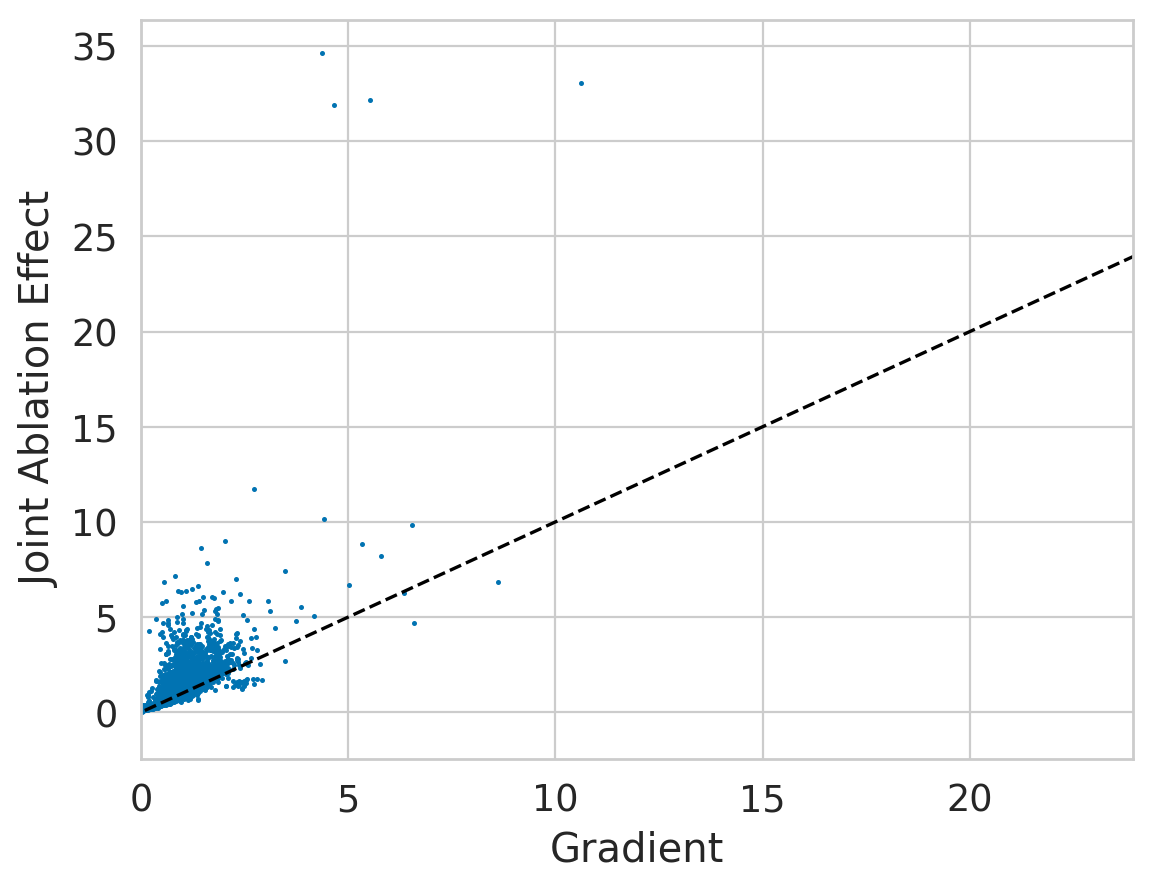}
         \includegraphics[width=\textwidth]{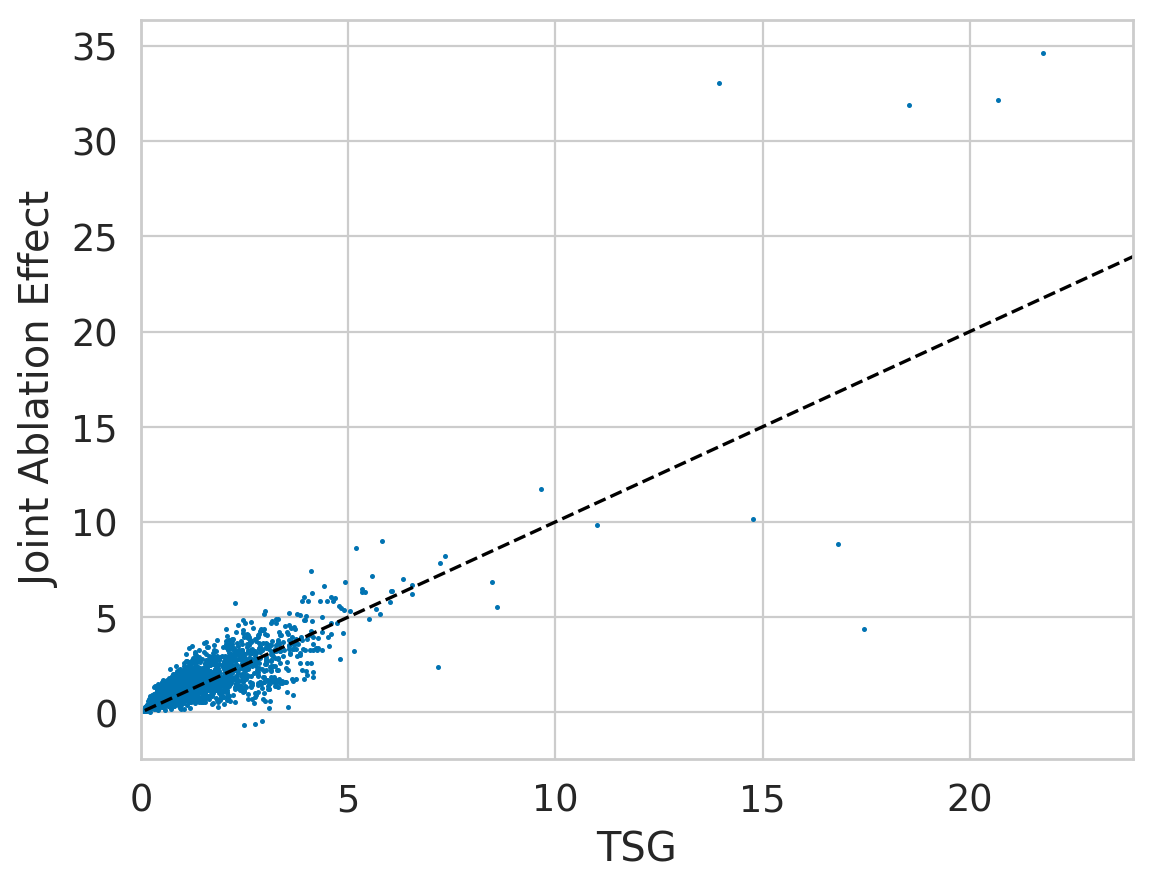}
         \caption{Movie}
     \end{subfigure}
     \hfill
     \begin{subfigure}[b]{0.25\textwidth}
         \centering
         \includegraphics[width=\textwidth]{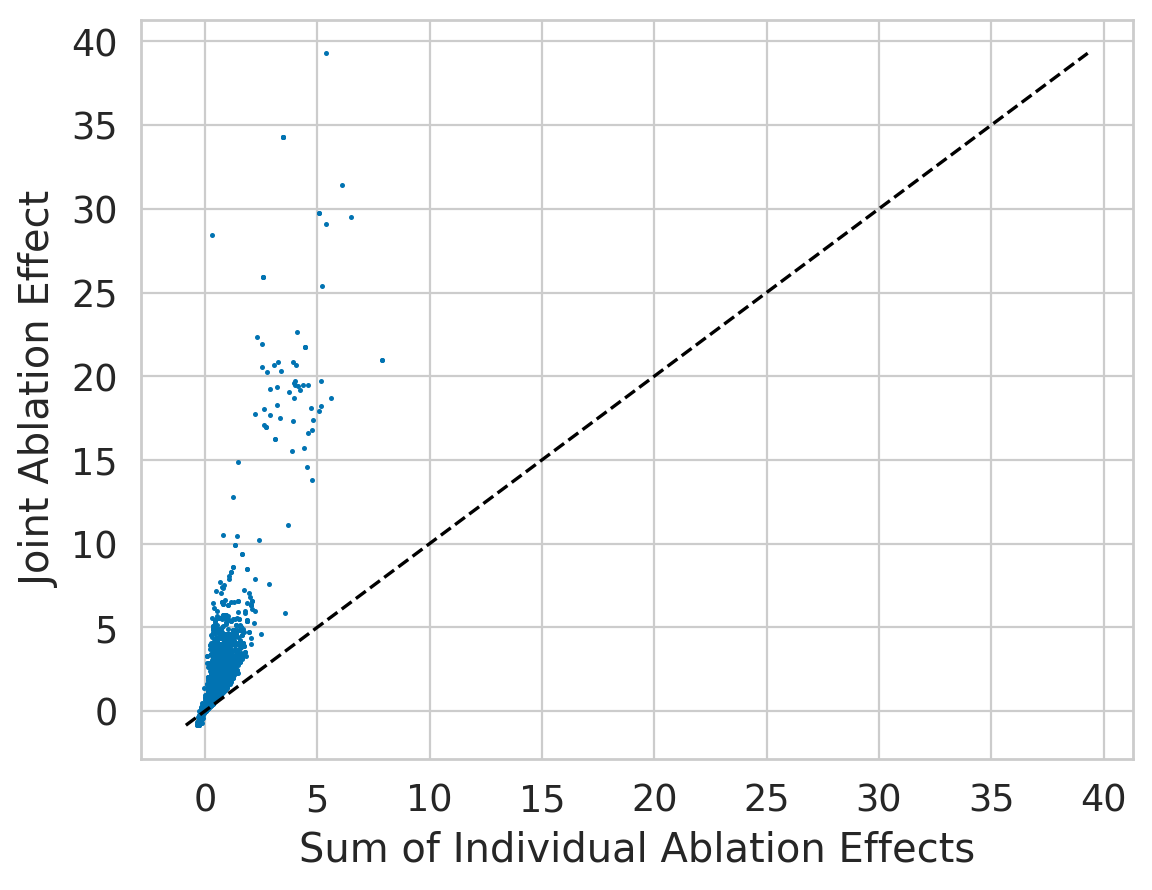}
         \includegraphics[width=\textwidth]{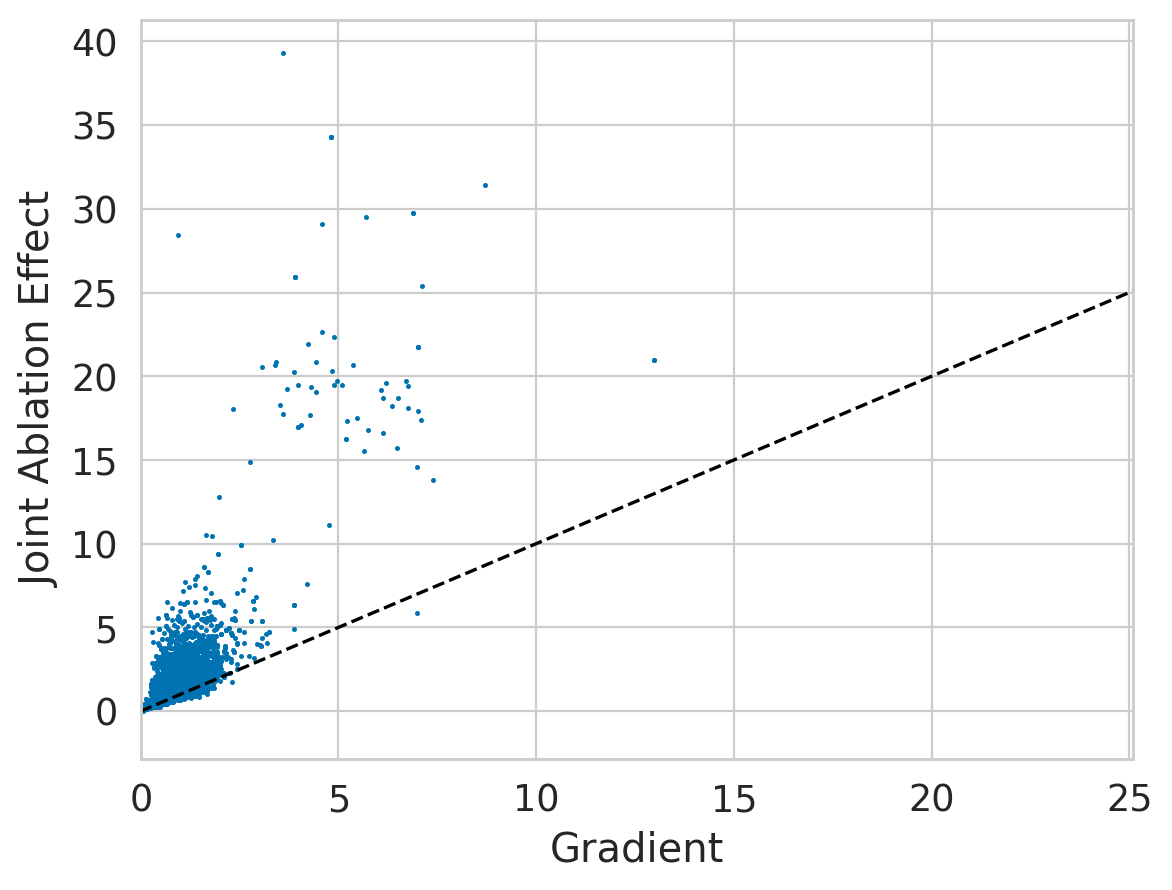}
         \includegraphics[width=\textwidth]{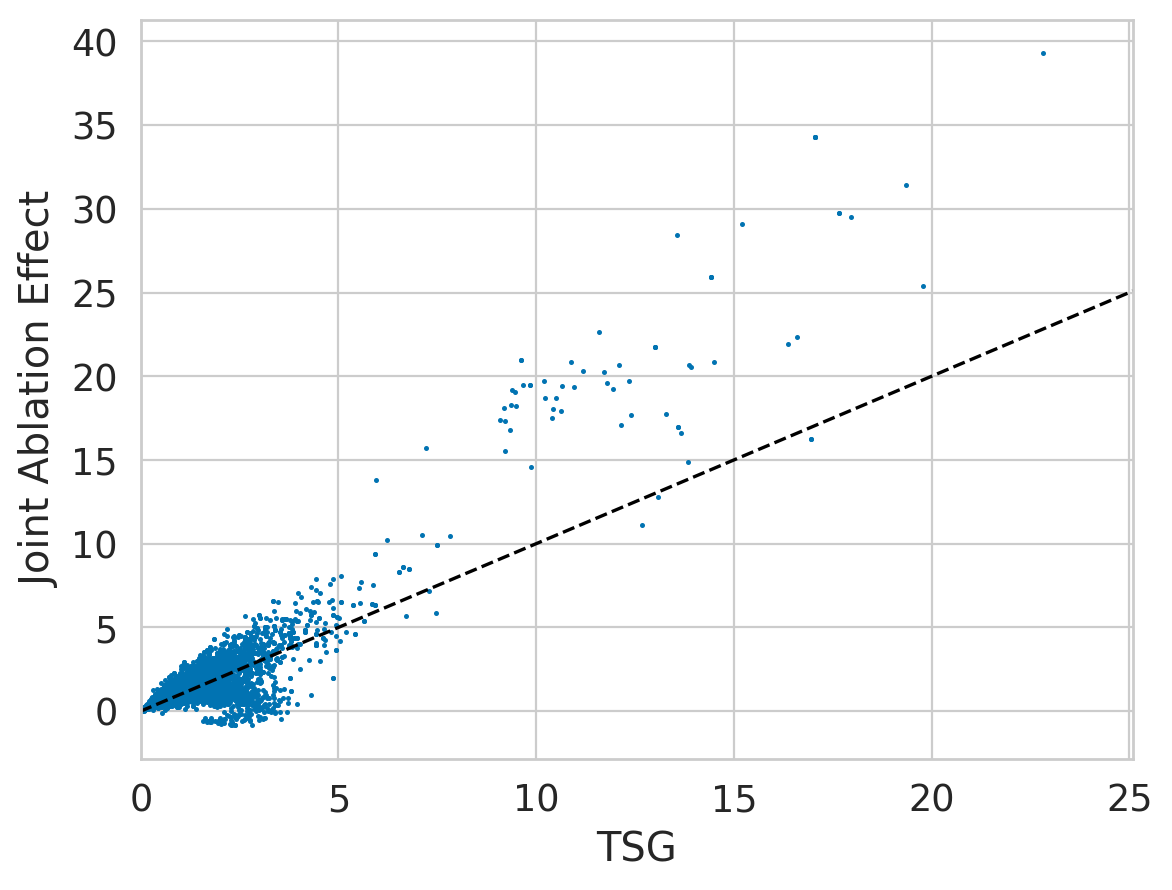}
         \caption{Fever}
     \end{subfigure}
        \caption{Self-repair in three attention scores for  \textbf{LLAMA-3.2 1B} and how TSG increases the attributions for the attention scores with the strongest self-repair effects}
        \label{fig:self-repair-llama1-three}
\end{figure*}

\subsection{Self-repair in three attention scores} \label{app:self-repair_three}
In the main paper, we only analyze self-repair for two attention scores; however, self-repair can occur for more attention scores. In \Cref{fig:self-repair-gemma2-three}, \Cref{fig:self-repair-llama1-three}, and \Cref{fig:self-repair-llama3-three}, we present additional empirical evidence for self-repair. We compare the joint ablation effect with the gradients, TSG, and the sum of individual ablation effects. TSG better approximates the joint ablation effect than the other two.

\subsection{Self-repair results with higher and lower threshold} \label{app:self-repair-threshold}
\begin{figure*}[h]
     \centering
     \begin{subfigure}[b]{0.25\textwidth}
         \centering
         \includegraphics[width=\textwidth]{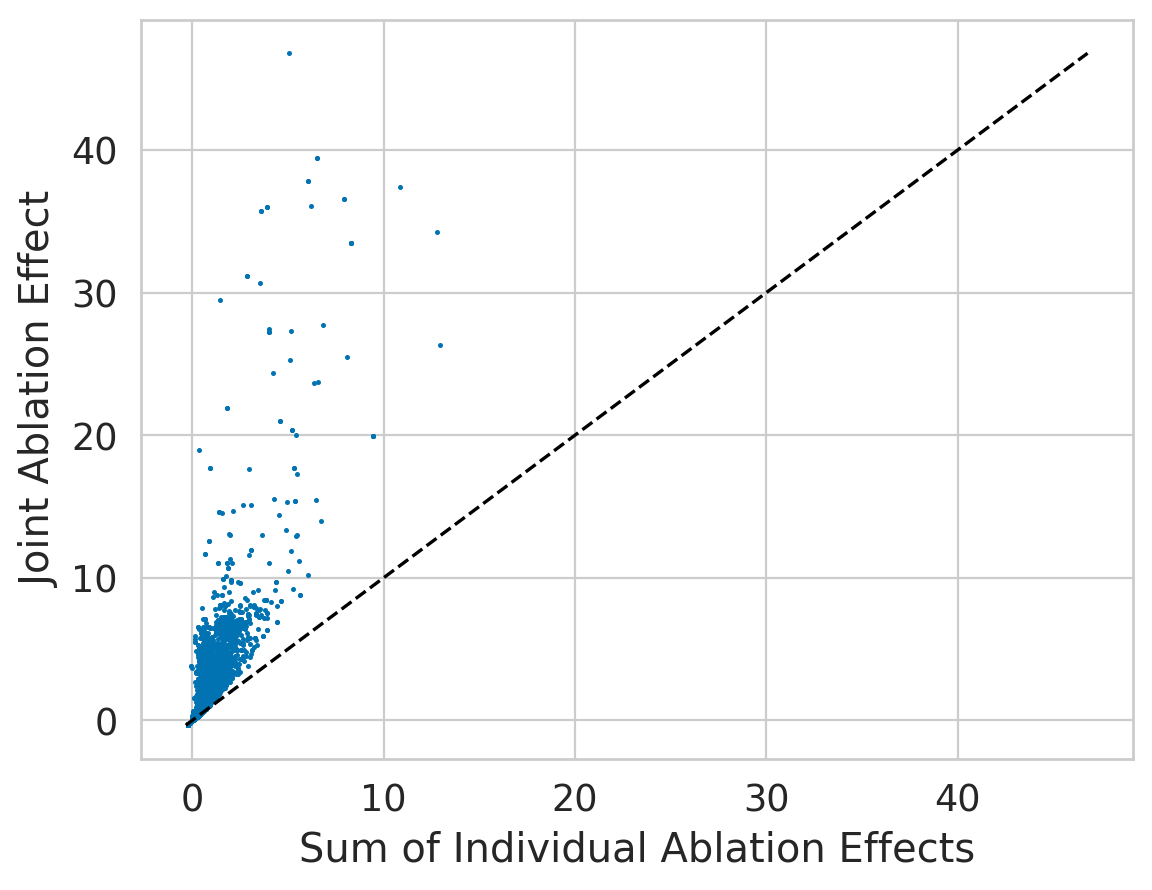}
         \includegraphics[width=\textwidth]{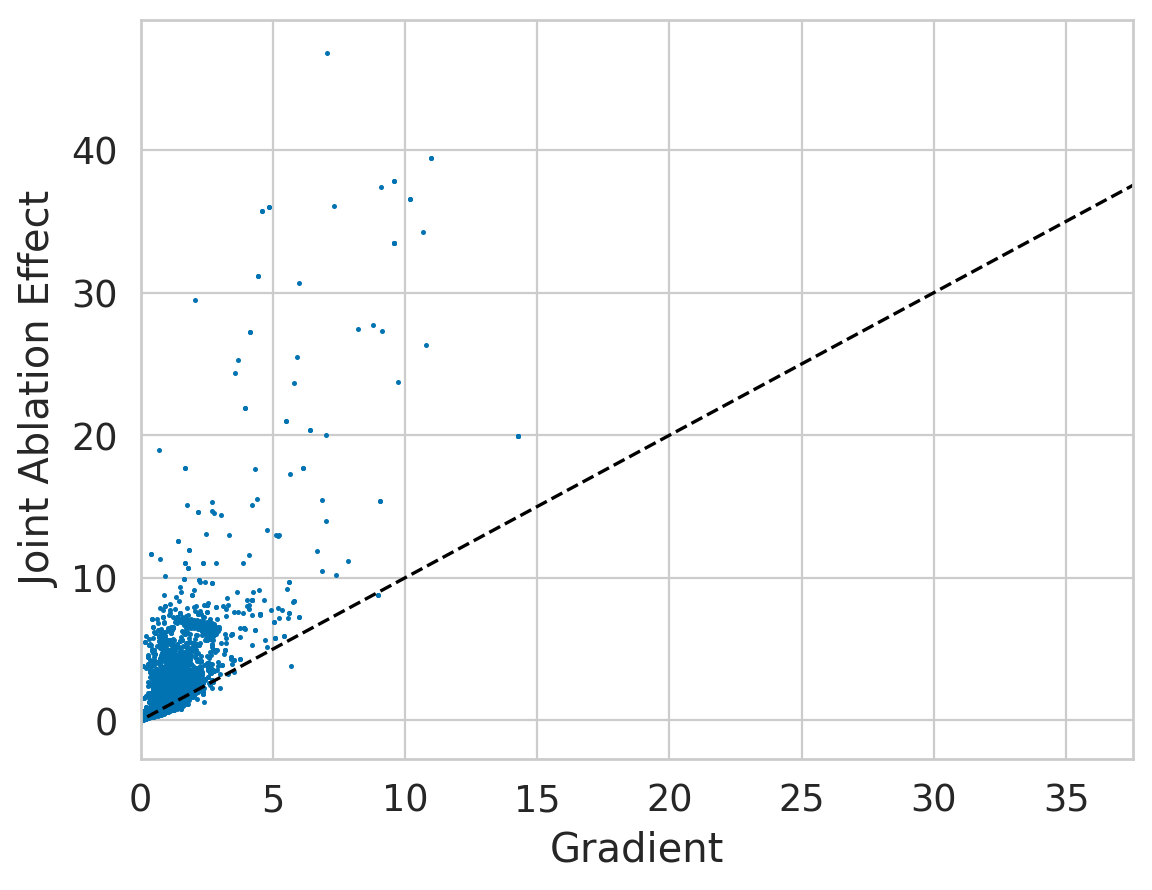}
         \includegraphics[width=\textwidth]{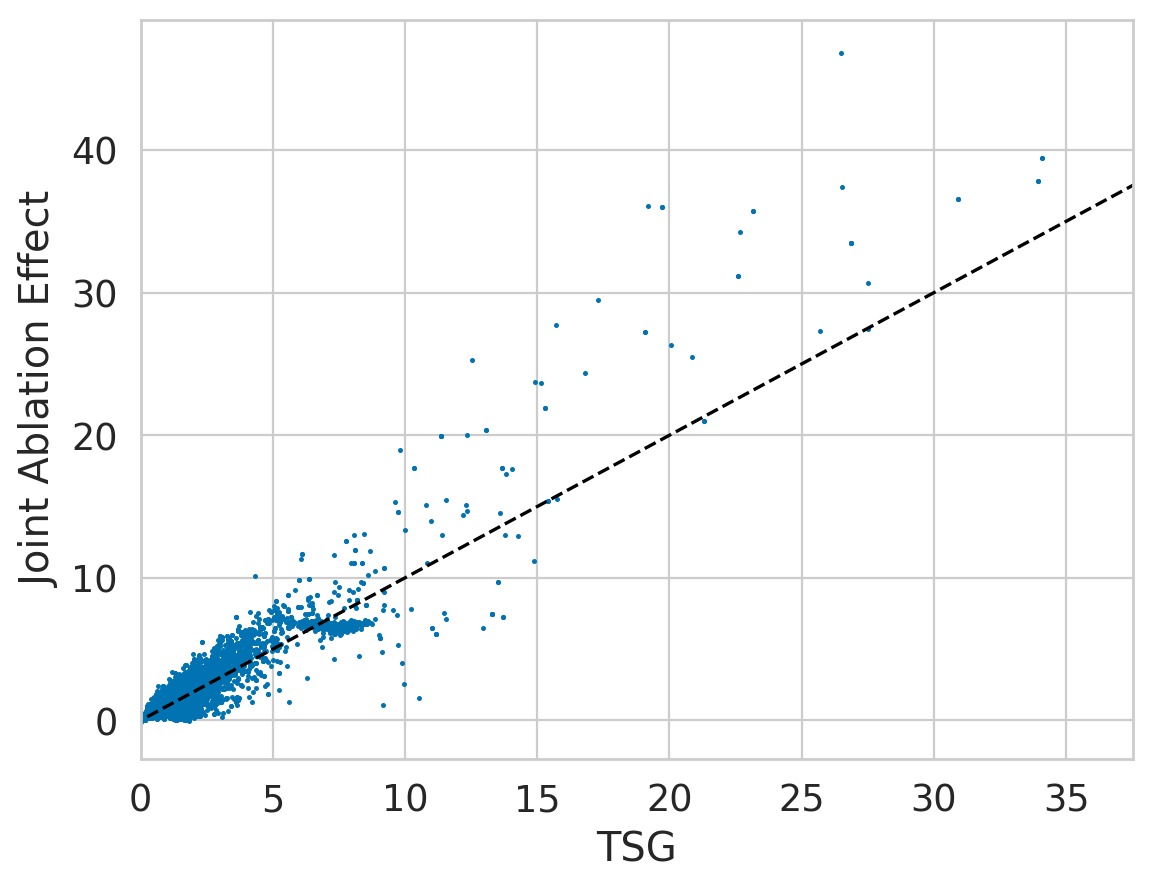}
         \caption{Fever DB=0.05}
     \end{subfigure}
     \hfill
     \begin{subfigure}[b]{0.25\textwidth}
         \centering
         \includegraphics[width=\textwidth]{figures/self_repair_0.1/fever/llama3-1b-it/ablation_effect.png}
         \includegraphics[width=\textwidth]{figures/self_repair_0.1/fever/llama3-1b-it/grad_vs_joint.png}
         \includegraphics[width=\textwidth]{figures/self_repair_0.1/fever/llama3-1b-it/tsg_vs_joint.png}
         \caption{Fever DB=0.1}
     \end{subfigure}
     \hfill
     \begin{subfigure}[b]{0.25\textwidth}
         \centering
         \includegraphics[width=\textwidth]{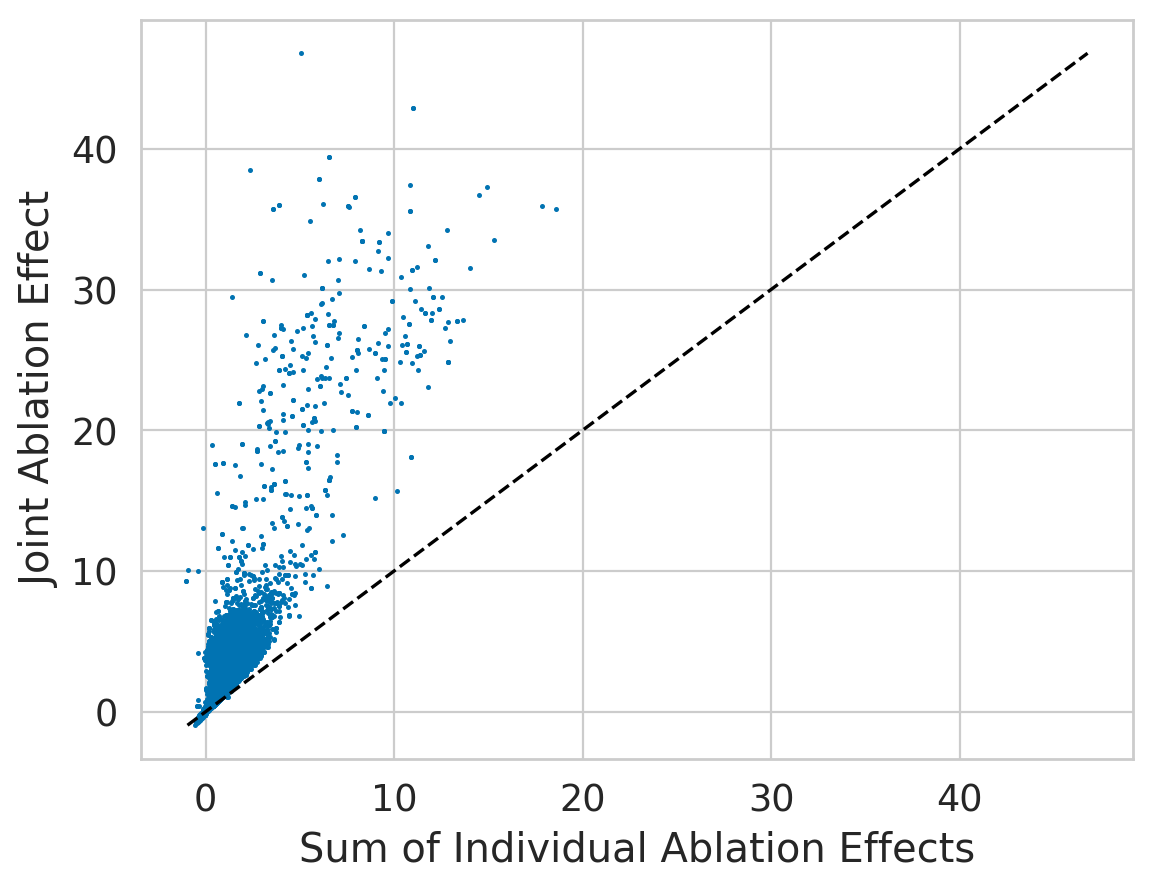}
         \includegraphics[width=\textwidth]{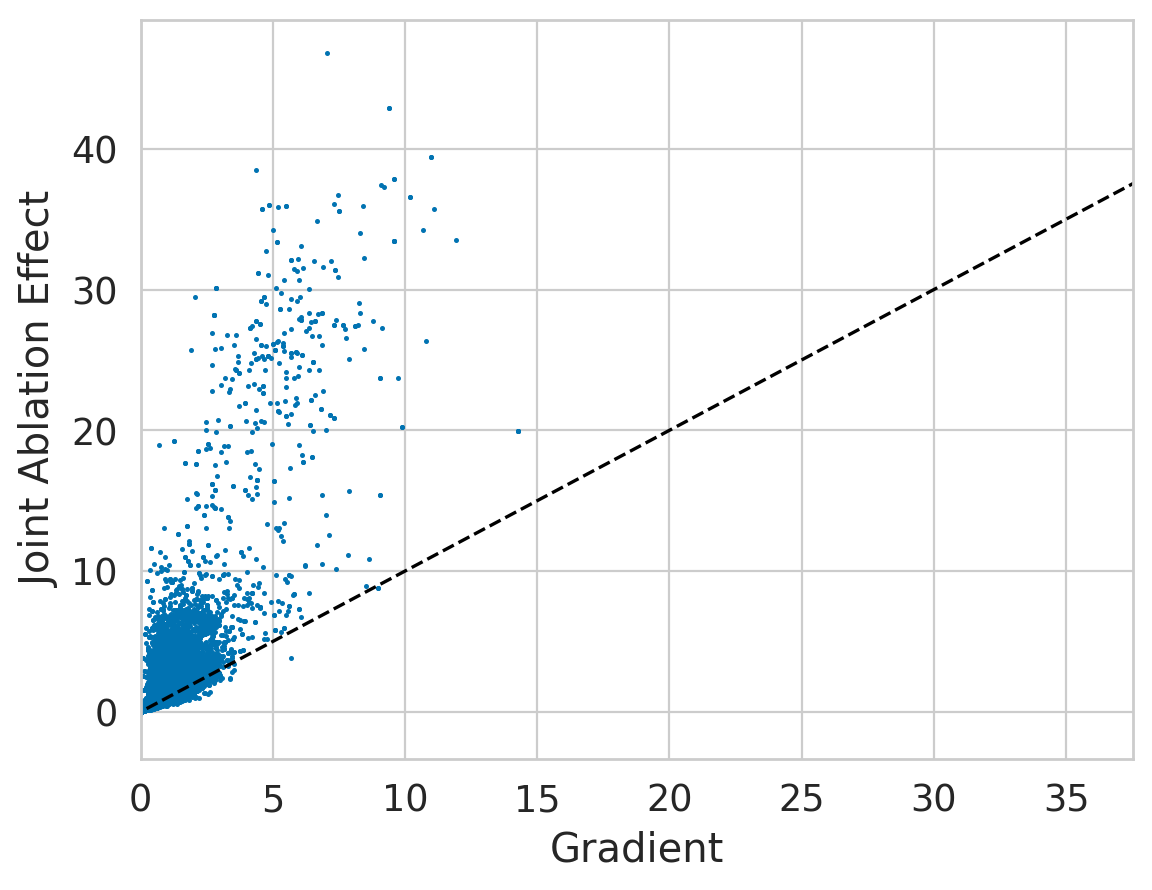}
         \includegraphics[width=\textwidth]{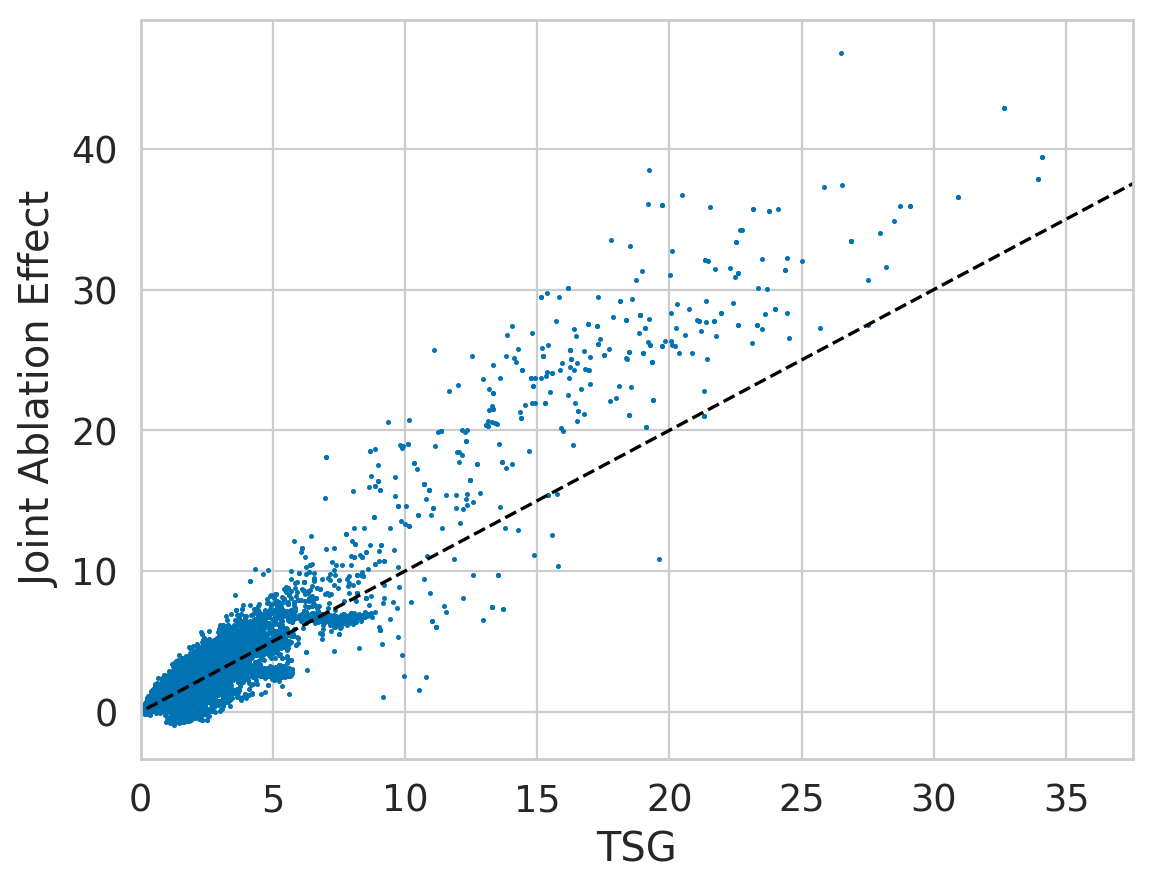}
         \caption{Fever DB=0.2}
     \end{subfigure}
  
     \begin{subfigure}[b]{0.25\textwidth}
         \centering
         \includegraphics[width=\textwidth]{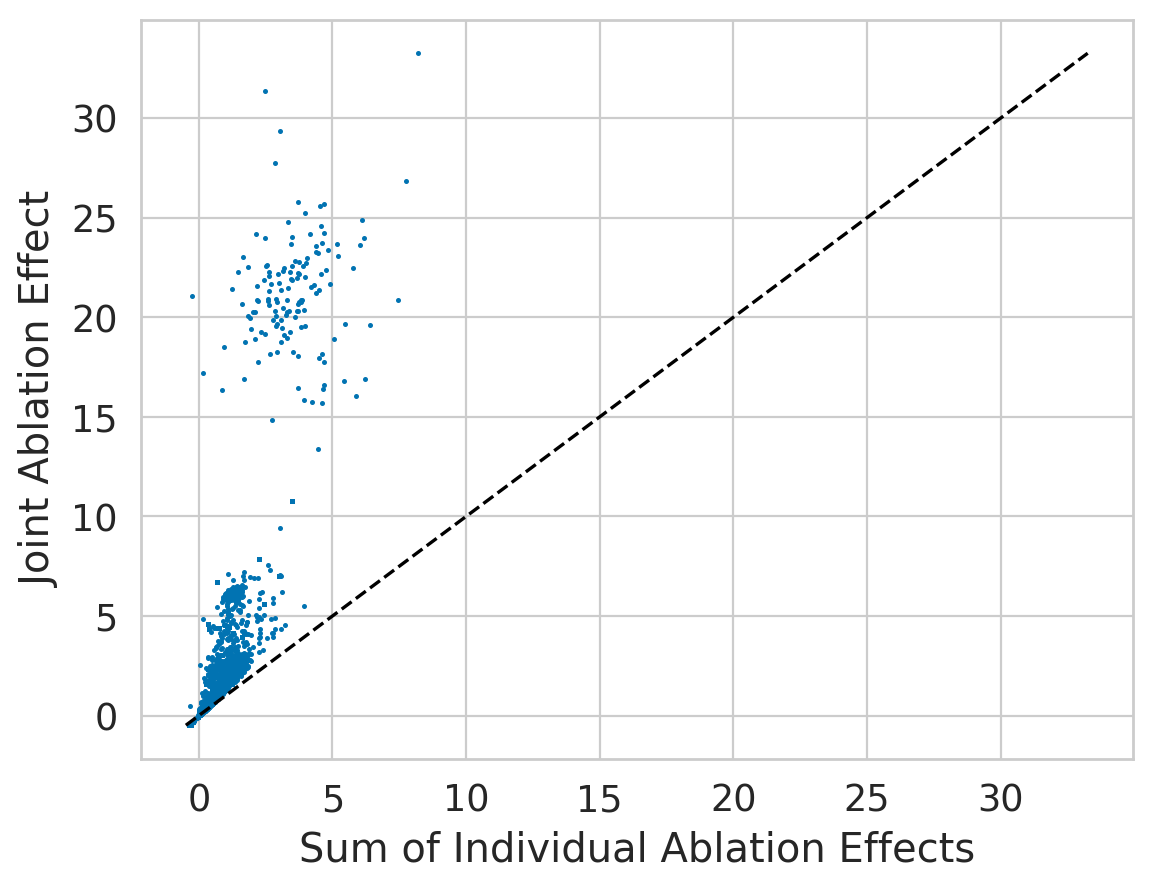}
         \includegraphics[width=\textwidth]{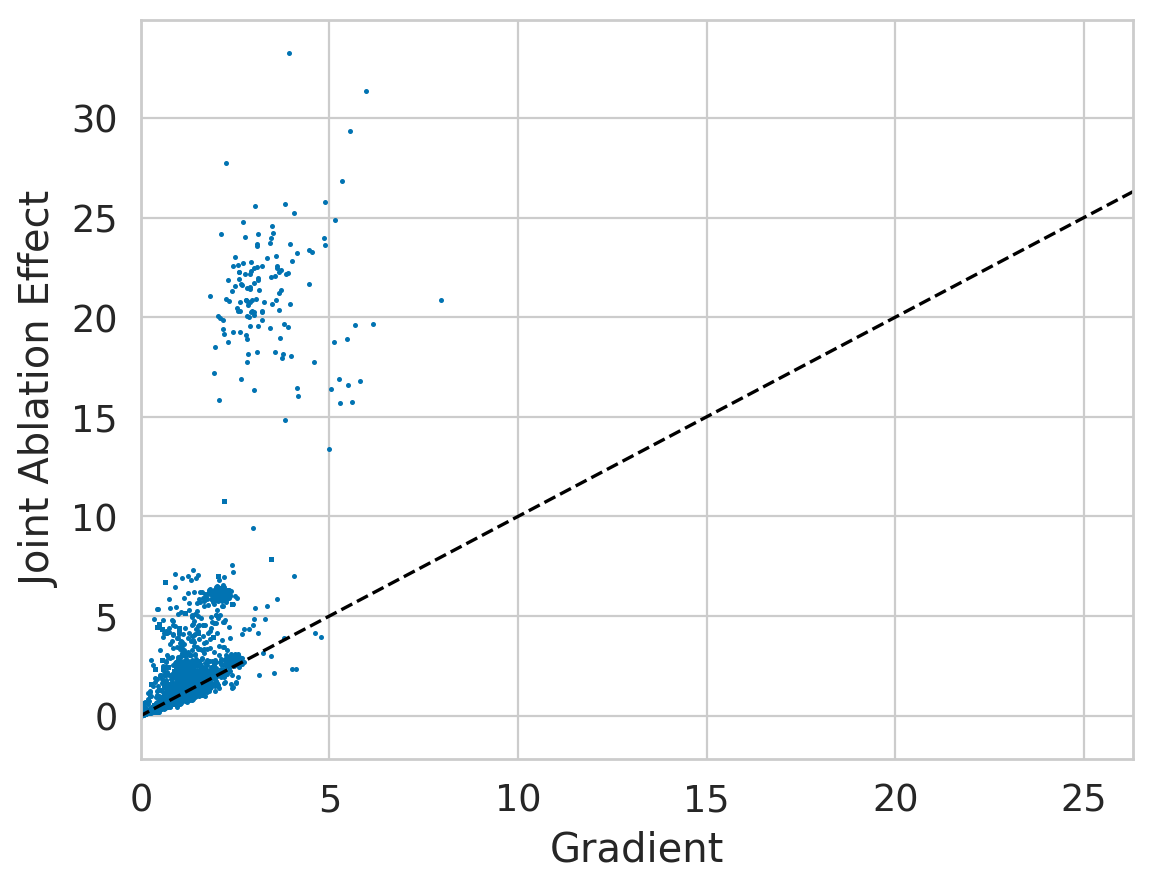}
         \includegraphics[width=\textwidth]{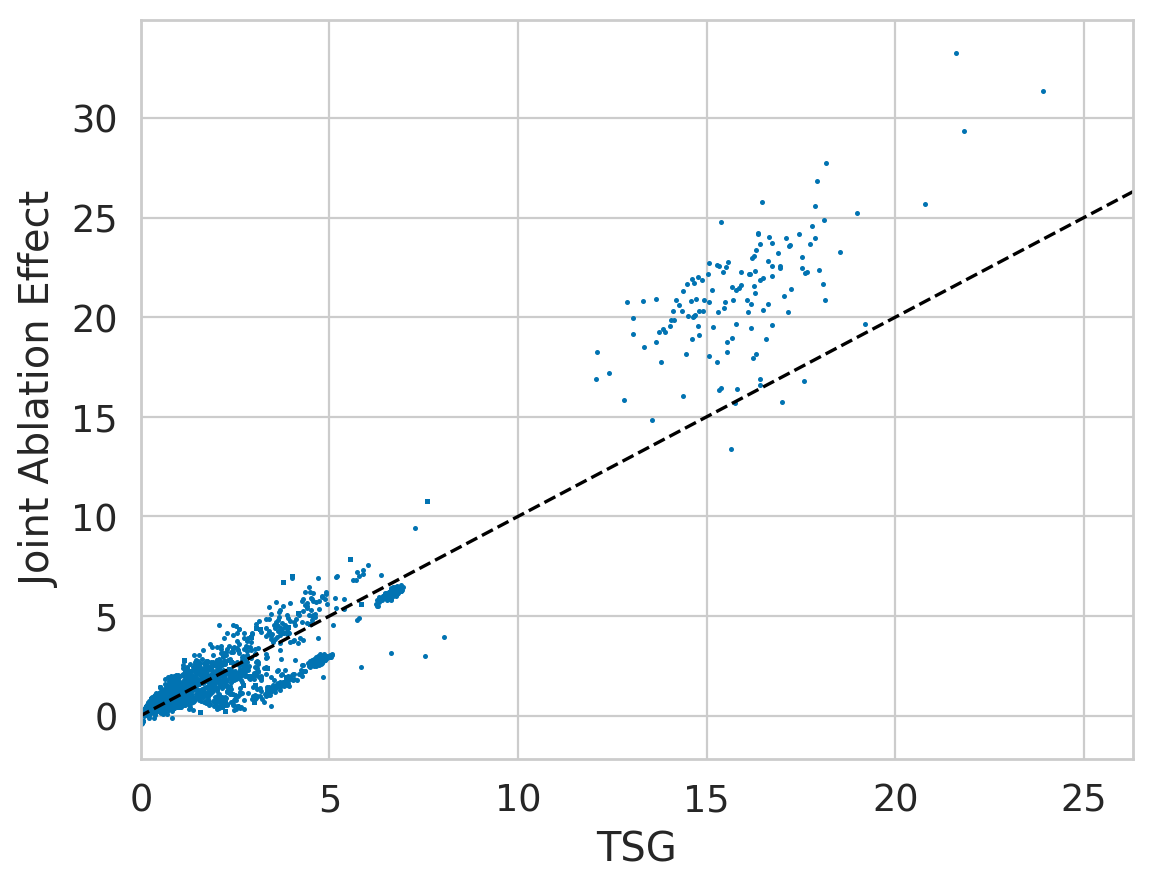}
         \caption{Hatexplain DB=0.05}
     \end{subfigure}
     \hfill
     \begin{subfigure}[b]{0.25\textwidth}
         \centering
         \includegraphics[width=\textwidth]{figures/self_repair_0.2/hatexplain/llama3-1b-it/ablation_effect.png}
         \includegraphics[width=\textwidth]{figures/self_repair_0.2/hatexplain/llama3-1b-it/grad_vs_joint.png}
         \includegraphics[width=\textwidth]{figures/self_repair_0.2/hatexplain/llama3-1b-it/tsg_vs_joint.png}
         \caption{Hatexplain DB=0.1}
     \end{subfigure}
     \hfill
     \begin{subfigure}[b]{0.25\textwidth}
         \centering
         \includegraphics[width=\textwidth]{figures/self_repair_0.2/hatexplain/llama3-1b-it/ablation_effect.png}
         \includegraphics[width=\textwidth]{figures/self_repair_0.2/hatexplain/llama3-1b-it/grad_vs_joint.png}
         \includegraphics[width=\textwidth]{figures/self_repair_0.2/hatexplain/llama3-1b-it/tsg_vs_joint.png}
         \caption{Hatexplain DB=0.2}
     \end{subfigure}
     
        \caption{Self-repair for \textbf{LLAMA-3.2 1B} using different thresholds for the coefficient of variation. The threshold determines the number of points in the figures, where the additional points have less self-repair.}
        \label{fig:self-repair-llama1-db}
\end{figure*}

To detect self-repair in the main paper, we used a threshold of 0.1 on the coefficient of variation. This threshold was chosen empirically. In \Cref{fig:self-repair-llama1-db}, we show the self-repair results using different thresholds (0.05, 0.1, and 0.2) for LLAMA 3-1B on the Fever and HateXplain datasets. Increasing the threshold results in more points in the figures, where the additional points have less self-repair. Using alternative thresholds does not alter the paper's findings.

\newpage
\clearpage

\section*{NeurIPS Paper Checklist}

\begin{enumerate}

\item {\bf Claims}
    \item[] Question: Do the main claims made in the abstract and introduction accurately reflect the paper's contributions and scope?
    \item[] Answer: \answerYes{}
    \item[] Justification: The abstract and introduction state three contributions: GIM achieving SOTA on the MIB circuit localization track, identifying and proving attention self-repair, and outperforming gradient-based feature attribution methods. Each of these claims are substantiated in \Cref{sec:self-repair-proof,sec:results} and \Cref{tab:cpr,tab:fa_results}.

\item {\bf Limitations}
    \item[] Question: Does the paper discuss the limitations of the work performed by the authors?
    \item[] Answer: \answerYes{}
    \item[] Justification: The Discussion section discusses limitations, including the assumption that perturbations affect all interacting attention scores (``Should gradients estimate the joint effect?''), the coarseness of dividing by 2 in Gradient Norm (``Gradient Norm is an oversimplification''), and the lack of a deeper theoretical understanding of why each modification helps (``Limitations'').

\item {\bf Theory assumptions and proofs}
    \item[] Question: For each theoretical result, does the paper provide the full set of assumptions and a complete (and correct) proof?
    \item[] Answer: \answerYes{}
    \item[] Justification: \Cref{prop:self-repair} states both \Cref{ass:mass} and \Cref{ass:repair} explicitly, derives the symmetric gradient form in \eqref{eq:softmax_grad}, and provides a complete proof in \Cref{sec:self-repair-proof}.

\item {\bf Experimental result reproducibility}
    \item[] Question: Does the paper fully disclose all the information needed to reproduce the main experimental results of the paper to the extent that it affects the main claims and/or conclusions of the paper (regardless of whether the code and data are provided or not)?
    \item[] Answer: \answerYes{}
    \item[] Justification: We describe models, datasets, metrics, and hyperparameters in the Experimental Setup section and \Cref{app:metrics,app:datasets,app:gim_eap_integration,app:gim_feature_attribution,app:self_repair_procedure}, and release anonymized code at the URL in the abstract.

\item {\bf Open access to data and code}
    \item[] Question: Does the paper provide open access to the data and code, with sufficient instructions to faithfully reproduce the main experimental results, as described in supplemental material?
    \item[] Answer: \answerYes{}
    \item[] Justification: An anonymized code repository with commands to reproduce every experiment is linked in the abstract; all datasets used are publicly available and listed in \Cref{sec:licenses}.

\item {\bf Experimental setting/details}
    \item[] Question: Does the paper specify all the training and test details (e.g., data splits, hyperparameters, how they were chosen, type of optimizer) necessary to understand the results?
    \item[] Answer: \answerYes{}
    \item[] Justification: Models, datasets, splits, the TSG temperature ($T=2$, selected on Gemma-2~2B/FEVER+HateXplain), counterfactual choices, and metric definitions are given in the Experimental Setup section and \Cref{app:metrics,app:datasets,app:gim_eap_integration,app:gim_feature_attribution}.

\item {\bf Experiment statistical significance}
    \item[] Question: Does the paper report error bars suitably and correctly defined or other appropriate information about the statistical significance of the experiments?
    \item[] Answer: \answerYes{}
    \item[] Justification: \Cref{tab:fa_results,tab:ablate_mod,tab:mod_results} report standard deviations in parentheses, the ablation tables additionally use independent two-sample $t$-tests at $\alpha=0.05$ to mark significant differences, and the per-layer figures (\Cref{fig:layer_aopc_llama1_2,fig:layer_aopc_gemma2,fig:layer_aopc_llama3}) show 95\% confidence intervals.

\item {\bf Experiments compute resources}
    \item[] Question: For each experiment, does the paper provide sufficient information on the computer resources (type of compute workers, memory, time of execution) needed to reproduce the experiments?
    \item[] Answer: \answerYes{}
    \item[] Justification: \Cref{sec:computation} reports that experiments were run on an A100 80GB GPU and gives wall-clock estimates per experiment type and model size.

\item {\bf Code of ethics}
    \item[] Question: Does the research conducted in the paper conform, in every respect, with the NeurIPS Code of Ethics \url{https://neurips.cc/public/EthicsGuidelines}?
    \item[] Answer: \answerYes{}
    \item[] Justification: The work uses publicly available models and datasets under their stated licenses (\Cref{sec:licenses}), involves no human subjects, and conforms with the NeurIPS Code of Ethics.

\item {\bf Broader impacts}
    \item[] Question: Does the paper discuss both potential positive societal impacts and negative societal impacts of the work performed?
    \item[] Answer: \answerYes{}
    \item[] Justification: The broader impacts are discussed in \Cref{sec:impact}. 

\item {\bf Safeguards}
    \item[] Question: Does the paper describe safeguards that have been put in place for responsible release of data or models that have a high risk for misuse (e.g., pre-trained language models, image generators, or scraped datasets)?
    \item[] Answer: \answerNA{}
    \item[] Justification: The paper releases an attribution method and analysis code only; it does not release pretrained models, generative systems, or scraped datasets that pose a high risk of misuse.

\item {\bf Licenses for existing assets}
    \item[] Question: Are the creators or original owners of assets (e.g., code, data, models), used in the paper, properly credited and are the license and terms of use explicitly mentioned and properly respected?
    \item[] Answer: \answerYes{}
    \item[] Justification: All datasets and model families are cited in the paper, and their licenses are listed in \Cref{sec:licenses}.

\item {\bf New assets}
    \item[] Question: Are new assets introduced in the paper well documented and is the documentation provided alongside the assets?
    \item[] Answer: \answerYes{}
    \item[] Justification: The released code repository (linked from the abstract) contains documentation, commands to reproduce each experiment, and prompt templates; no new datasets or pretrained models are introduced.

\item {\bf Crowdsourcing and research with human subjects}
    \item[] Question: For crowdsourcing experiments and research with human subjects, does the paper include the full text of instructions given to participants and screenshots, if applicable, as well as details about compensation (if any)?
    \item[] Answer: \answerNA{}
    \item[] Justification: The paper does not involve crowdsourcing or research with human subjects.

\item {\bf Institutional review board (IRB) approvals or equivalent for research with human subjects}
    \item[] Question: Does the paper describe potential risks incurred by study participants, whether such risks were disclosed to the subjects, and whether Institutional Review Board (IRB) approvals (or an equivalent approval/review based on the requirements of your country or institution) were obtained?
    \item[] Answer: \answerNA{}
    \item[] Justification: The paper does not involve human subjects research.

\item {\bf Declaration of LLM usage}
    \item[] Question: Does the paper describe the usage of LLMs if it is an important, original, or non-standard component of the core methods in this research?
    \item[] Answer: \answerNA{}
    \item[] Justification: LLMs are studied as the object of analysis but are not used as a non-standard component of the core methodology; any use of LLMs by the authors was limited to writing/editing assistance, which does not require declaration.

\end{enumerate}

\end{document}